\documentclass[sigconf]{acmart}
\usepackage{graphicx}
\usepackage{multirow}
\usepackage{subfigure}
\usepackage{amsmath}
\usepackage{color}

\AtBeginDocument{%
  \providecommand\BibTeX{{%
    \normalfont B\kern-0.5em{\scshape i\kern-0.25em b}\kern-0.8em\TeX}}}

\begin{document}

\author{Yiyang Shen}
\affiliation{%
\institution{Nanjing University of Aeronautics and Astronautics}
}
\email{shenyiyang114@gmail.com}

\author{Yidan Feng}
\affiliation{%
\institution{Nanjing University of Aeronautics and Astronautics}
}
\email{fengyidan1995@126.com}

\author{Sen Deng}
\affiliation{%
\institution{Nanjing University of Aeronautics and Astronautics}
}
\email{sendeng@nuaa.edu.cn}

\author{Dong Liang}
\affiliation{%
\institution{Nanjing University of Aeronautics and Astronautics}
}
\email{liangdong@nuaa.edu.cn}

\author{Jing Qin}
\affiliation{%
\institution{The Hong Kong Polytechnic University}
}
\email{harry.qin@polyu.edu.hk}
\author{Haoran Xie}
\affiliation{%
\institution{Lingnan University}
}
\email{hrxie@ln.edu.hk}
\author{Mingqiang Wei}
\affiliation{%
\institution{Nanjing University of Aeronautics and Astronautics}
}
\email{mingqiang.wei@gmail.com}

\title{MBA-RainGAN: Multi-branch Attention Generative Adversarial Network for Mixture of Rain Removal from Single Images}

\begin{abstract}

Rain severely hampers the visibility of scene objects when images are captured through glass in heavily rainy days.
We observe three intriguing phenomenons that, 1) rain is a mixture of  \textit{raindrops}, \textit{rain streaks} and \textit{rainy haze}; 2) the depth from the camera determines the degrees of object visibility, where objects nearby and faraway are visually blocked by rain streaks and rainy haze, respectively; and 3) raindrops on the glass randomly affect the object visibility of the whole image space. We for the first time consider that, the overall visibility of objects is determined by the mixture of rain (MOR). However, existing solutions and established datasets lack full consideration of the MOR. In this work, we first formulate a new rain imaging model; by then, we enrich the popular RainCityscapes by considering raindrops, named RainCityscapes++. Furthermore, we propose a multi-branch attention generative adversarial network (termed an MBA-RainGAN) to fully remove the MOR. The experiment shows clear visual and numerical improvements of our approach over the state-of-the-arts on RainCityscapes++. The code and dataset will be available.

\end{abstract}


\maketitle
\section{Introduction}
Rain is one of the commonest dynamic weather phenomena. Images captured in rainy conditions generally undergo
degradations such as low visibility and distortion in local areas, which directly impair the visual perception quality and render these images worthless for sharing and usage \cite{li2018non,fan2018residual}. Moreover, artifacts caused by rainwater may severely hamper the performance of many computer vision applications, including intelligent driving and outdoor surveillance systems \cite{yu2019gradual,yang2019single}.
\begin{figure}[htbp]
	\centering{
		\subfigure[Window]{
			\includegraphics[width=0.3\linewidth]{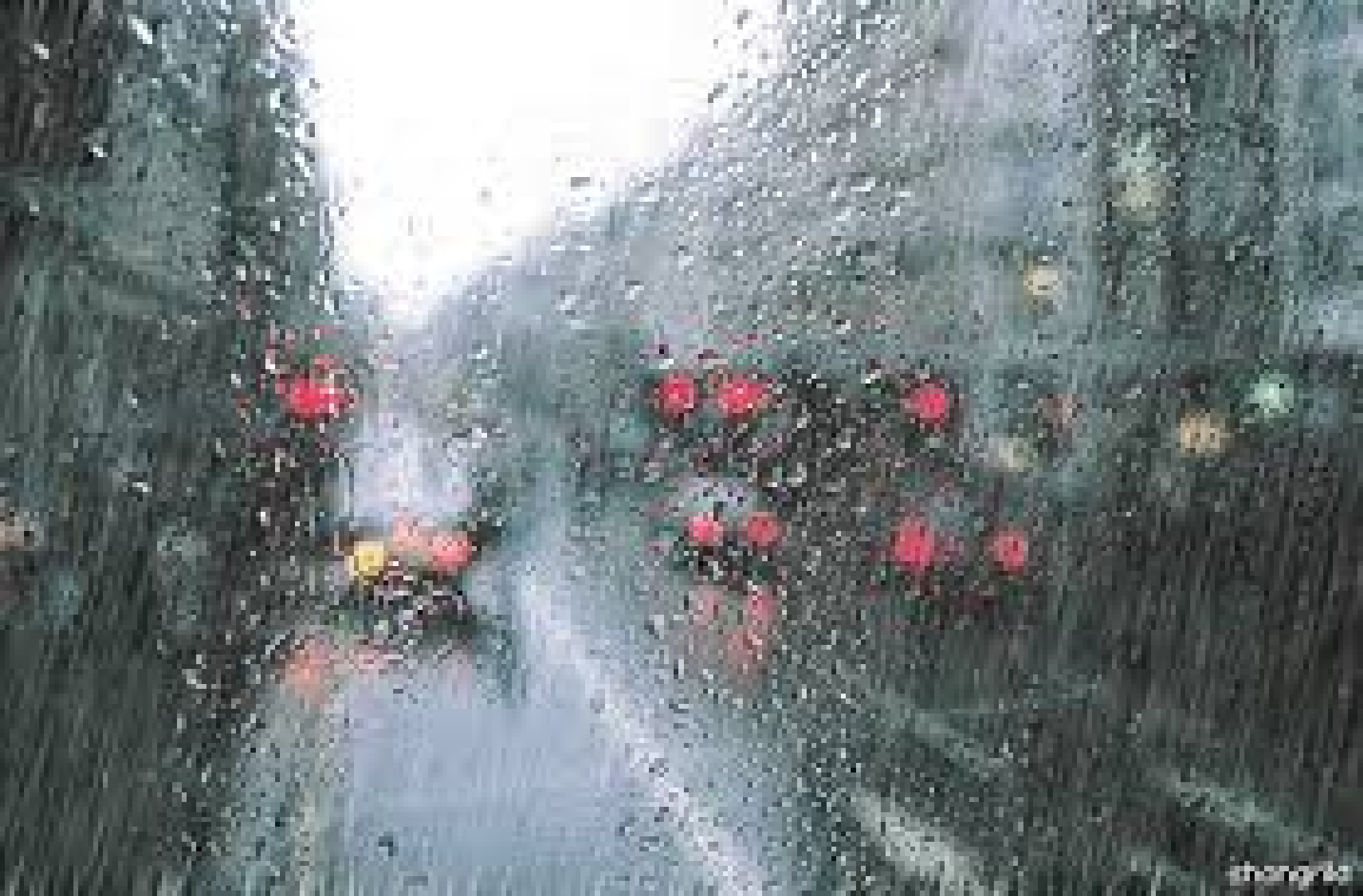}}
		\subfigure[Windshield]{
			\includegraphics[width=0.3\linewidth]{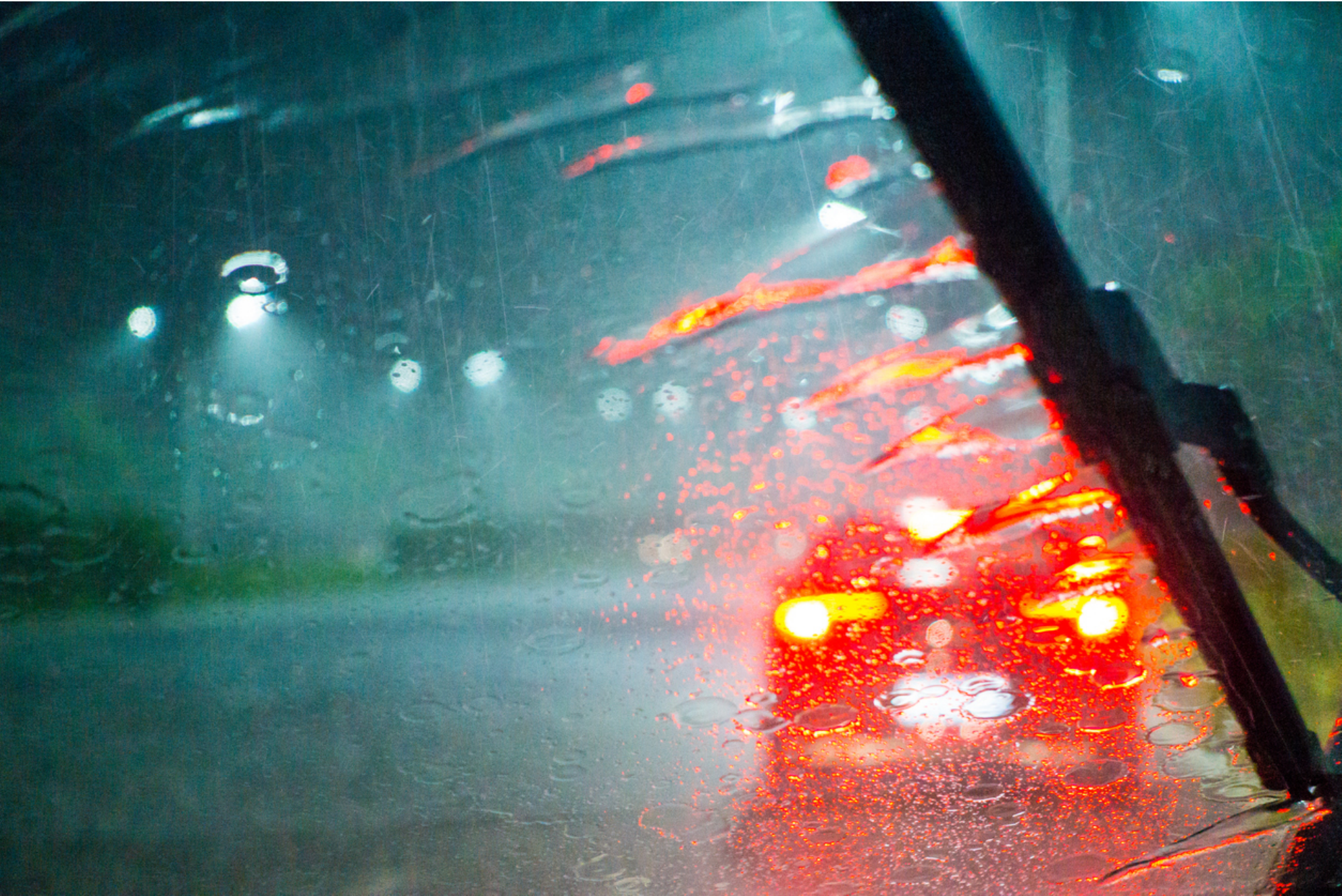}}
		\subfigure[Surveillance]{
			\includegraphics[width=0.3\linewidth]{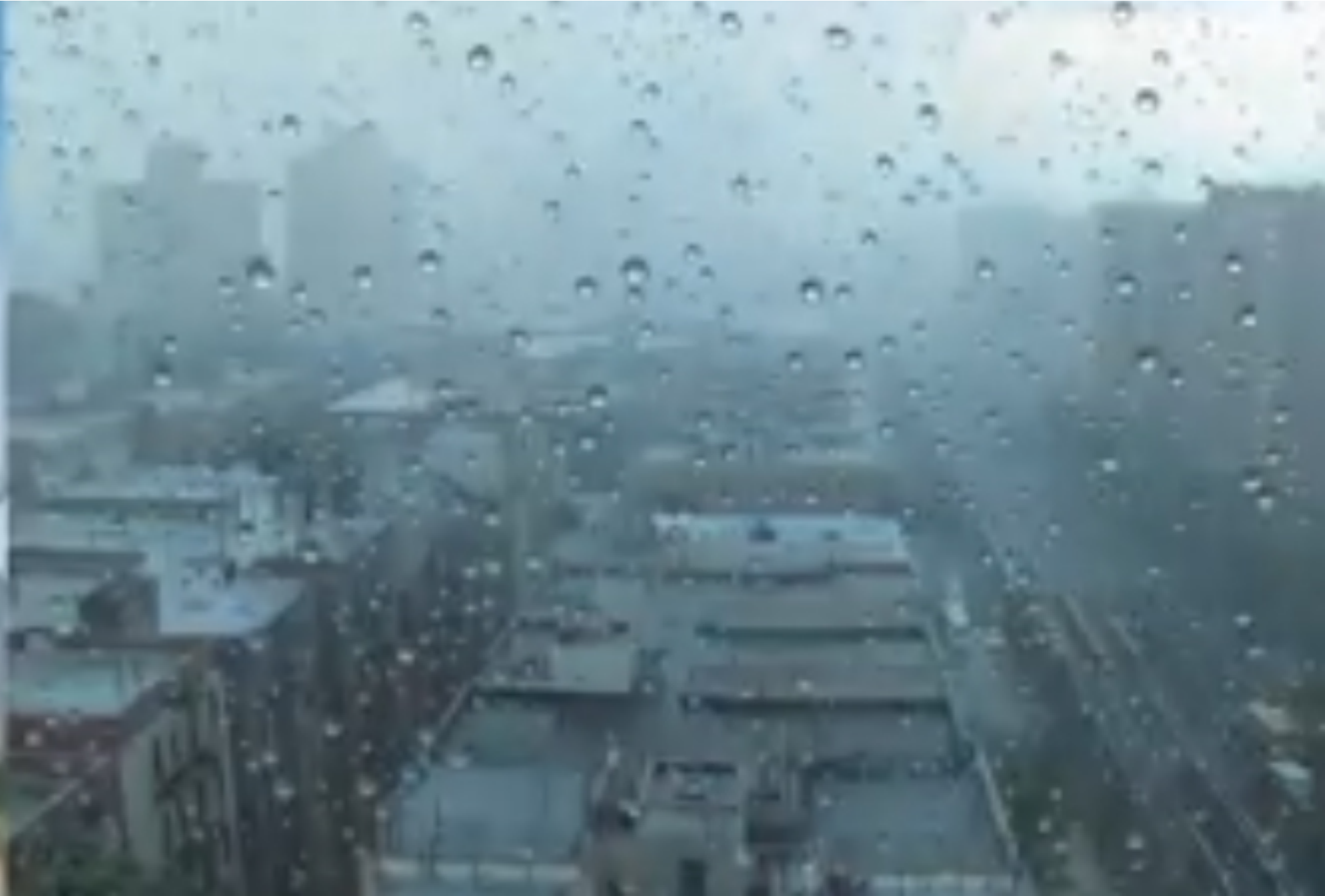}}
		
		\caption{Real rainwater artifacts.}
		\label{real}
	}
\end{figure}
Compared with video-based rain removal, single image deraining is more challenging due to the lack of sequence information. Conventional methods mainly rely on various image priors, such as Gaussian Mixture Model \cite{p_li2016gmm}, sparse coding \cite{p_kang2011automatic,p_kang2012self} and the low rank representation \cite{p_chen2013generalized}, and solve an optimization problem. Recently, deep neural networks have shown its power in coping with this ill-posed problem. By learning a complex model from massive human-selected data, the performance of image deraining is substantially boosted. However, most researches only focus on single types of rainwater artifacts, i.e., rain streaks only or raindrops only, regardless of the fact that rainwater is transformable and can appear in various forms under different shooting conditions. Hu et al. consider that heavy rain is a mixture of dynamic rain streaks and diffusing rainy haze \cite{hu2019depth}. These two rainwater artifacts are complexly correlated and difficult to be removed by existing methods. This difficulty can be eased by a proper formulation of the mixture degradation. Thus, they formulate these two effects by the scene depth, which endorses the effectiveness of a depth attentional network and guides  the construction of their RainCityscapes dataset. In addition, several different formulations are proposed in \cite{d_haze_shen2018joint,d_haze_wang2019two-branch,li2019heavy} and are integrated into the network by guiding either the learning of deep embedding or the design of the whole architecture.

In this paper, we study a more comprehensive rain removal problem by taking photographing conditions into account, which includes outdoor camera lens without protection, indoor photographing through windows and driver assistance systems behind the windshield (see Fig. \ref{real} for a reference). Images taken under these scenarios suffer from the mixture of rain (MOR), that is, the effect of rainwater from close to far captured as raindrops, rain streaks and rainy haze. Based on this observation, we enhance the formulation in \cite{hu2019depth} by considering the existence of raindrops, where we regard the raindrop location as an important prior knowledge for dissolving the MOR problem. Accordingly, we construct an improved version of the RainCityscapes dataset, named RainCityscapes++, by compositing real raindrop layers on the images affected by both rain streaks and rainy haze to reproduce realistic MOR affected scenes.

To remove the entangled MOR effect, we adopt a three-stage decomposition strategy. First, based on the distinctivity in the frequency domain, we separate rain streaks and rainy haze by a low-pass filter. Next, we learn separate attention maps for each form of rainwater artifacts using a multi-branch structure. These attention maps are extracted by collateral recurrent networks in a coarse-to-fine manner, which progressively guide the final image decomposition in the contextual autoencoder. Furthermore, we adopt an attentive discriminator for image-level constraint to ensure the fidelity of the output. In conclusion, our contributions are:

1) We for the first time consider the mixture of rain (MOR) in the image rain removal problem and formulate this entangled effect by integrating the scene depth and raindrop location.

2) We enrich the popular RainCityscapes dataset, named RainCityscapes++, by superimposing real raindrop layers on the images affected by both rain streaks and rainy haze, which accommodates the randomness in raindrop distribution and thus reproduces realistic MOR affected scenes.

3) We propose the multi-branch attention scheme especially for mixture artifact removal, based on which we present our MBA-RainGAN for MOR removal. Extensive experiments show the superiority of MBA-RainGAN in removing the intricately entangled degradations in rainy scenes.

\section{Related Work}
Current studies on image rain removal can be categorized into three groups according to the rainwater forms: \textit{rain streak removal}, \textit{rain streak} \& \textit{rainy haze removal} and \textit{raindrop removal}. The majority of them emphasize on tackling rain streaks of disparate directions and densities, while Hu et al. \cite{hu2019depth} first investigate the mixture of rain streaks and rainy haze. Meanwhile, there have been researches on raindrop removal, but are independent from the rainy scene.

\subsection{Rain Streak Removal}
Rainwater dynamically falling in the air is imaged as streaks, which are in similar directions and sparsely distributed in the whole image \cite{garg2007vision}. Many researches exploit these physical properties to build various image priors in an optimization framework. In \cite{p_kang2011automatic,p_fu2011single,p_kang2012self}, it is assumed that rain streaks are sparse and of high frequency in the image, and thus the problem is translated into progressive image decomposition via dictionary learning. Sun et al. \cite{p_sun2014exploiting} and Chen et al. \cite{p_chen2014visual} propose to introduce structural similarity and depth of field for further regularization, respectively. In \cite{p_chen2013generalized,p_zhang2017convolutional,p_du2018single}, the low rank representations are incorporated into rain streak removal. Luo et al. \cite{p_luo2015removing} propose a novel sparse coding scheme considering the discrimination of background and rain streaks. Li et al. \cite{p_li2016gmm} adopt Gaussian Mixture Model (GMM) to explore patch-based priors for both the clean and rain layers. In \cite{p_wang2017hierarchical} and \cite{p_gu2017joint}, guided filter and analysis sparse representation are integrated into sparse coding for finer layer decomposition.

The introduction of neural networks brings a leap of the performance in image deraining tasks. Fu et al. \cite{d_detail_layer} first adopt a residual network to learn negative rain streaks after extracting the high frequency layer from the original image. Thereafter, many efforts are made either to introduce advancing network modules and structures, or to integrate problem-related knowledge into network design. Network modules, such as dense block \cite{d_wang2019erl,d_li2018non,d_fan_residual}, recursive block \cite{d_fan_residual,d_ren2019progressive} and dilated convolution
 \cite{d_deng2019drd, yang2017deep}, and structures, such as RNN \cite{d_rescan,d_ren2019progressive}, GAN \cite{li2019heavy,d_pu2018cyclegan,d_jin2019ai,d_zhang_gan,zhang2019image} and multi-stream networks \cite{yang2017deep,d_liu2019dual,d_qin2019nasnet,d_wei2019coarse,d_deng2019drd}, are validated to be effective in rain streak removal. Auxiliary information, including rain density \cite{zhang2018density}, streak position \cite{yang2017deep}, gradient information \cite{d_wang2019gradient} and motion blur kernel \cite{d_wang2018kernel}, are leveraged to improve the robustness and performance of deraining networks. Particularly, Wang et al. \cite{wang2019spatial} propose to build a real rain dataset using video-based deraining results and Li et al. \cite{d_li2019bench} present a comprehensive benchmark named MPID for evaluation of various deraining methods.

\subsection{Rain Streak \& Rainy Haze Removal}
Studies on image dehazing \cite{h_cai2016dehazenet,h_he2010single,h_yang2018towards,h_zhang2018densely} are independent to image deraining until Hu et al. \cite{hu2019depth} observe that rainwater faraway in the scene actually generates haze-like effects and thus combine rain and haze removal in a unified framework. Specifically, the depth information is utilized to produce a spatial attention map that guides a deep convolutional network. Besides, Li et al. \cite{li2019heavy} design a complex composite network which contains three sub-networks to learn the unknowns in the formulation of mixed rain streak and rainy haze and then refine the calculated result via a depth-guided GAN. A similar scheme is utilized by Wang et al. \cite{d_haze_wang2019two-branch} but with a different mixture formulation. In \cite{d_haze_shen2018joint}, Shen et al. set two disparate objective functions for only one dense block, which aim to minimize the distance between dark channels for dehazing and between DWT features for rain streak removal.

\subsection{Raindrop Removal}
Single-image raindrop removal is a challenging task since complex reflection effects caused by raindrops lead to the loss of background information in local areas of unknown shape and location.
Eigen et al. \cite{eigen2013restoring} first explores the data-driven method via a three-layer CNN, which is incapable to preserve image details and remove large raindrops. Subsequently, Qian et al. \cite{qian2018attentive} construct a raindrop dataset where the raindrop is produced by spraying water on the glass. They solve the raindrop occlusion problem by integrating spatial attention of raindrop location into an inpainting network and have achieved promising results on their test dataset. Raindrop location is also exploited in \cite{hao2019learning}, but is blended with a preliminary raindrop removal result to build input features for a refinement network.


\begin{figure*}[ht]
	\centering
	\subfigure[Background layer A]{
		\begin{minipage}[b]{0.18\linewidth}
			\includegraphics[width=1\linewidth]{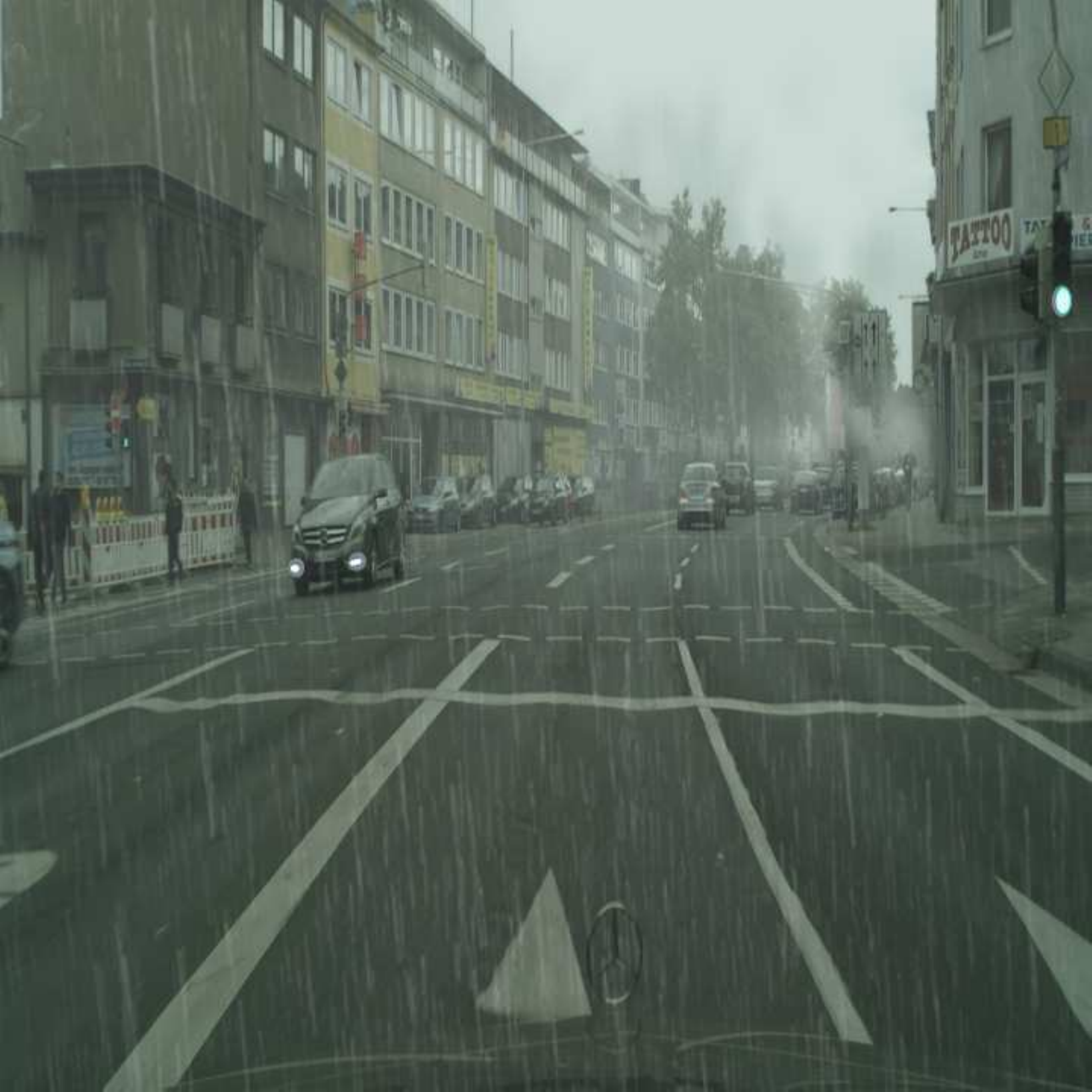}\vspace{2pt}
			\includegraphics[width=1\linewidth]{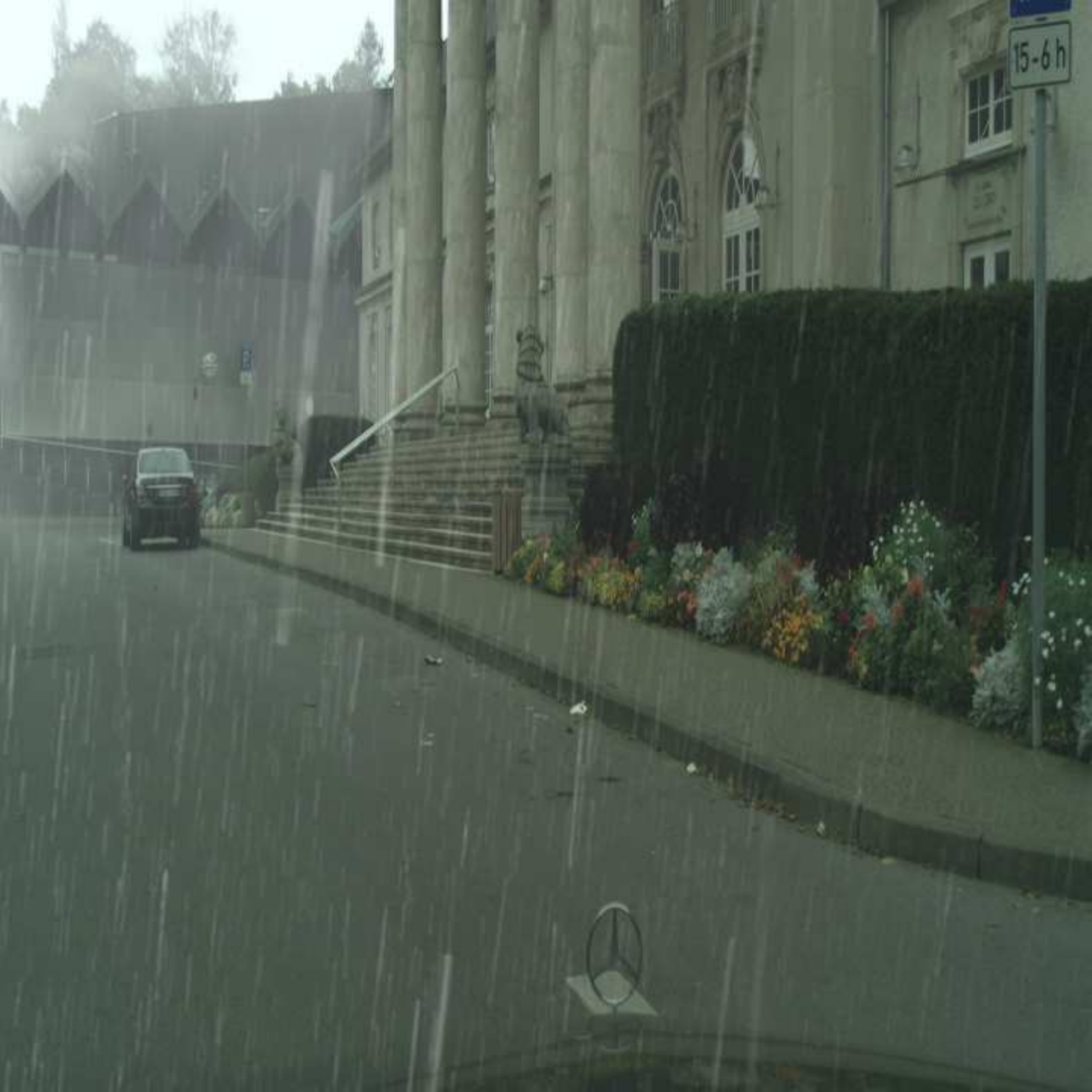}\vspace{2pt}
			\includegraphics[width=1\linewidth]{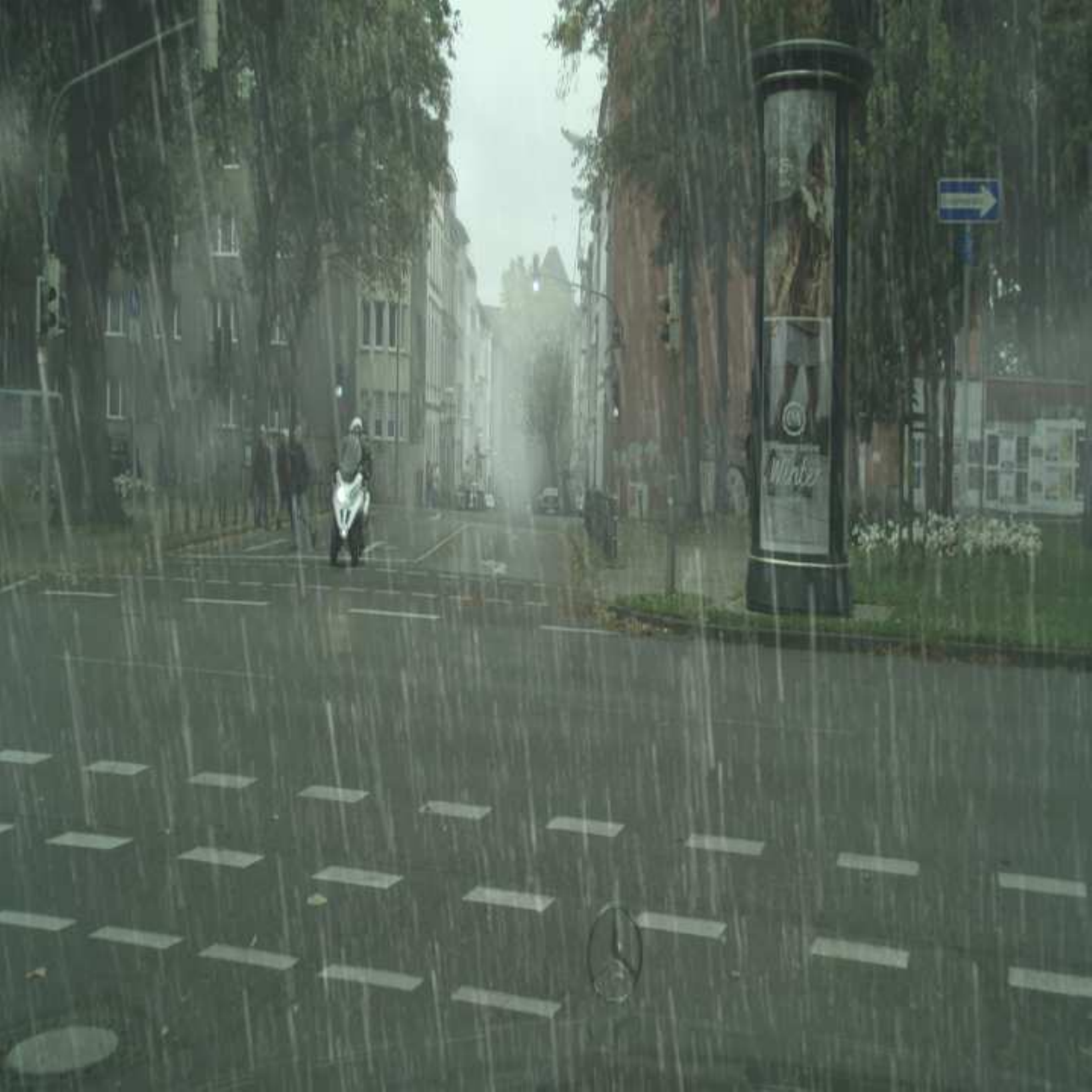}\vspace{2pt}
	\end{minipage}}
	\subfigure[Cover layer B]{
		\begin{minipage}[b]{0.18\linewidth}
			\includegraphics[width=1\linewidth]{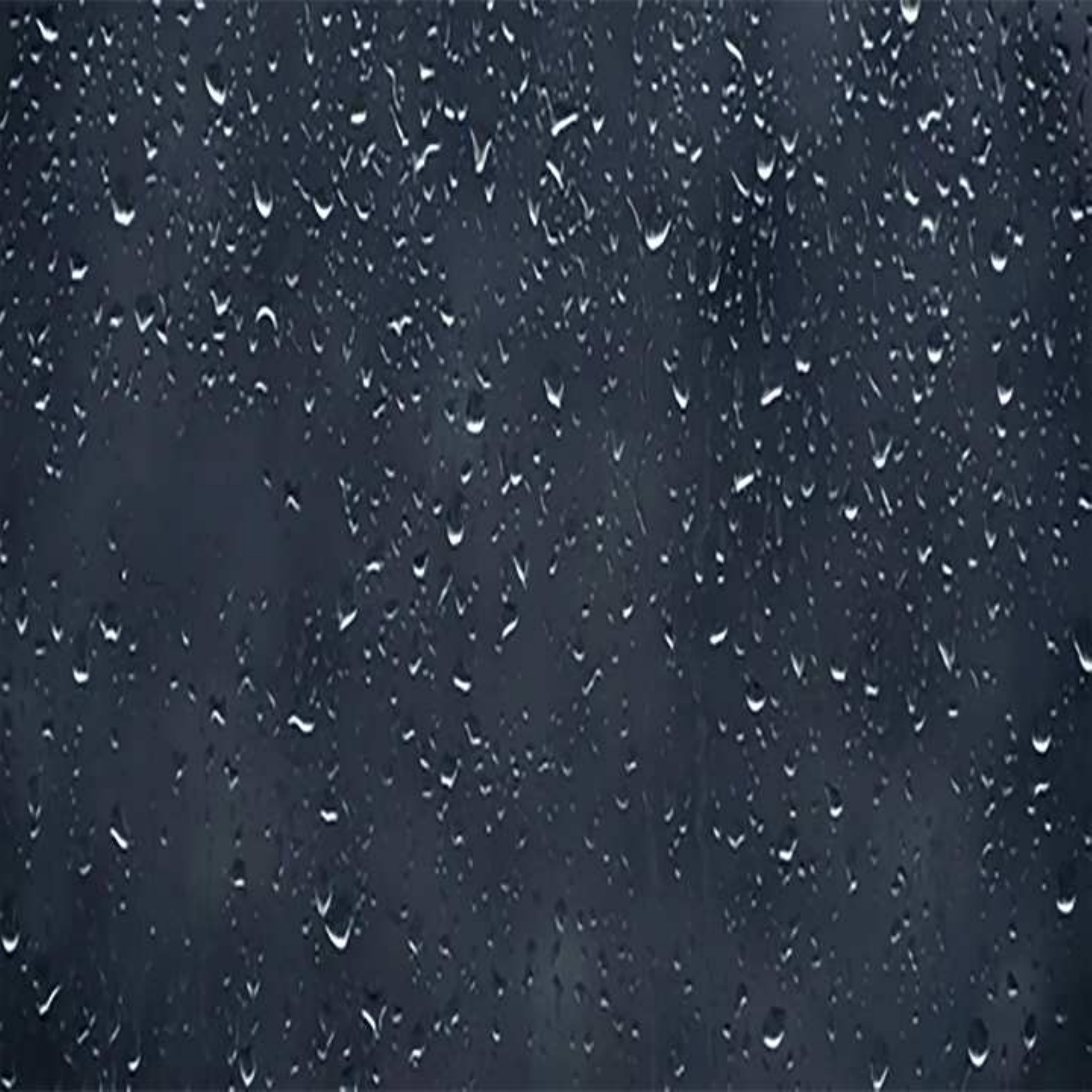}\vspace{2pt}
			\includegraphics[width=1\linewidth]{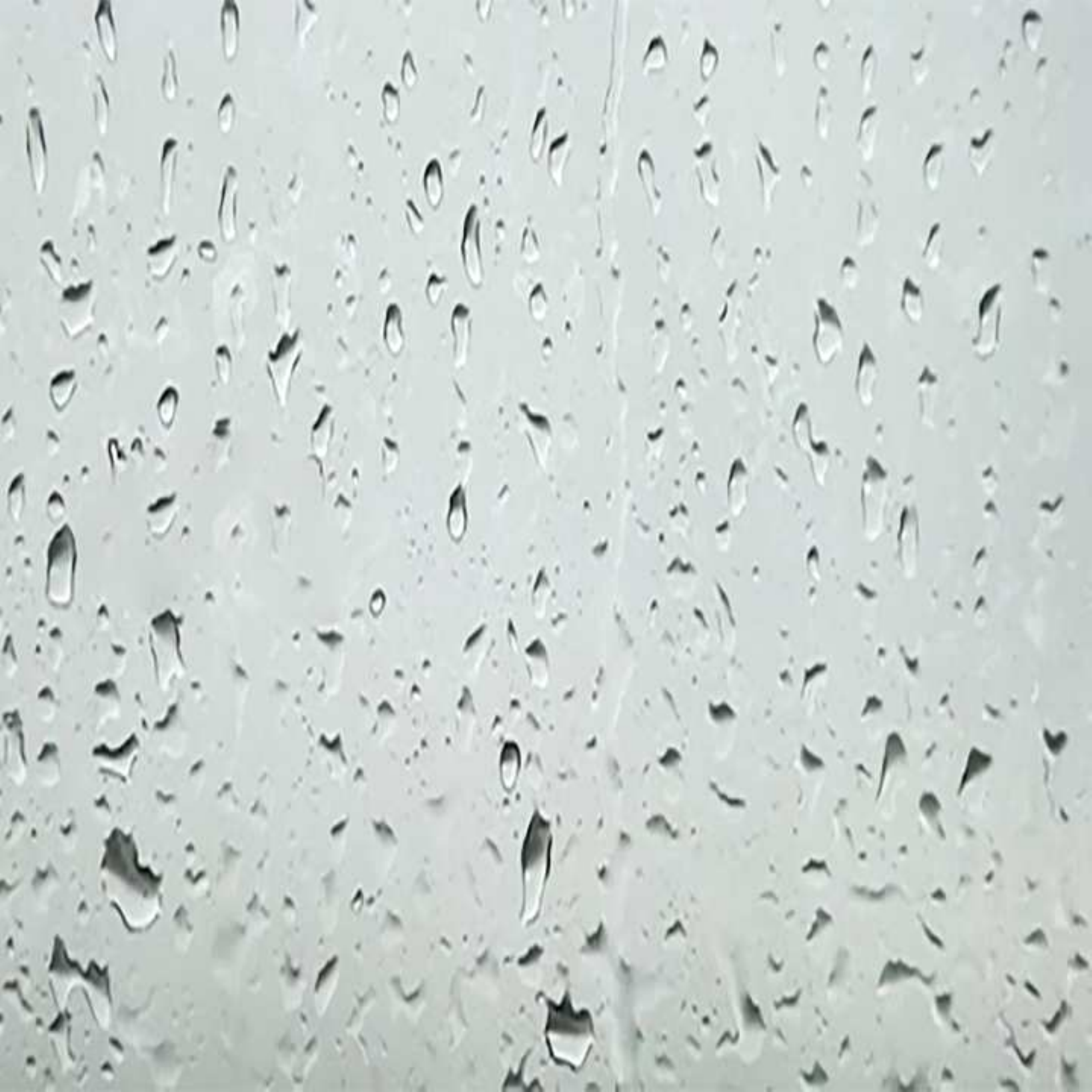}\vspace{2pt}
			\includegraphics[width=1\linewidth]{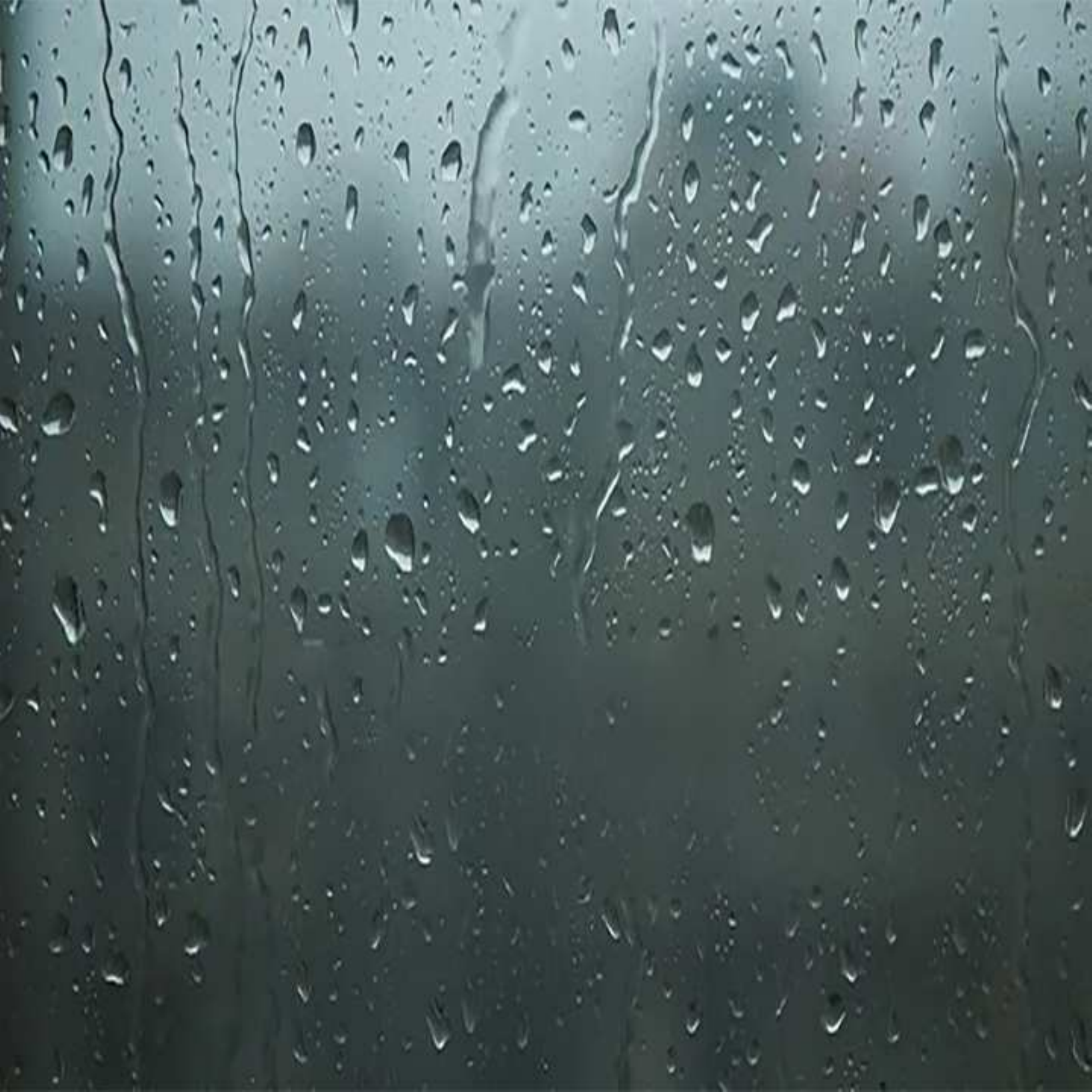}\vspace{2pt}
	\end{minipage}}
	\subfigure[Overlay mode $C_1$]{
		\begin{minipage}[b]{0.18\linewidth}
			\includegraphics[width=1\linewidth]{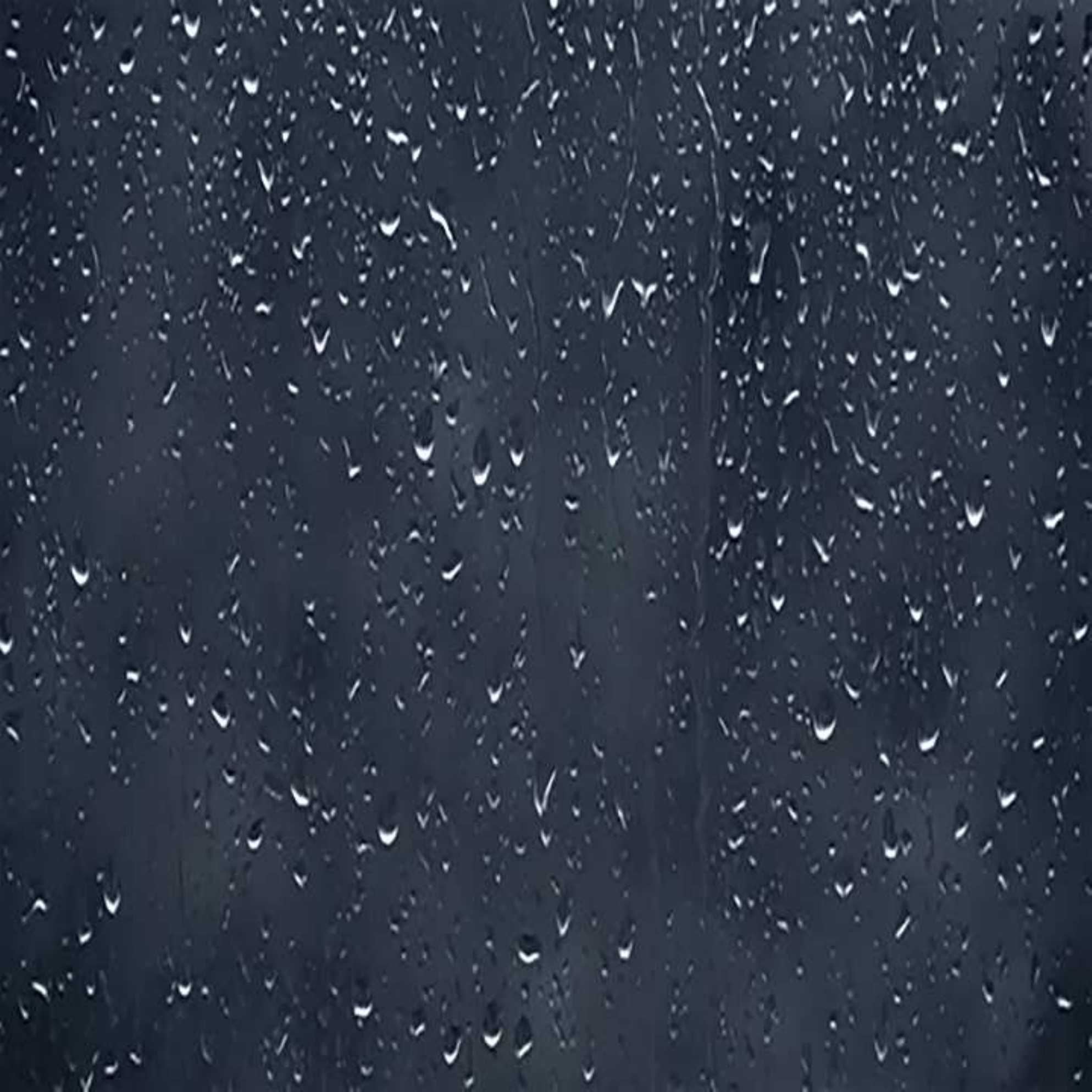}\vspace{2pt}
			\includegraphics[width=1\linewidth]{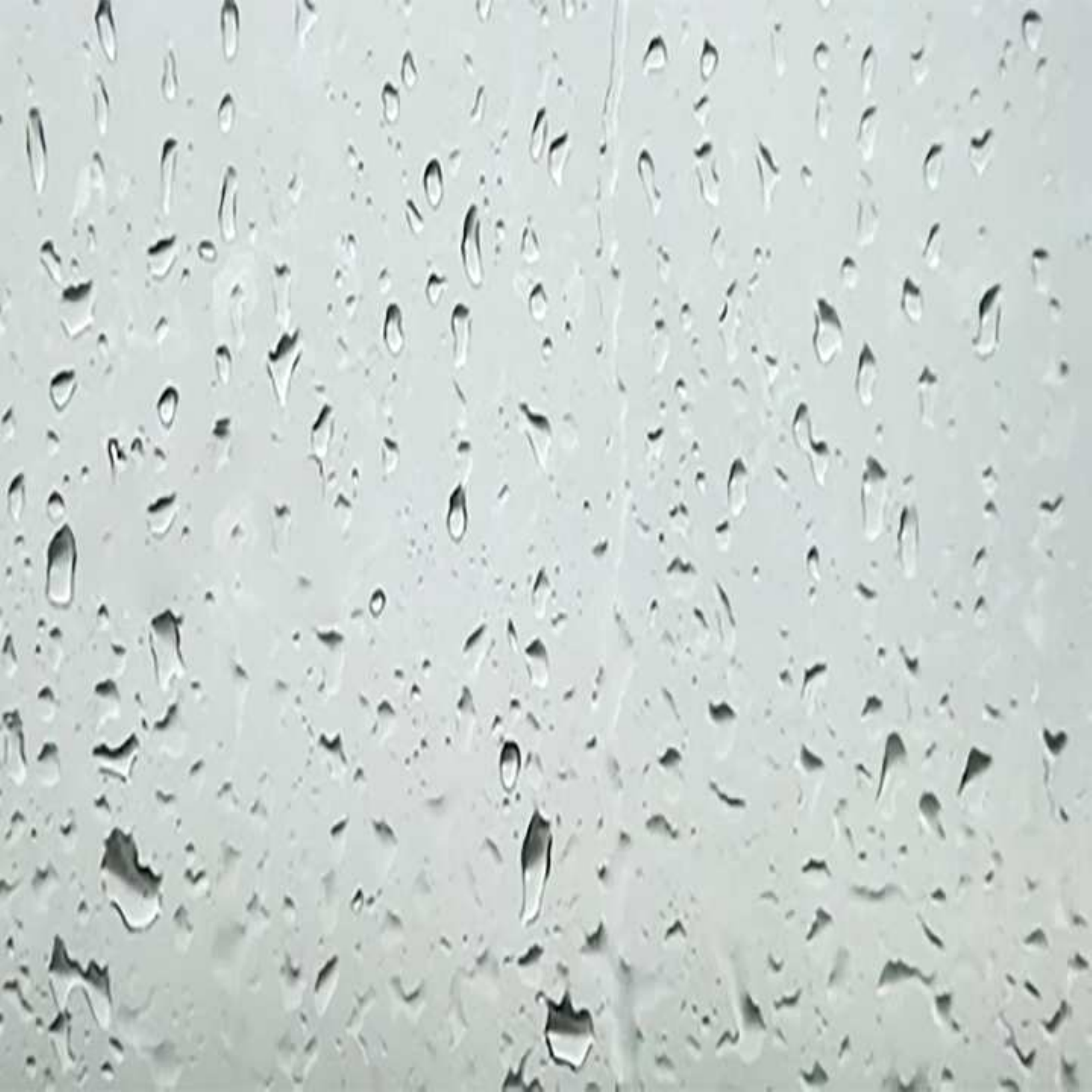}\vspace{2pt}
			\includegraphics[width=1\linewidth]{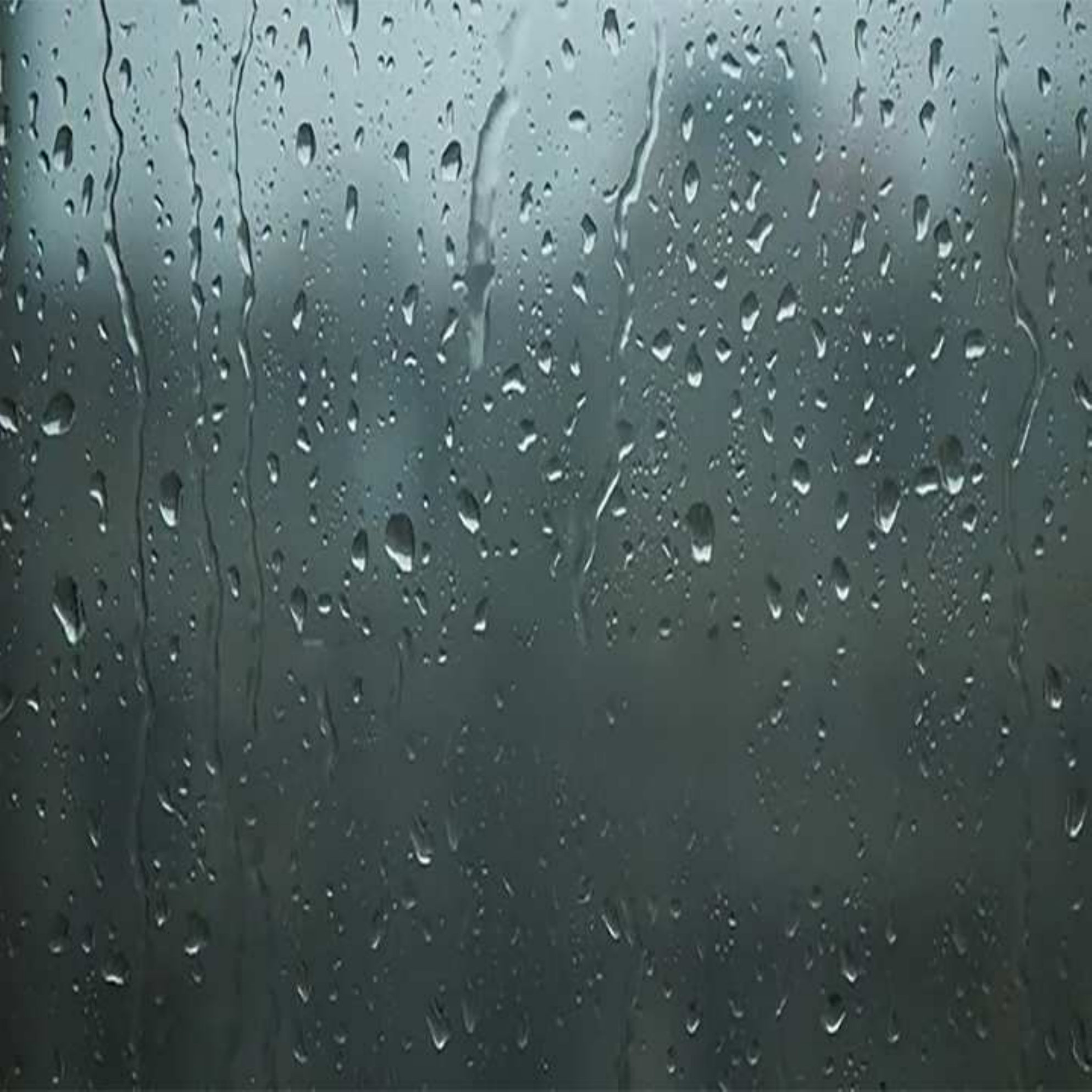}\vspace{2pt}
	\end{minipage}}
	\subfigure[Highlight mode $C_2$]{
		\begin{minipage}[b]{0.18\linewidth}
			\includegraphics[width=1\linewidth]{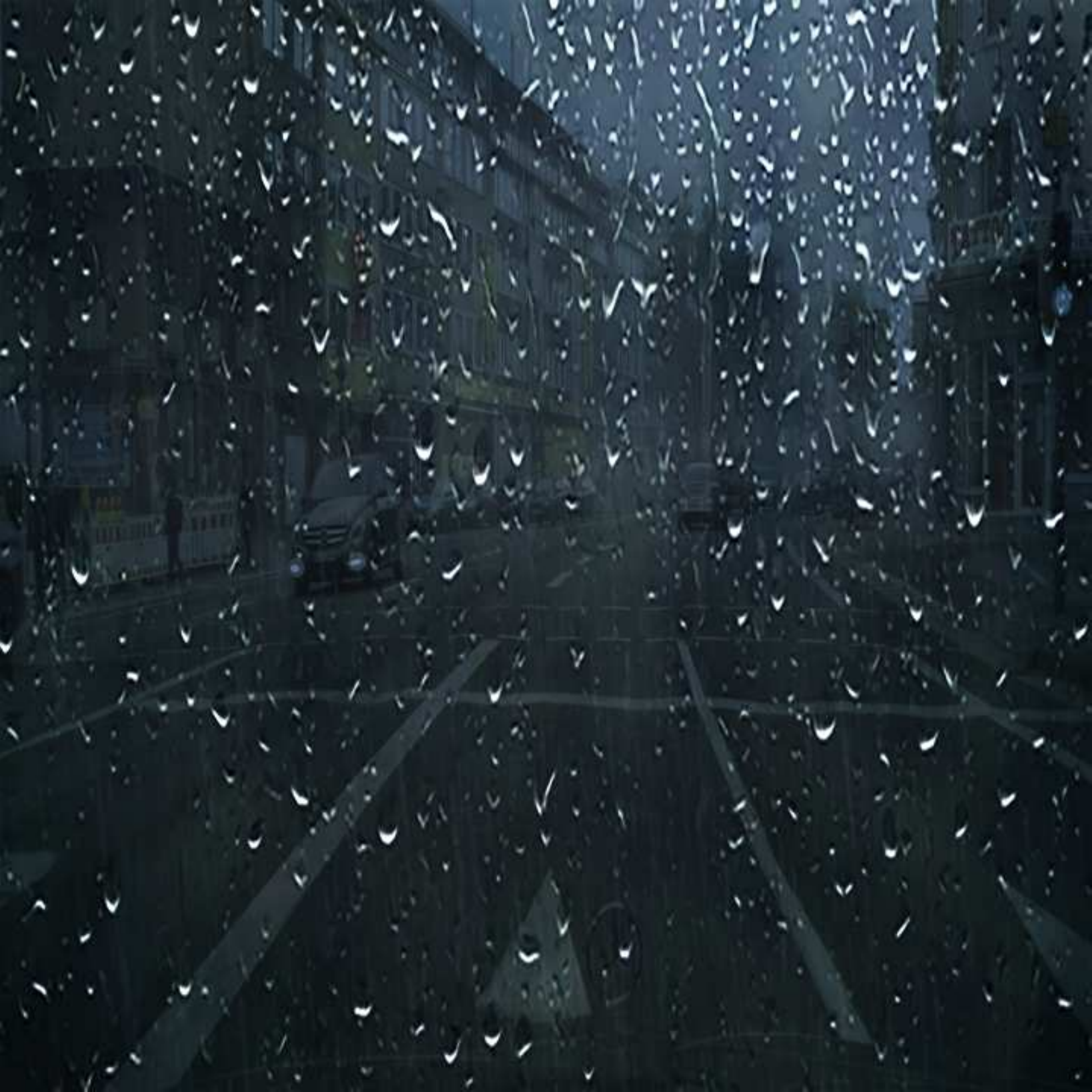}\vspace{2pt}
			\includegraphics[width=1\linewidth]{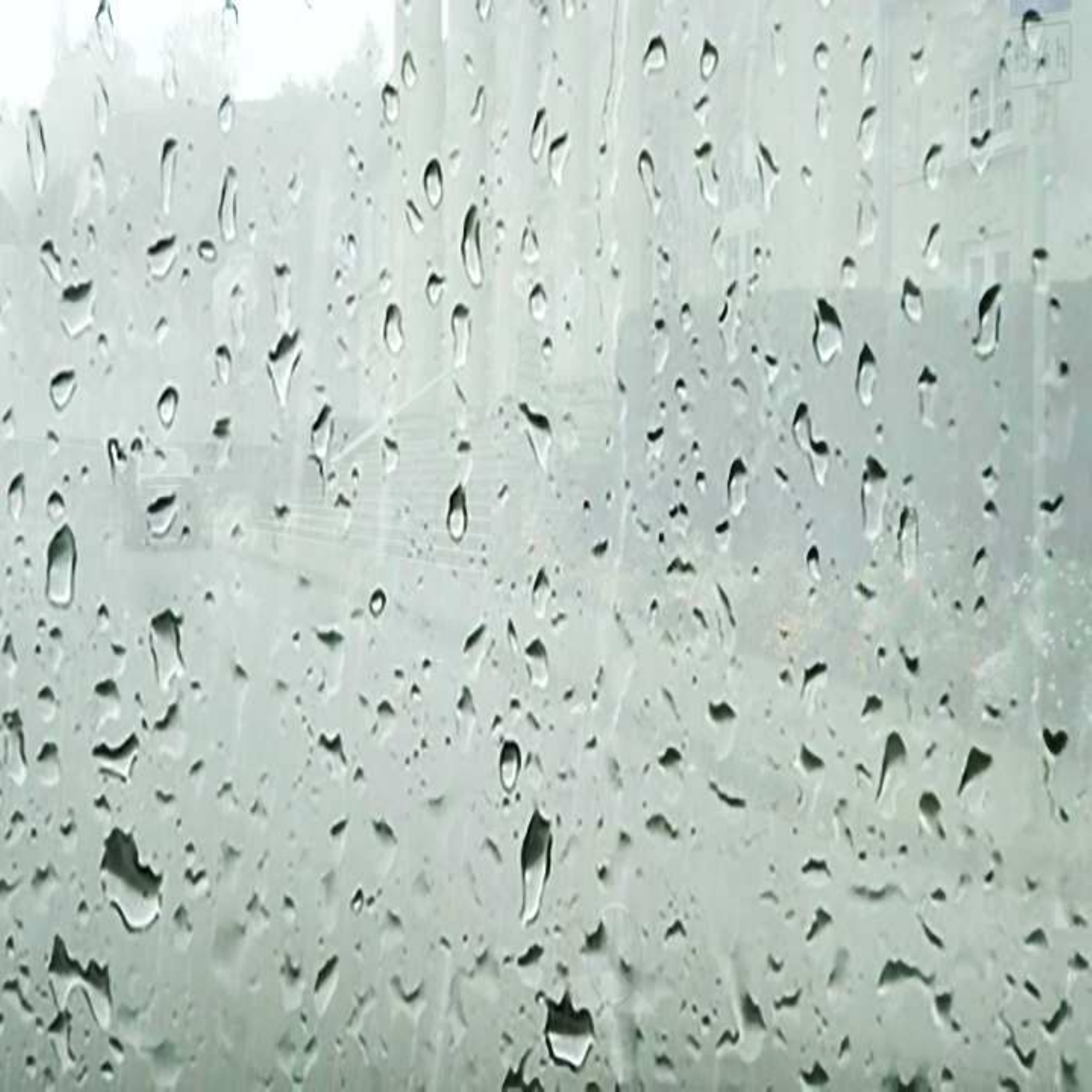}\vspace{2pt}
			\includegraphics[width=1\linewidth]{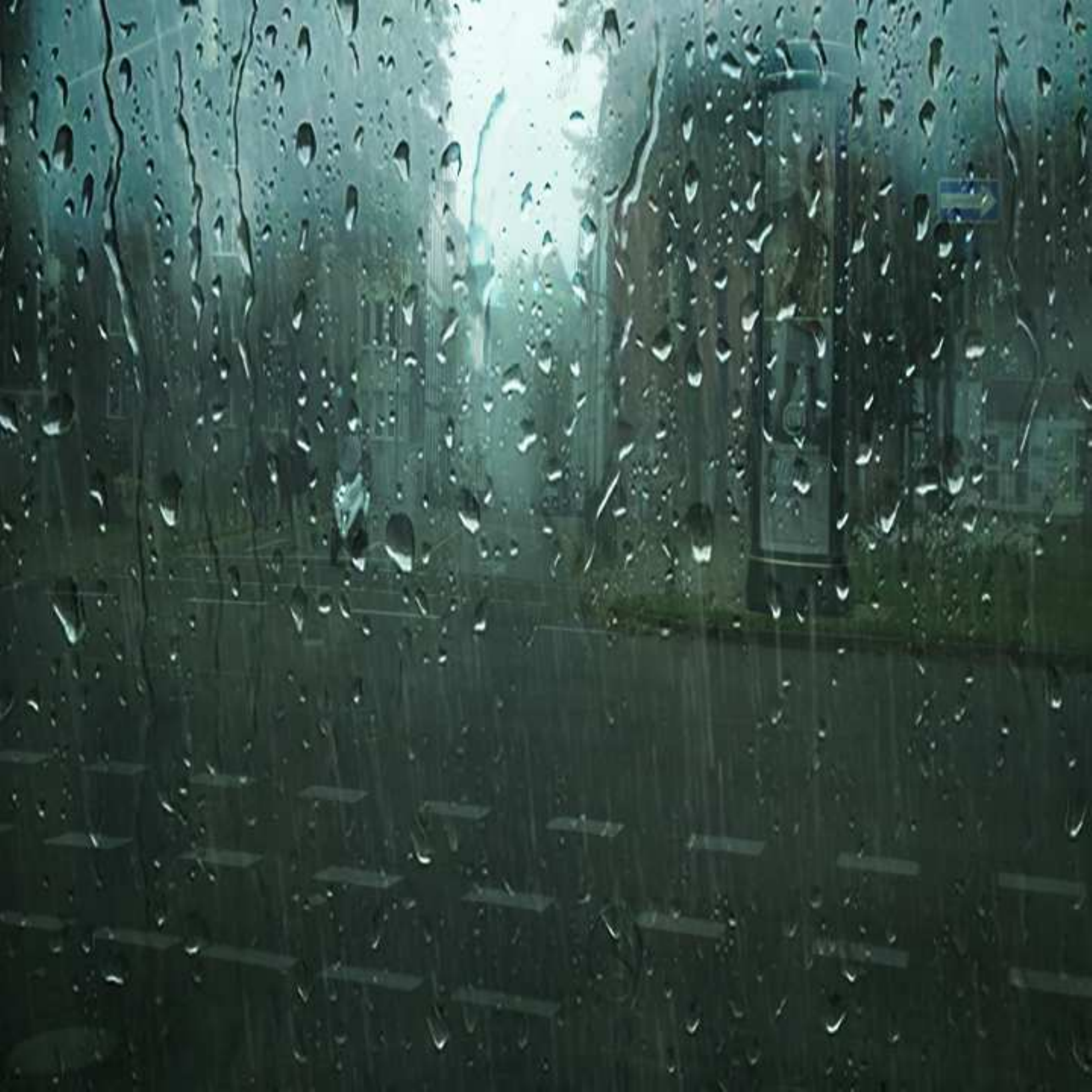}\vspace{2pt}
	\end{minipage}}
	\subfigure[Final mode $C_3$/$C_4$]{
		\begin{minipage}[b]{0.18\linewidth}
			\includegraphics[width=1\linewidth]{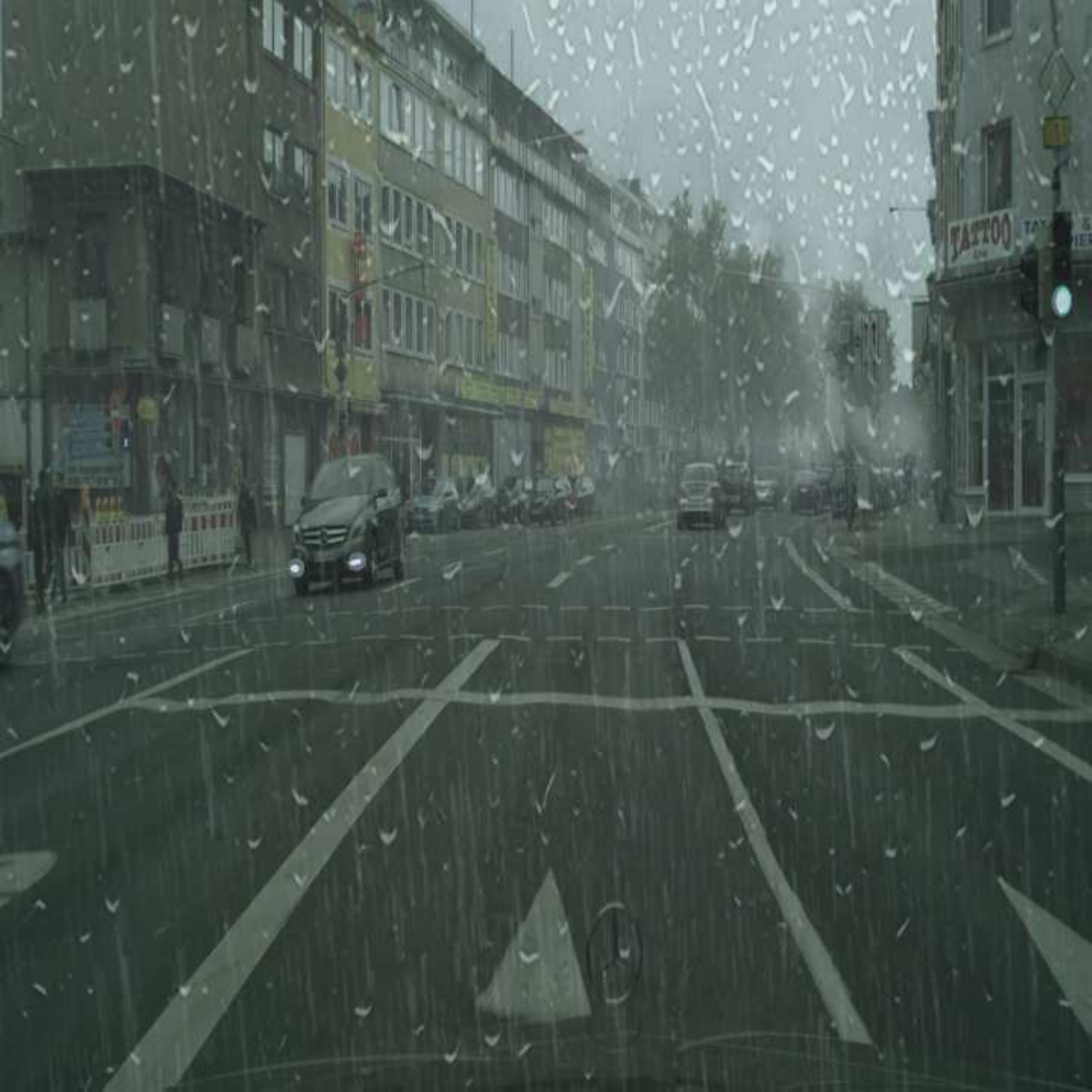}\vspace{2pt}
			\includegraphics[width=1\linewidth]{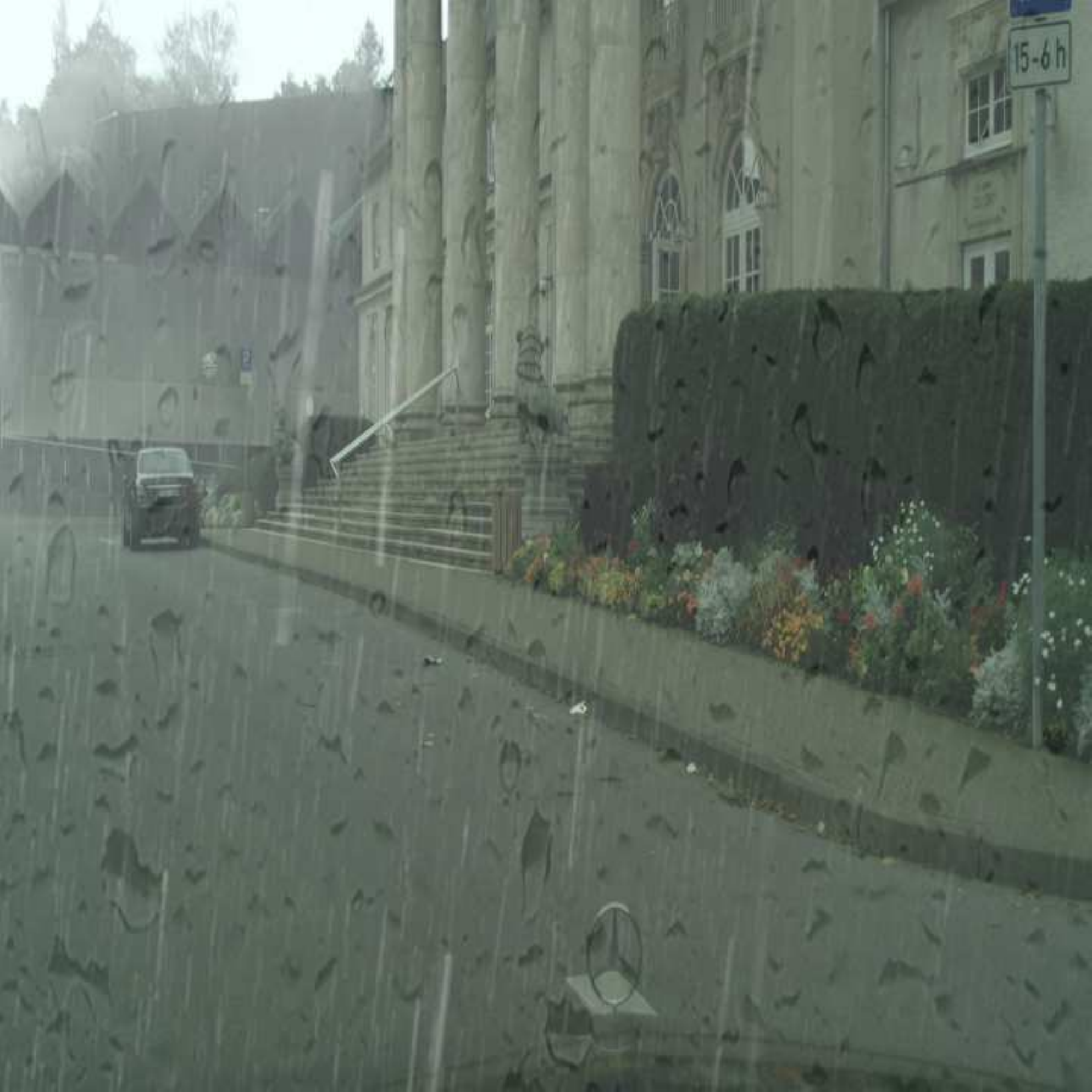}\vspace{2pt}
			\includegraphics[width=1\linewidth]{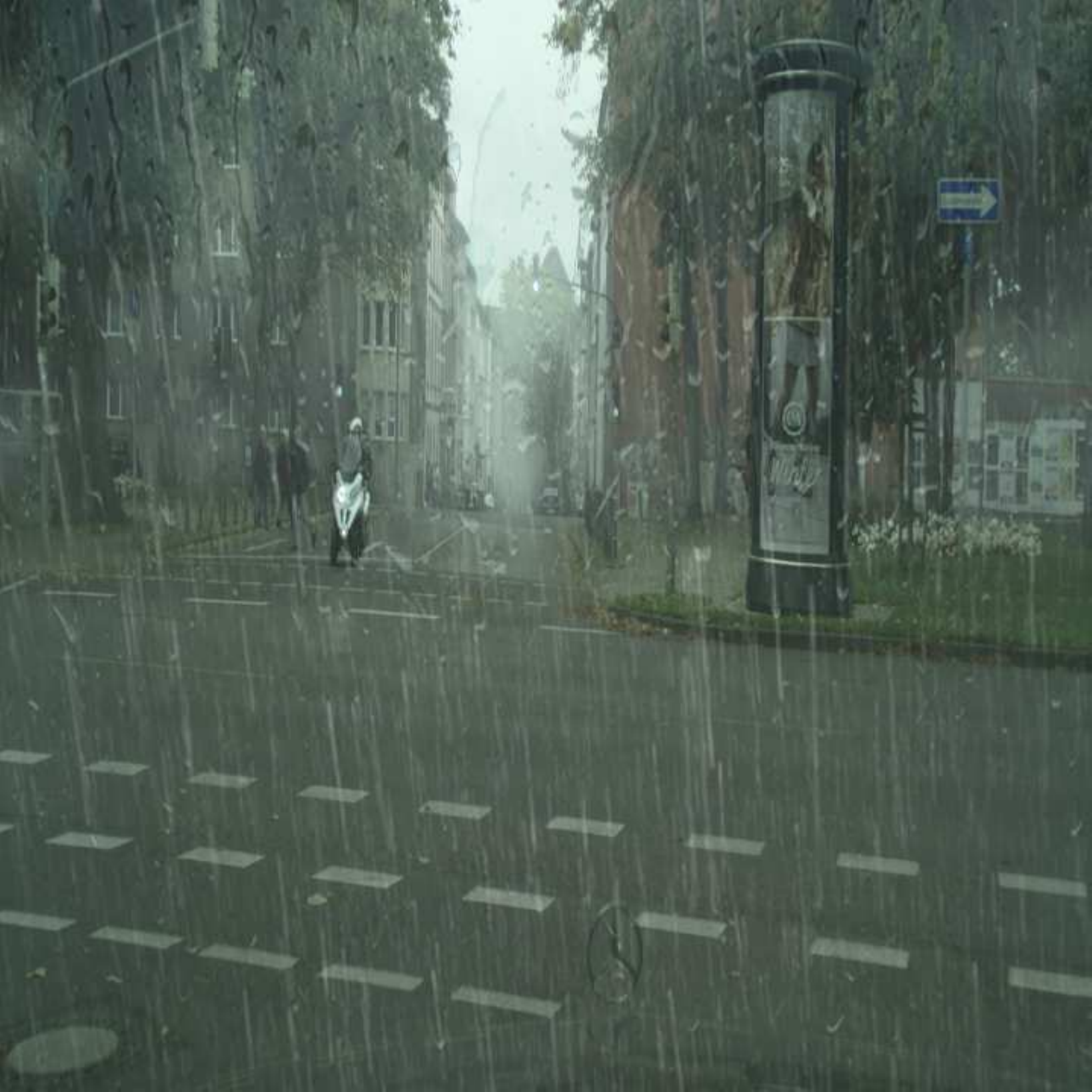}\vspace{2pt}
	\end{minipage}}
	\caption{ Three sets of example images in our dataset RainCityscapes++.}
	\label{fig:label1}
\end{figure*}
\section{Formulation and Dataset}
\subsection{Rain Image Formulation}

We consider that an image degraded by rain composes of a background image and a mixture of three layers, i.e., the raindrop layer, the rainy haze layer and the rain streak layer.
Differently from existing image rain formulations, we devise the captured rain image $I(x)$ at a pixel $x$ as:
\begin{equation}\label{eq:mixture_rain}
\begin{aligned}
 I(x)= &(1-M_{d}(x))\cdot[B(x)(1-S(x)-A(x))+S(x)
 \\&+A_{0}A(x)]+D(x)
\end{aligned}    
\end{equation}
where $M_{d}(x)\in \{0,1\}$ means whether the pixel $x$ is corrupted by raindrops ($1$ is Yes, and $0$ is No), $B(x)$ denotes the rain-free background image with the clear scene radiance, $S(x)\in[0,1]$ and $A(x)\in[0,1]$ are the rain streak layer and the rainy haze layer in accordance with \cite{hu2019depth}, and $D(x)\in[0,1]$ denotes the raindrop layer. $A_0$ is the atmospheric light, which is assumed to be a global constant following \cite{sakaridis2018semantic}. Thus, Eq. \ref{eq:mixture_rain} represents the complex mixture of the background scene, rain streaks, rainy haze and the light reflected by the environment and passing through the raindrops.

According to Garg and Nayar \cite{garg2006photorealistic}, the scene depth from the camera to the underlying scene objects behind the rain determines the visual intensity of the rain streak and the rainy haze layers. Hu et al. \cite{hu2019depth} have established the RainCityscapes dataset which contains both the rain streaks and the rainy haze. However, we observe that the raindrop layer $D(x)$ is independent to the scene depth, since raindrops distribute on the glass randomly. This motivates us to create a more natural rain dataset containing the mixture of rain.

\subsection{RainCityscapes++ Dataset}\label{section:nfnet}
\textbf{Motivation.} When observing rain through the glass, we found that the distant rain streaks will condense into rainy haze, while the nearby ones hit the glass into raindrops. To this end, we consider that rain is a mixture of rainy haze, rain streaks and raindrops. Constructing the new dataset containing the three forms of rain will benefit a series of vision-based applications, such as the driver assistance system. As known, it is challenging to obtain the real rain images and the ground-truth rain-free images simultaneously, we attempt to blend real raindrops into the existing RainCitysnapes \cite{hu2019depth} where the depth of image is kindly available.



\textbf{Overlay.} We collect \textit{14783} real outdoor photos of raindrops which are randomly distributed on the glass as the \textit{cover layer}, and collect \textit{8580} images from the training and validation sets of RainCityscapes \cite{hu2019depth} as our \textit{background layer}. We design an overlay model to superimpose the cover layers on the background layers to simulate the scene of raindrops randomly scattered on the glass:
\begin{equation}
\label{blend_1}
C_{1}=\begin{cases}
\frac{A \times B}{128},A \leq 128\\
255-\frac{A_t \times B_t}{128},A > 128 \\
\end{cases}
\end{equation}
where $A$ is the background layer, $B$ is the cover layer. $A_t$ and $B_t$ represent the anti-phase of $A$ and $B$. By Eq. \ref{blend_1}, the composition layer $C_{1}$ is obtained but mingled with unreal occlusion brought by the cover layer (see Fig. \ref{fig:label1}).

\textbf{Highlight.} To solve the problem of visual occlusion, we evolve the overlay mode into highlight mode to enhance the color contrast between the two layers, which is defined as:
\begin{equation} \label{eq:highlight}
C_{2}=\begin{cases}
\frac{A \times B}{128},B \leq 128\\
255-\frac{A_t \times B_t}{128},B > 128 \\
\end{cases}
\end{equation}

\textbf{Transparency.} To approximate the real scene, we further emphasize the background layer by increasing the transparency of the cover layer:
\begin{equation}\label{eq:tr}
C_{3}=t \times A +(1-t)\times B
\end{equation}
where $t$ denotes the transparency. Based on Eq. \ref{eq:highlight} and Eq. \ref{eq:tr}, the final blending model can be expressed as:

\begin{equation}\label{fig:c3}
C_{4}=\begin{cases}
\frac{t \times A \times (1-t)\times B}{128},B \leq 128\\
255-\frac{t \times A_t \times (1-t)\times B_t}{128},B > 128 \\
\end{cases}
\end{equation}
After highlighting and adjusting the transparency, we obtain rain images $C_{4}$ that form our RainCityscapes++.

\begin{figure*}[htb]
\centering
\includegraphics[width=0.8\linewidth]{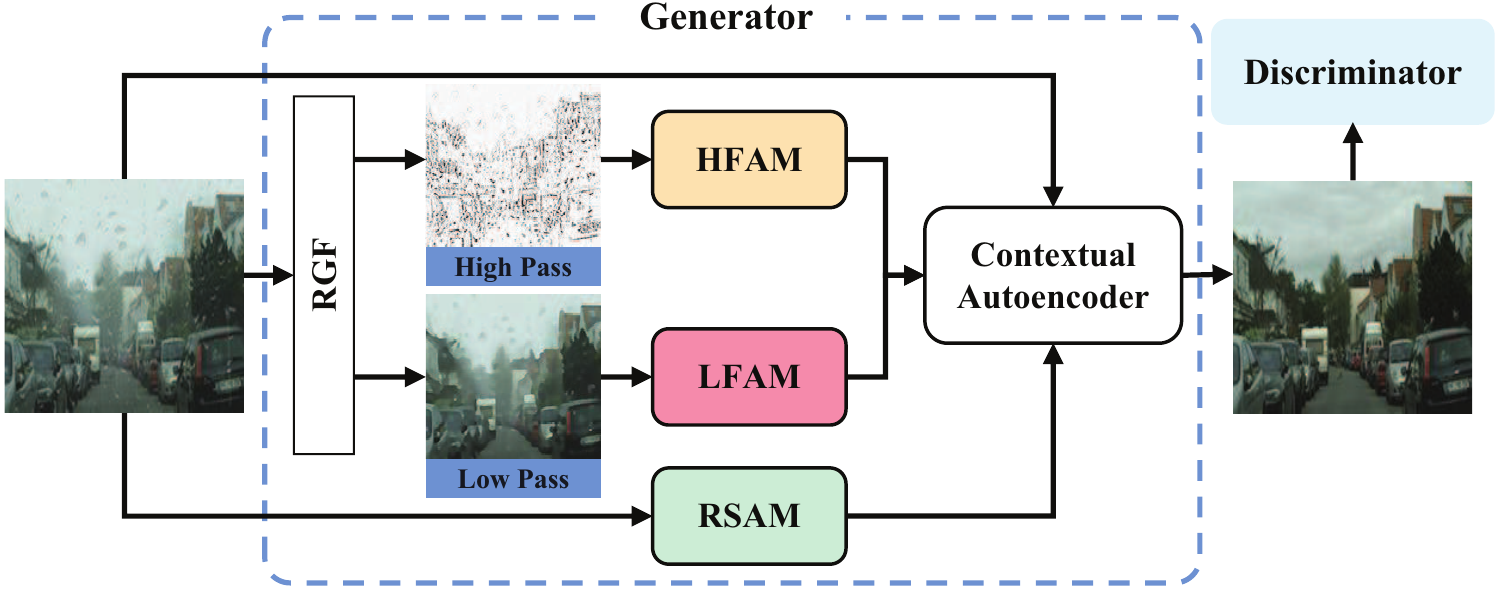}
\caption{Overall architecture of the multi-branch attention generative adversarial network (MBA-RainGAN).}
\label{fig:label2}
\end{figure*}
\section{MBA-RainGAN}
We introduce the multi-branch attention generative adversarial network (MBA-RainGAN) for image deraining, especially for the removal of entangled mixture of different rainwater forms. The overall architecture is illustrated in Fig. \ref{fig:label2}. Within the generator, a novel multi-branch attention module (MAM) constitutes the contextual autoencoder, which takes full advantage of our multi-branch attention scheme. In this section, we elaborate our generative network and discriminative network respectively to demonstrate how we dissolve the intractable MOR problem.
%
\subsection{Generative Network}
As shown in Fig. \ref{fig:label2}, to remove the mixture of rain (MOR), we adopt a three-stage decomposition strategy in the generative network: first, a guided image filter is applied to extract rain streaks while preserving the main image structure; then, the separated rain streak layer and rainy haze layer are fed into two individual attention learning modules, i.e., high frequency attentive module (HFAM) and low frequency attentive module (LFAM), while the raindrop attention map is directly learned from the input using raindrop spatial attentive module (RSAM); lastly, a contextual autoencoder is progressively guided by these three attention maps to detach the MOR.

\textbf{Guided Image Decomposition}
In the heavy rain images, the visual appearances of rainy haze and rain streaks are entangled with each other. Obtaining the rainy haze attention map directly from the input image is challenging due to the strong presence of rain streaks and vice versa. Hence, we propose to first separate these two effects to ease the burden of multiple attention learning.
%

Rain streaks are of high-frequency. Decomposing the input image into high- and low-frequency components in advance benefits the rain removal task. Since our goal is to extract rain streaks while preserving rainy haze and main image structures, we adopt the scale-aware Rolling Guidance Filter (RGF) \cite{zhang2014rolling} which eliminates small-scale structural features such as rain streaks and recovers the image structures iteratively. Thereafter, we obtain the high pass component for predicting the rain streak attention and the low pass component for the rainy haze attention, as shown in Fig. \ref{fig:label2}.

\begin{figure}[ht]
\centering
\includegraphics[width=0.8\linewidth]{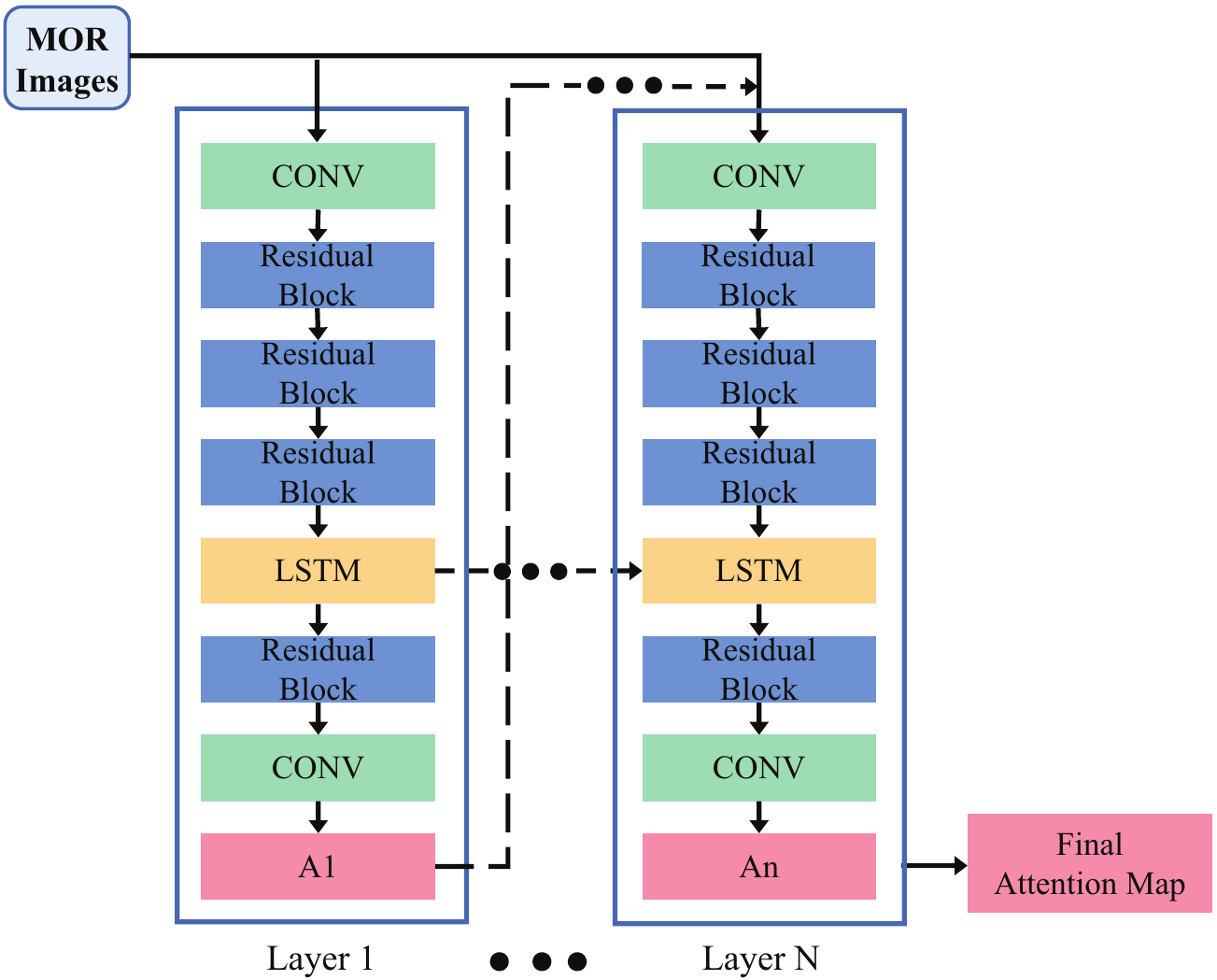}
\caption{The architecture of our HFAM and LFAM, $A_N$ represents the final attention map.}
\label{fig:hfam}
\end{figure}
\begin{figure}[ht]
\centering
\includegraphics[width=1\linewidth]{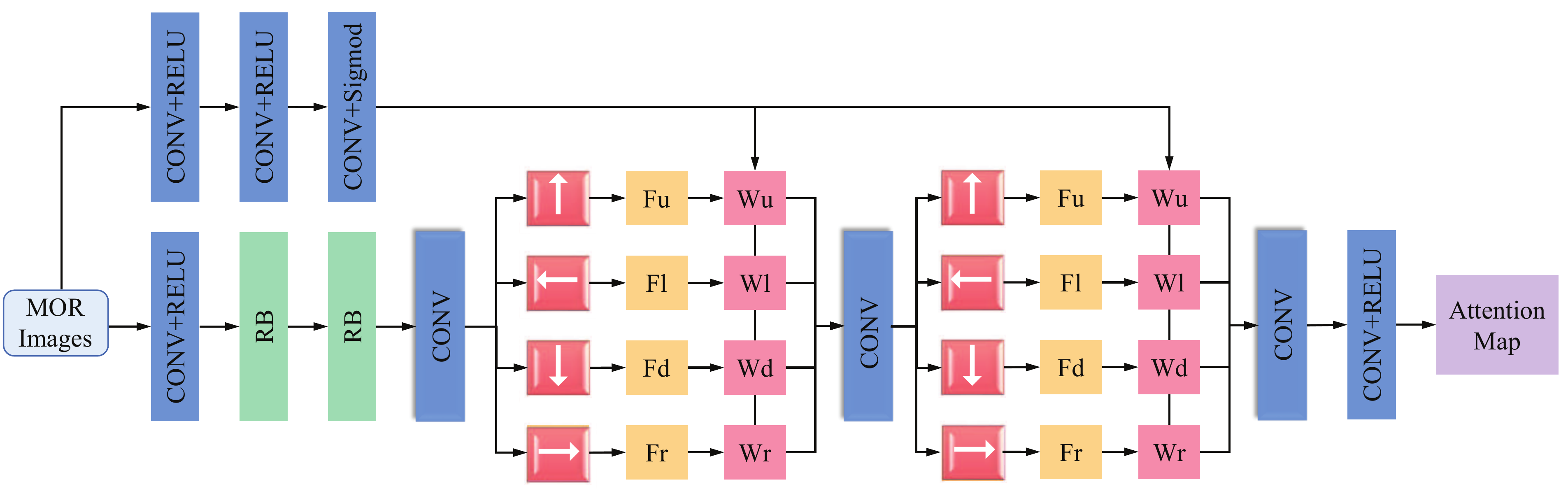}
\caption{The architecture of our RSAM.}
\label{fig:rsam}
\end{figure}

\textbf{Multi-branch Attention}
The attention mechanism has been proved effective for deep networks to focus on interested areas \cite{gregor2015draw,mnih2014recurrent,zhao2017diversified}. In rain removal tasks, this idea has been applied to emphasize raindrop regions \cite{qian2018attentive}, dense haze regions \cite{qin2019ffa} and rain streak regions \cite{wang2019spatial}, especially, Hu et al. design the depth-guided attention mechanism, and train an end-to-end network to learn the depth attentional features to remove rain streaks and rainy haze in rain images. Based on these researches, we design a multi-branch attention scheme to generate attention maps for individual degradations in MOR, from which the multi-attentive module (MAM) we proposed plays the key role in removing mixture artifacts.

To predict the rain streak attention map $A_{HN}$ and the rainy haze attention map $A_{FN}$, the high frequency attentive module (HFAM) and the low frequency attentive module (LFAM) accept the high pass component and the low pass component, respectively. The loss functions are written as:
\begin{equation}\label{HFAM}
\begin{aligned}
\textbf{L}_ \textbf{HFAM}=\textbf{L}_ \textbf{MSE}(A_{HN},M_{s})
\end{aligned}
\end{equation}
\begin{equation}\label{LFAM}
\textbf{L}_ \textbf{LFAM}=\textbf{L}_ \textbf{MSE}(A_{FN},A)
\end{equation}
where $M_{s}$ is the binary map that indicates the locations of rain streaks, obtained by setting a threshold to the rain streak layer. $A\in \{0,1\}$ refers to the rainy hazy layer in Eq. \ref{eq:mixture_rain}, computed according to the depth image. As shown in Fig. \ref{fig:hfam}, HFAM and LFAM share the network structure, where four residual blocks \cite{he2016deep} extract features from the input image, a convolutional LSTM unit \cite{xingjian2015convolutional} bridges the recurrent stages and a convolutional layer generates the 2D attention map.

As for raindrops, due to the randomness of its distribution and the complex reflection effect in the contaminated areas, we adopt directional recurrent neural network with ReLU and identity matrix initialization (IRNN) to build the raindrop spatial attentive module (RSAM) and learn the raindrop attention map $A_{RN}$ by minimizing:
\begin{equation}\label{RSAM}
\textbf{L}_ \textbf{RSAM}=\textbf{L}_ \textbf{MSE}(A_{RN},M_{d})
\end{equation}
where $M_{d}$ refers to the binary map in Eq. \ref{eq:mixture_rain} that marks the existence of raindrops. The detailed structure of RSAM is illustrated in Fig. \ref{fig:rsam}. Similar to \cite{bell2016inside,hu2018direction, wang2019spatial}, we apply the two-round four-directional IRNN to accumulate the global contextual information among stages, which substantially enlarges the receptive field for extracting raindrop features.

To take full advantage of these three attention maps, we propose the novel multi-attentive module (MAM), where the prediction of network embeddings is progressively guided by different attention maps. As shown in Fig. \ref{fig:mam}, the attention maps of rainy haze, rain streaks and raindrops are sequentially added into MAM, which benefit in two aspects: first, it avoids the confusion caused by simultaneously introducing the information from multiple disparate attention maps; second, it considers the superposition of MOR that rainy haze, rain streak and raindrops happen in the bottom, medium and top layer respectively. As suggested in Sec. \ref{ablation}, the proposed multi-branch attention scheme is empirically validated to be effective in decomposing mixture degradations.
\begin{figure}[ht]
\centering
\includegraphics[width=1.0\linewidth]{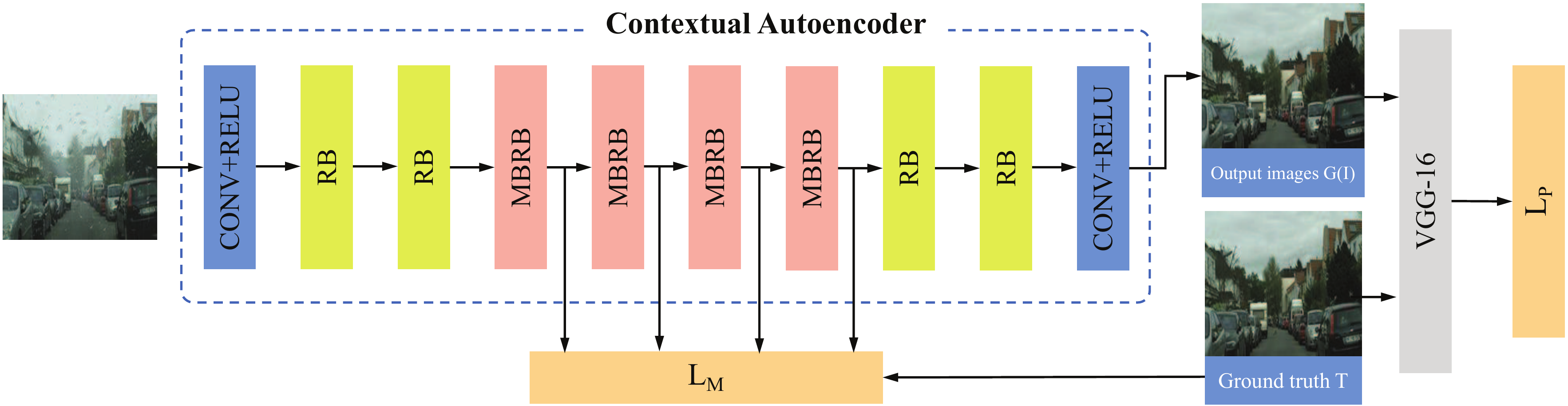}
\caption{The architecture of our contextual autoencoder.}
\label{fig:auto}
\end{figure}
\begin{figure}[ht]
\centering
\includegraphics[width=1\linewidth]{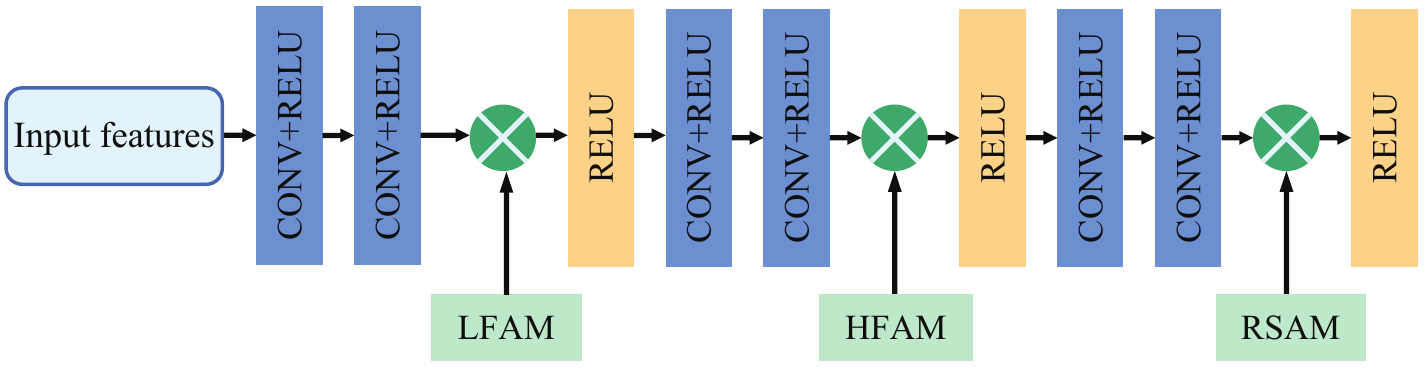}
\caption{The architecture of our MAM.}
\label{fig:mam}
\end{figure}

\textbf{Contextual Autoencoder}
Our contextual autoencoder generates images that are free of MOR. As shown in Fig. \ref{fig:auto}, the first convolutional layer extracts the image features, followed by two residual blocks to obtain deeper embeddings. The key part of the contextual autoencoder consists of four MAMs that reflects the complex entanglement of MOR. The multi-scale features extracted from each MAM are exploited by the multi-scale losses $L_{M}$ as:
\begin{equation}\label{eq:tensorvoting}
\textbf{L}_ \textbf{M}=\sum_{i=1}^{M}\lambda_i \textbf{L}_ \textbf{MSE}(S_i,T_i)
\end{equation}
where $S_i$ indicates the ith output from the MAM, and $T_i$ indicates the ith ground-truth multi-scale features. $\lambda_i$ are set to be 0.4, 0.6, 0.8, 1.0.

Finally, the prediction $O$ is made after two more residual blocks and a convolutional layer, which together with the ground truth $T$ are fed into VGG-16 to form the perceptual loss \cite{johnson2016perceptual} that measures the global discrepancy between the features. The perceptual loss is expressed as:
\begin{equation}\label{eq:tensorvoting}
\textbf{L}_ \textbf{P}=\textbf{L}_ \textbf{MSE}(VGG(O),VGG(T))
\end{equation}

Therefore, the overall loss for the generative network is as:
\begin{equation}\label{overallloss}
   \textbf{L}_ \textbf{GN}=\textbf{L}_ \textbf{HFAM}+\textbf{L}_ \textbf{LFAM}+\textbf{L}_ \textbf{RSAM}+\textbf{L}_ \textbf{P}+\textbf{L}_ \textbf{M}
\end{equation}

\begin{figure}[ht]
\centering
\includegraphics[width=0.4\textwidth]{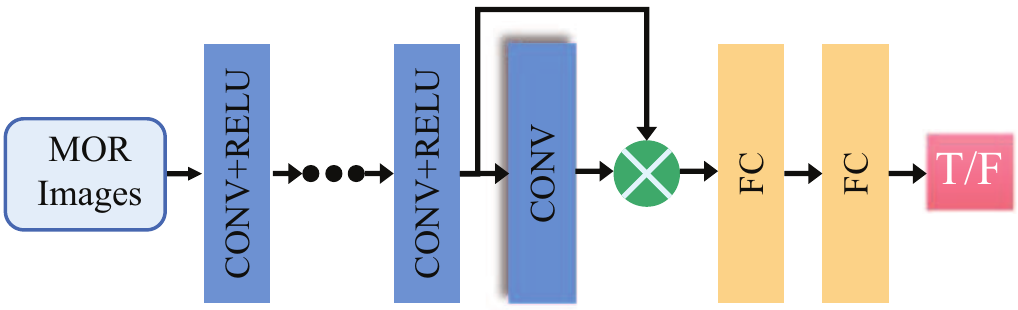}
\caption{The architecture of our discriminator.}
\label{fig:label7}
\end{figure}

\subsection{Discriminative Network}\label{section:nfnet}
The discriminative network accepts the output of generative network and checks if it looks like the ground-truth photos. Similar to \cite{qian2018attentive}, we adopt an attentive discriminator to leverage the abundant information provided by our multi-branch attention scheme, which allows the discriminator to focus on local areas that are likely to be fake. Specifically, the features extracted from interior layers of the discriminator are constrained by minimizing the distance between the attention maps and the output after feeding them into another CNN. This loss function is written as:
\begin{equation}\label{eq:lmap}
\begin{aligned}
\textbf{L}_ \textbf{map}=&\textbf{L}_ \textbf{MSE}(D_{map}(O),A_{HN})+
\textbf{L}_\textbf{MSE}(D_{map}(O),A_{LN})\\
&+\textbf{L}_ \textbf{MSE}(D_{map}(O),A_{RN})+\textbf{L}_ \textbf{MSE}(D_{map}(R),0)
\end{aligned}
\end{equation}
where $D_{map}$ represents the process of producing a 2D map by the discriminative network and $0$ is the map only containing the values of 0.

As illustrated in Fig. \ref{fig:label7}, our discriminative network consists of six convolutional layers and two fully connected layers, the features extracted from the fifth convolutional layers are multiplied back in element-wise. The whole loss function of the discriminative network can be expressed as:
\begin{equation}\label{eq:lds}
\begin{aligned}
 \textbf{L}_ \textbf{DS}=-\log(D(R))-\log(1-D(O))+\gamma\textbf{L}_ \textbf{map}
\end{aligned}
\end{equation}
where $R$ represents the clean image and the balancing weight $\gamma$ is set to be 0.10.
\begin{figure*}
\centering
\subfigure[Input image]{
\begin{minipage}[b]{0.1\linewidth}
\includegraphics[width=1\linewidth]{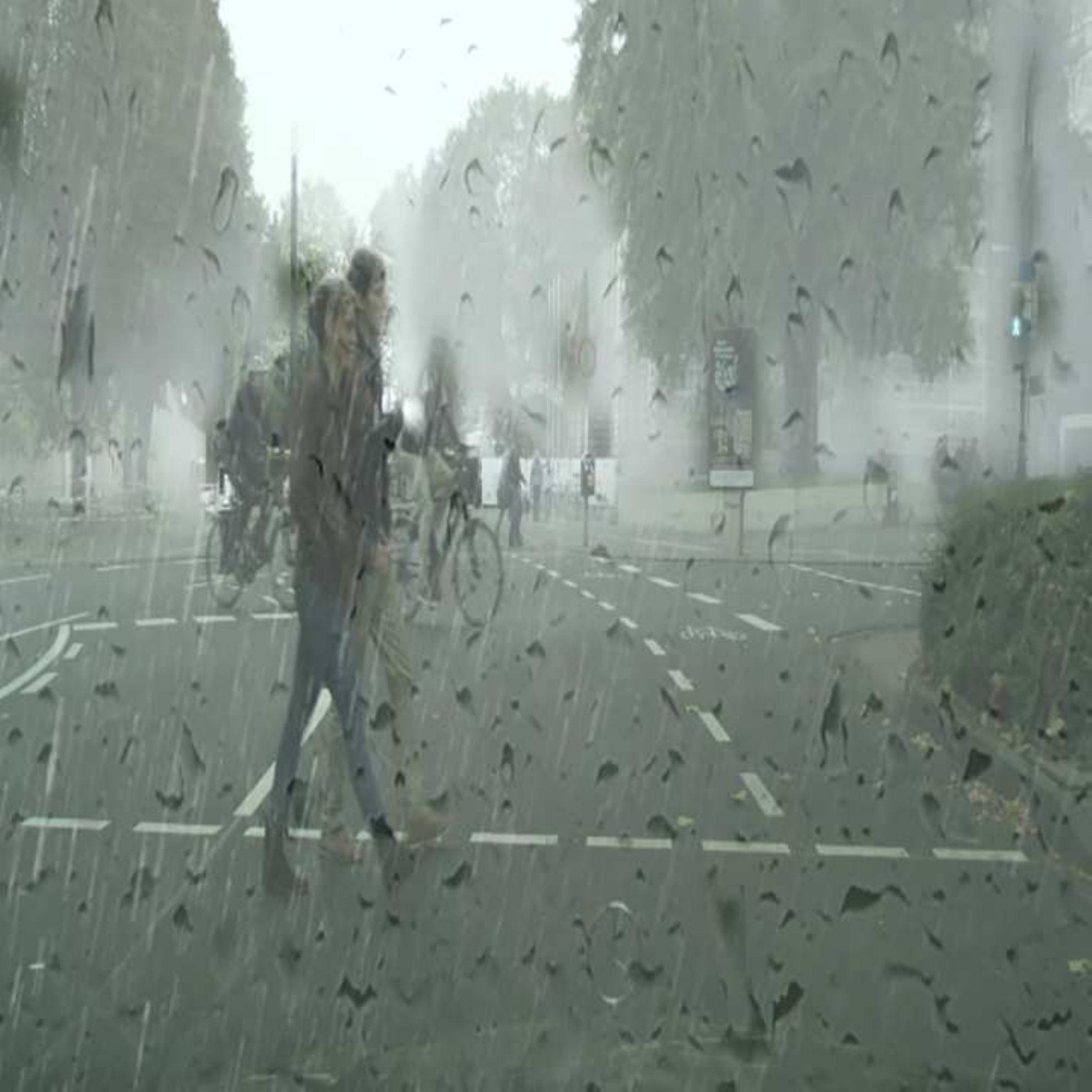}\vspace{2pt}
\includegraphics[width=1\linewidth]{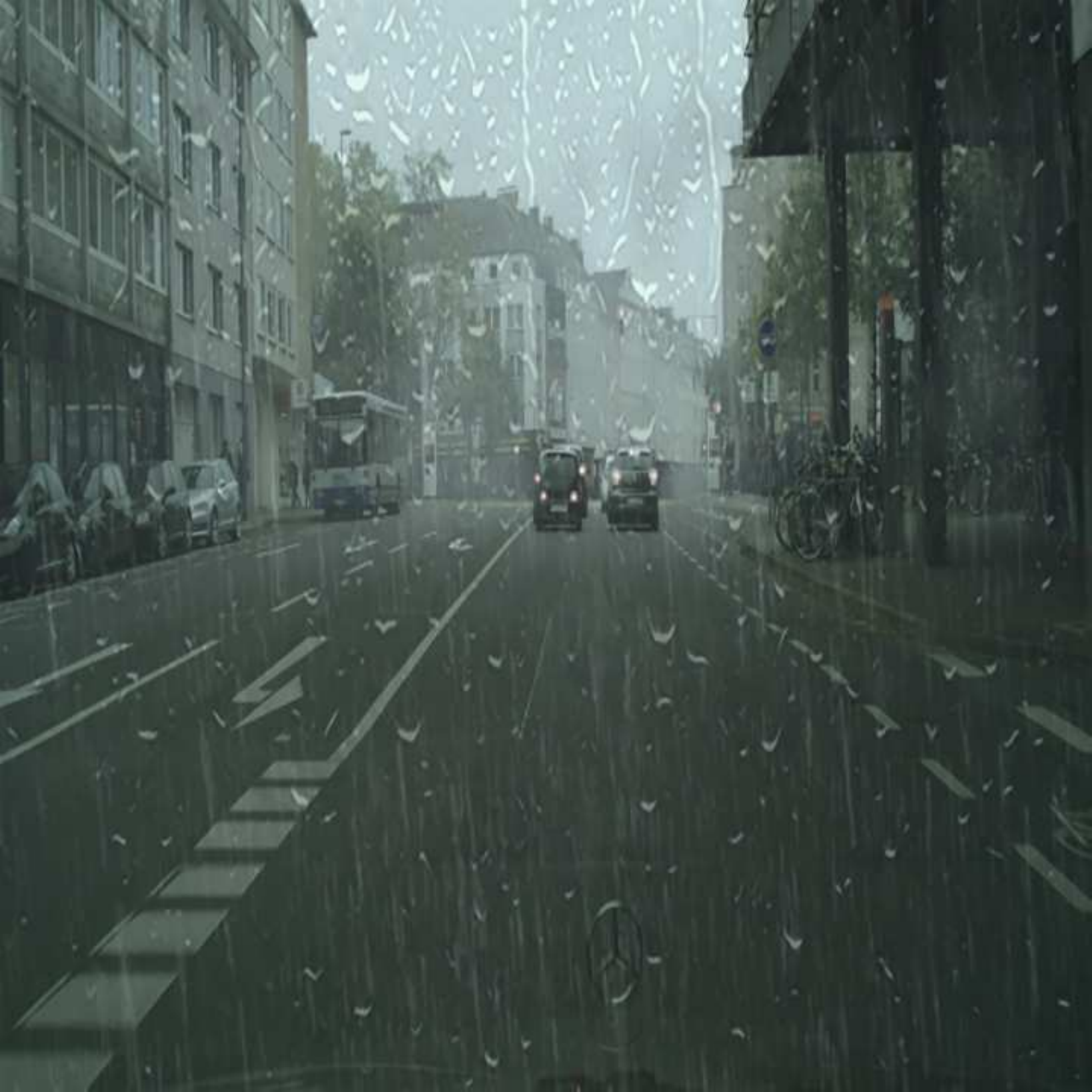}\vspace{2pt}
\includegraphics[width=1\linewidth]{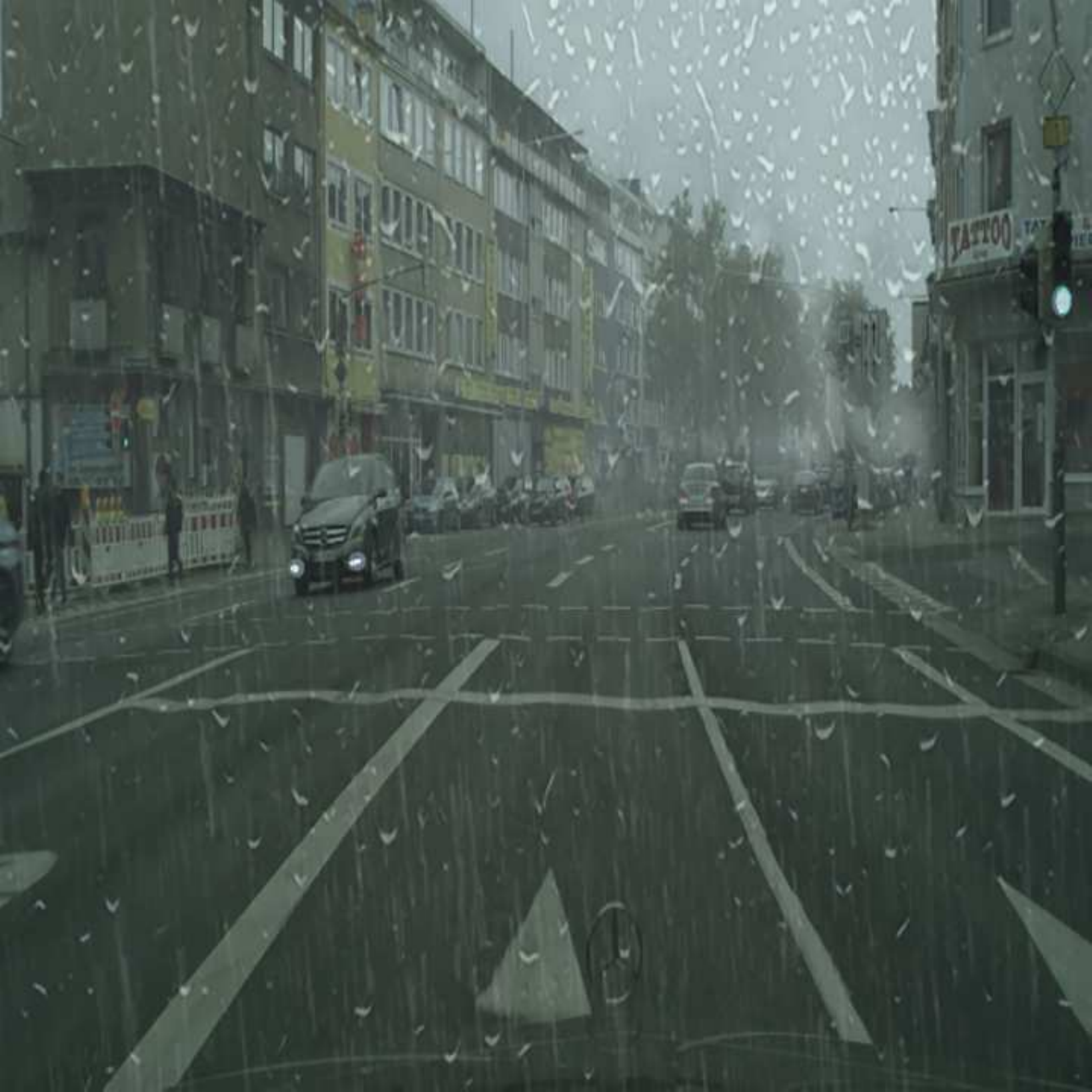}\vspace{2pt}
\includegraphics[width=1\linewidth]{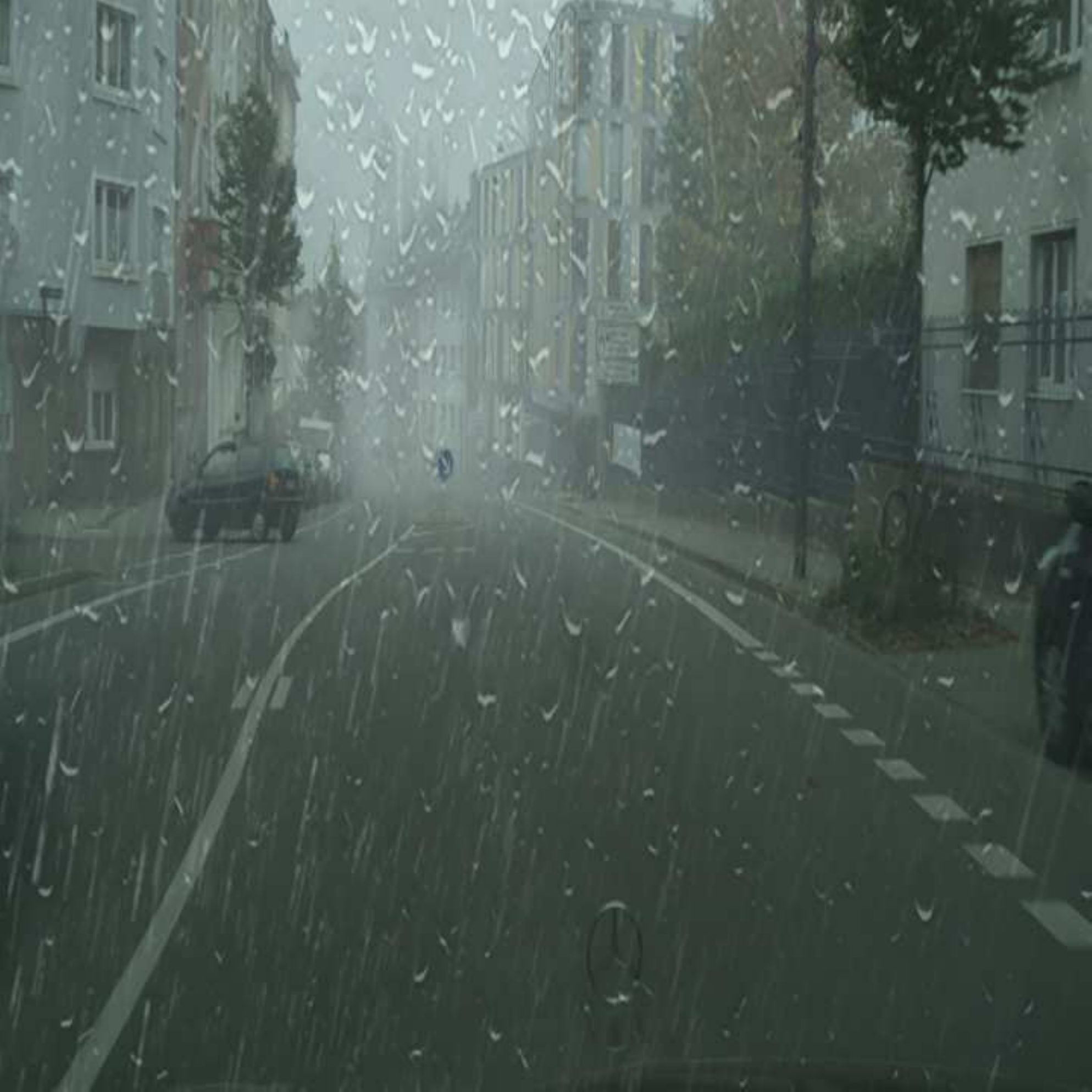}\vspace{2pt}
\includegraphics[width=1\linewidth]{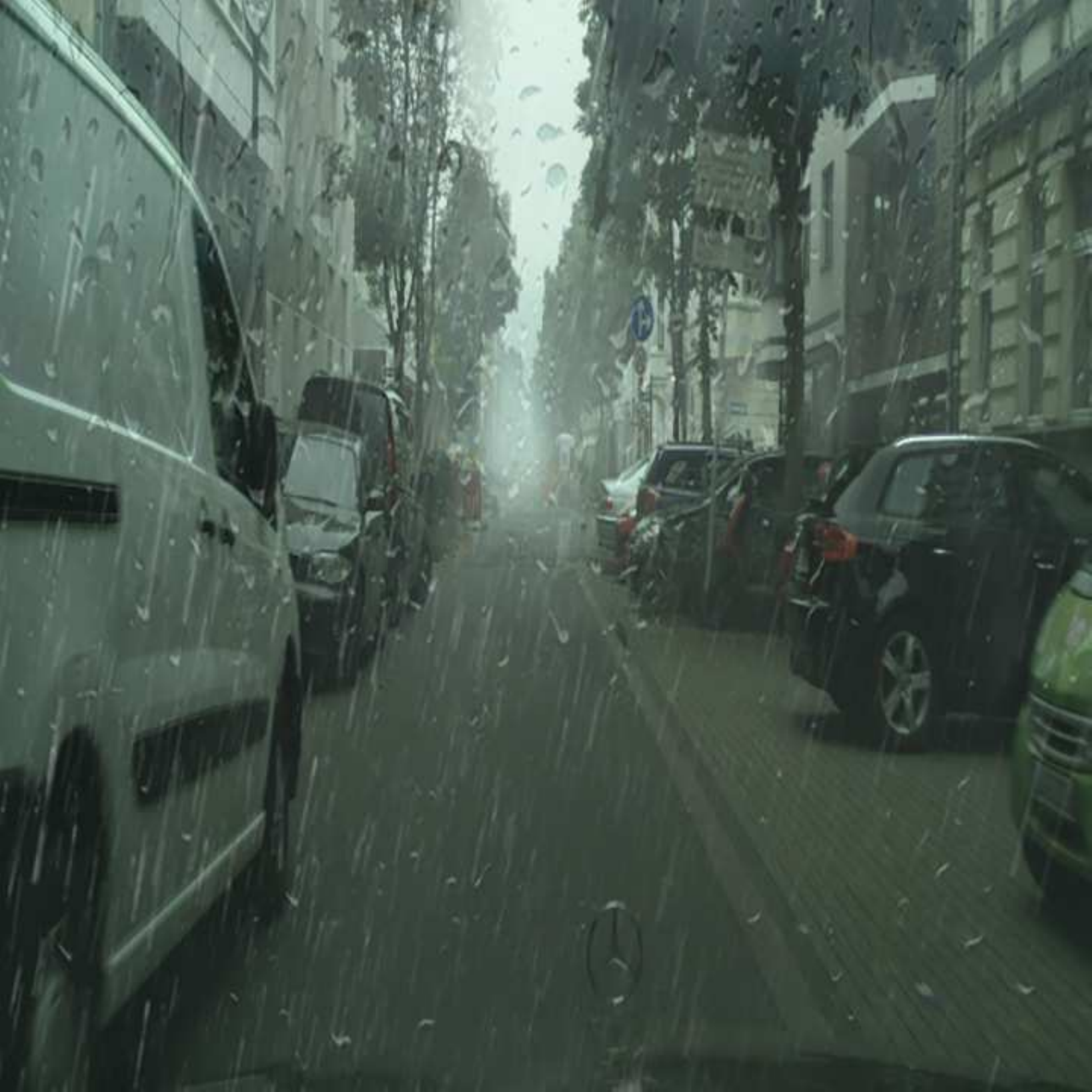}\vspace{2pt}
\end{minipage}}
\subfigure[GT]{
\begin{minipage}[b]{0.1\linewidth}
\includegraphics[width=1\linewidth]{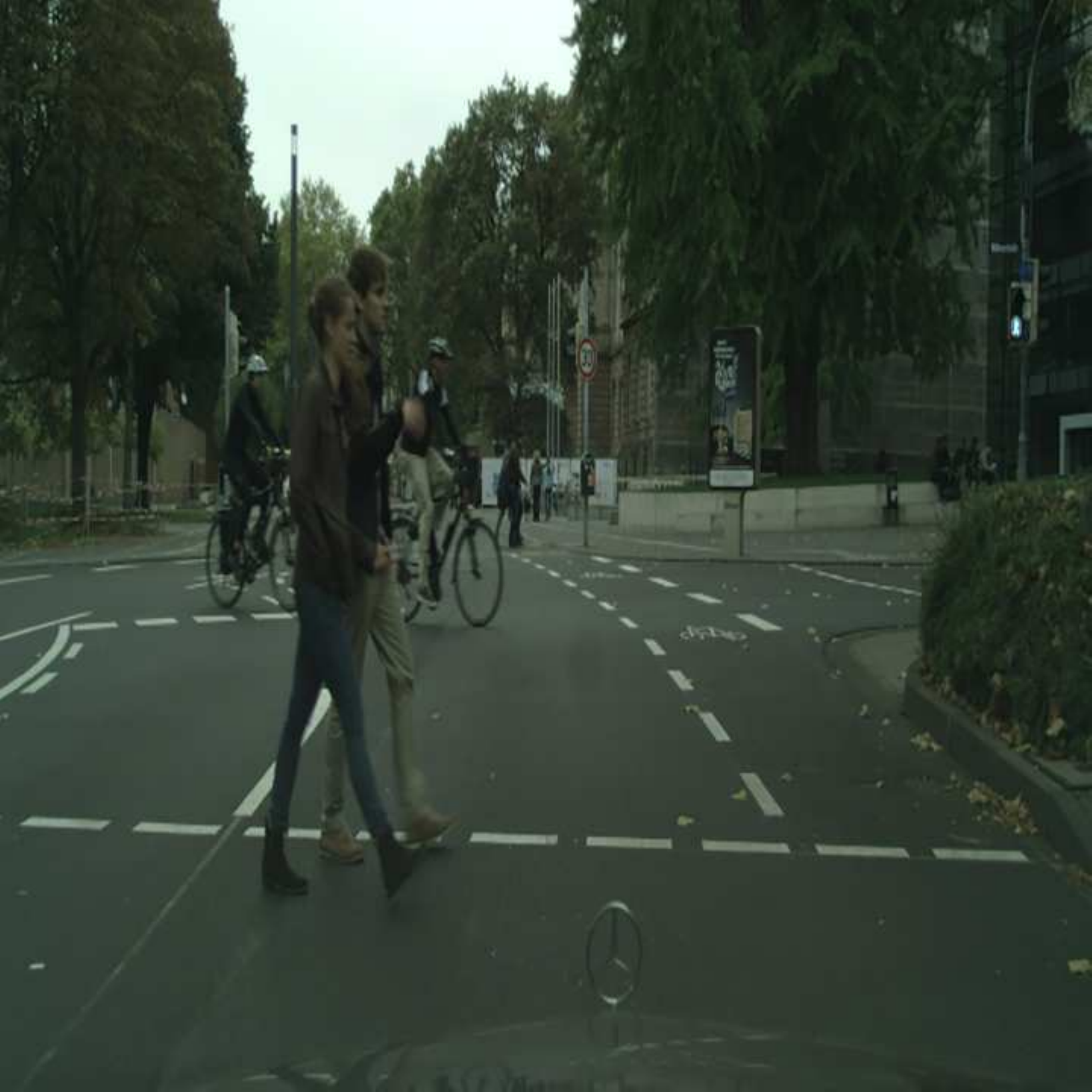}\vspace{2pt}
\includegraphics[width=1\linewidth]{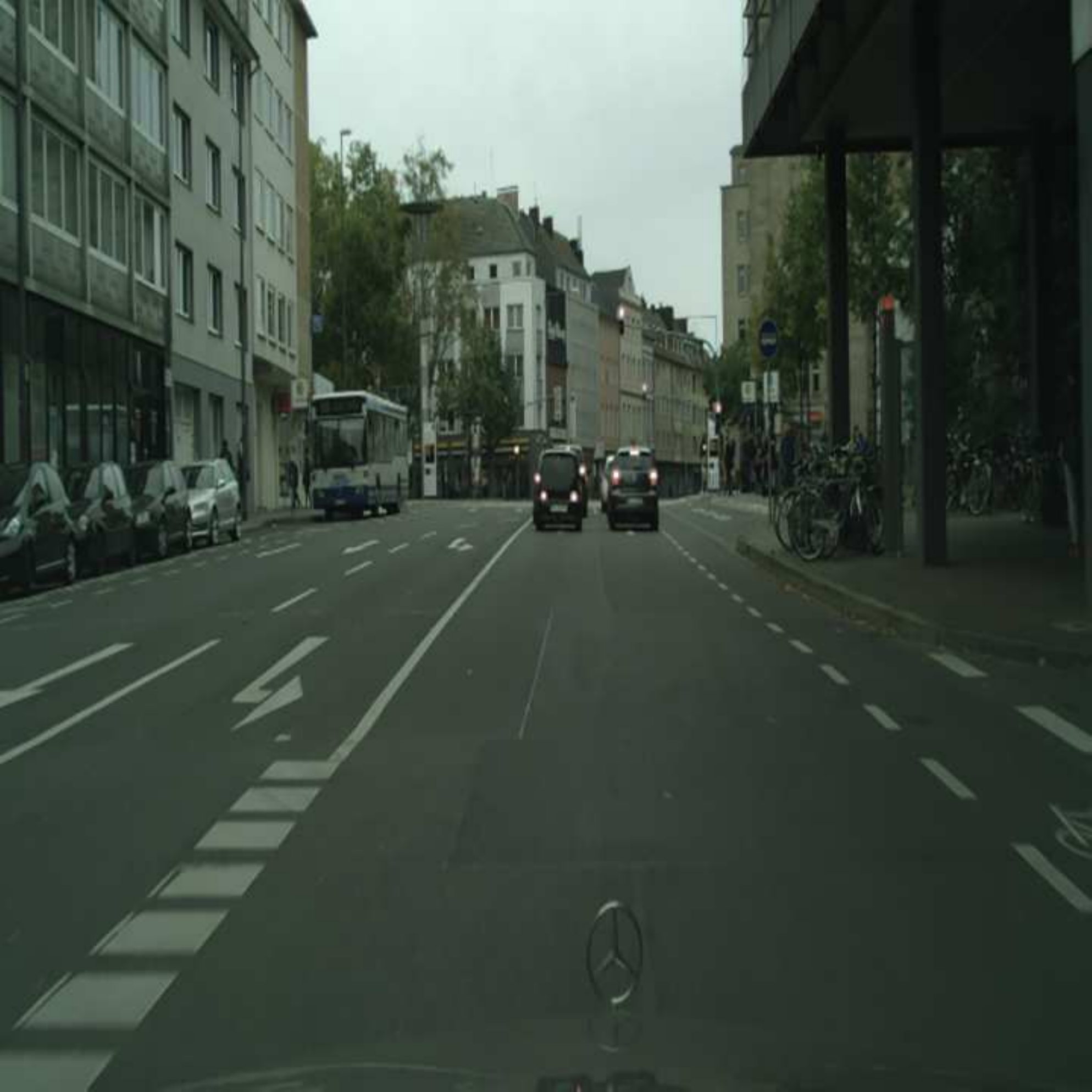}\vspace{2pt}
\includegraphics[width=1\linewidth]{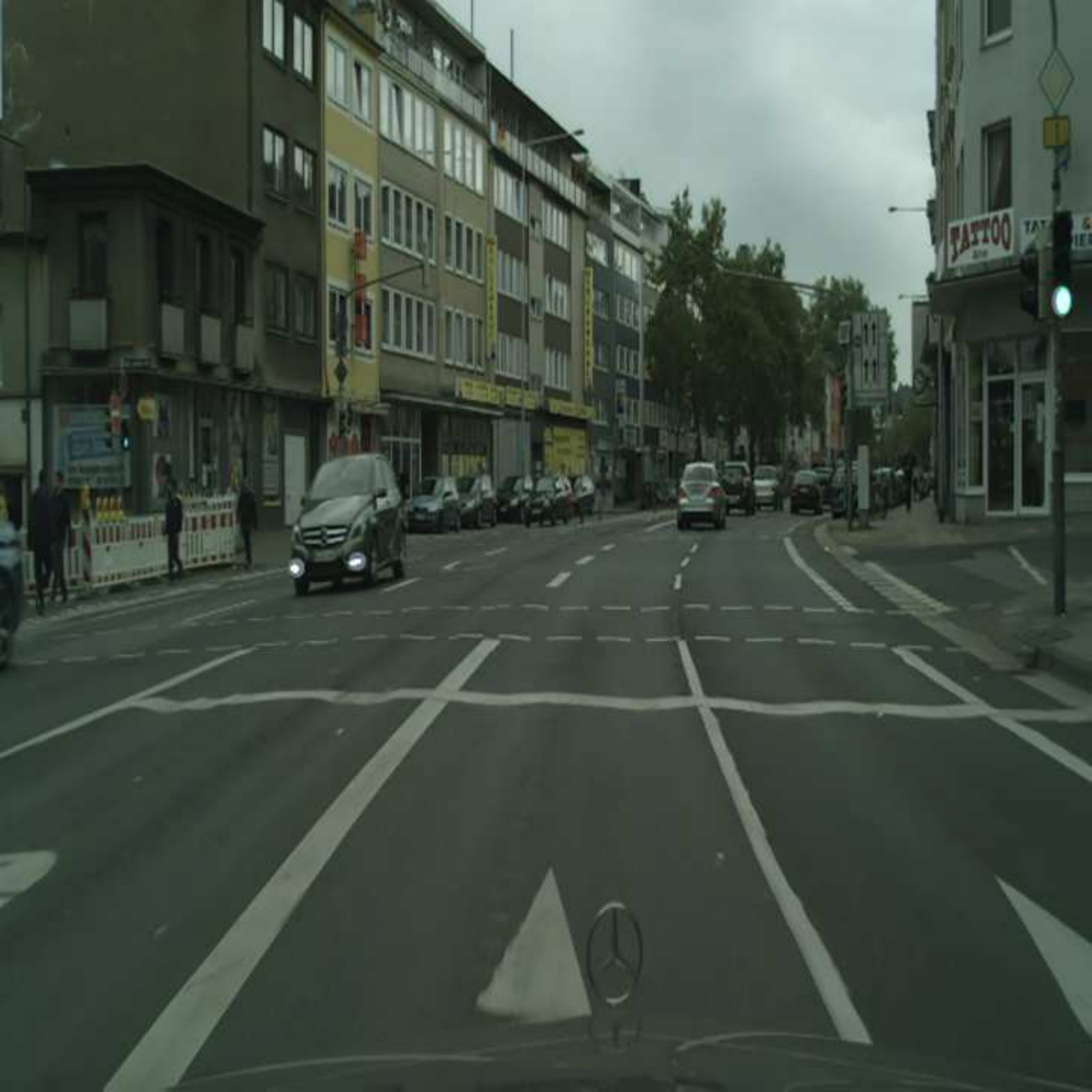}\vspace{2pt}
\includegraphics[width=1\linewidth]{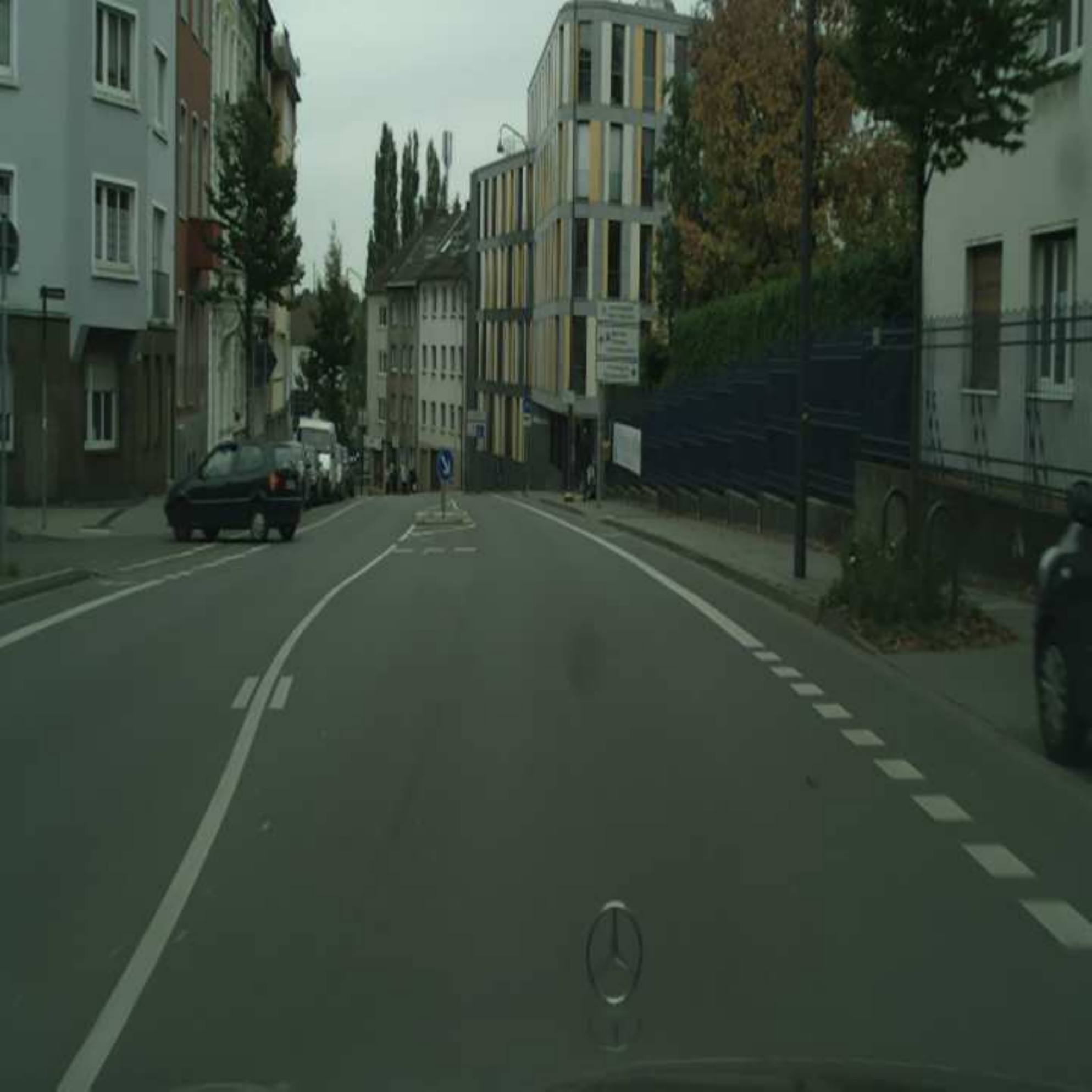}\vspace{2pt}
\includegraphics[width=1\linewidth]{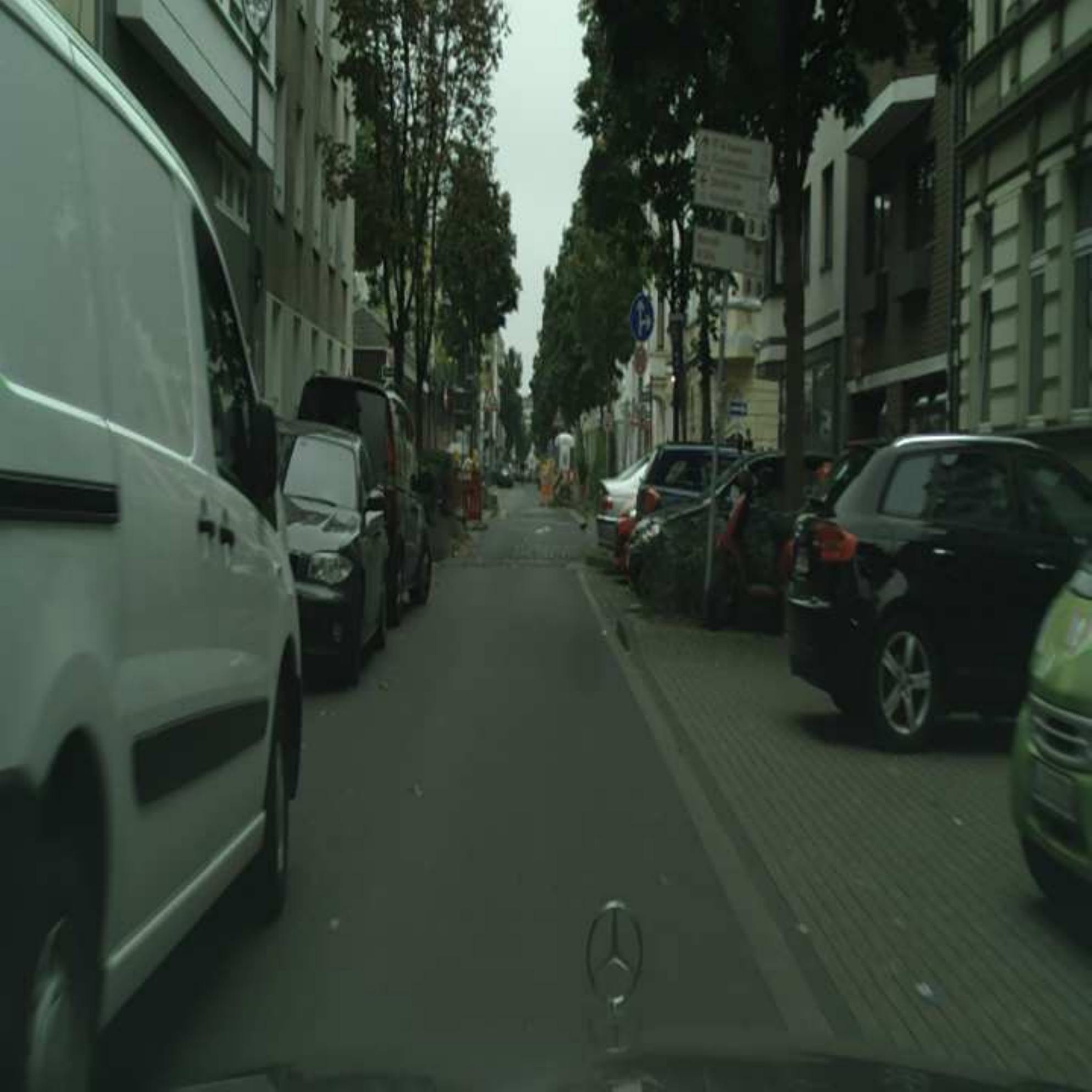}\vspace{2pt}
\end{minipage}}
\subfigure[Li et al.]{
\begin{minipage}[b]{0.1\linewidth}
\includegraphics[width=1\linewidth]{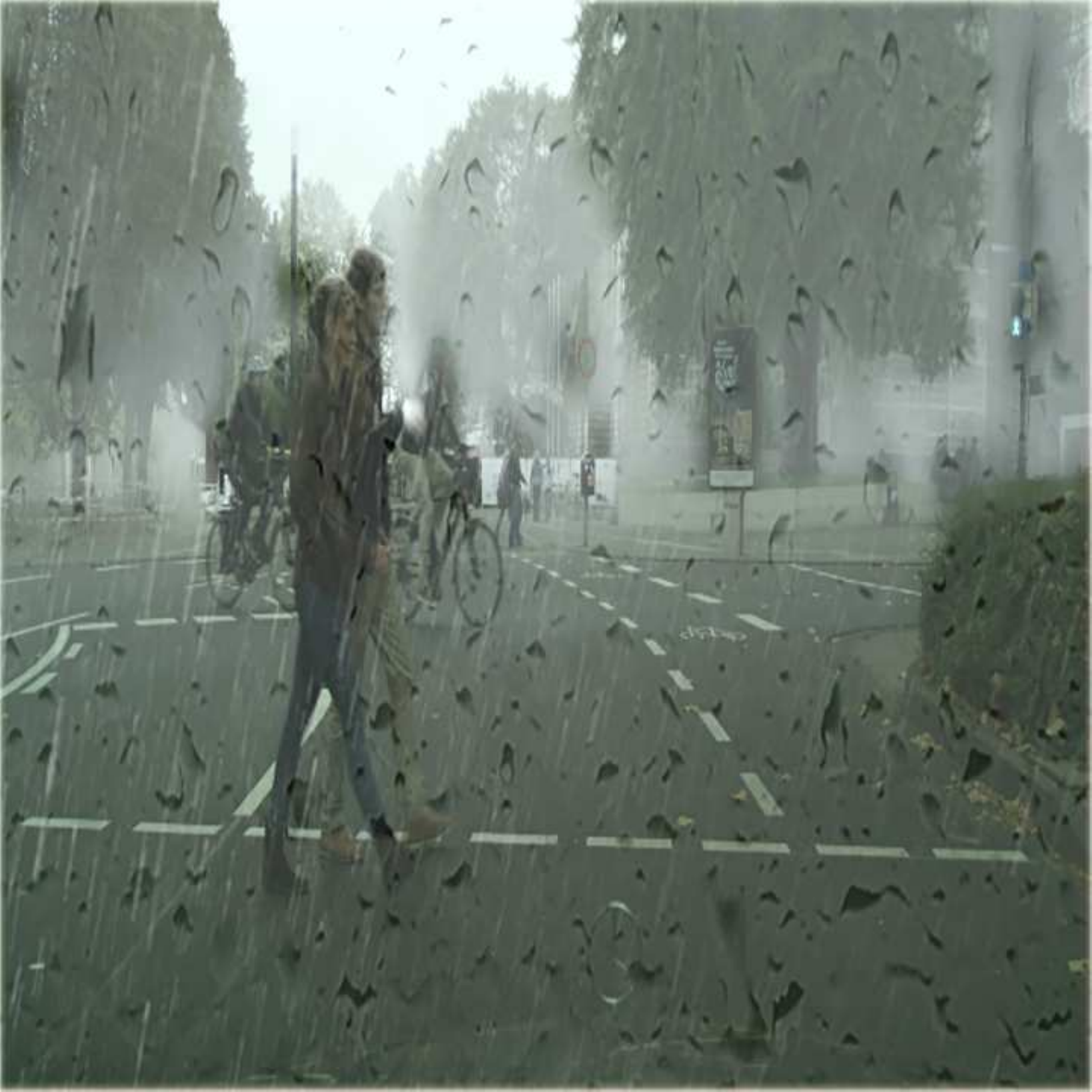}\vspace{2pt}
\includegraphics[width=1\linewidth]{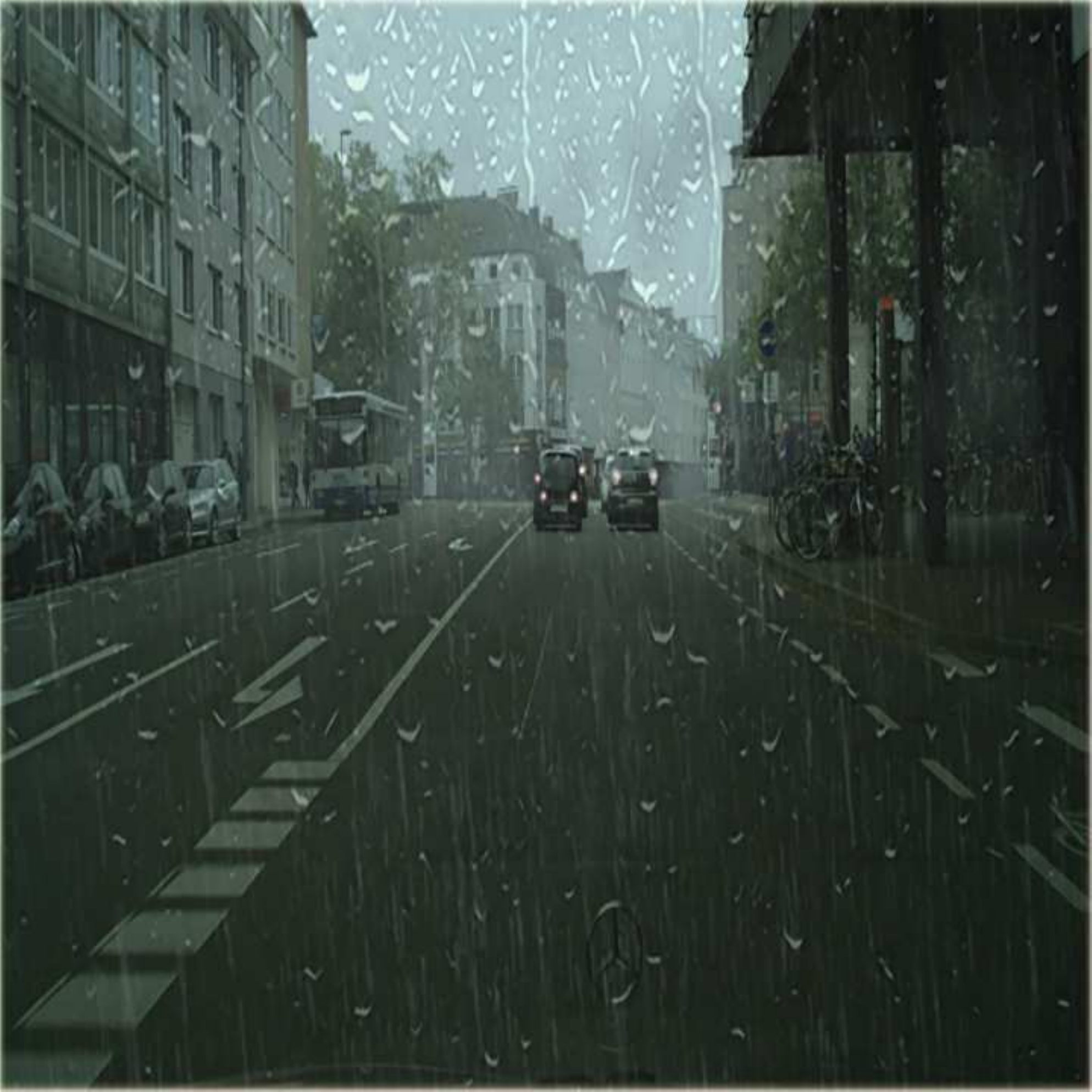}\vspace{2pt}
\includegraphics[width=1\linewidth]{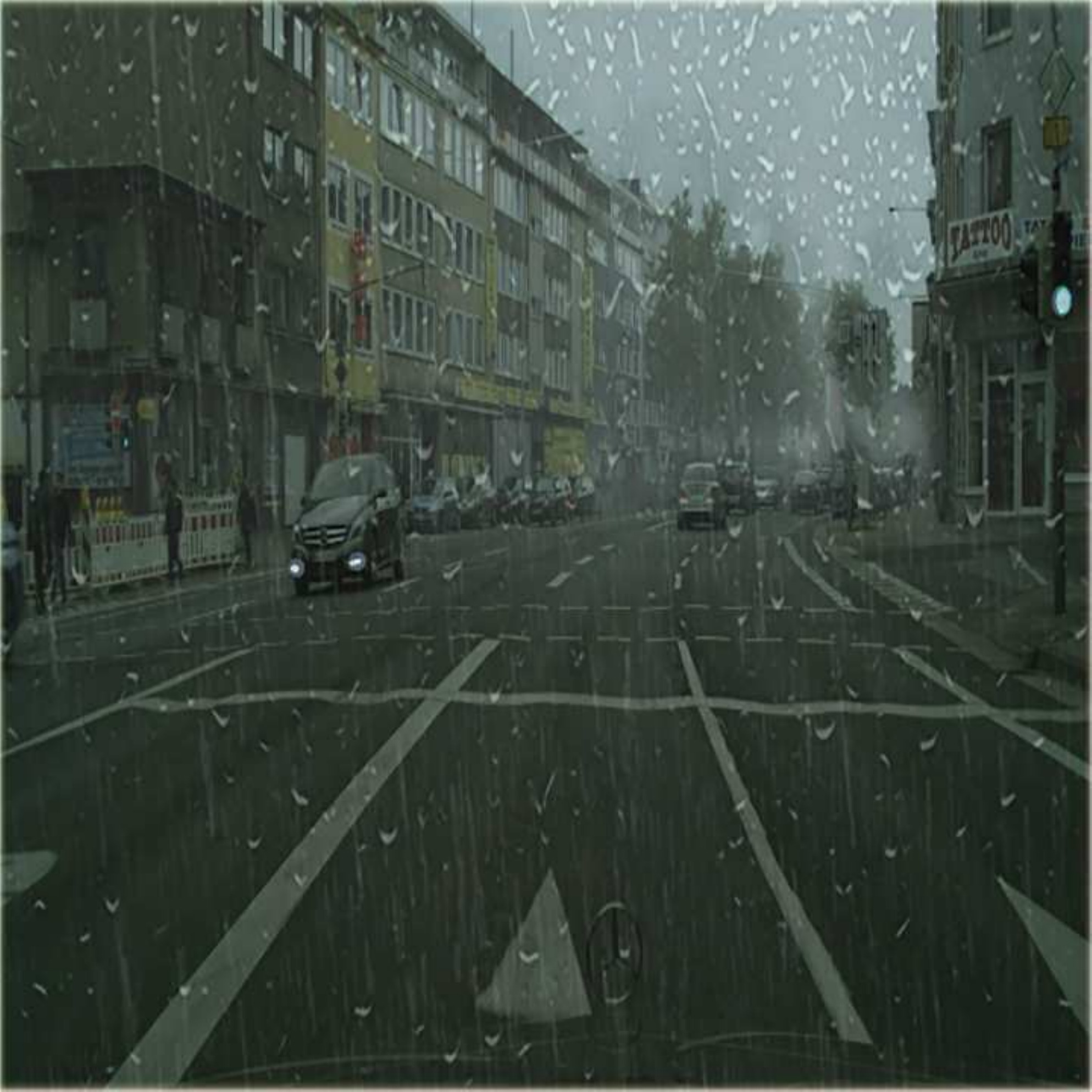}\vspace{2pt}
\includegraphics[width=1\linewidth]{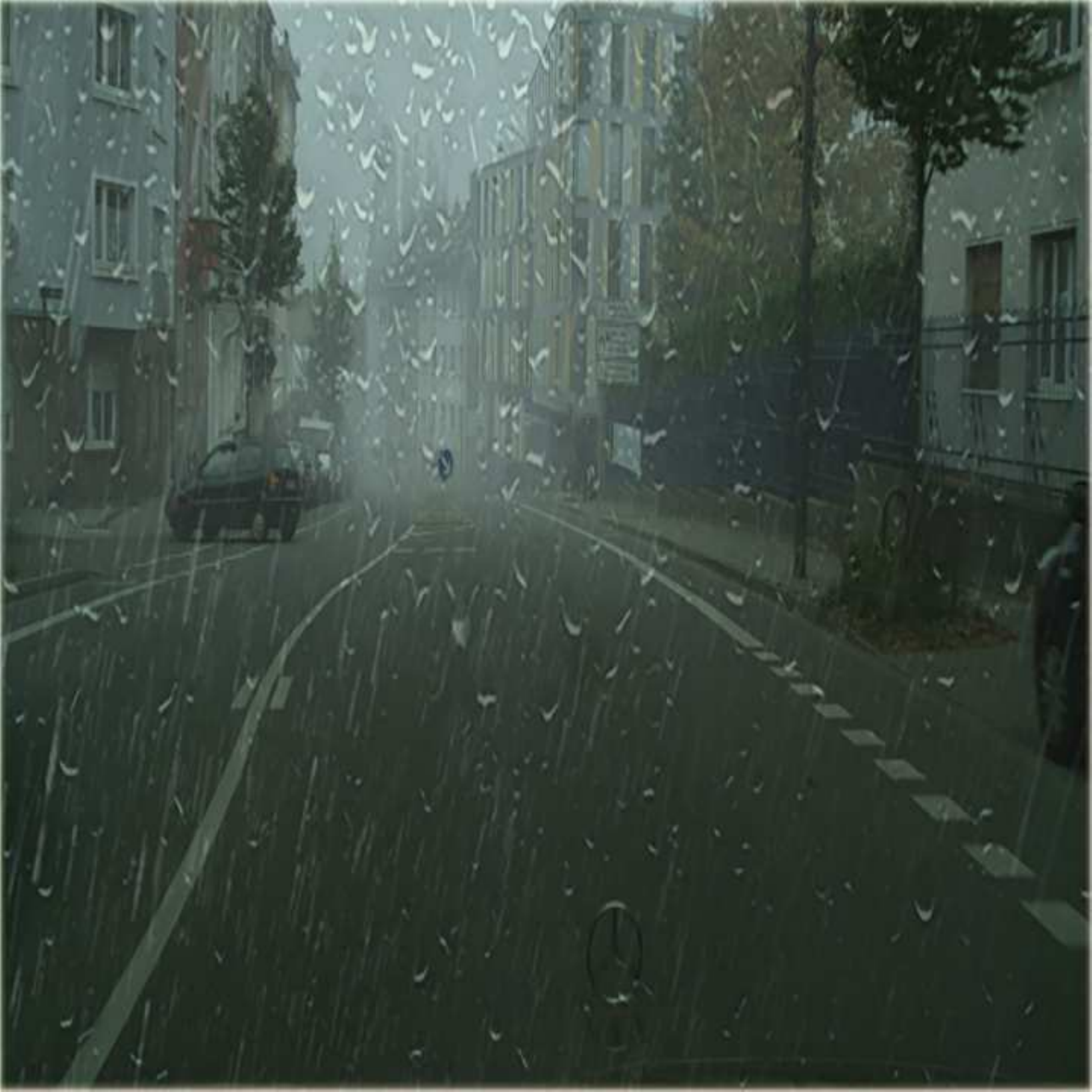}\vspace{2pt}
\includegraphics[width=1\linewidth]{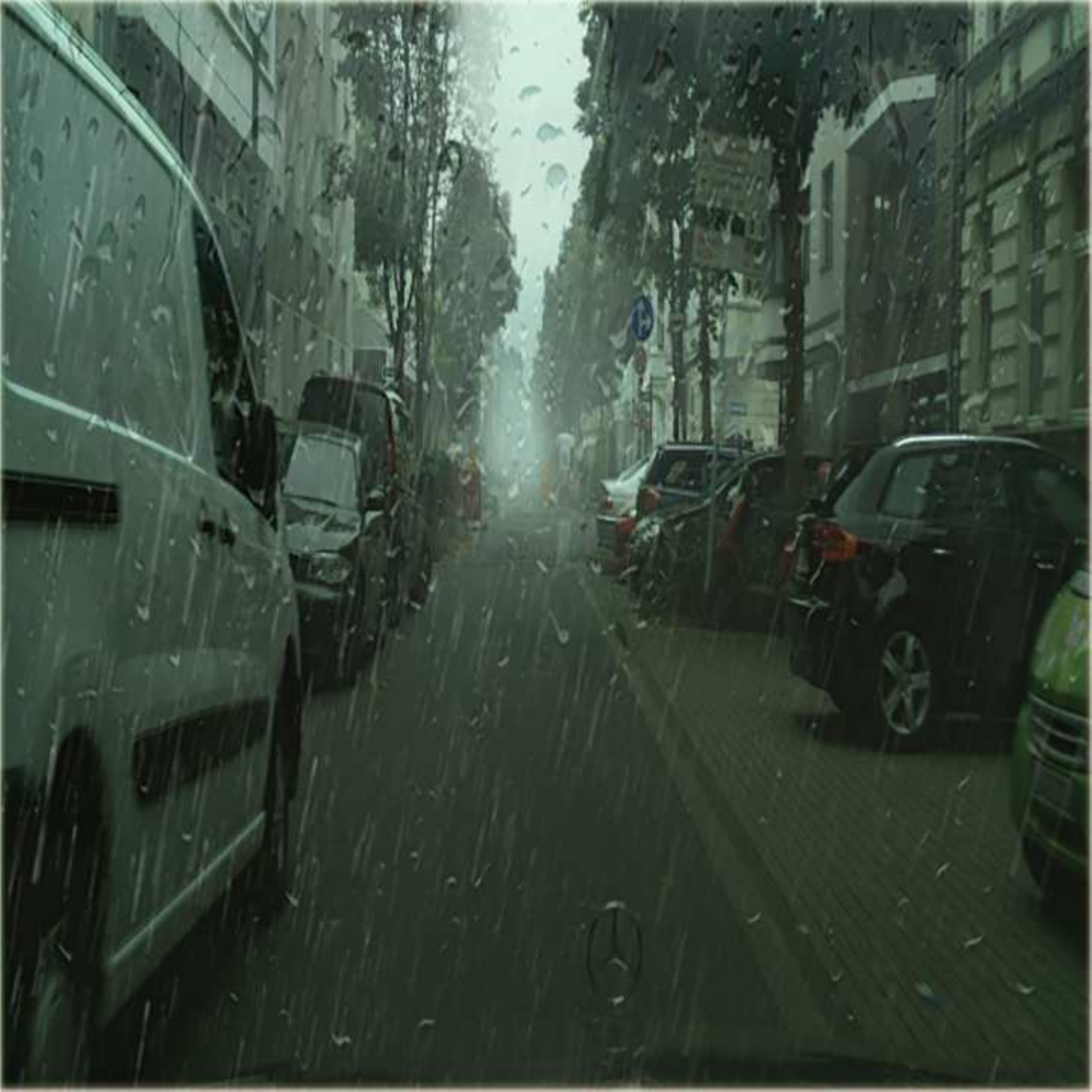}\vspace{2pt}
\end{minipage}}
\subfigure[Ren et al.]{
\begin{minipage}[b]{0.1\linewidth}
\includegraphics[width=1\linewidth]{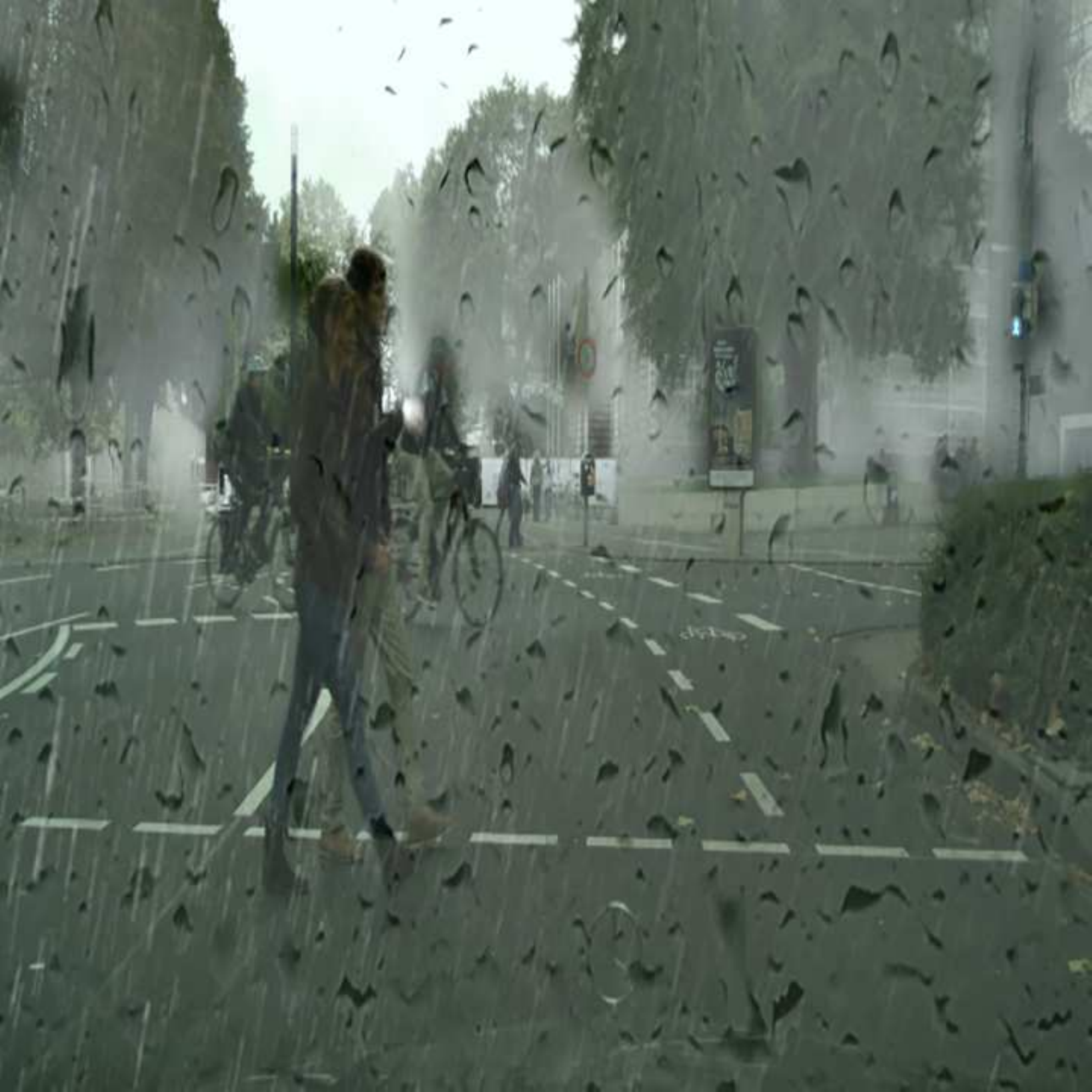}\vspace{2pt}
\includegraphics[width=1\linewidth]{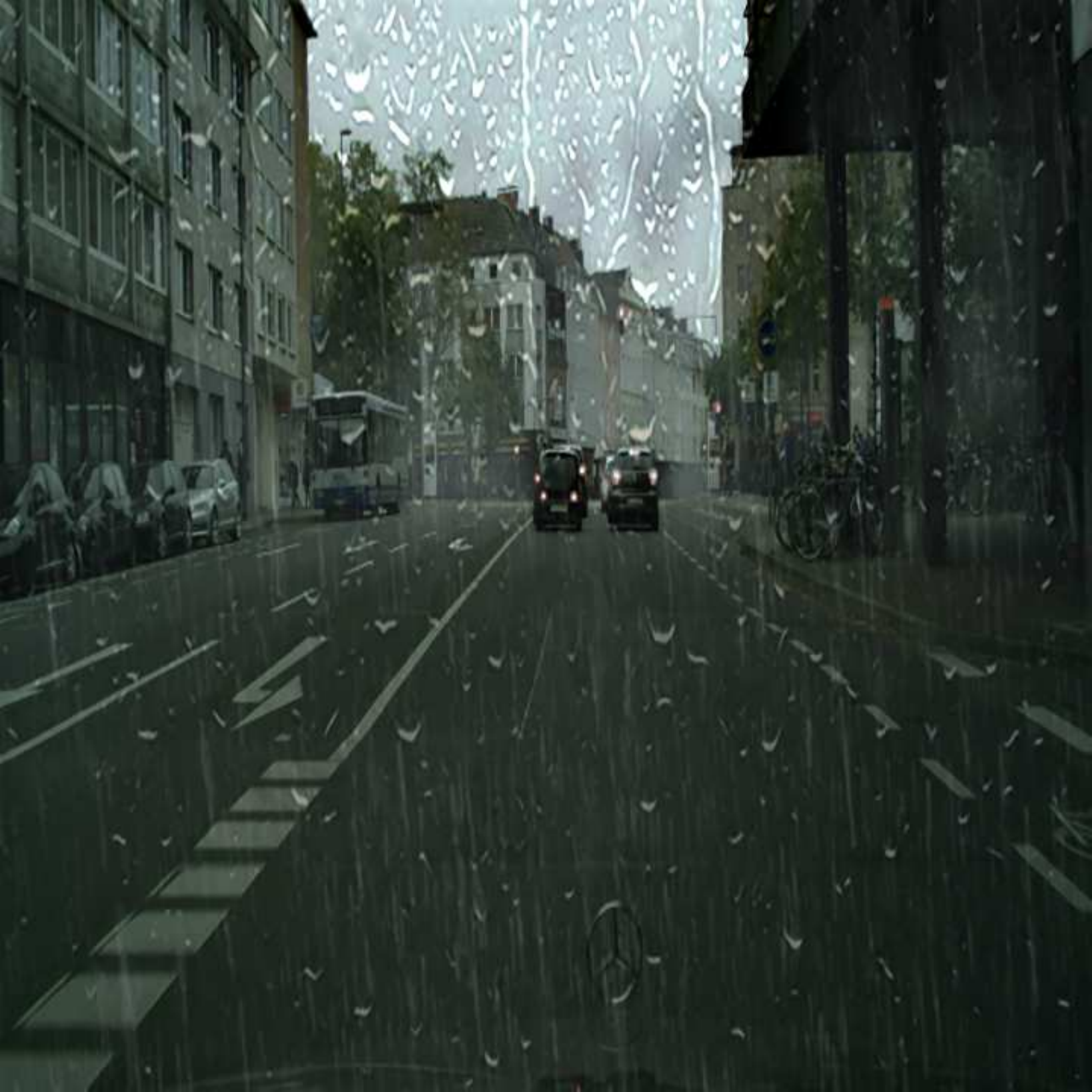}\vspace{2pt}
\includegraphics[width=1\linewidth]{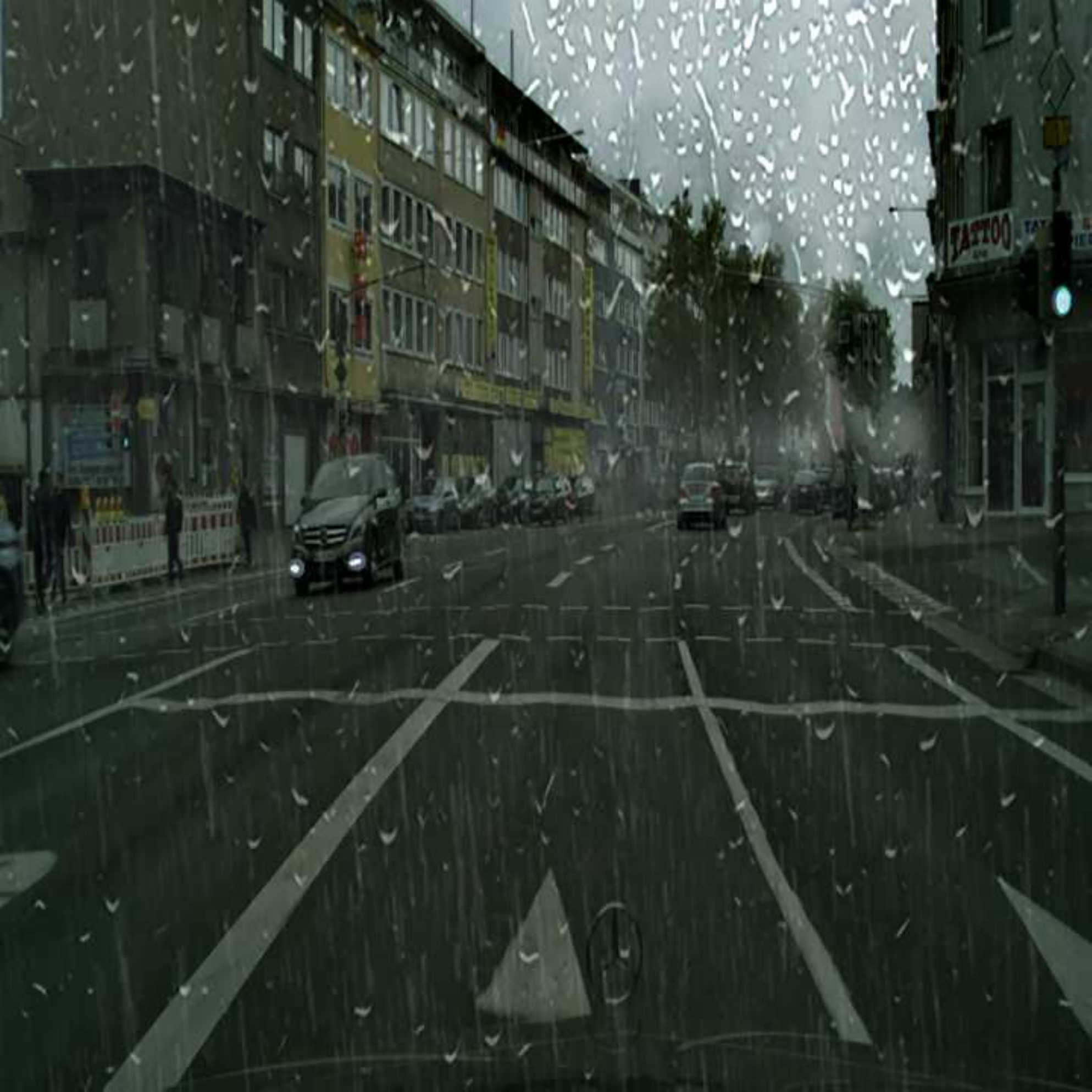}\vspace{2pt}
\includegraphics[width=1\linewidth]{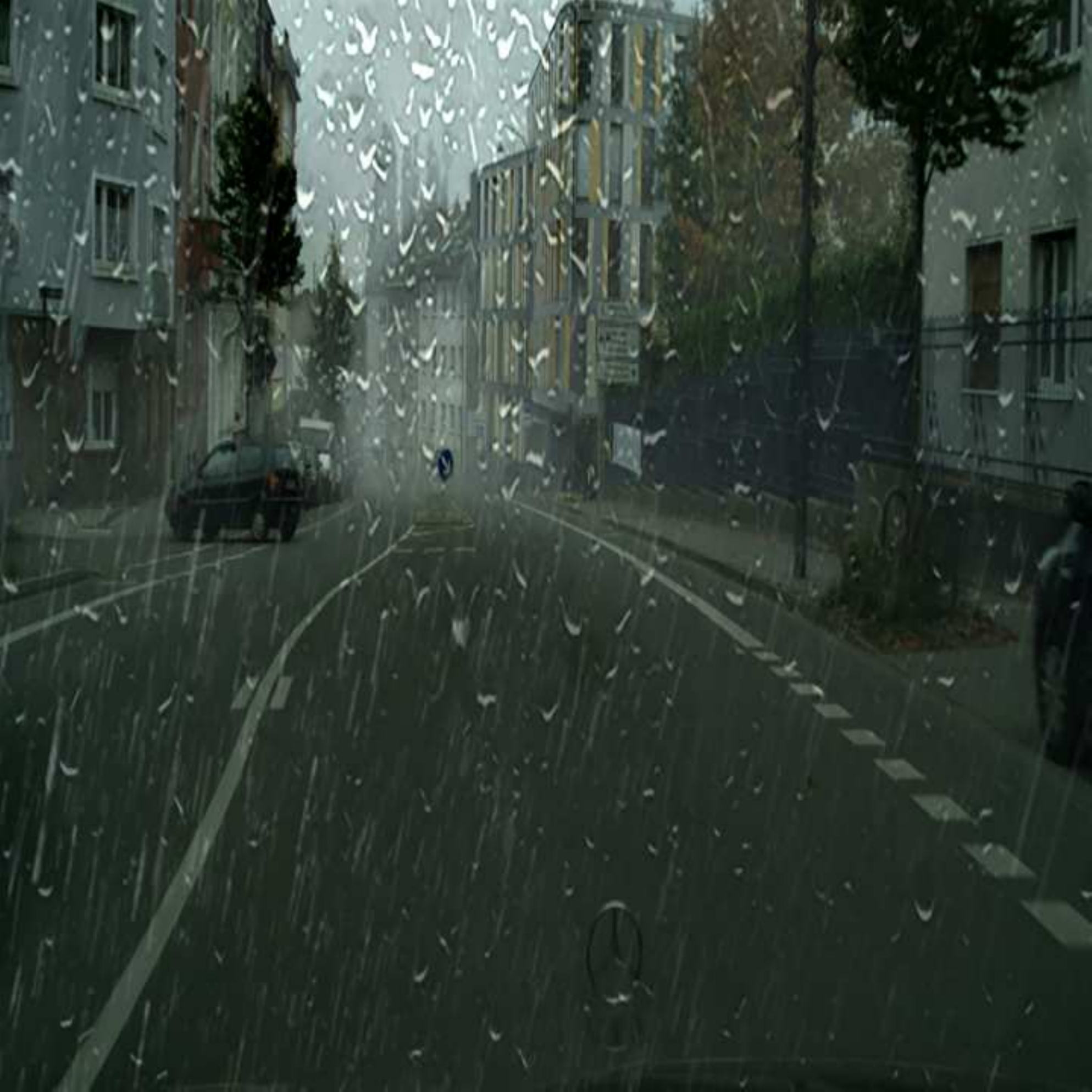}\vspace{2pt}
\includegraphics[width=1\linewidth]{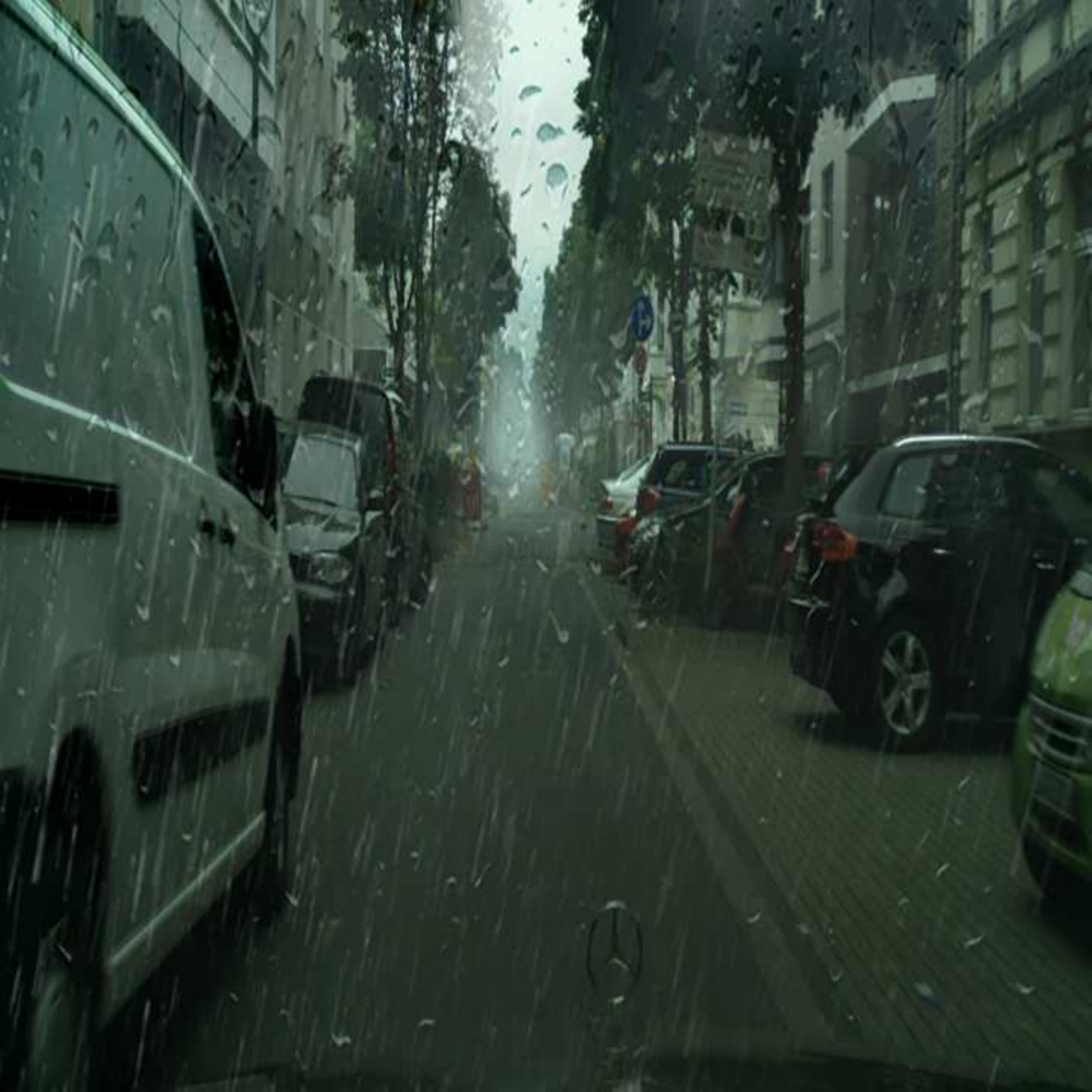}\vspace{2pt}
\end{minipage}}
\subfigure[Eigen et al.]{
\begin{minipage}[b]{0.1\linewidth}
\includegraphics[width=1\linewidth]{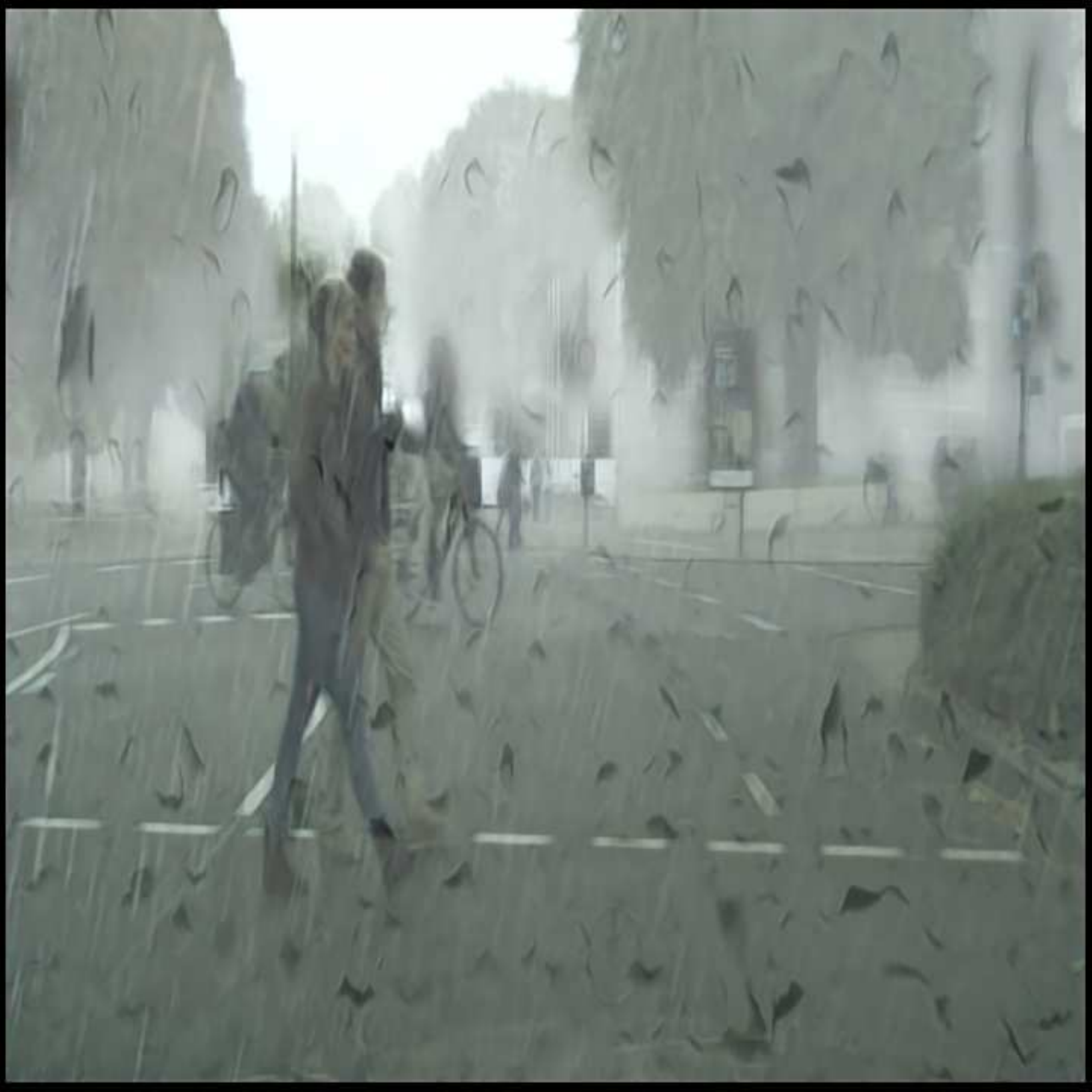}\vspace{2pt}
\includegraphics[width=1\linewidth]{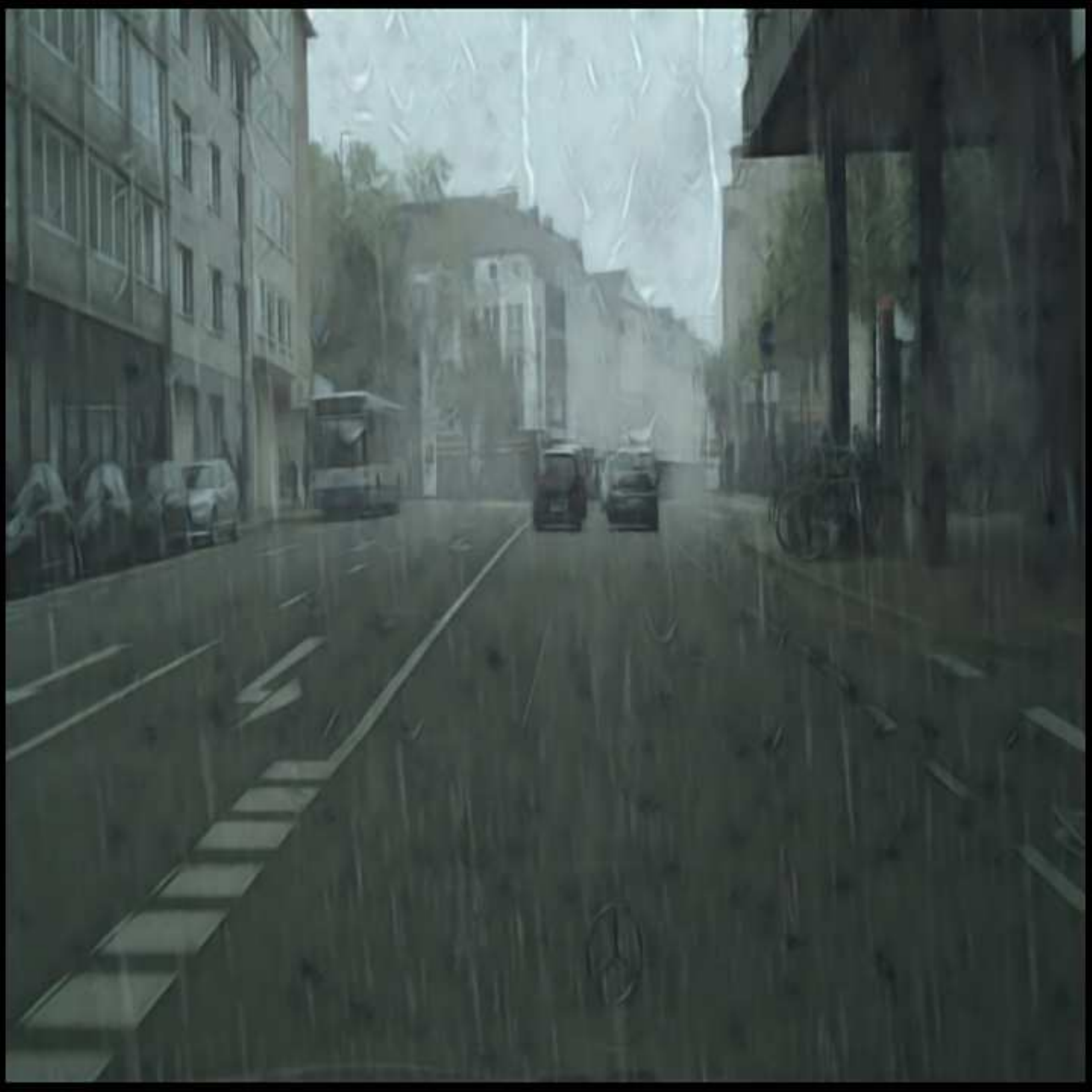}\vspace{2pt}
\includegraphics[width=1\linewidth]{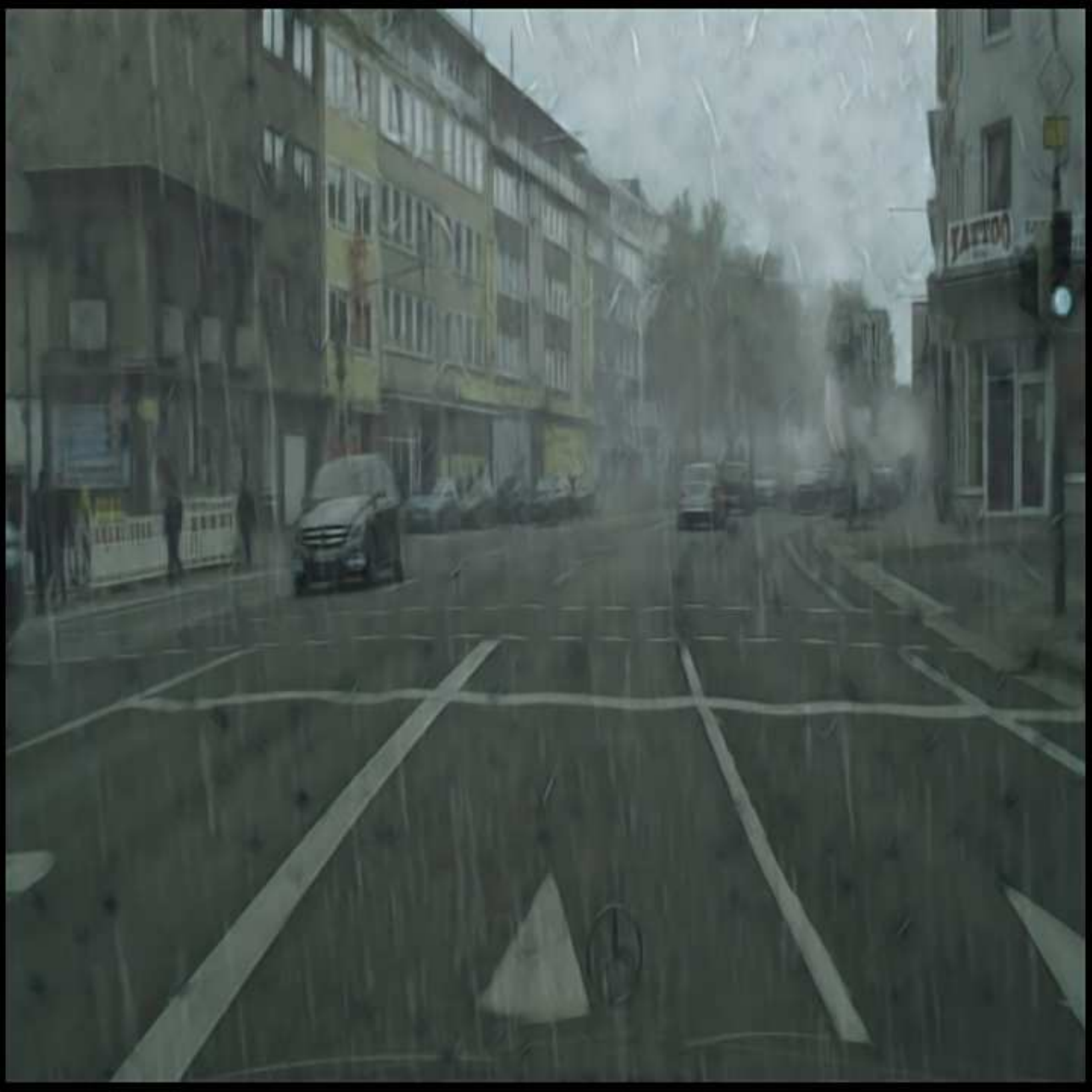}\vspace{2pt}
\includegraphics[width=1\linewidth]{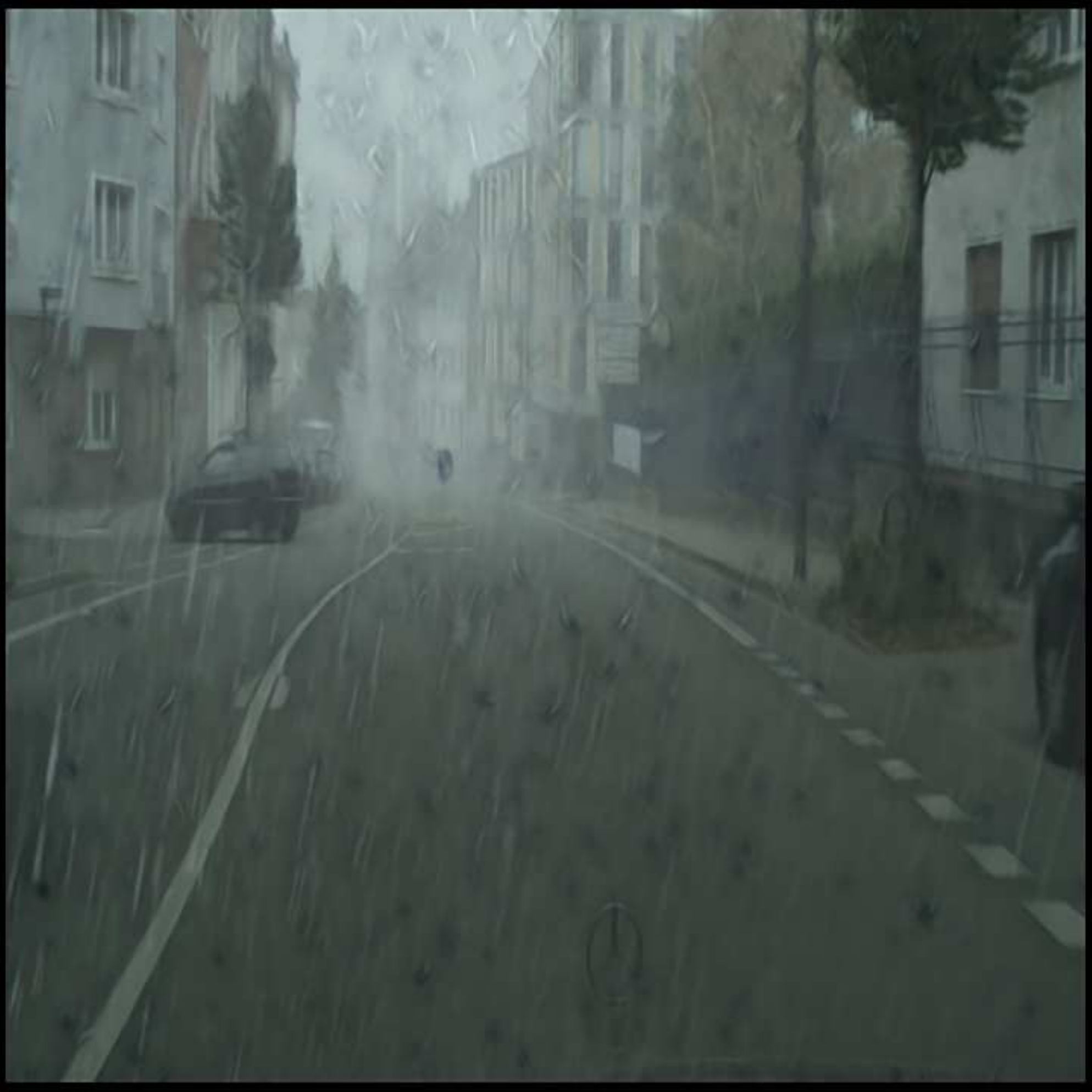}\vspace{2pt}
\includegraphics[width=1\linewidth]{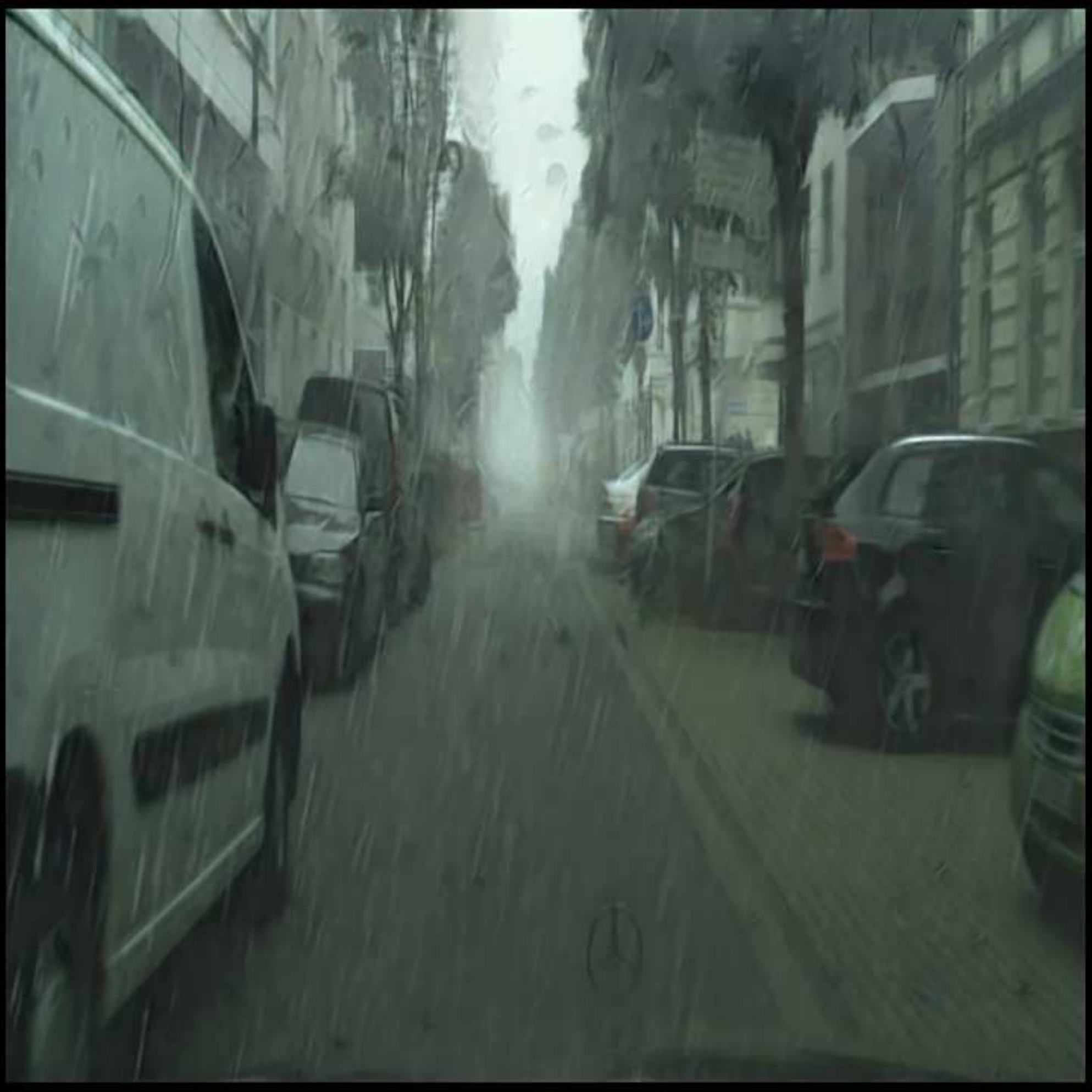}\vspace{2pt}
\end{minipage}}
\subfigure[Wang et al.]{
\begin{minipage}[b]{0.1\linewidth}
\includegraphics[width=1\linewidth]{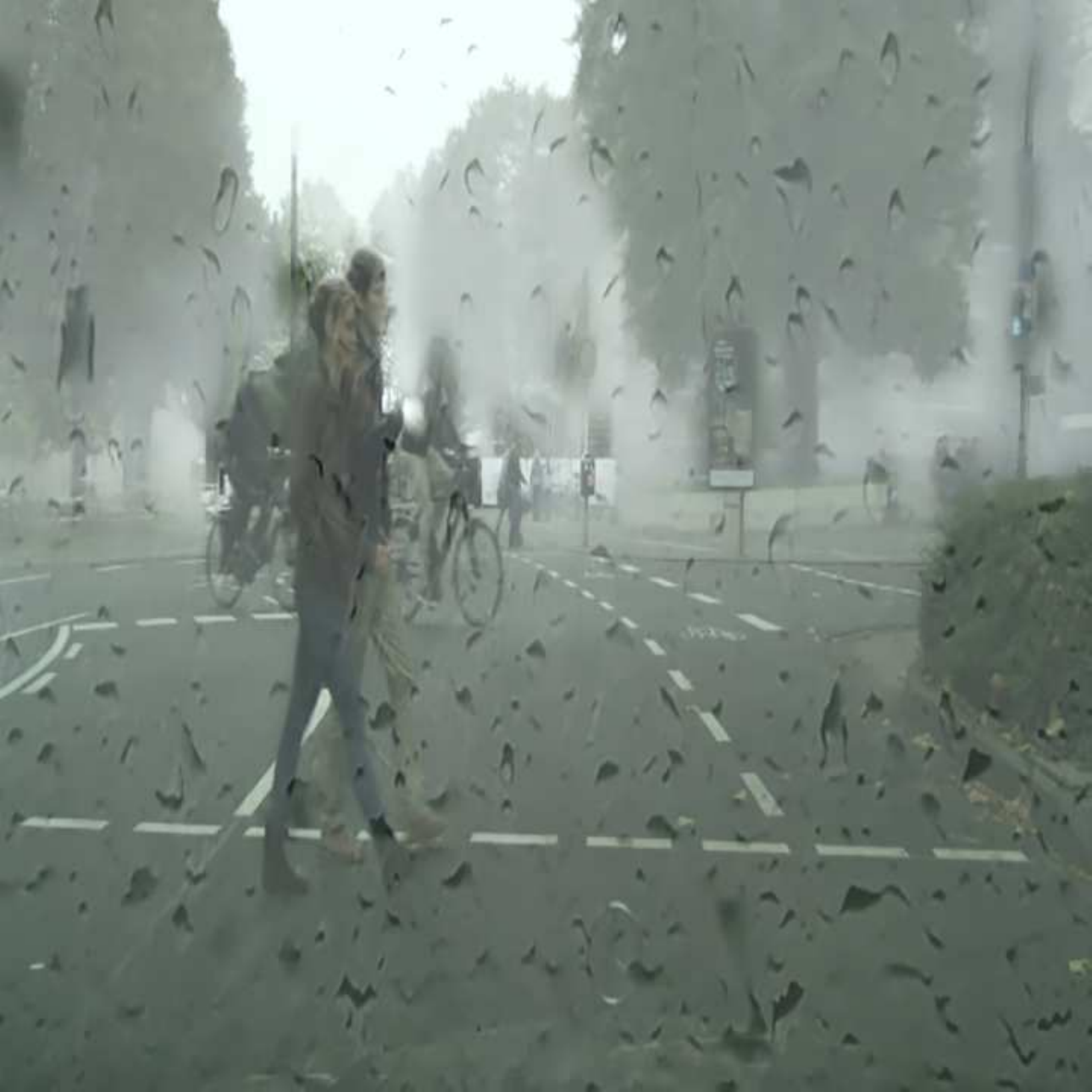}\vspace{2pt}
\includegraphics[width=1\linewidth]{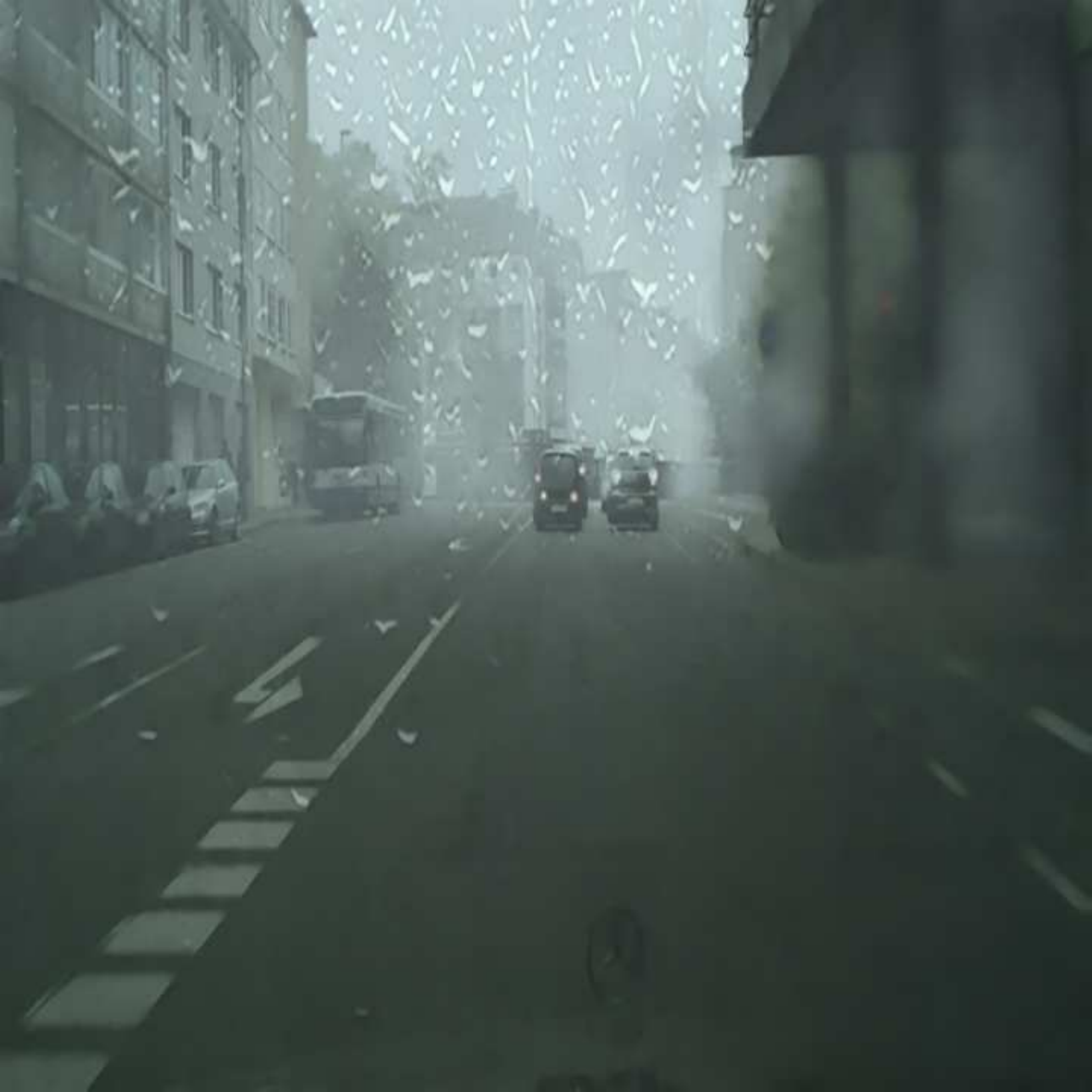}\vspace{2pt}
\includegraphics[width=1\linewidth]{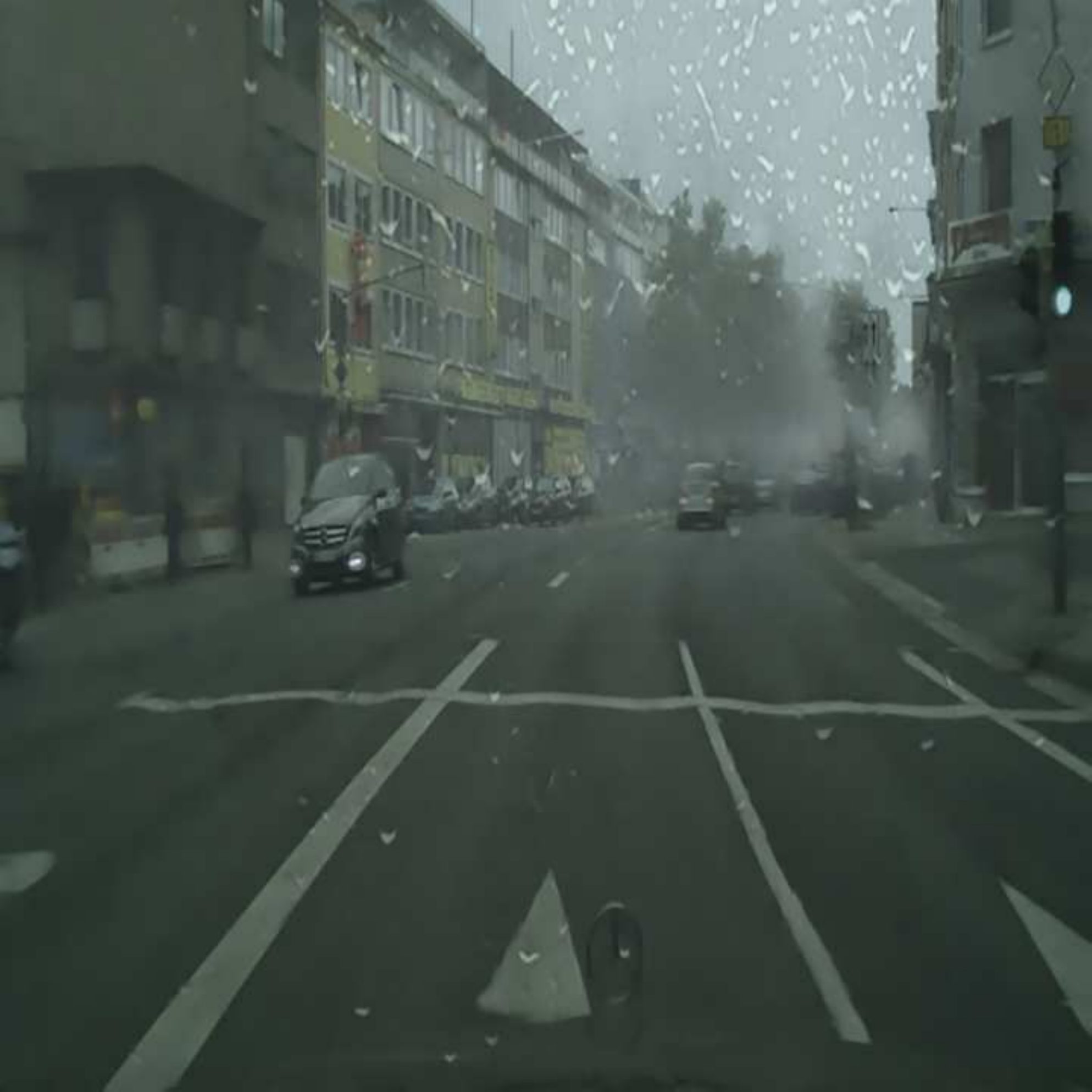}\vspace{2pt}
\includegraphics[width=1\linewidth]{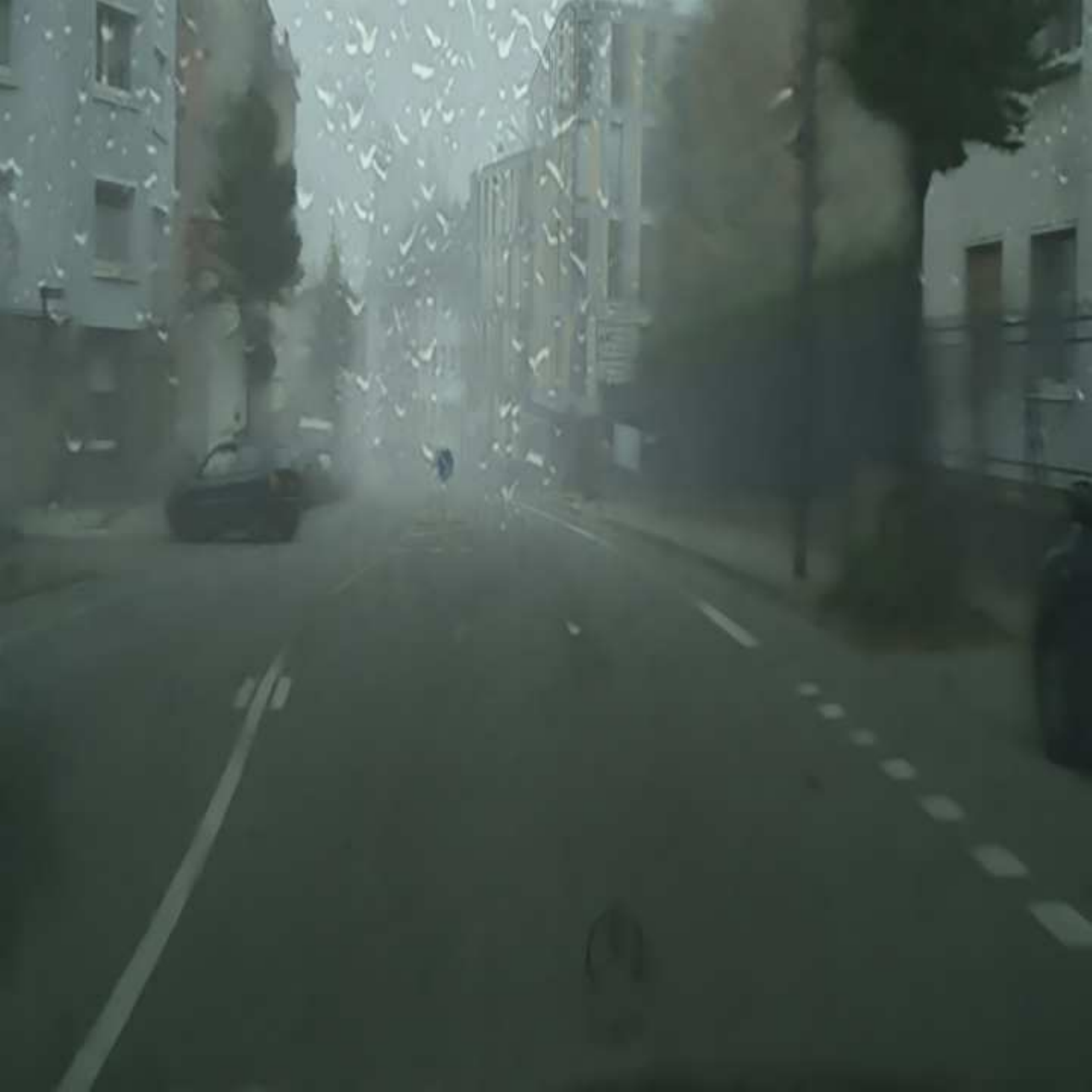}\vspace{2pt}
\includegraphics[width=1\linewidth]{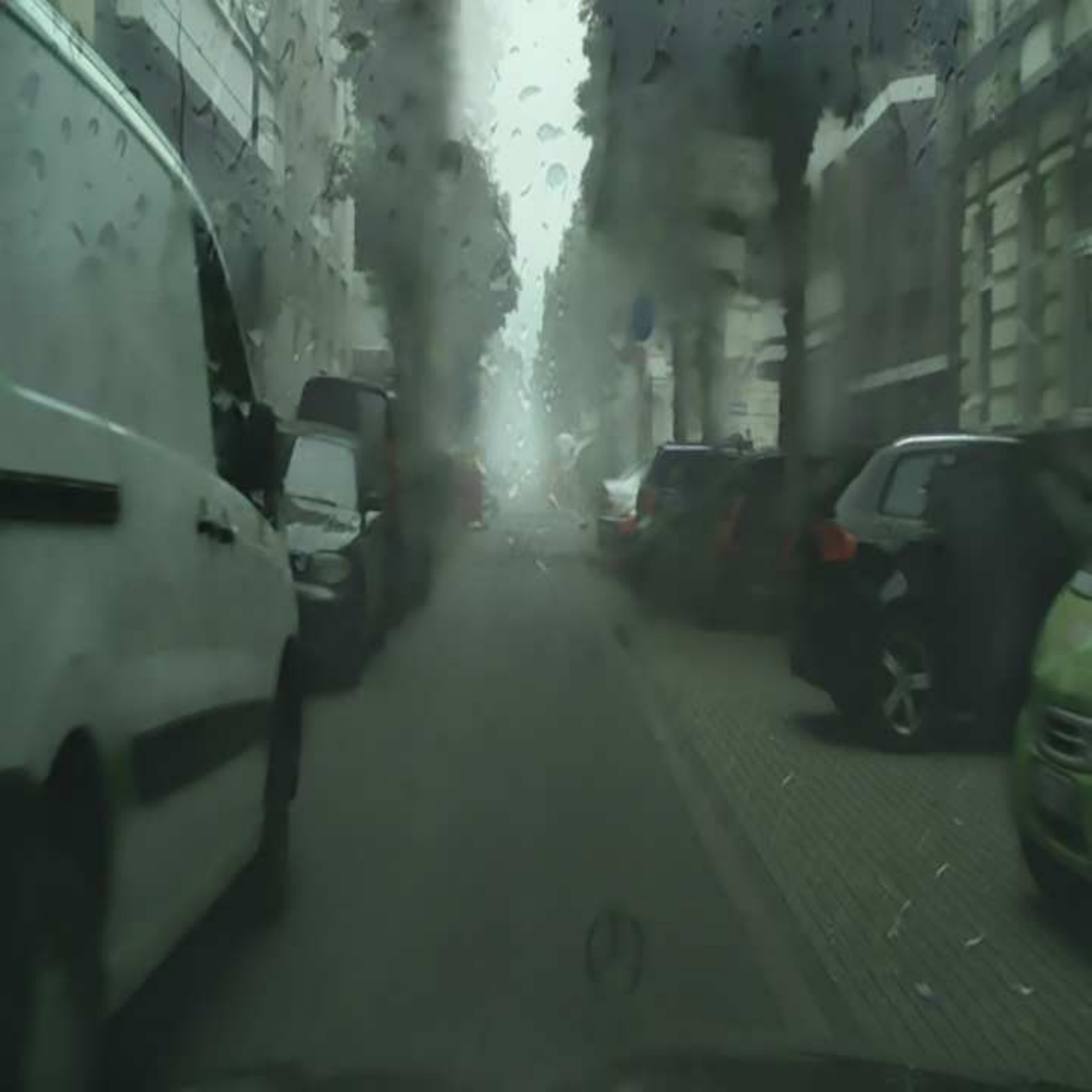}\vspace{2pt}
\end{minipage}}
\subfigure[Qian et al.]{
\begin{minipage}[b]{0.1\linewidth}
\includegraphics[width=1\linewidth]{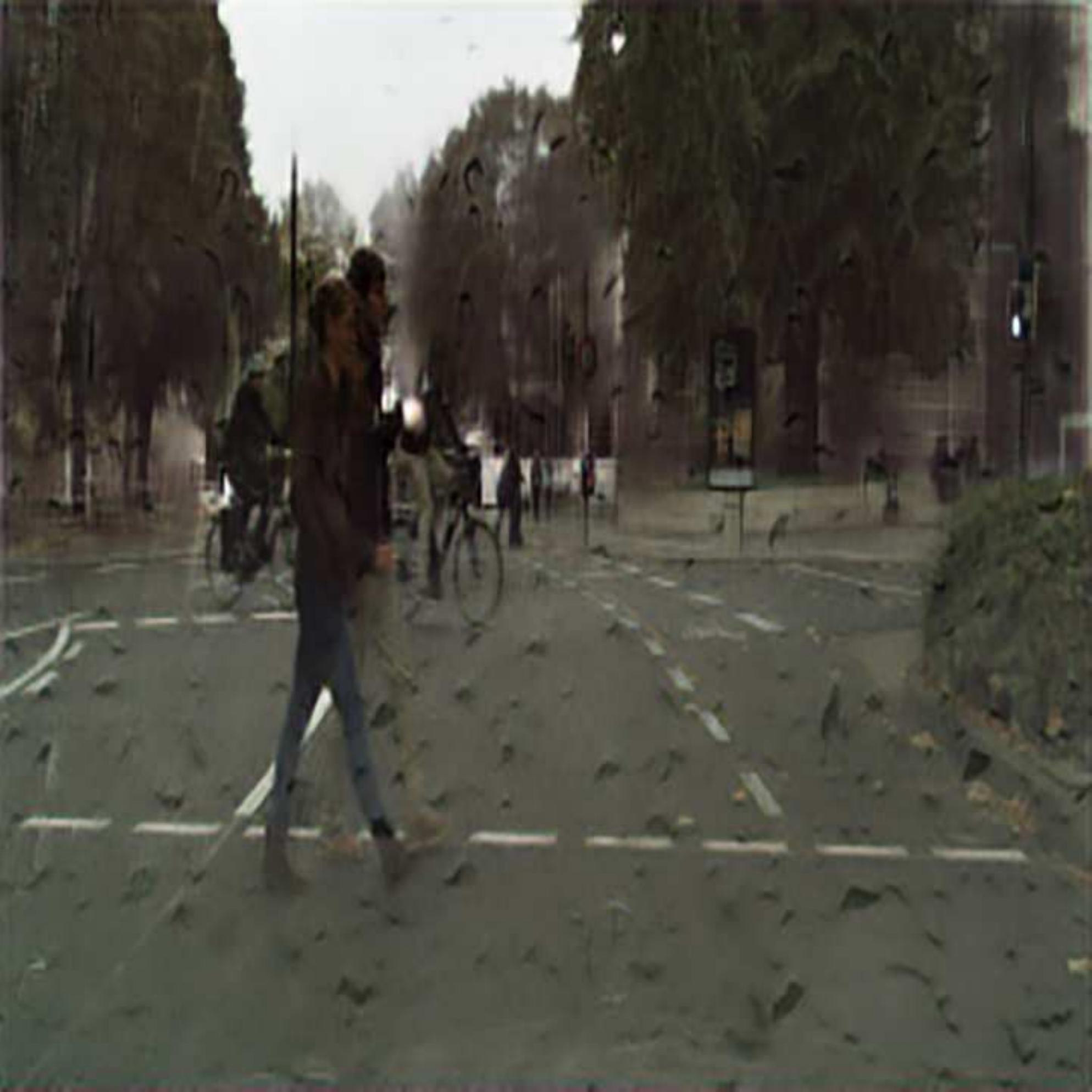}\vspace{2pt}
\includegraphics[width=1\linewidth]{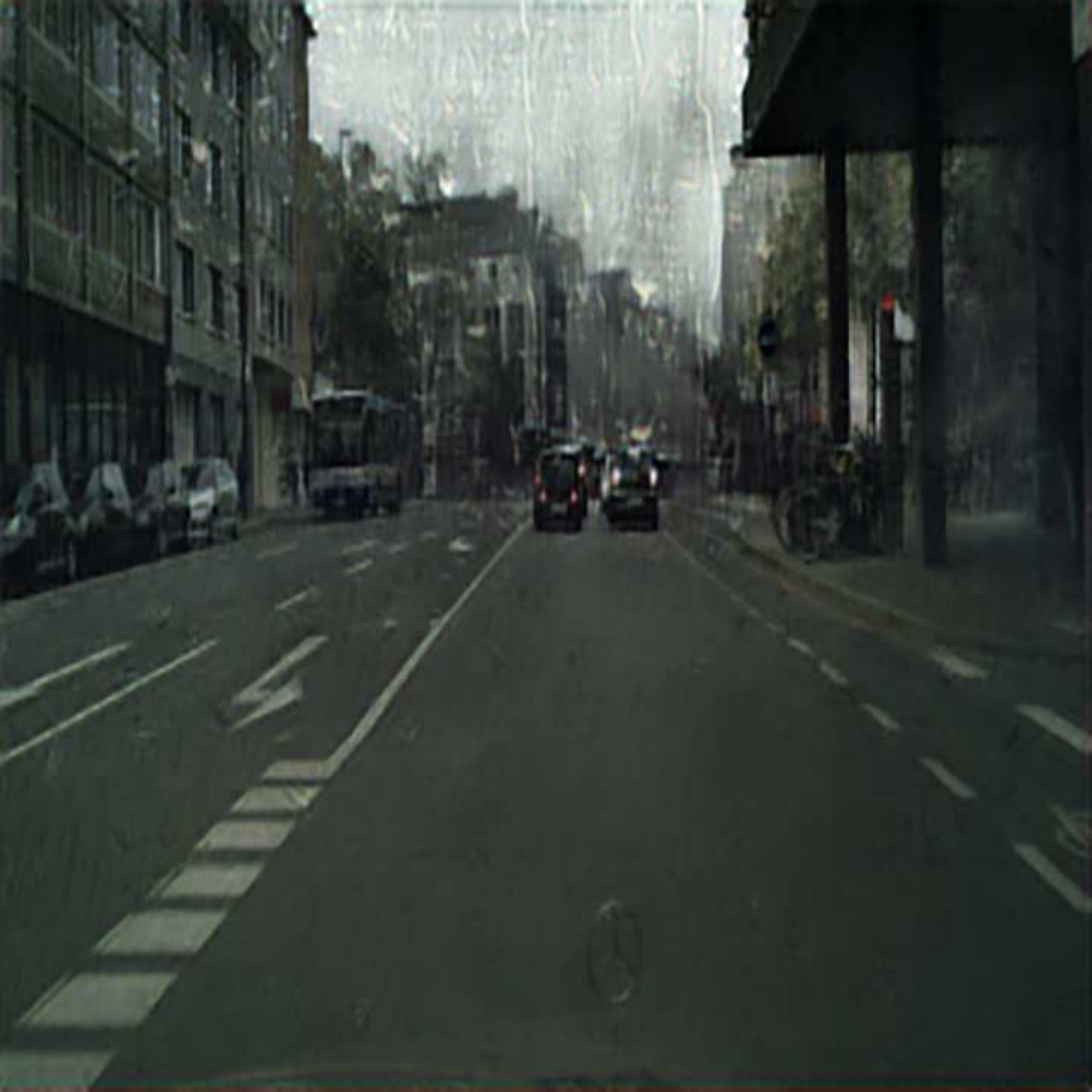}\vspace{2pt}
\includegraphics[width=1\linewidth]{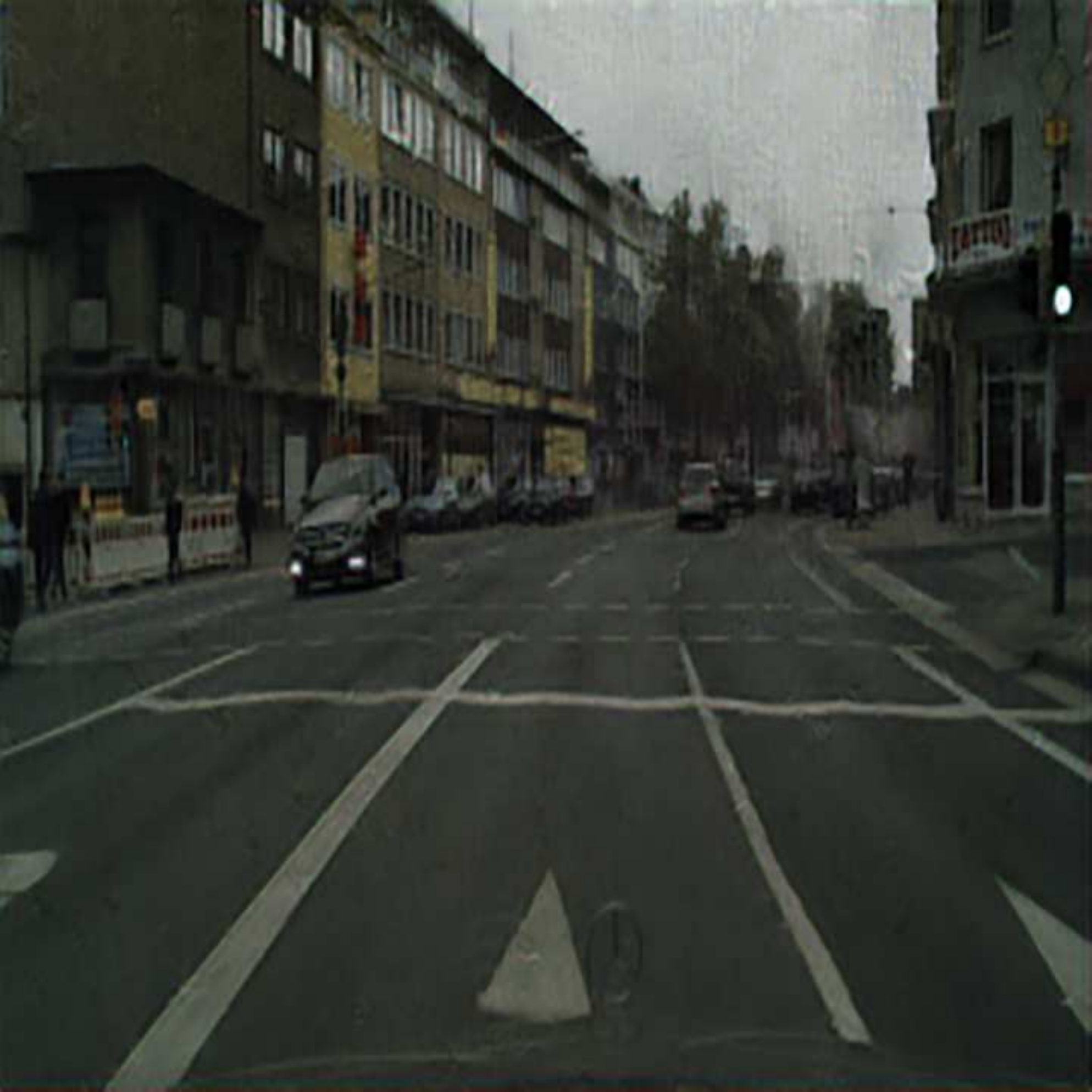}\vspace{2pt}
\includegraphics[width=1\linewidth]{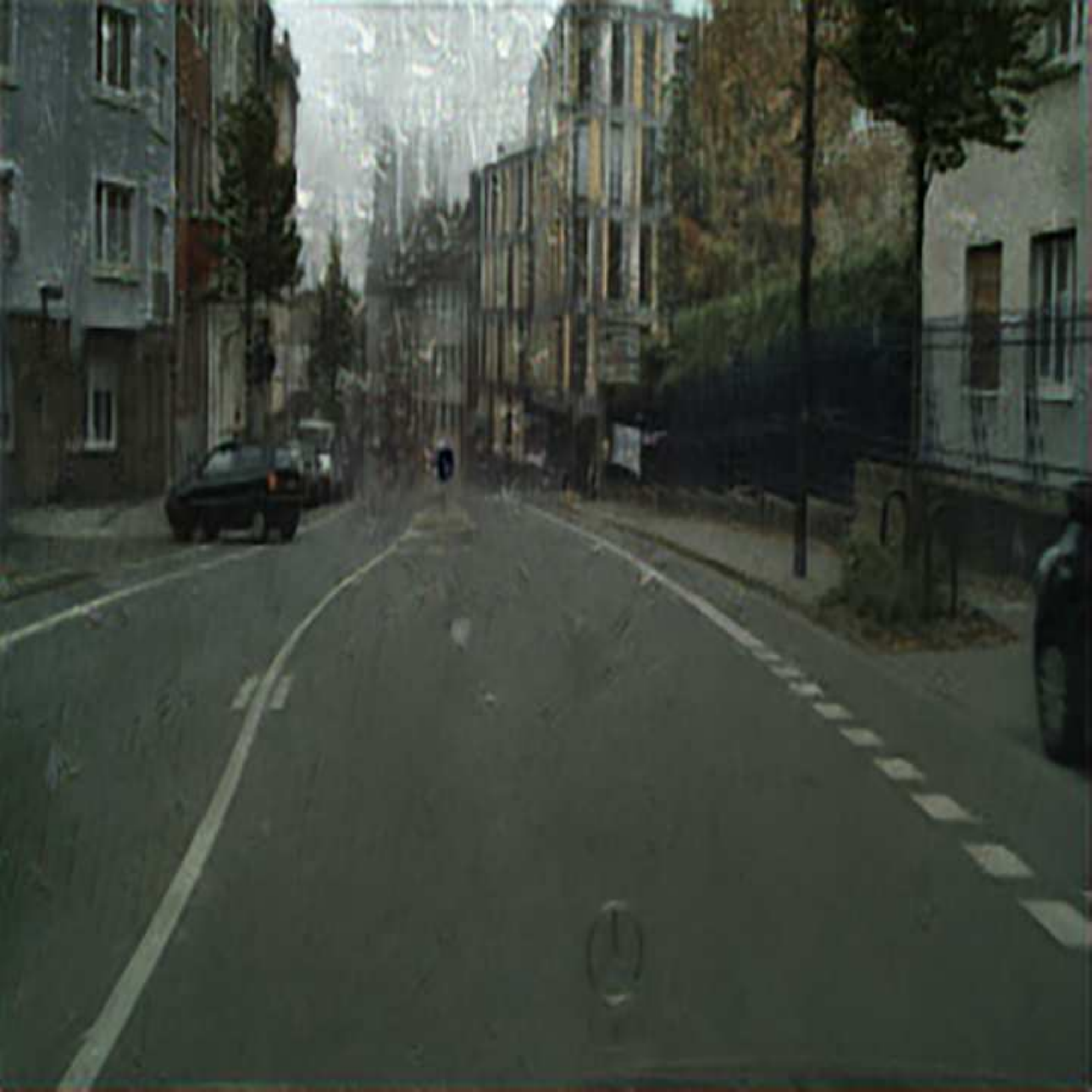}\vspace{2pt}
\includegraphics[width=1\linewidth]{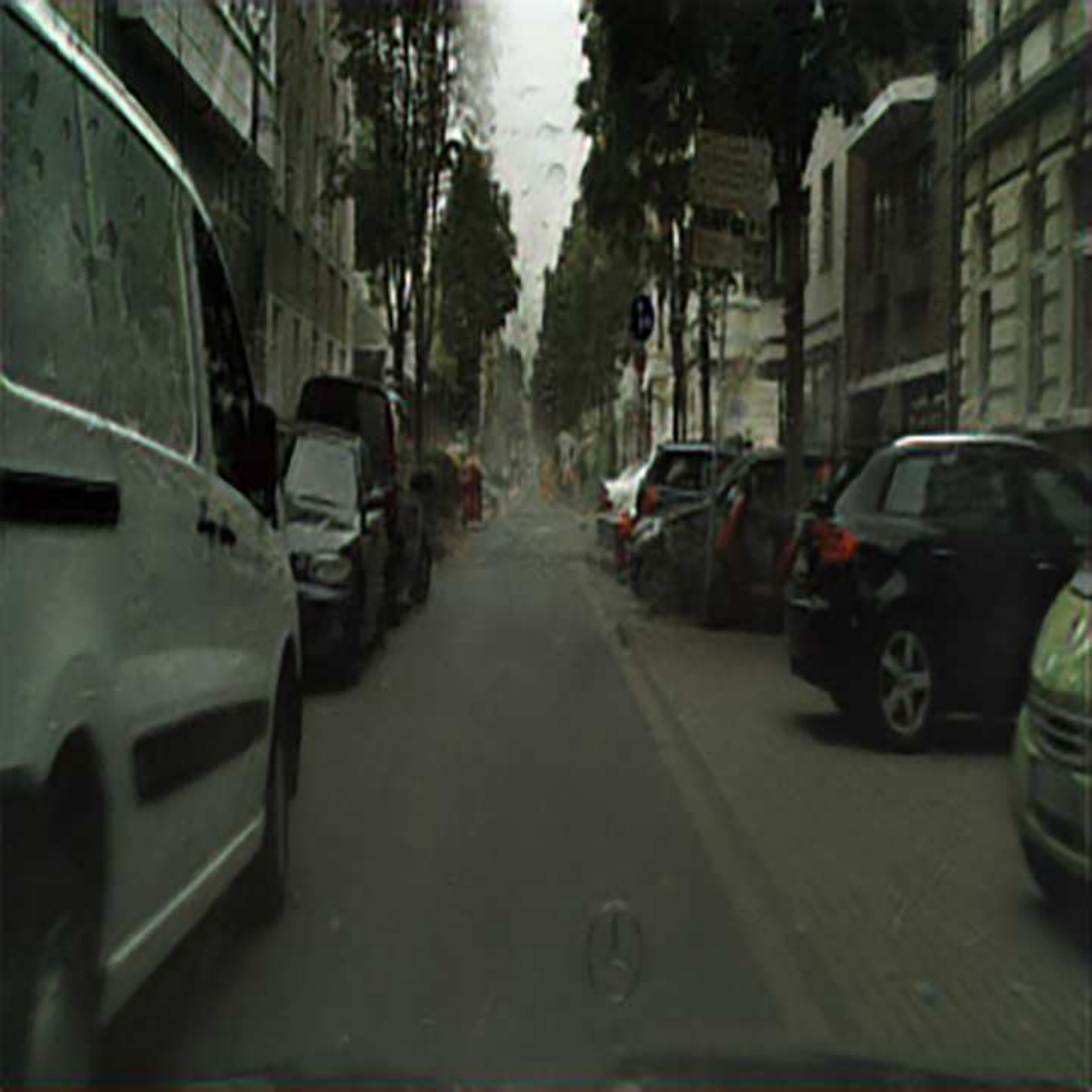}\vspace{2pt}
\end{minipage}}
\subfigure[Hu et al.]{
\begin{minipage}[b]{0.1\linewidth}
\includegraphics[width=1\linewidth]{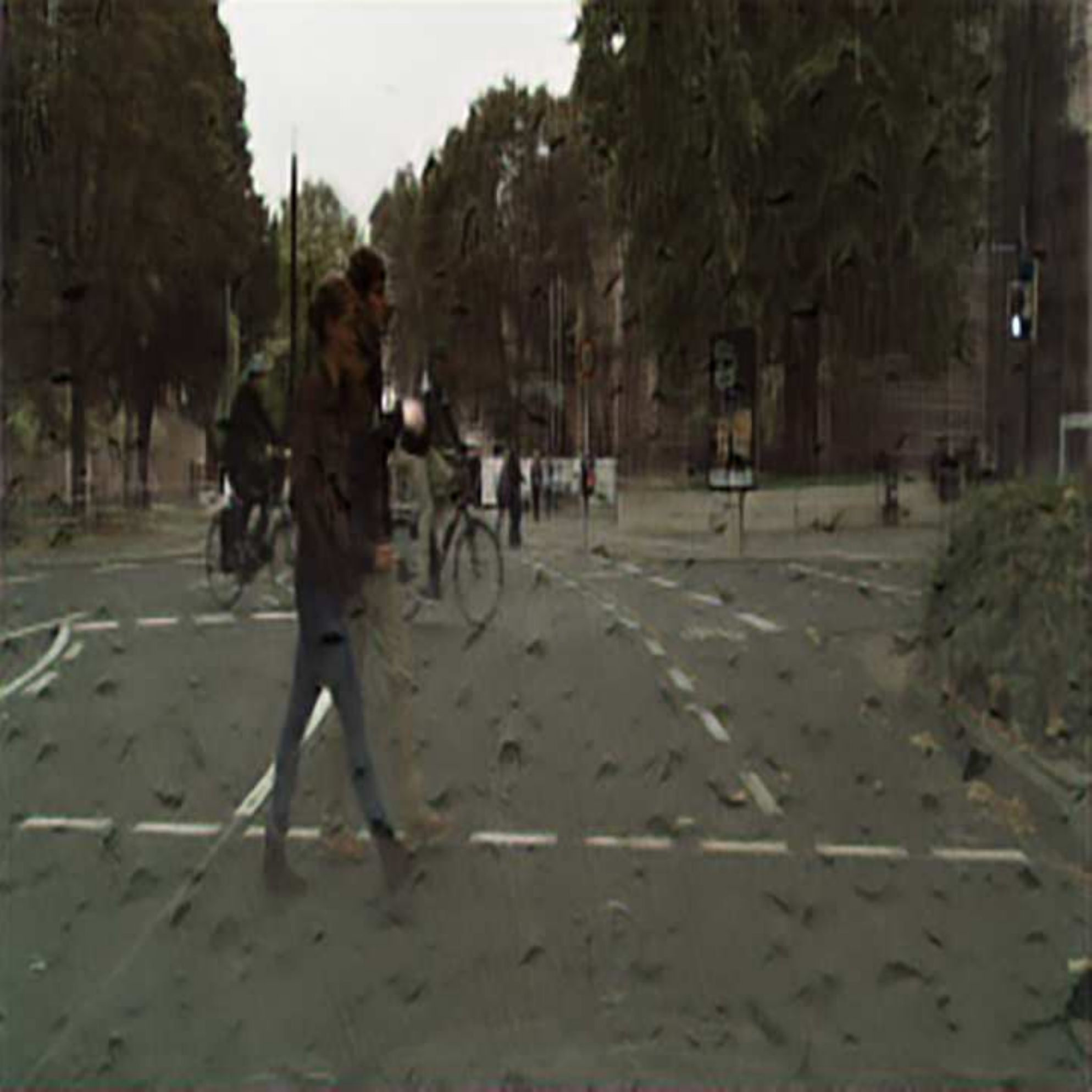}\vspace{2pt}
\includegraphics[width=1\linewidth]{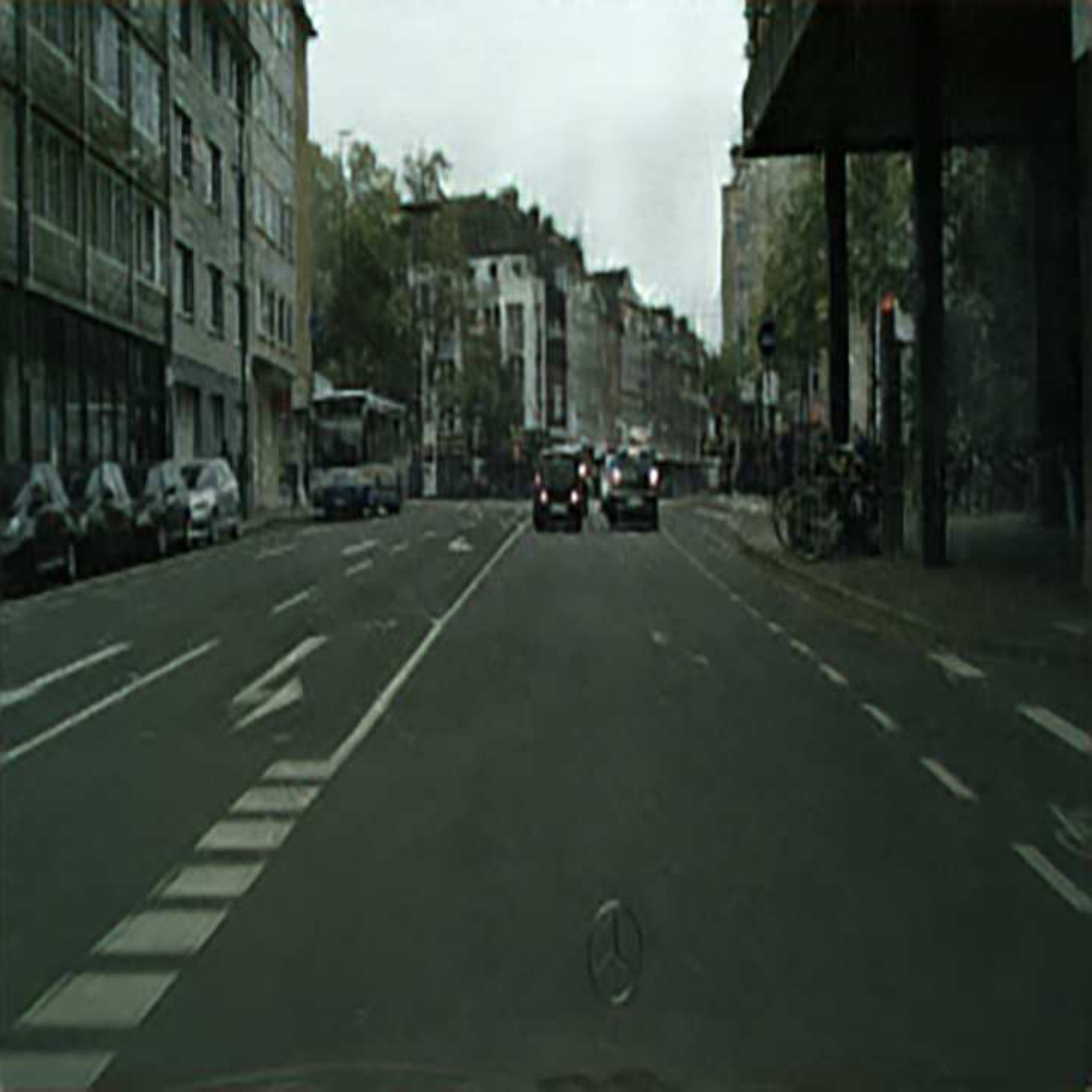}\vspace{2pt}
\includegraphics[width=1\linewidth]{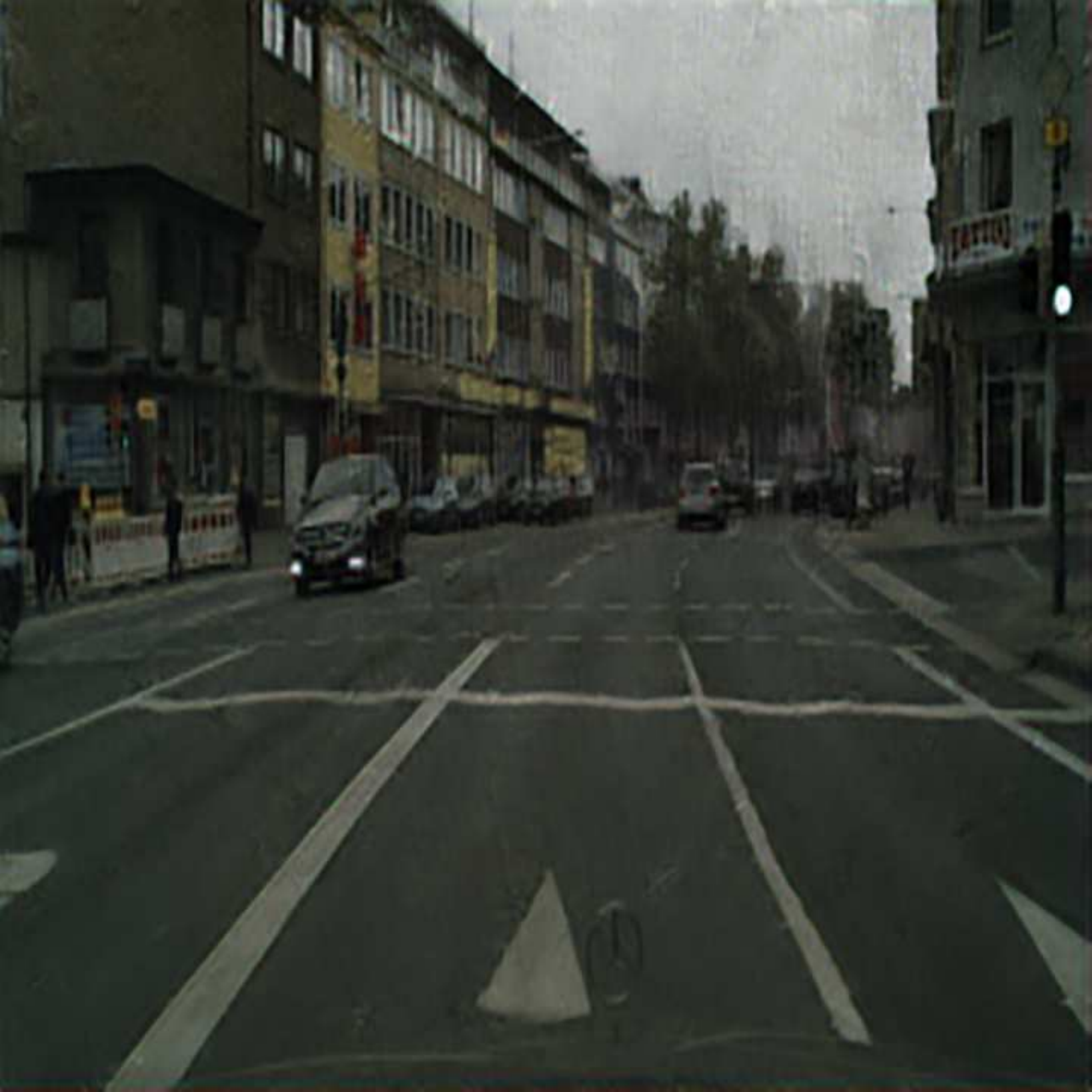}\vspace{2pt}
\includegraphics[width=1\linewidth]{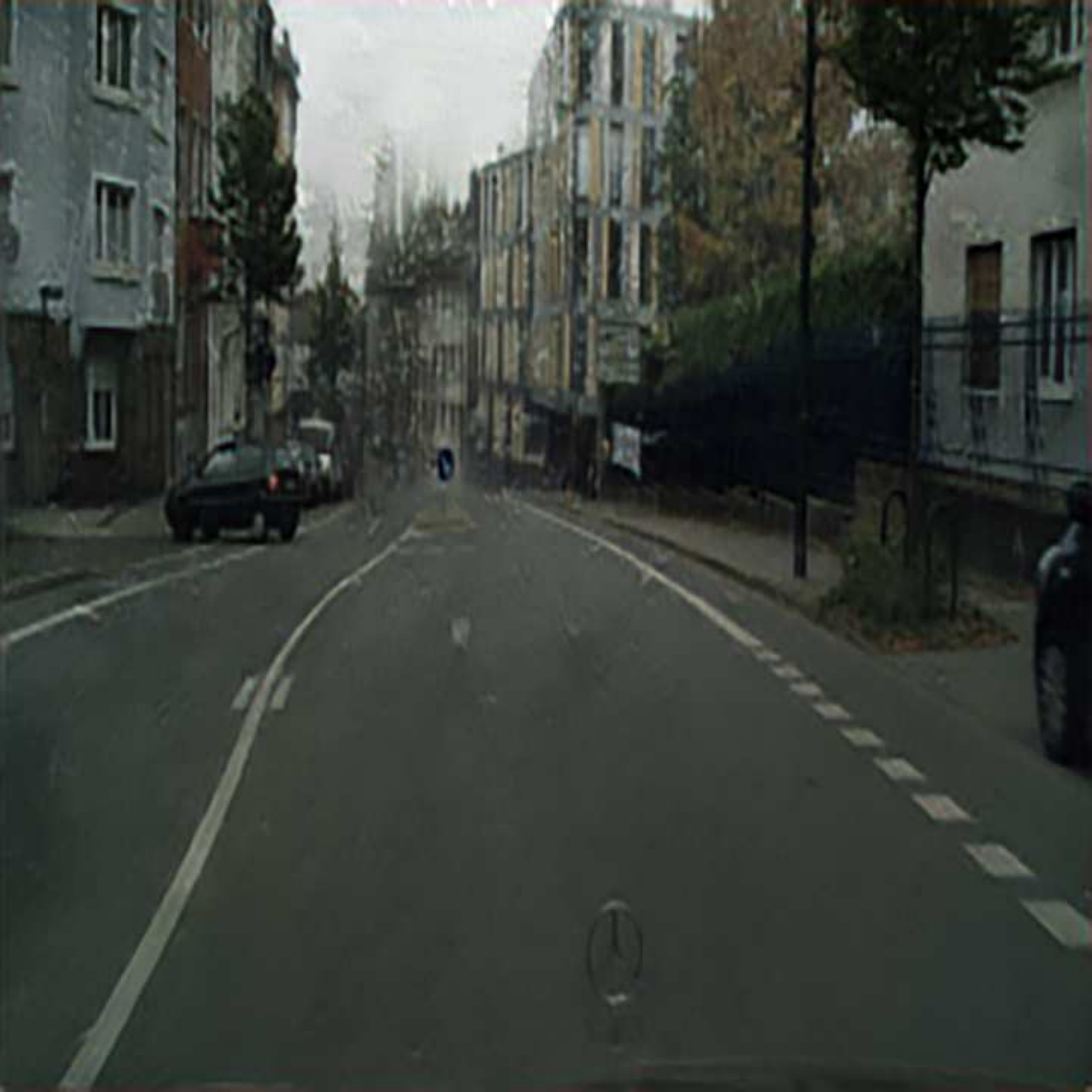}\vspace{2pt}
\includegraphics[width=1\linewidth]{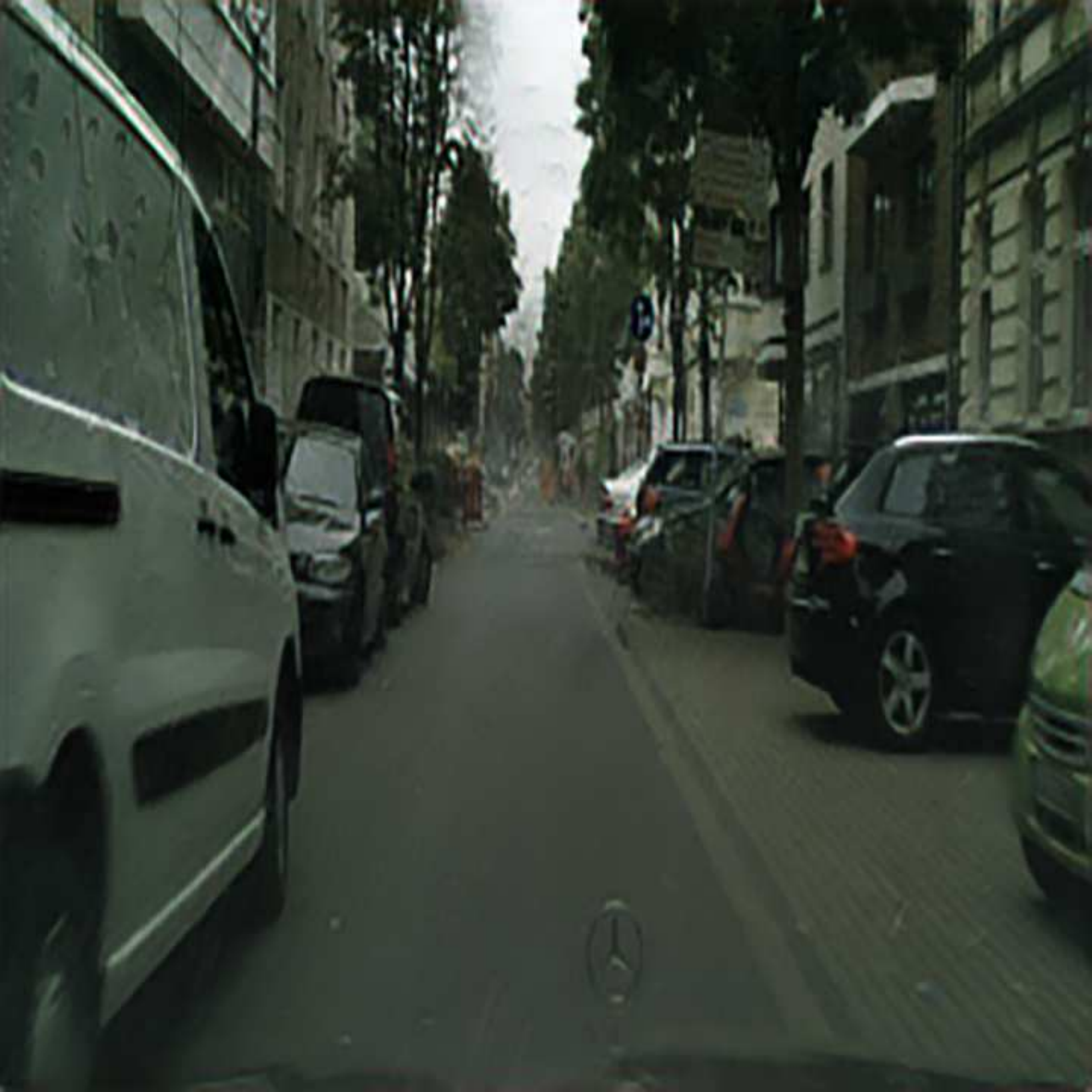}\vspace{2pt}
\end{minipage}}
\subfigure[Ours]{
\begin{minipage}[b]{0.1\linewidth}
\includegraphics[width=1\linewidth]{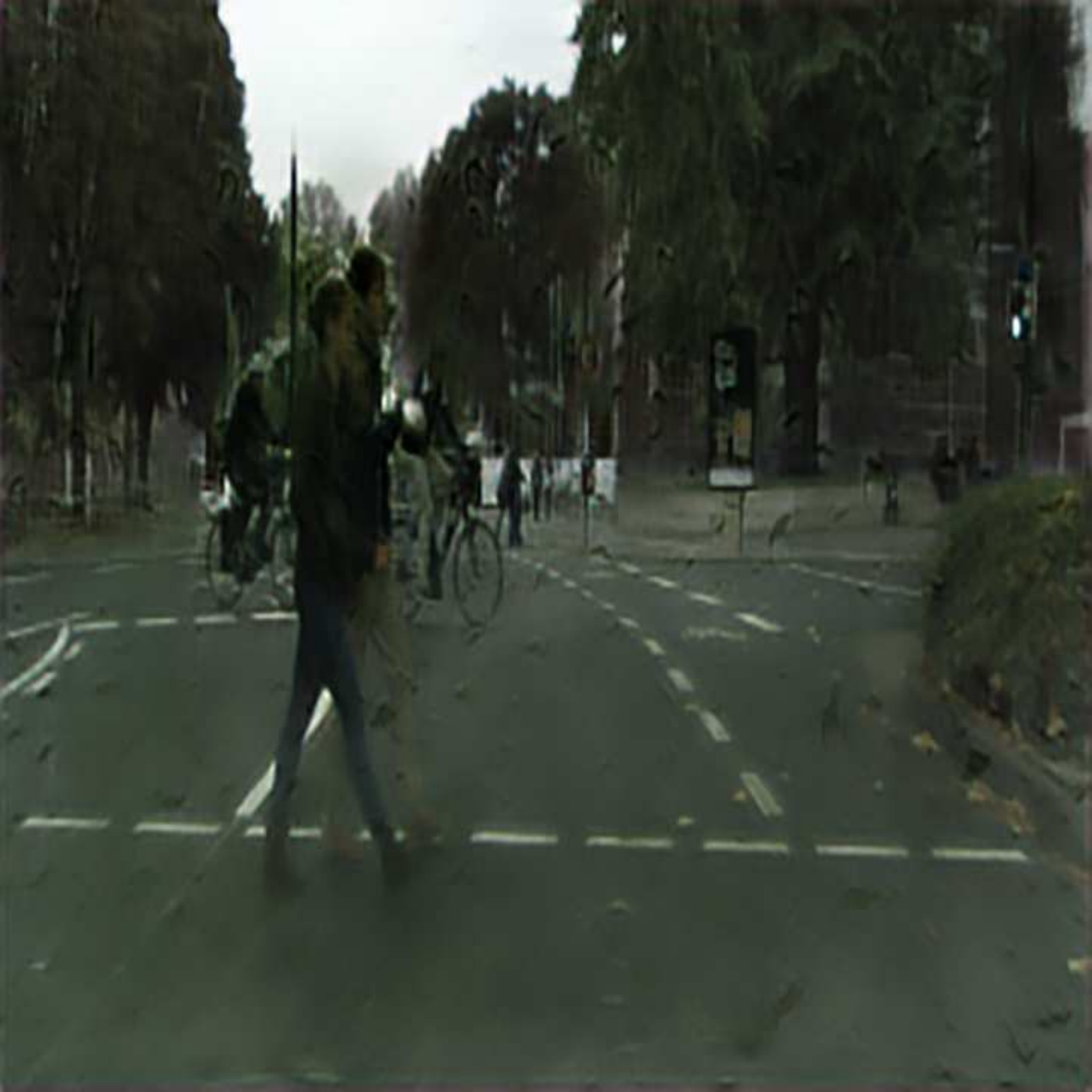}\vspace{2pt}
\includegraphics[width=1\linewidth]{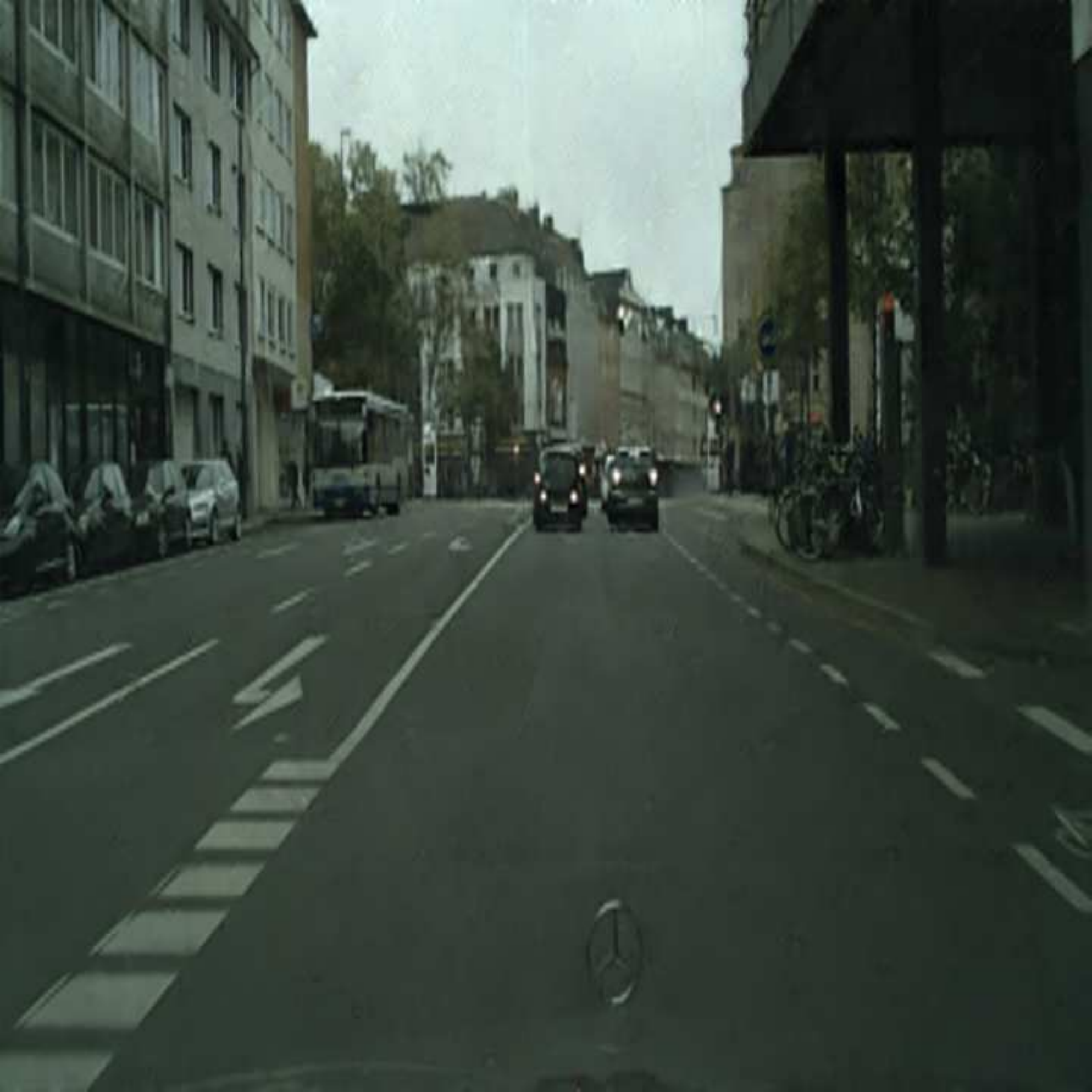}\vspace{2pt}
\includegraphics[width=1\linewidth]{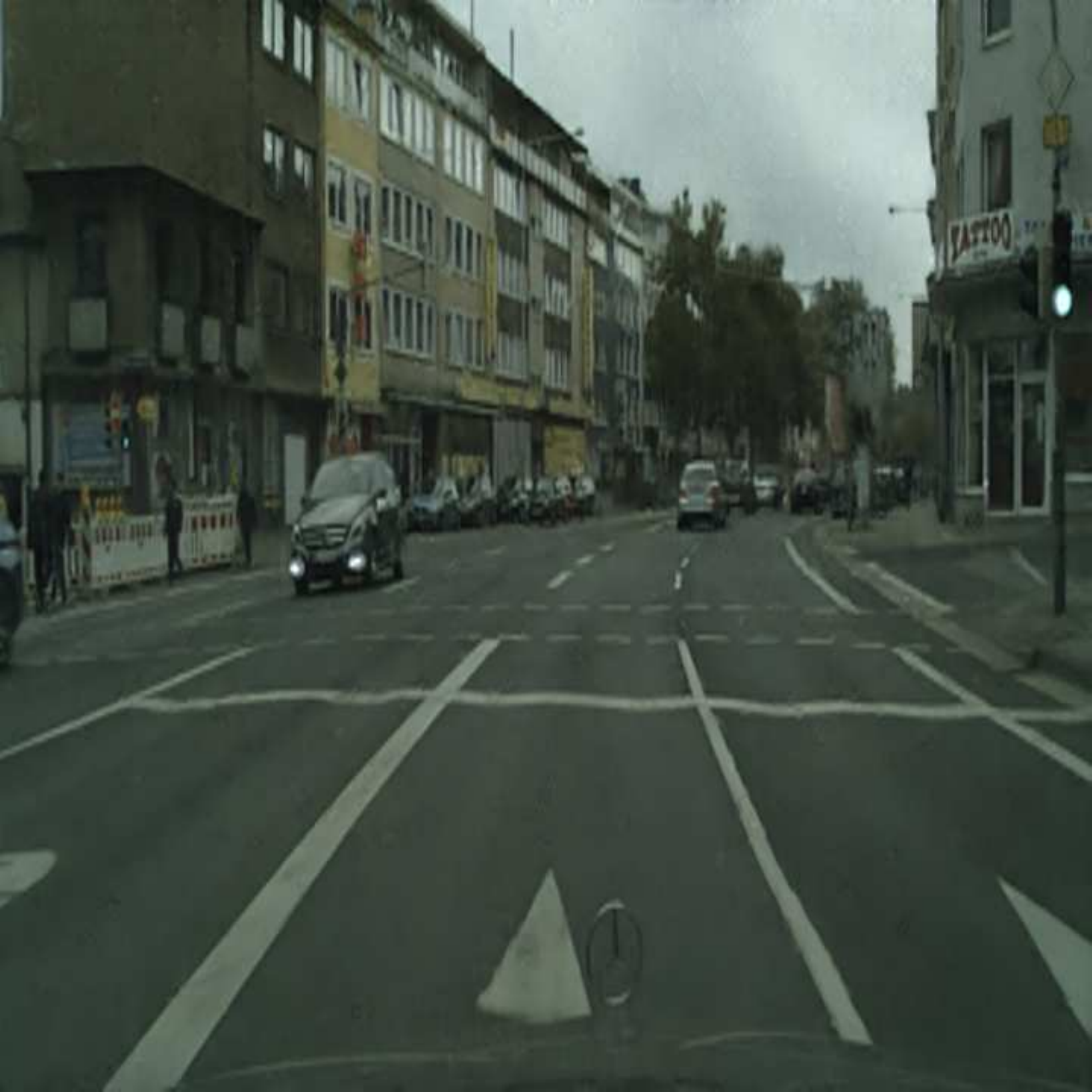}\vspace{2pt}
\includegraphics[width=1\linewidth]{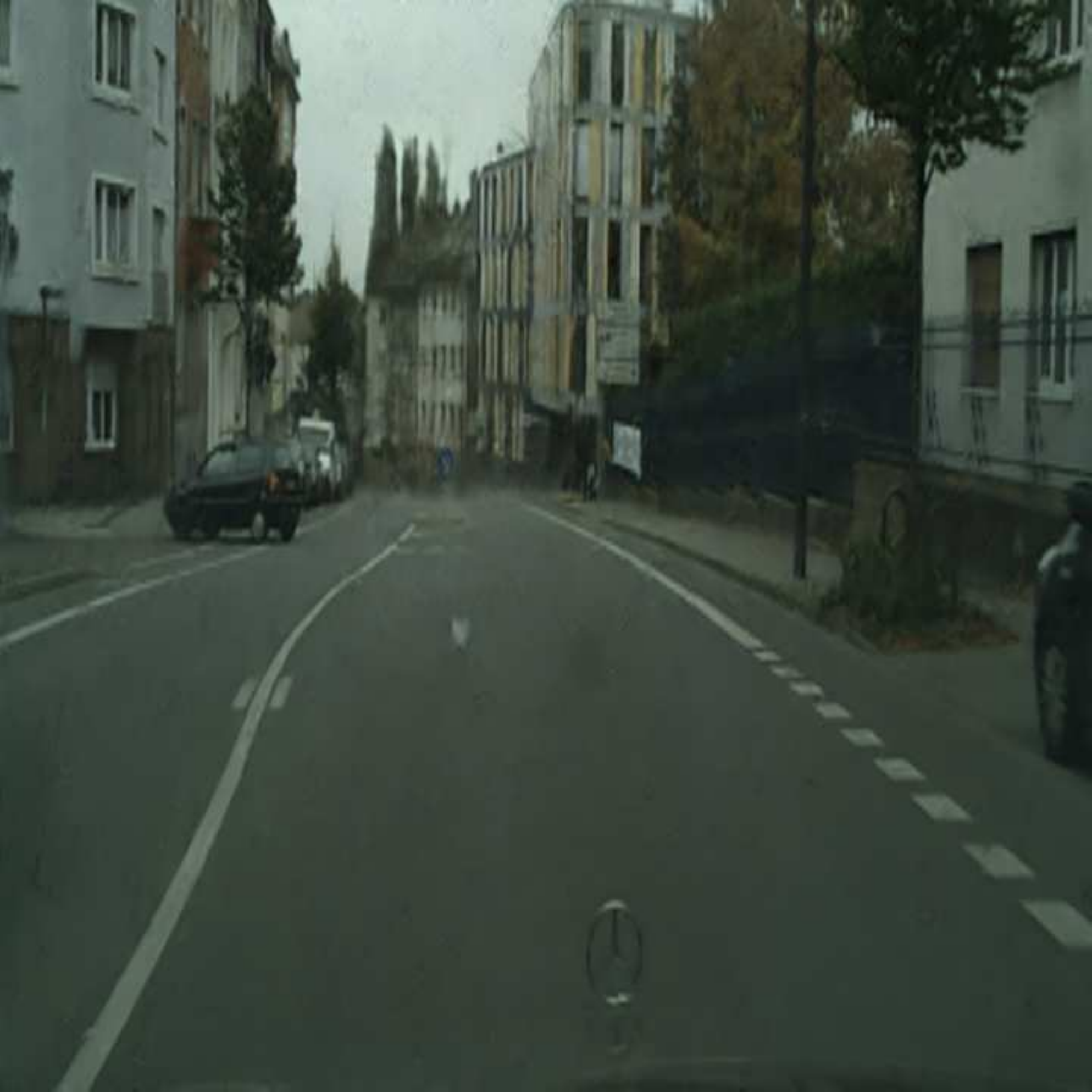}\vspace{2pt}
\includegraphics[width=1\linewidth]{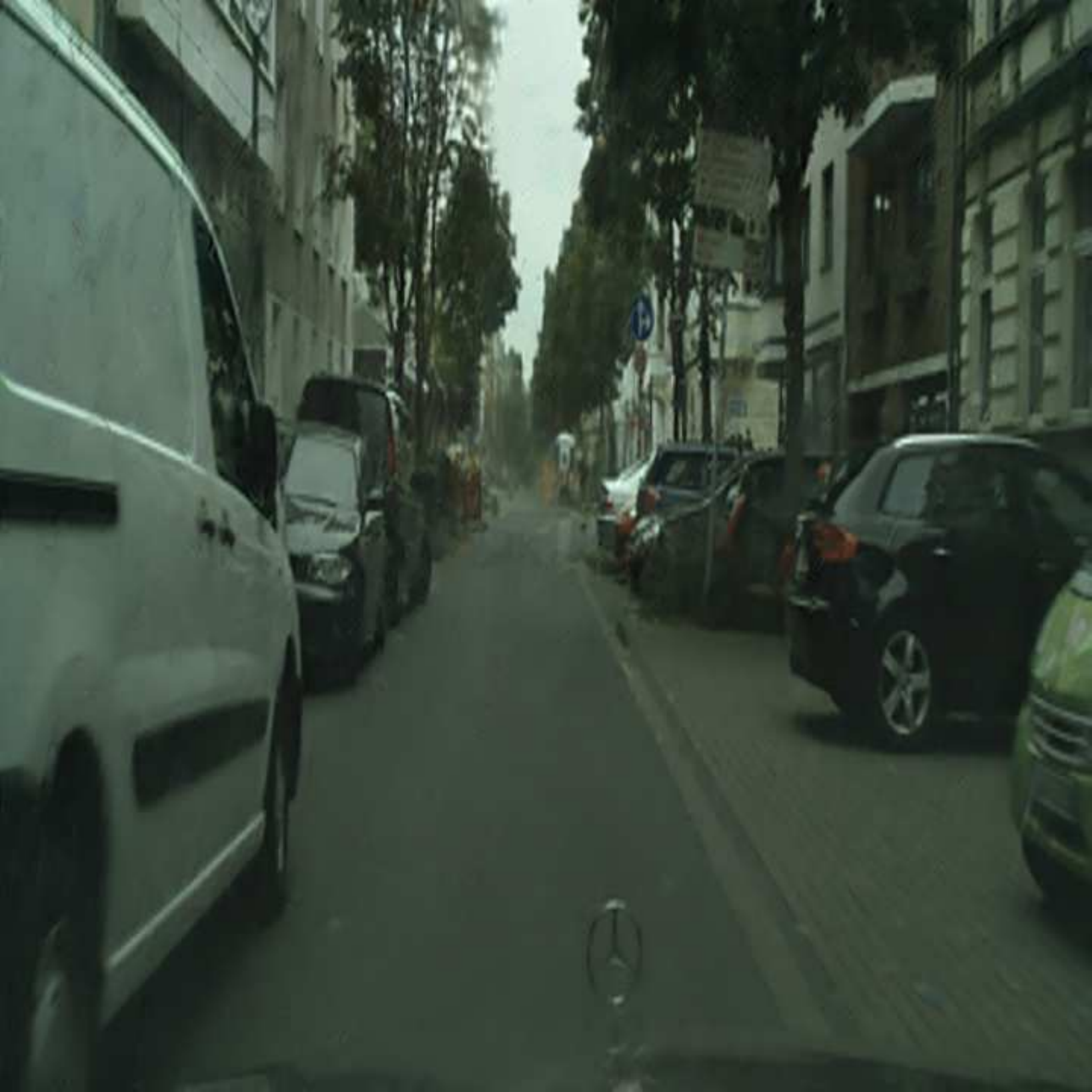}\vspace{2pt}
\end{minipage}}
\caption{Qualitative evaluation on RainCityscapes++. From left to right: input image, ground truth (GT), and results of Li et al. \cite{li2017aod}, Ren et al. \cite{ren2016single}, Eigen et al. \cite{eigen2013restoring}, Wang et al. \cite{wang2019spatial}, Qian et al. \cite{qian2018attentive}, Hu et al. \cite{hu2019depth} and our method, respectively. In MOR, nearly all raindrops, rain streaks and rainy haze are removed by our method despite the diversity of their colors, shapes and transparency, while others are commonly not.}
\label{fig:f1}
\end{figure*}

\begin{figure*}
\centering
\subfigure[Input image]{
\begin{minipage}[b]{0.11\linewidth}
\includegraphics[width=1\linewidth]{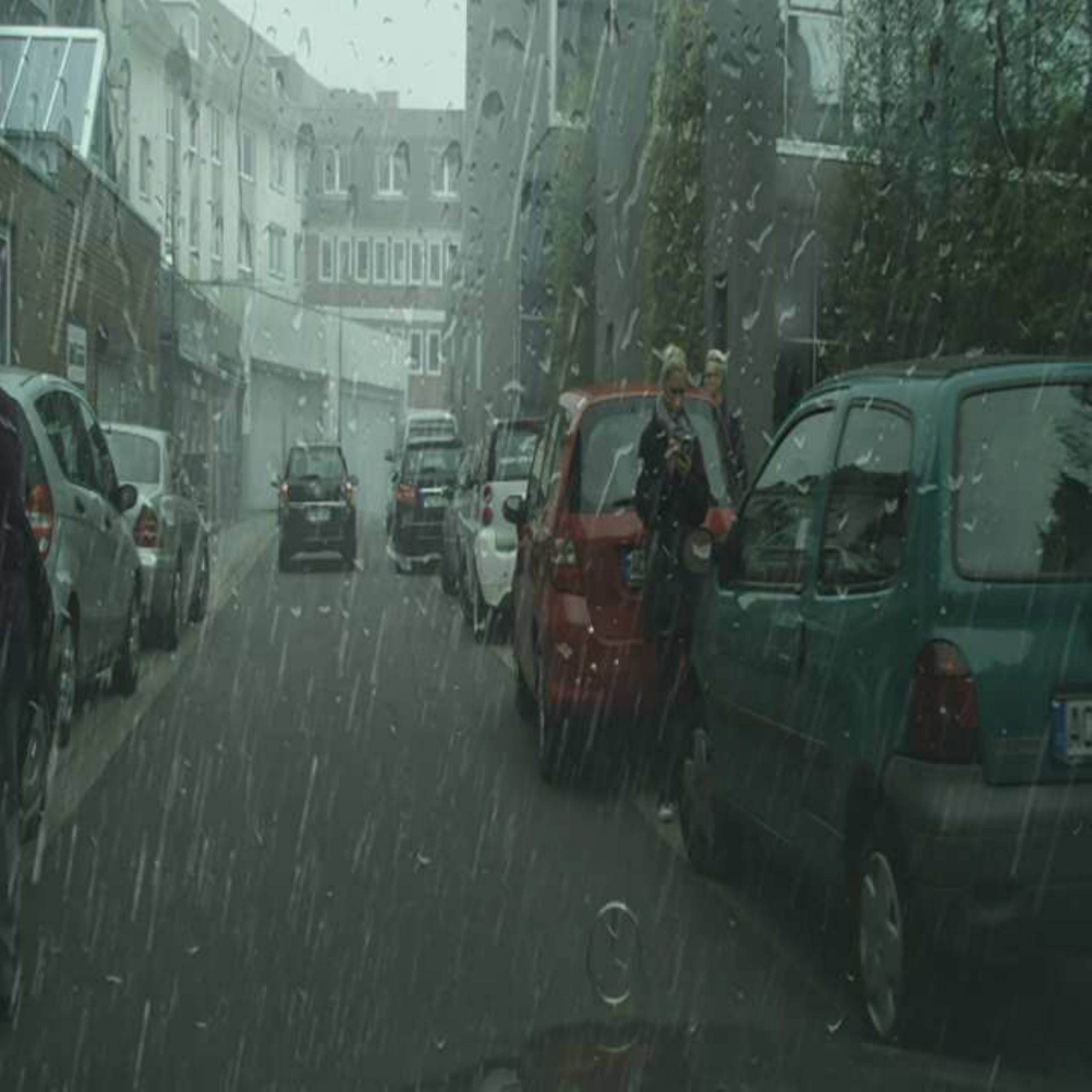}\vspace{2pt}
\includegraphics[width=1\linewidth]{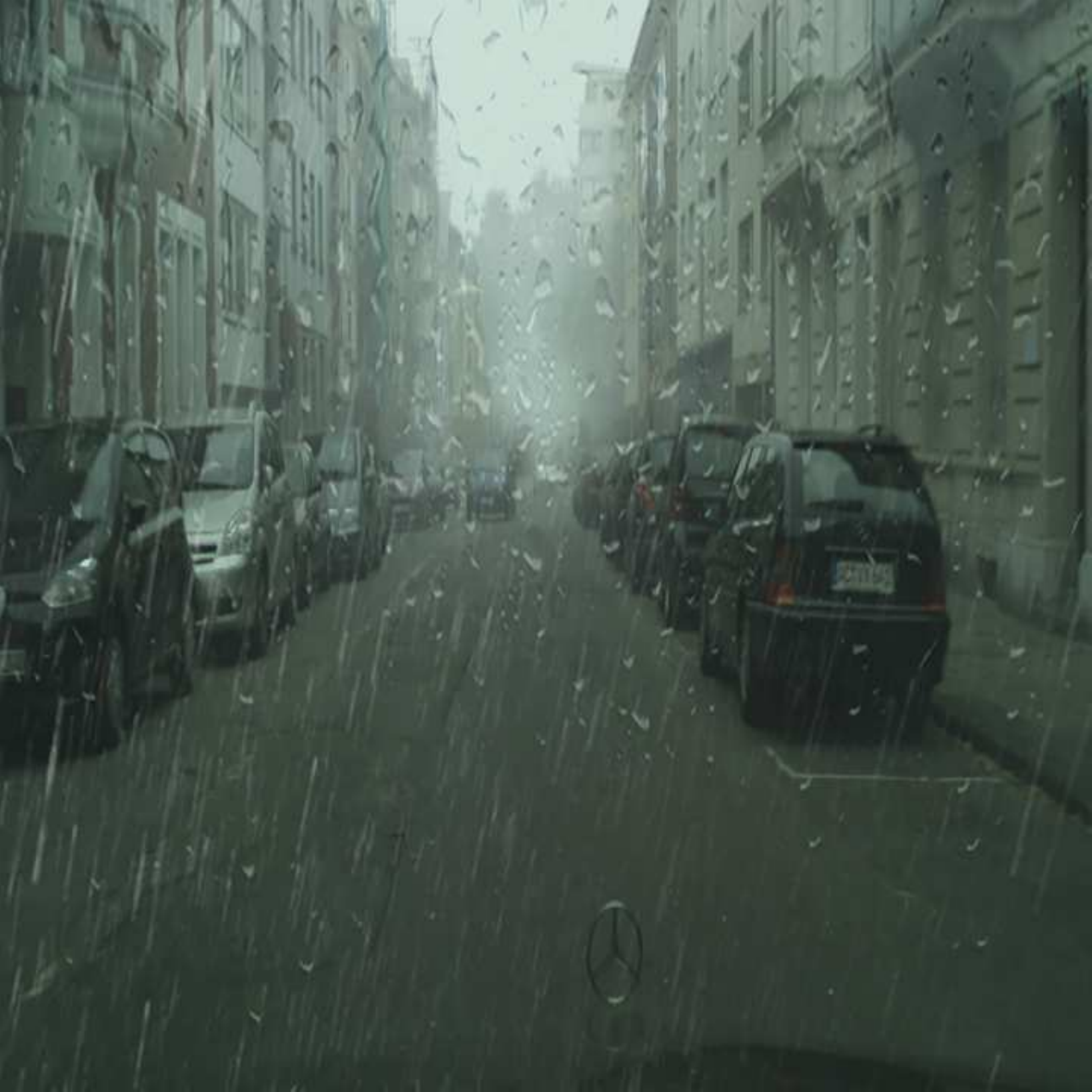}\vspace{2pt}
\includegraphics[width=1\linewidth]{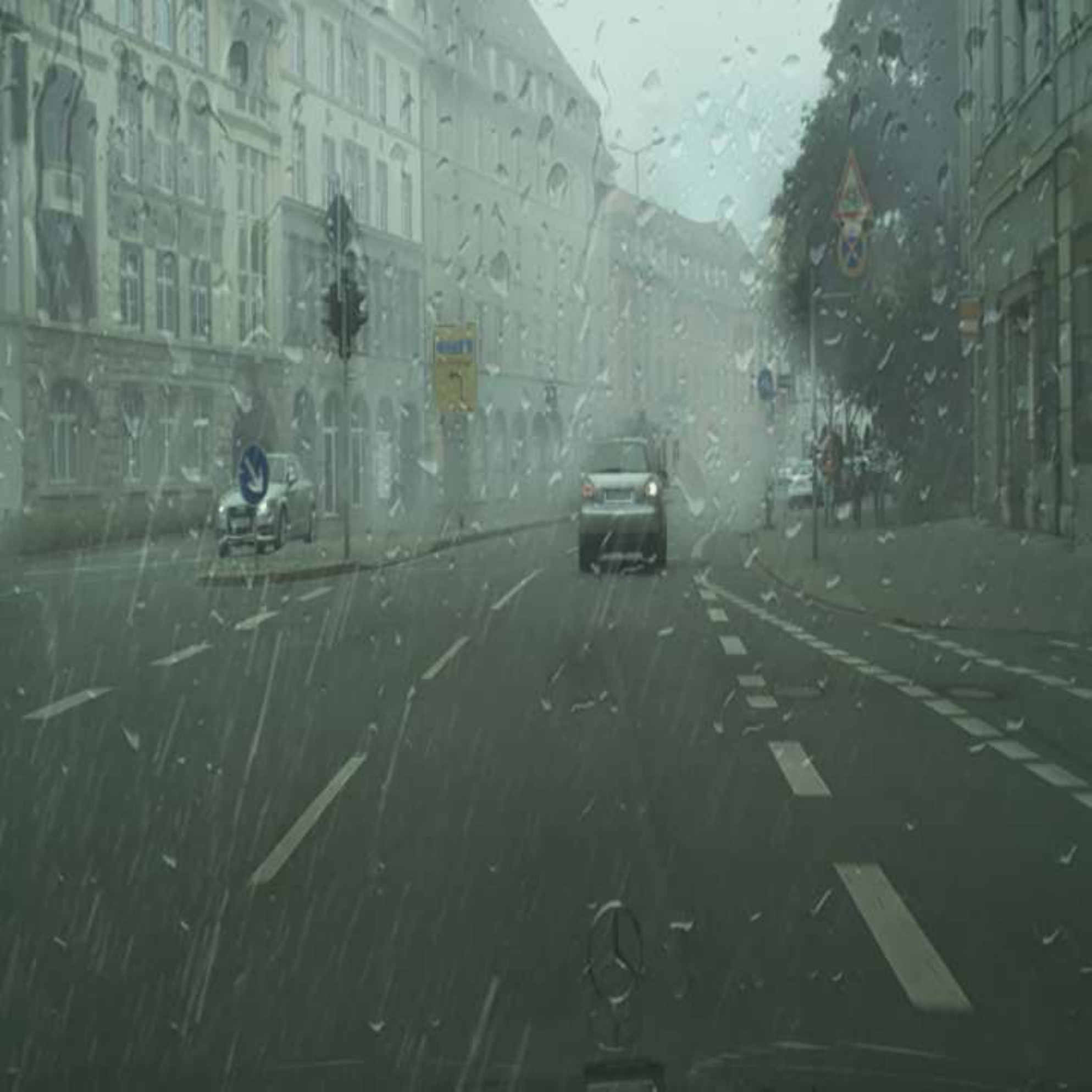}\vspace{2pt}
\end{minipage}}
\subfigure[GT]{
\begin{minipage}[b]{0.11\linewidth}
\includegraphics[width=1\linewidth]{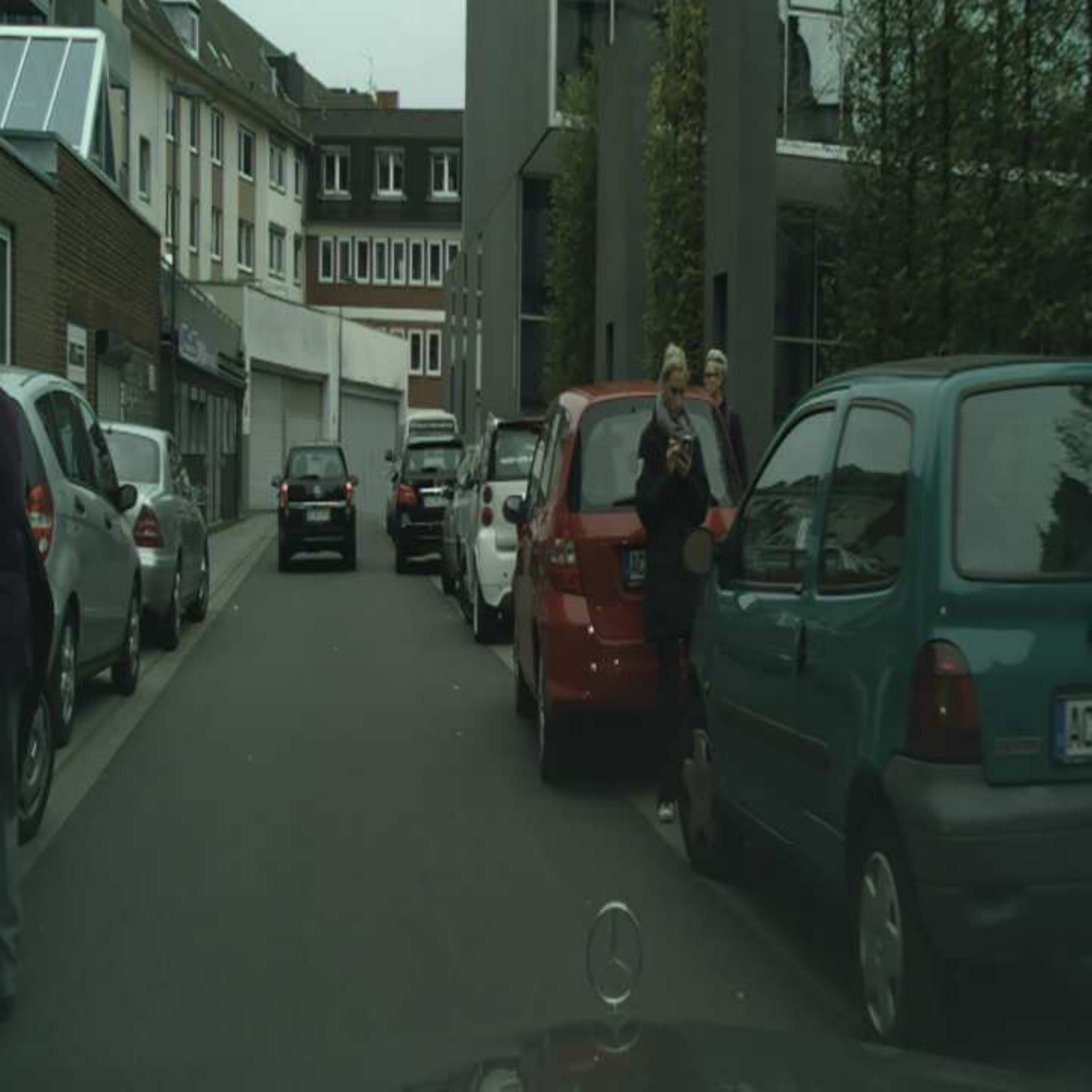}\vspace{2pt}
\includegraphics[width=1\linewidth]{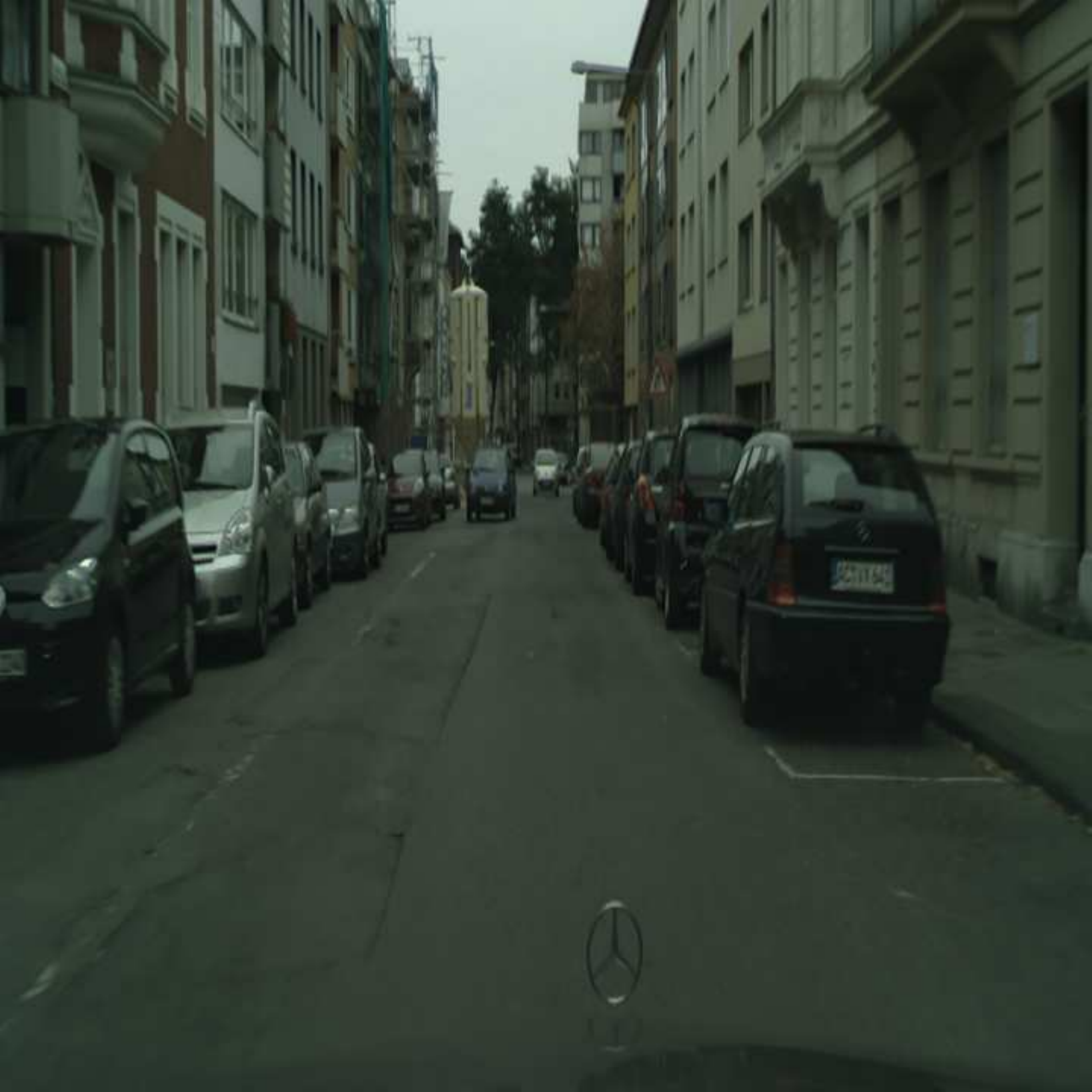}\vspace{2pt}
\includegraphics[width=1\linewidth]{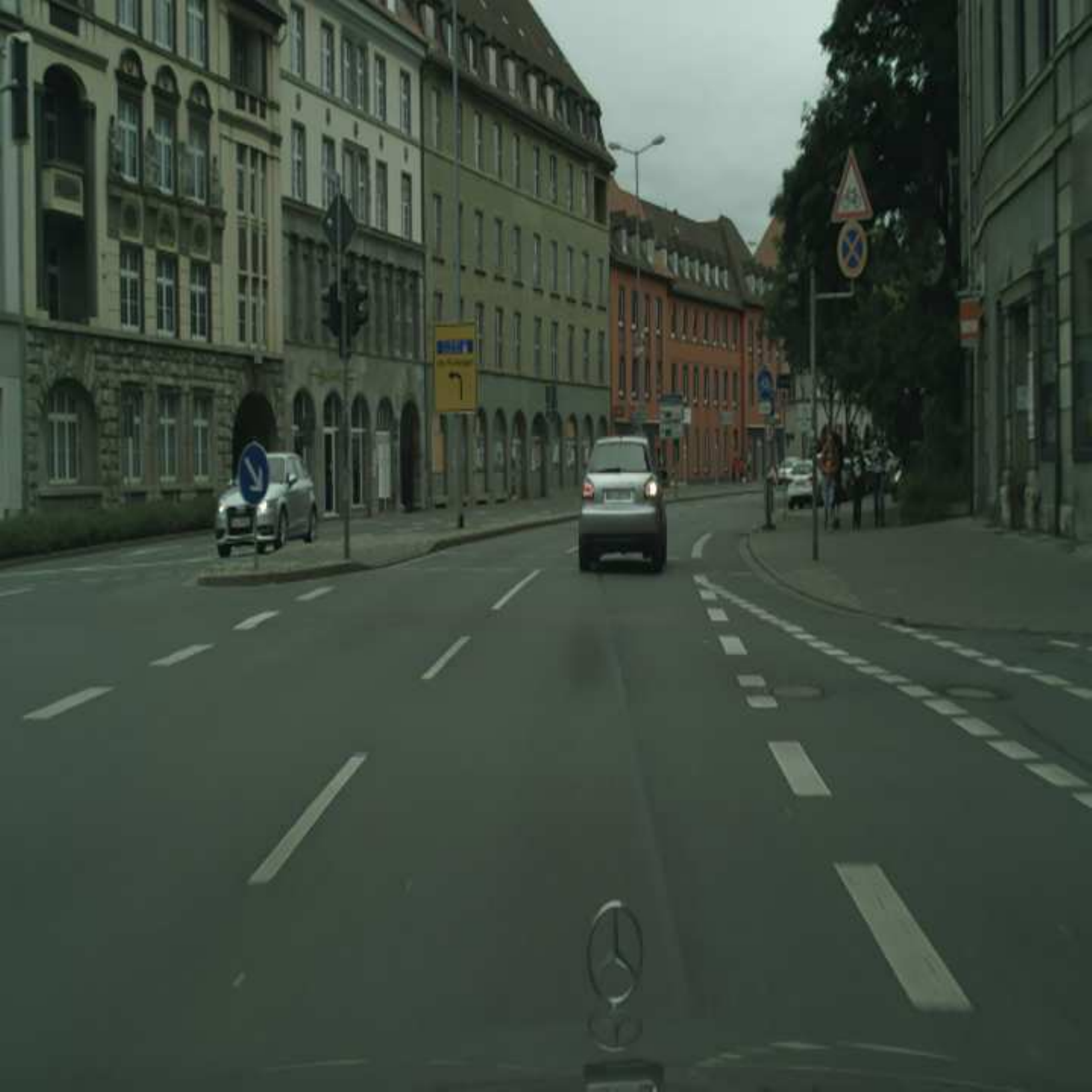}\vspace{2pt}
\end{minipage}}
\subfigure[Li+Qian]{
\begin{minipage}[b]{0.11\linewidth}
\includegraphics[width=1\linewidth]{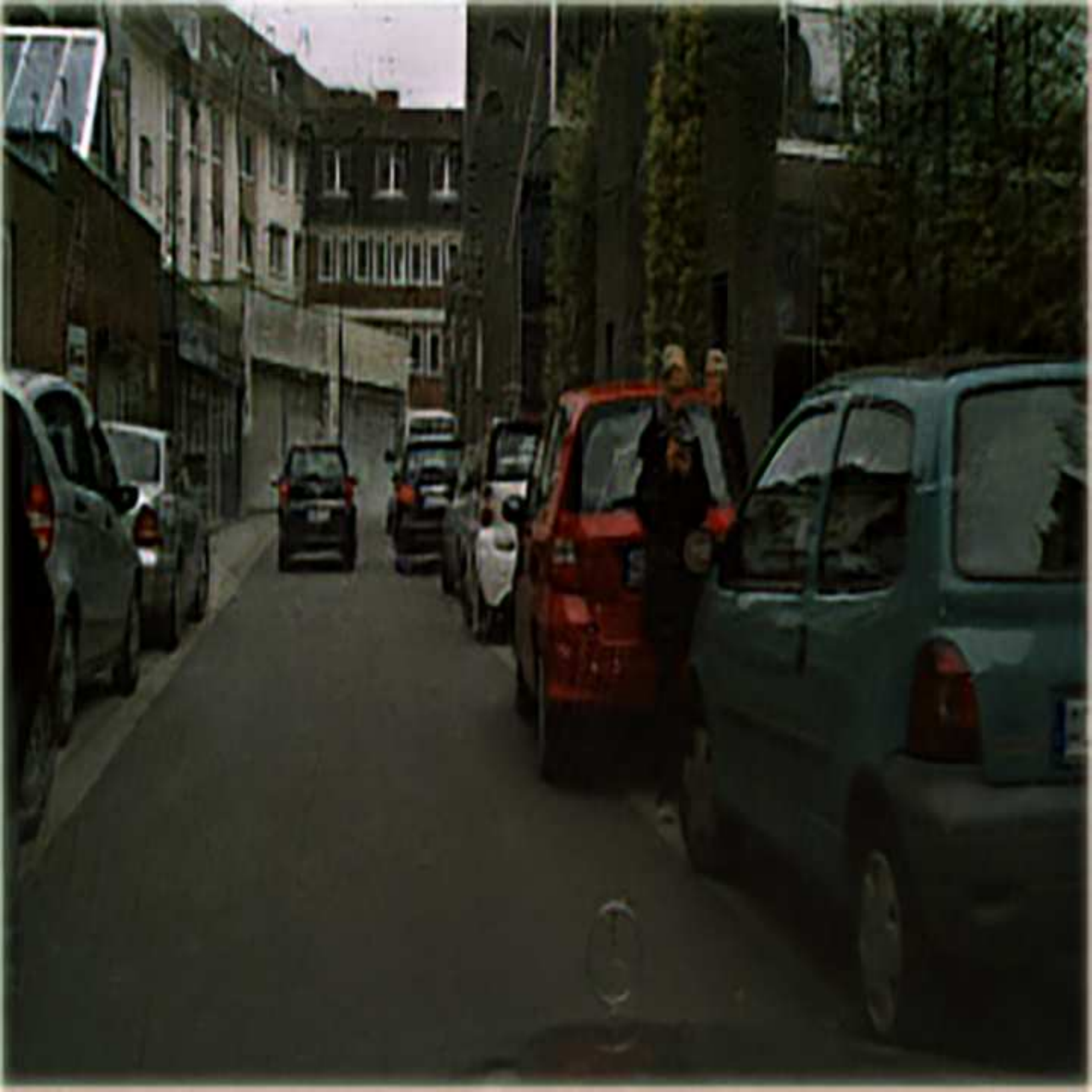}\vspace{2pt}
\includegraphics[width=1\linewidth]{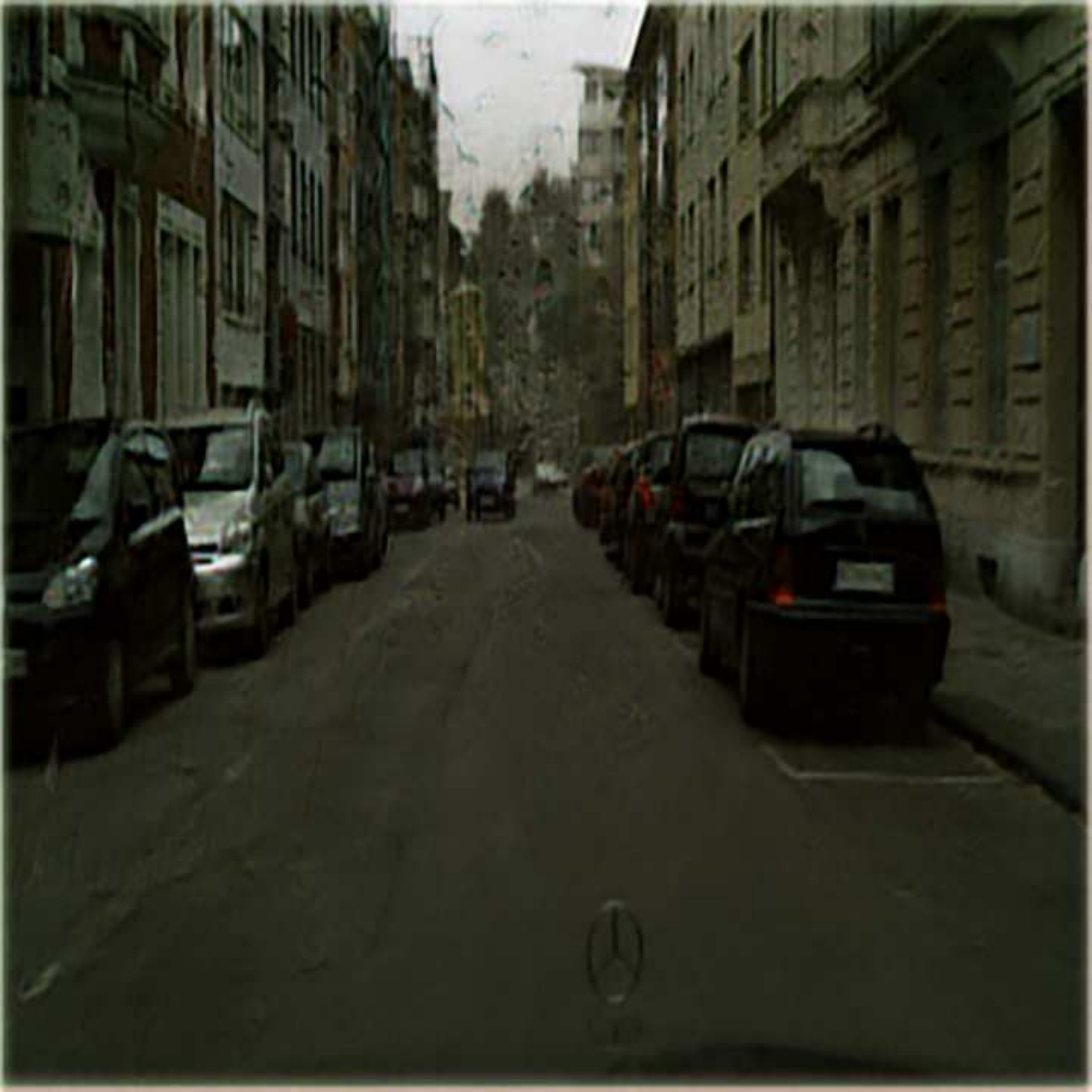}\vspace{2pt}
\includegraphics[width=1\linewidth]{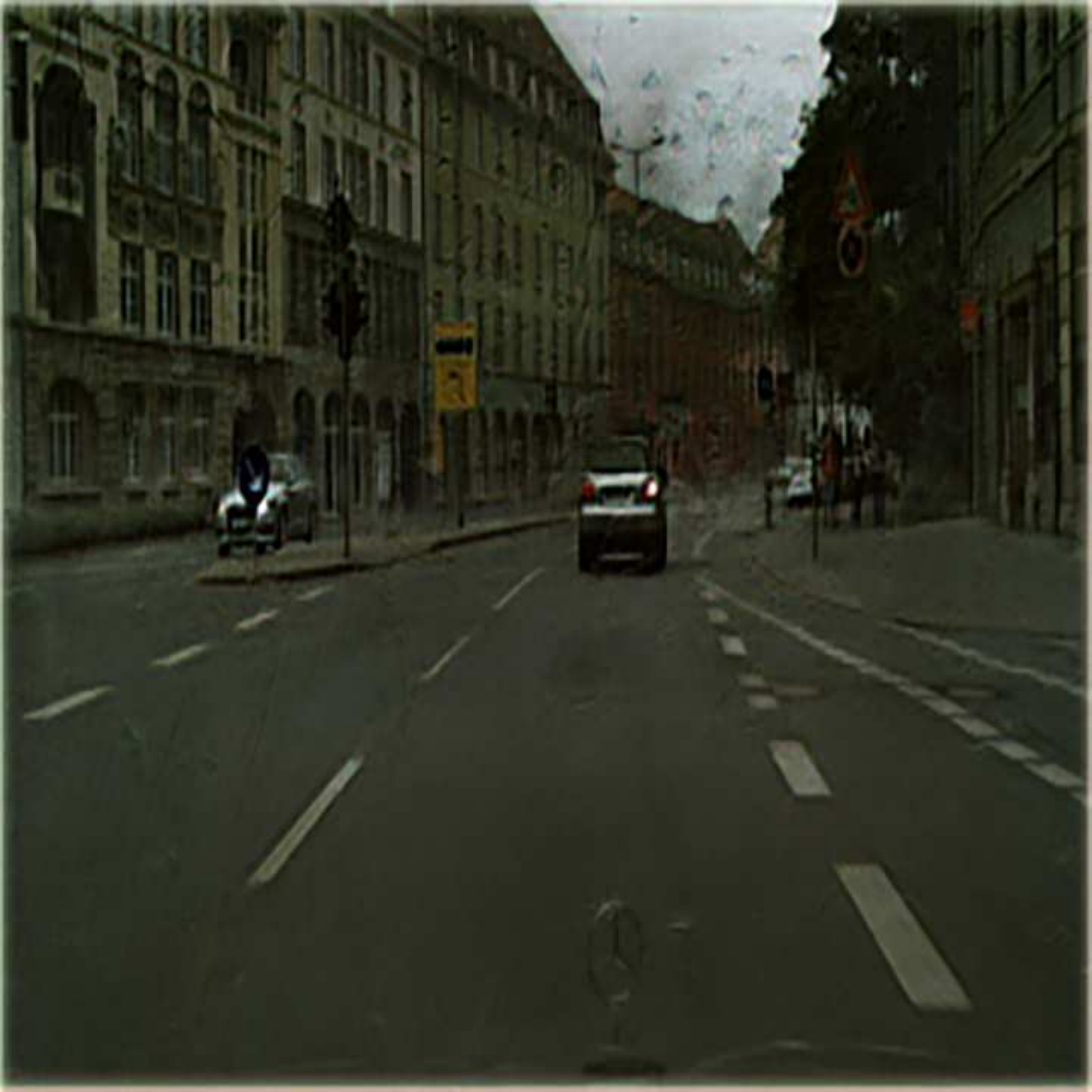}\vspace{2pt}
\end{minipage}}
\subfigure[Ren+Qian]{
\begin{minipage}[b]{0.11\linewidth}
\includegraphics[width=1\linewidth]{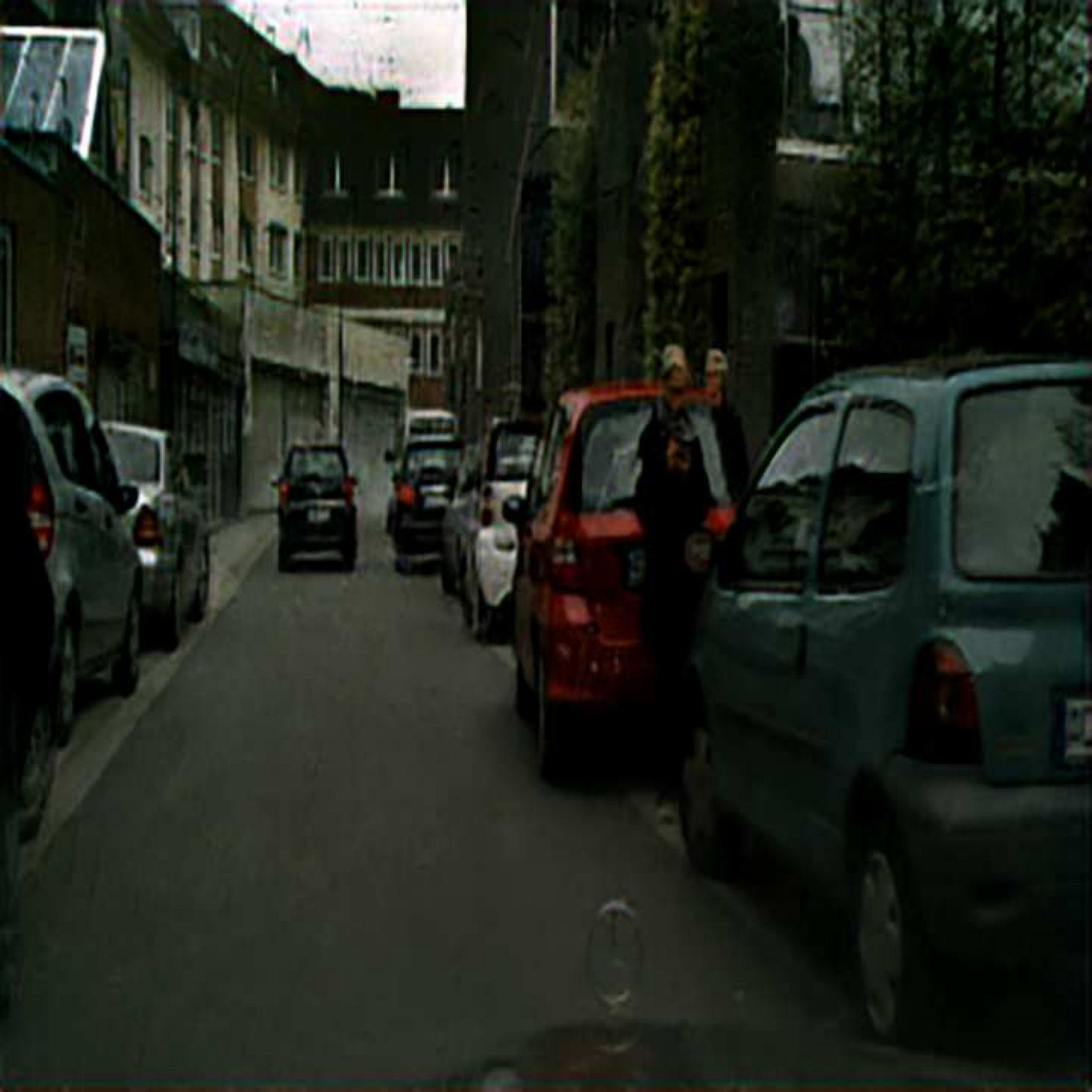}\vspace{2pt}
\includegraphics[width=1\linewidth]{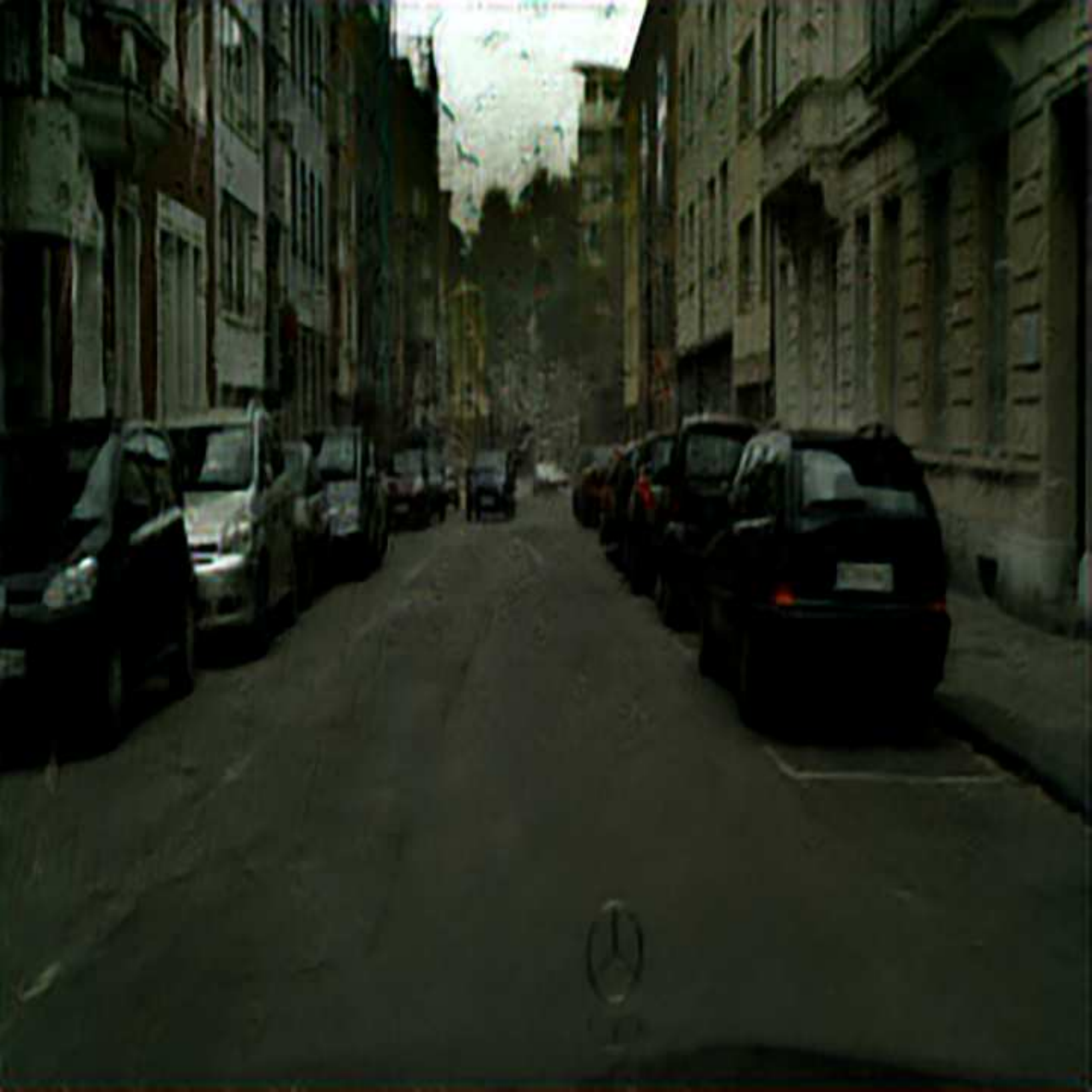}\vspace{2pt}
\includegraphics[width=1\linewidth]{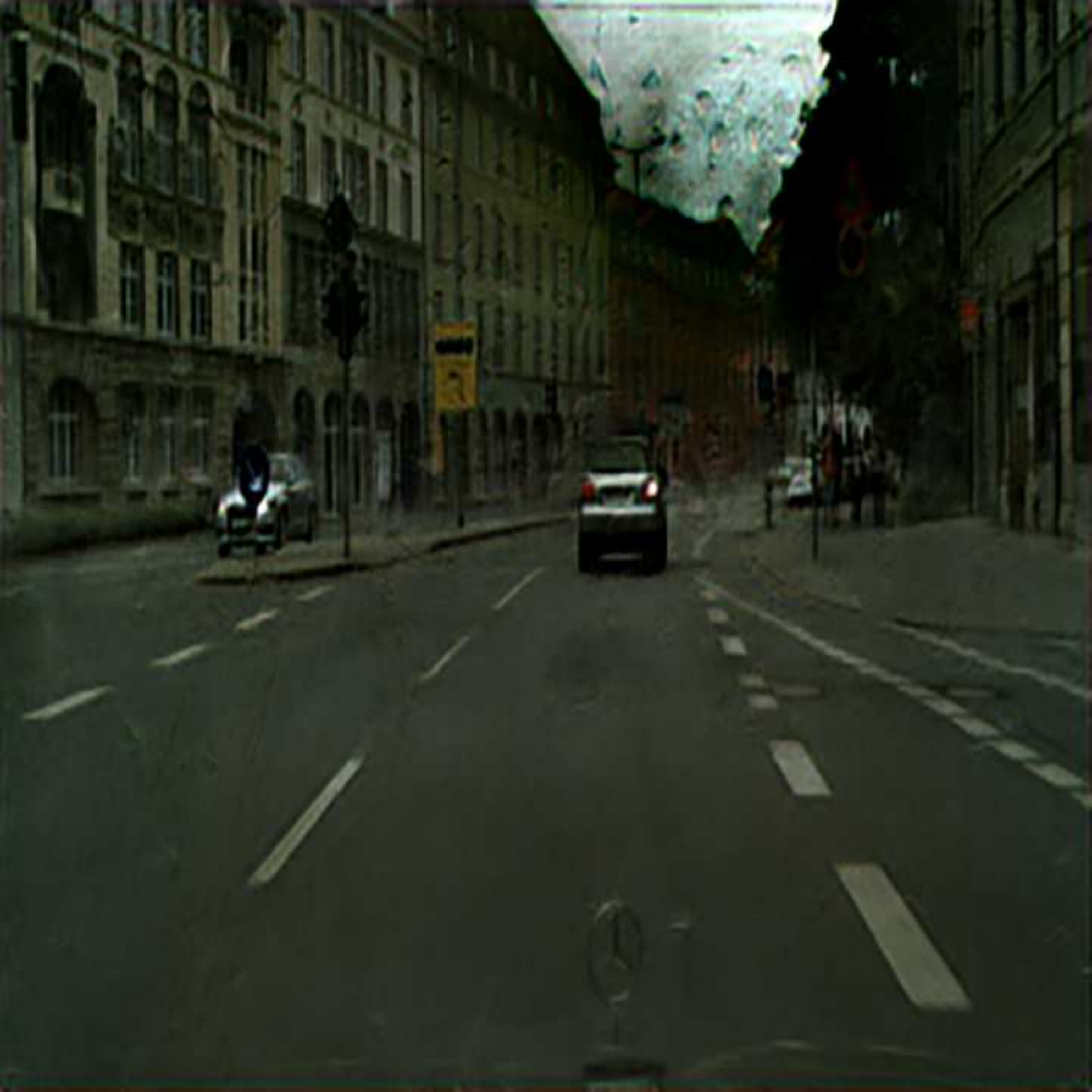}\vspace{2pt}
\end{minipage}}
\subfigure[Eigen+Ren]{
\begin{minipage}[b]{0.11\linewidth}
\includegraphics[width=1\linewidth]{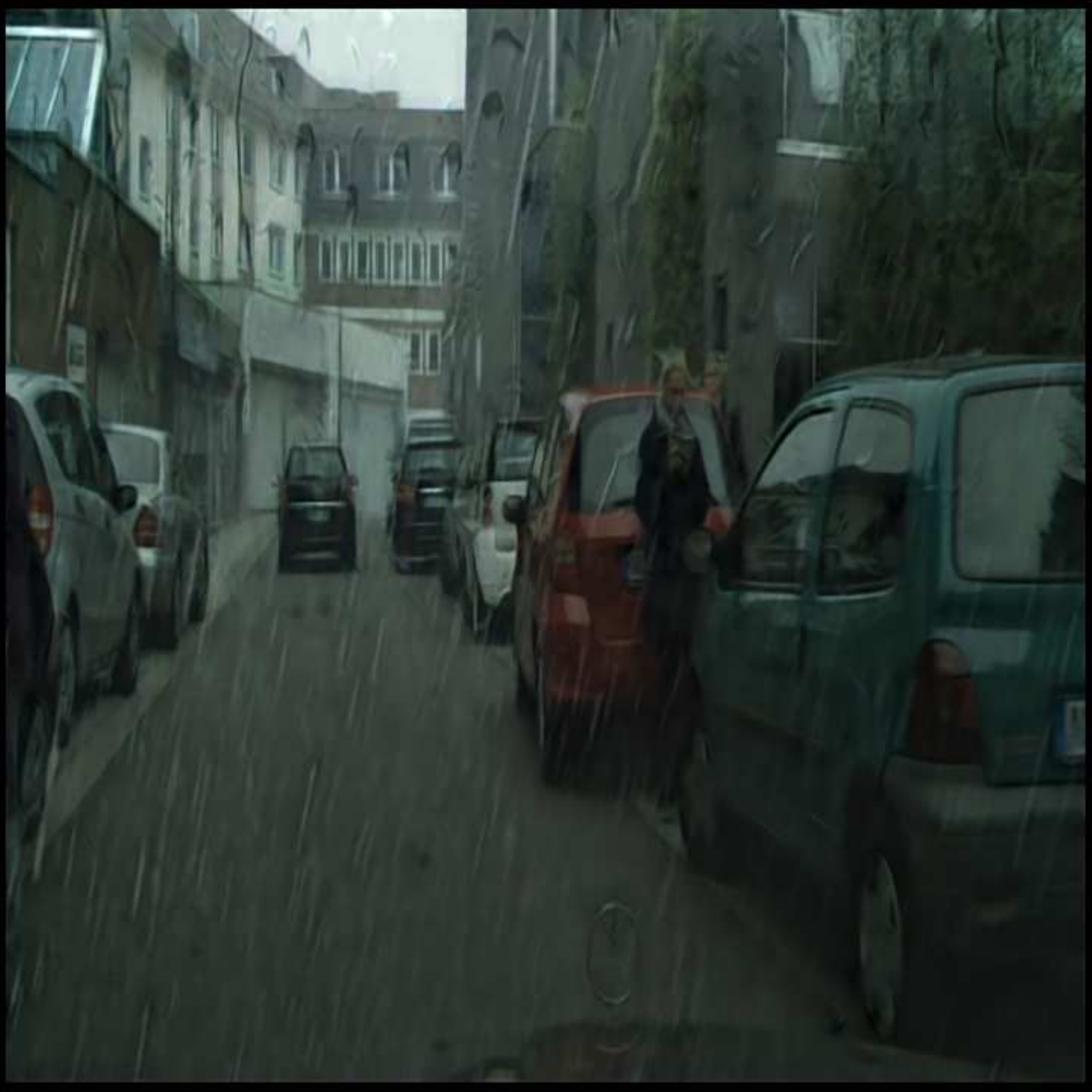}\vspace{2pt}
\includegraphics[width=1\linewidth]{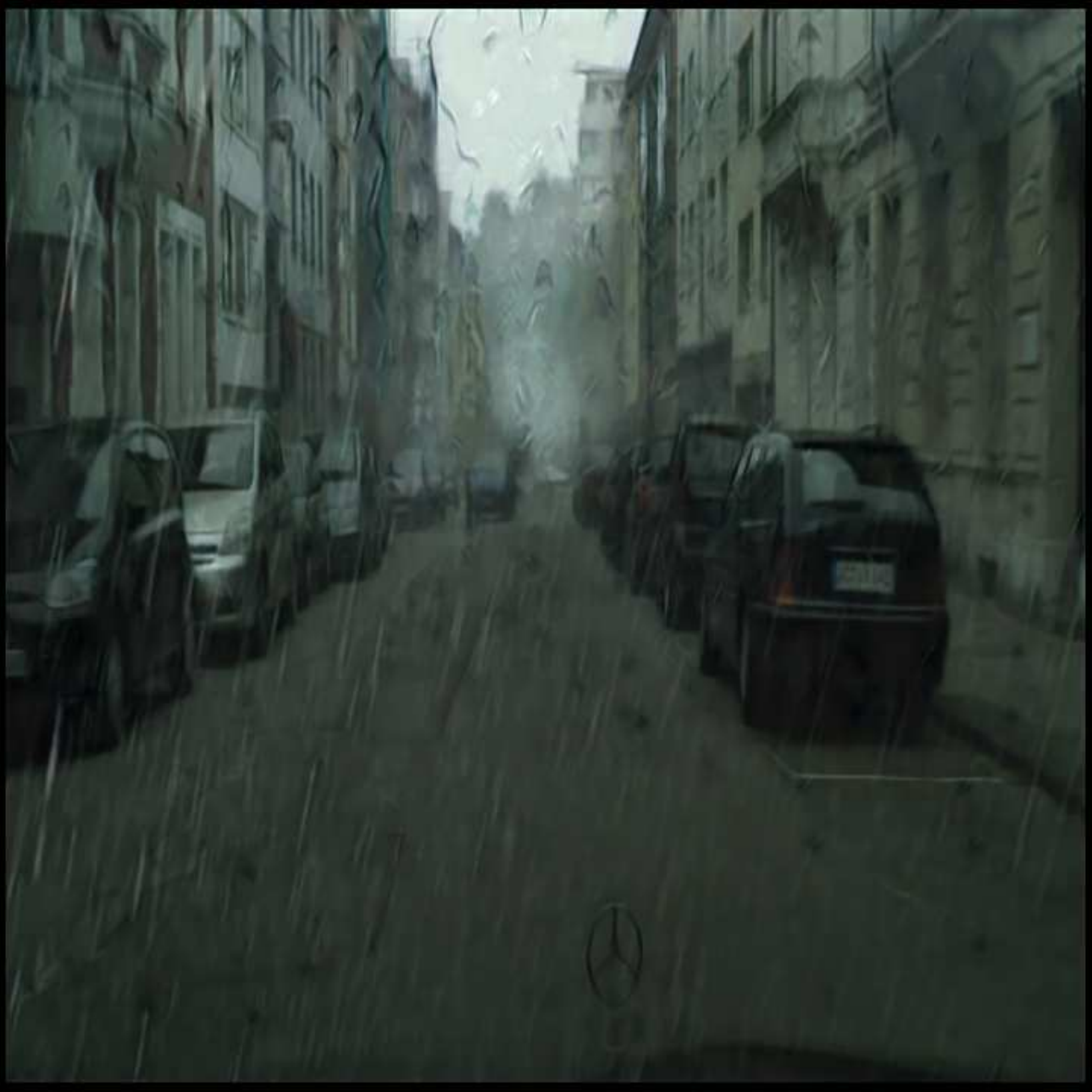}\vspace{2pt}
\includegraphics[width=1\linewidth]{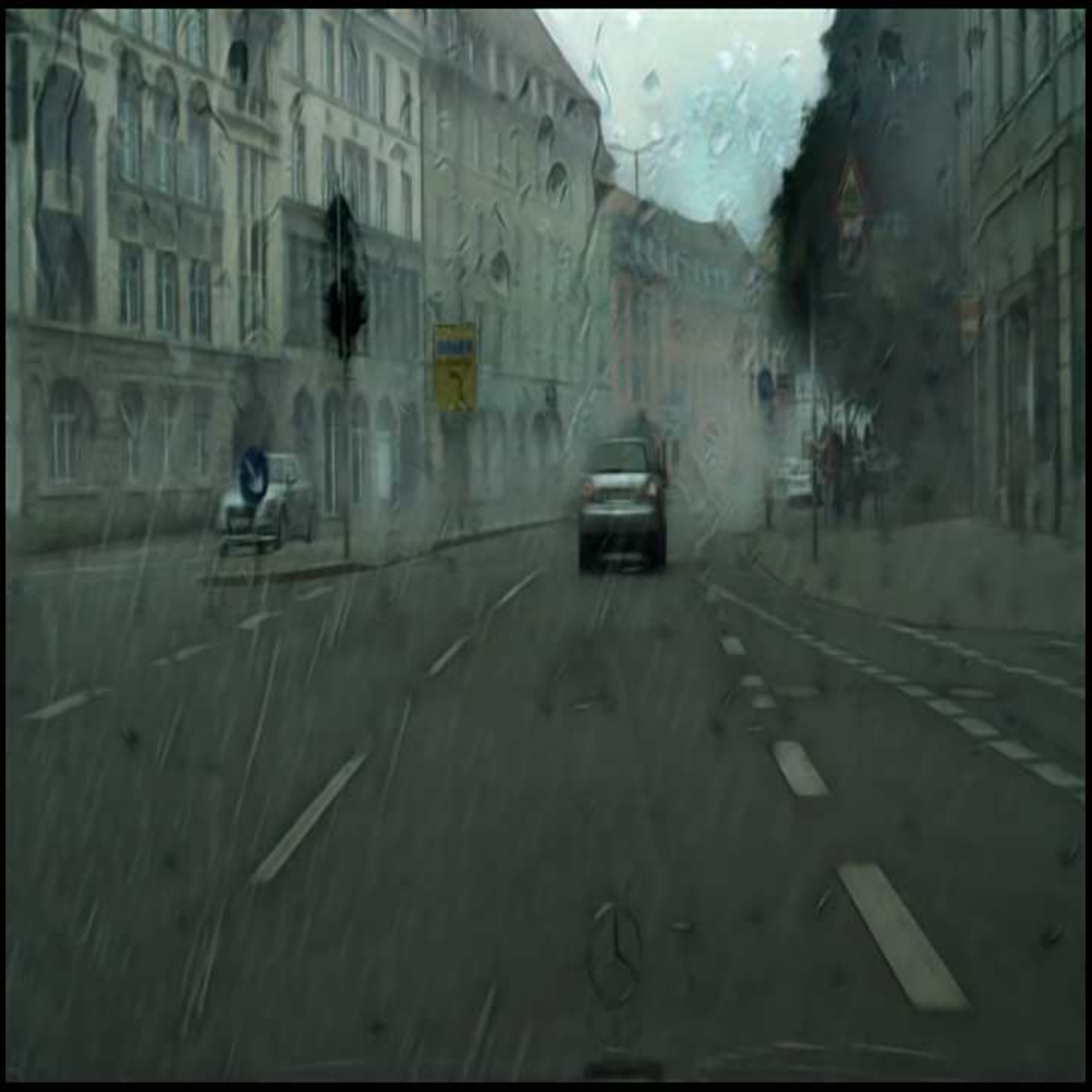}\vspace{2pt}
\end{minipage}}
\subfigure[Wang+Li]{
\begin{minipage}[b]{0.11\linewidth}
\includegraphics[width=1\linewidth]{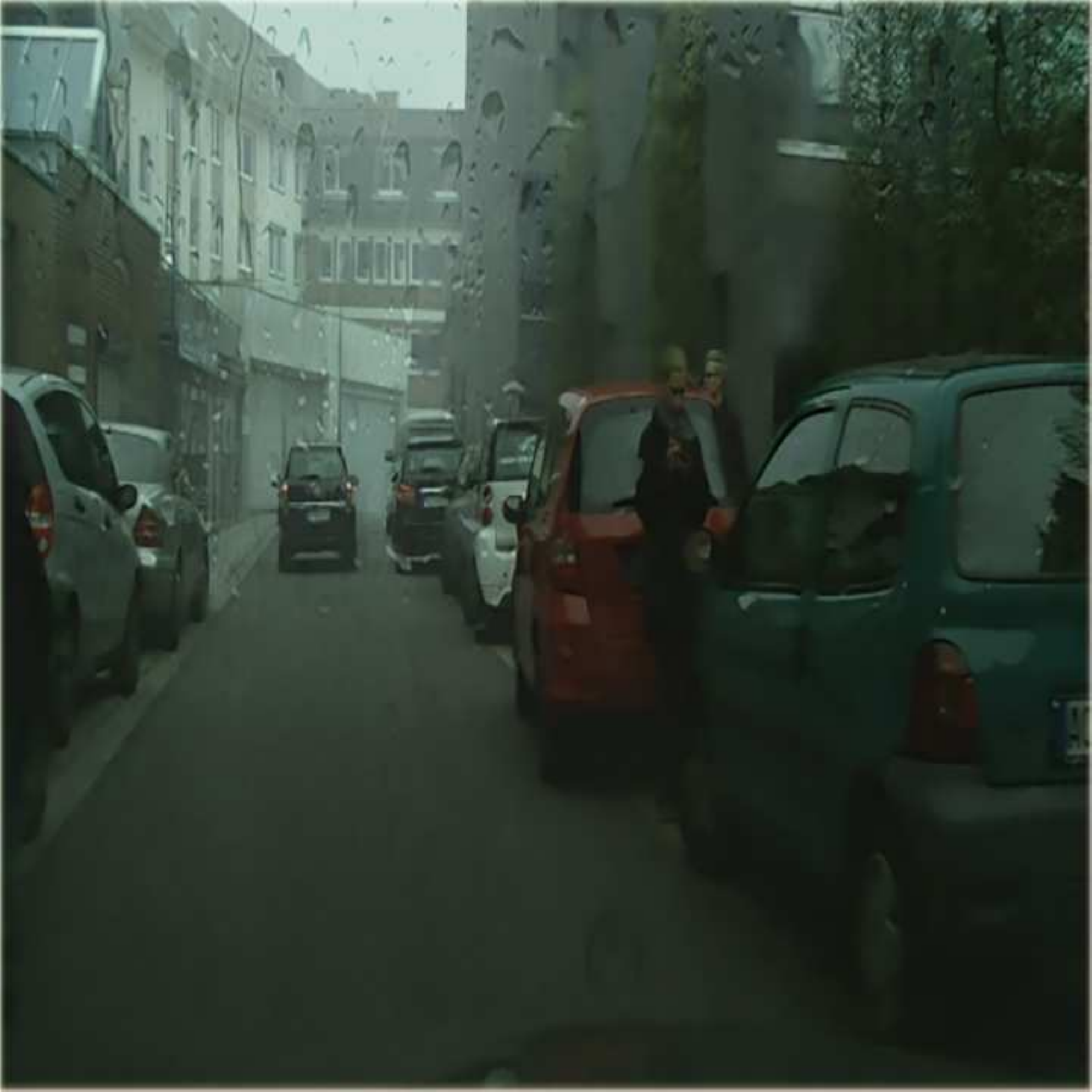}\vspace{2pt}
\includegraphics[width=1\linewidth]{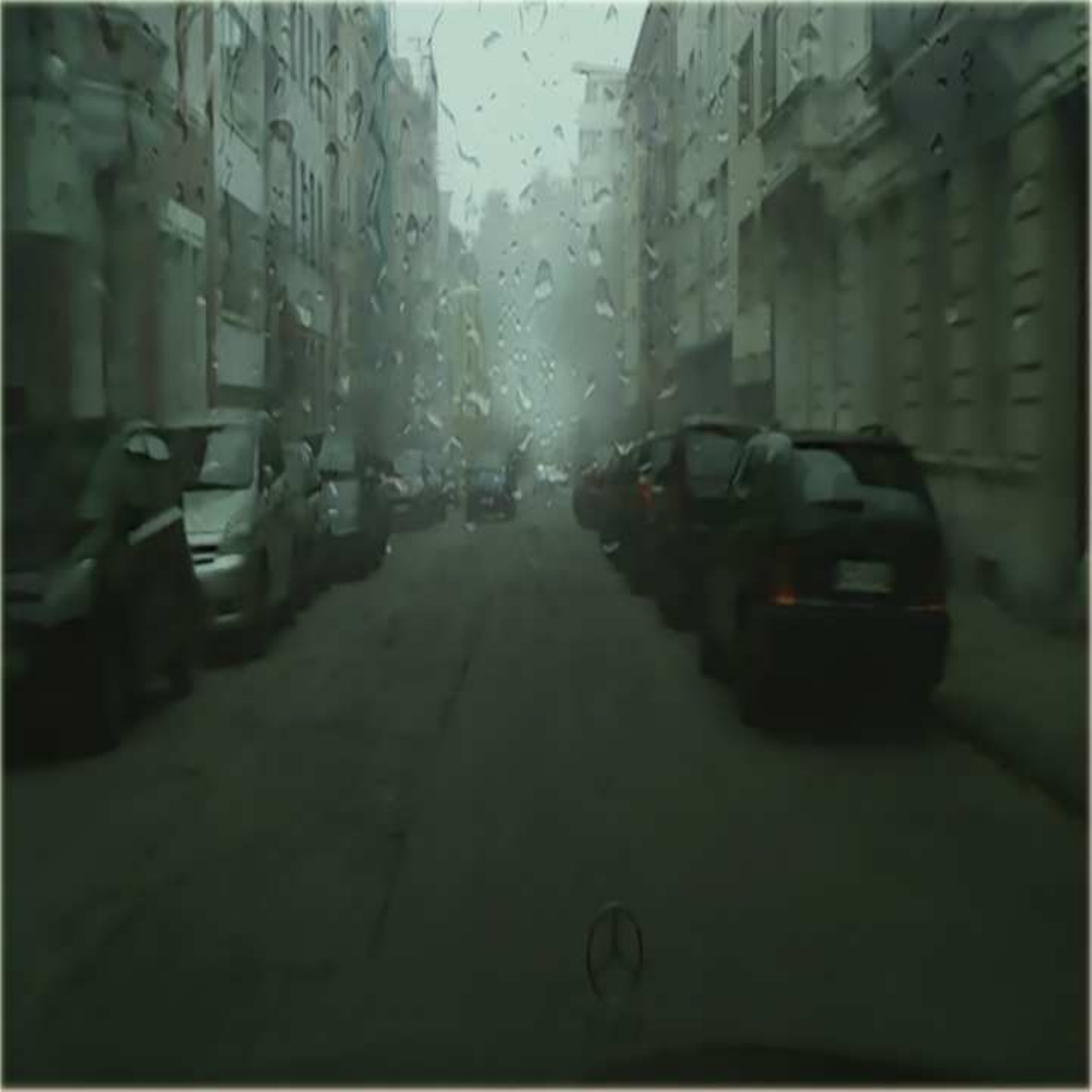}\vspace{2pt}
\includegraphics[width=1\linewidth]{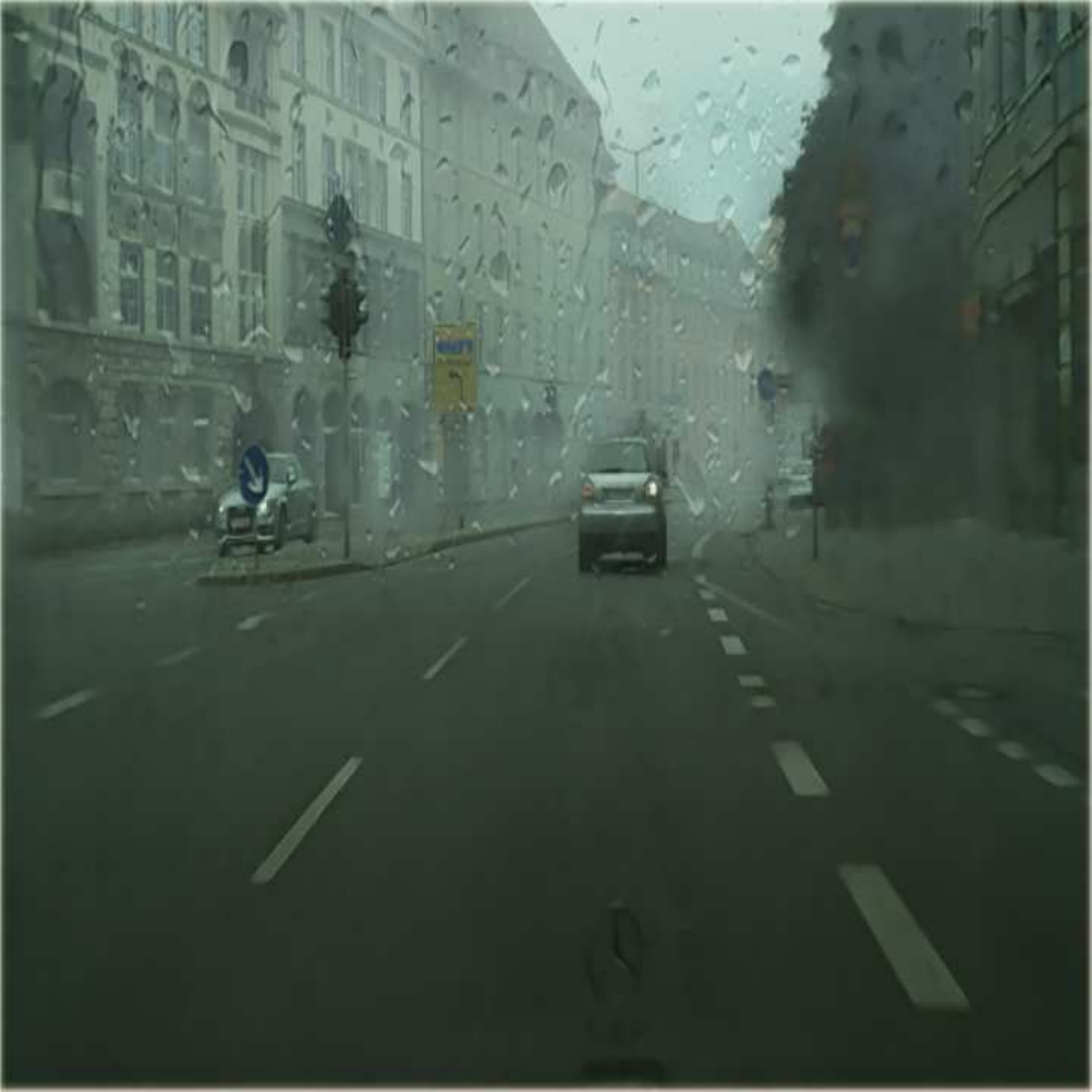}\vspace{2pt}
\end{minipage}}
\subfigure[Hu+Qian]{
\begin{minipage}[b]{0.11\linewidth}
\includegraphics[width=1\linewidth]{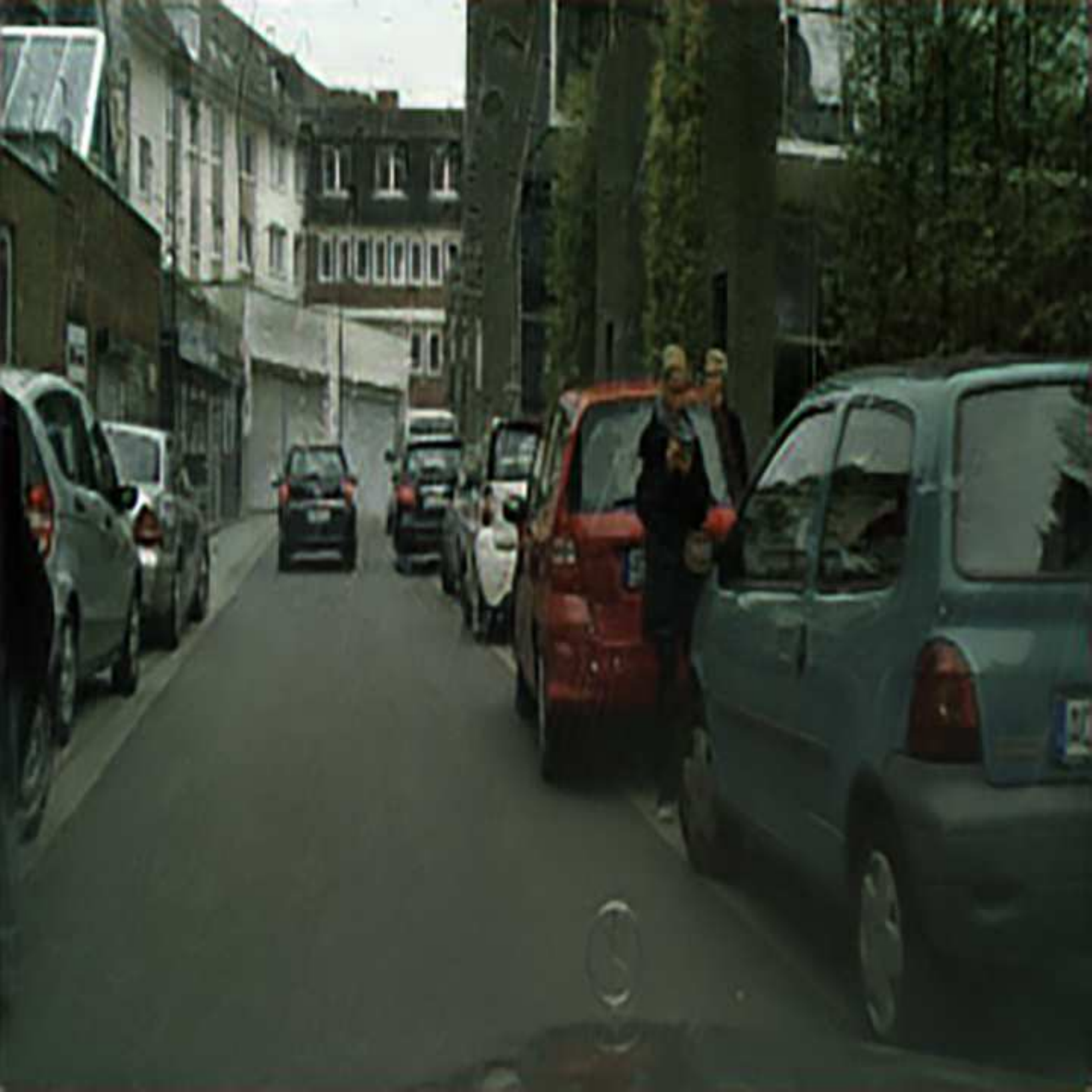}\vspace{2pt}
\includegraphics[width=1\linewidth]{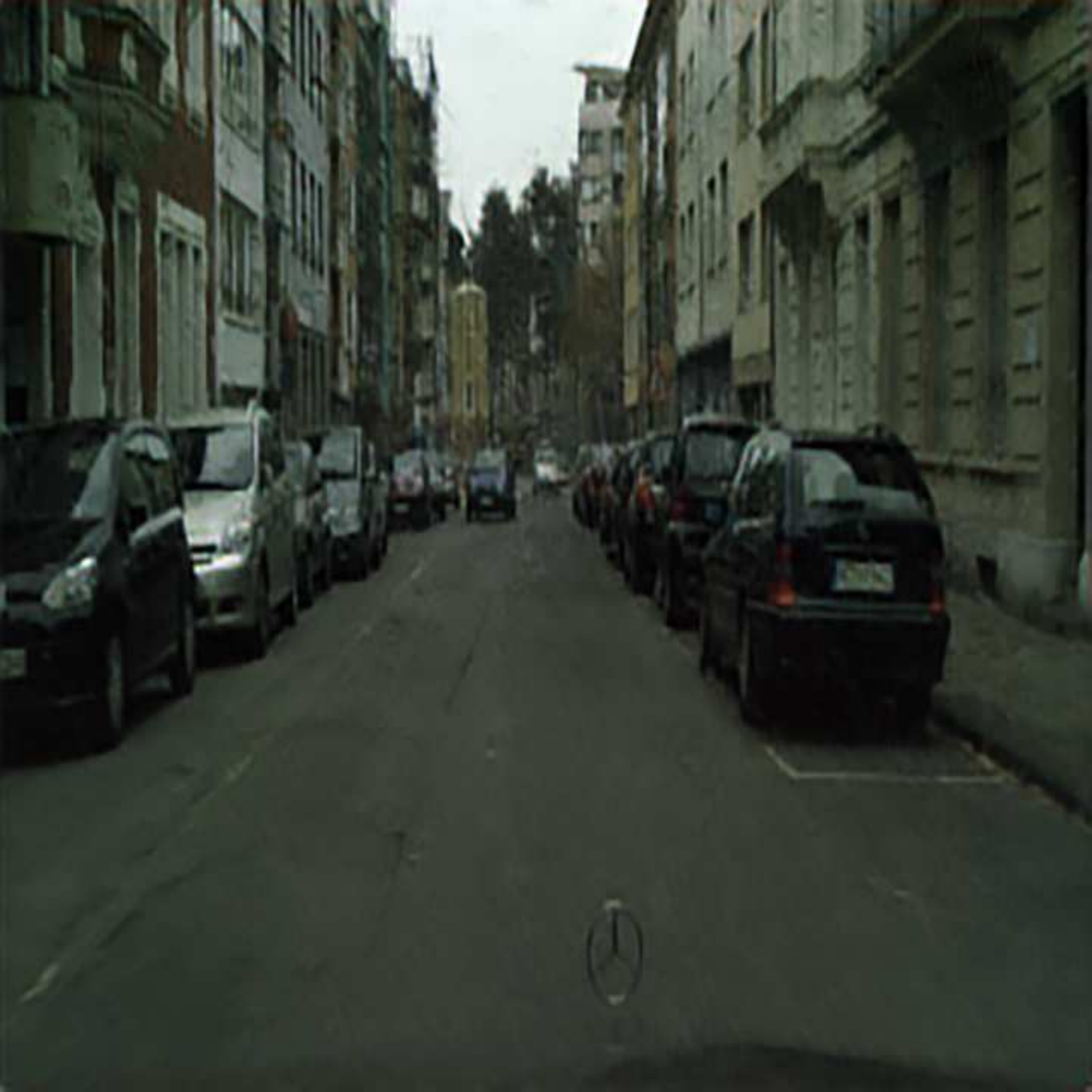}\vspace{2pt}
\includegraphics[width=1\linewidth]{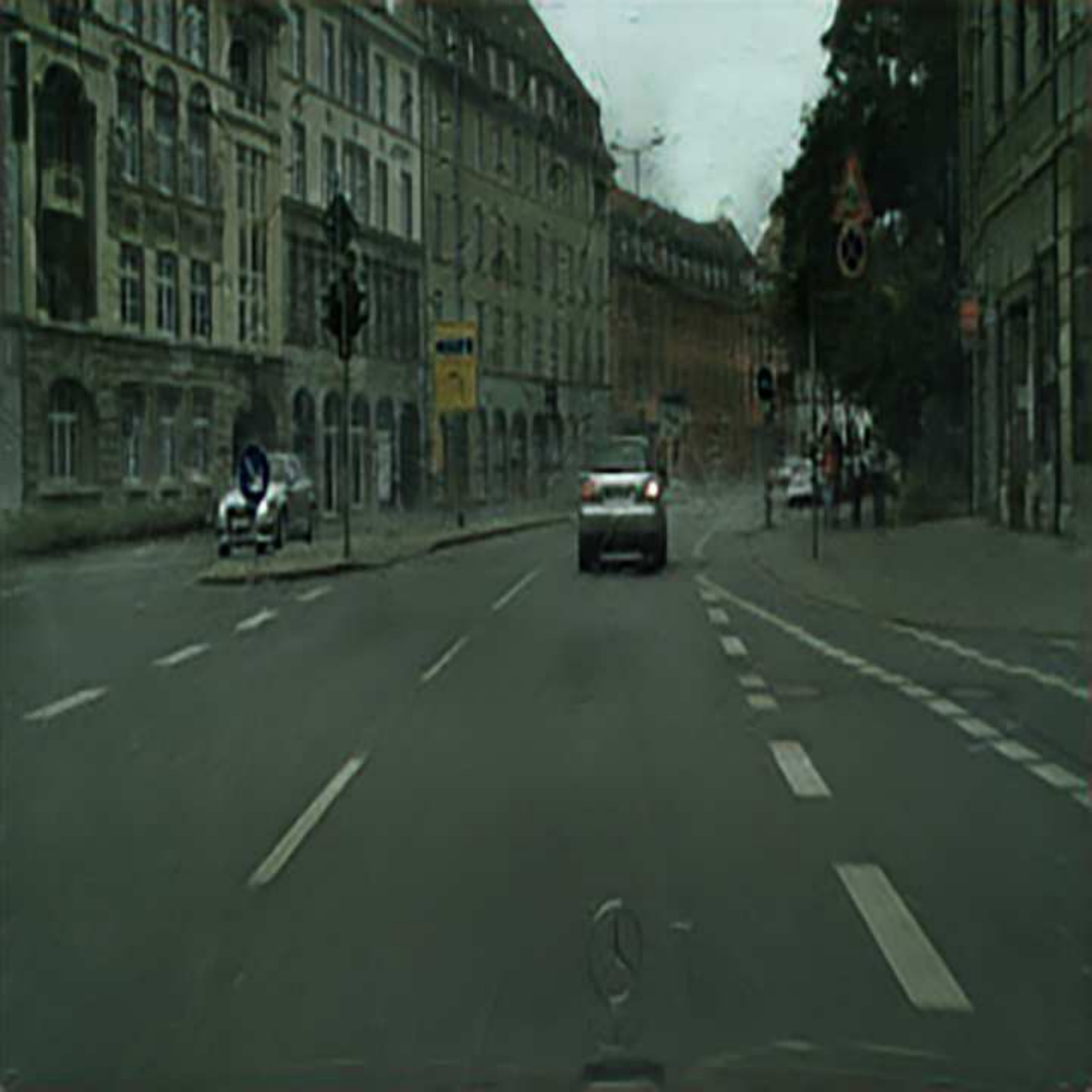}\vspace{2pt}
\end{minipage}}
\subfigure[Ours]{
\begin{minipage}[b]{0.11\linewidth}
\includegraphics[width=1\linewidth]{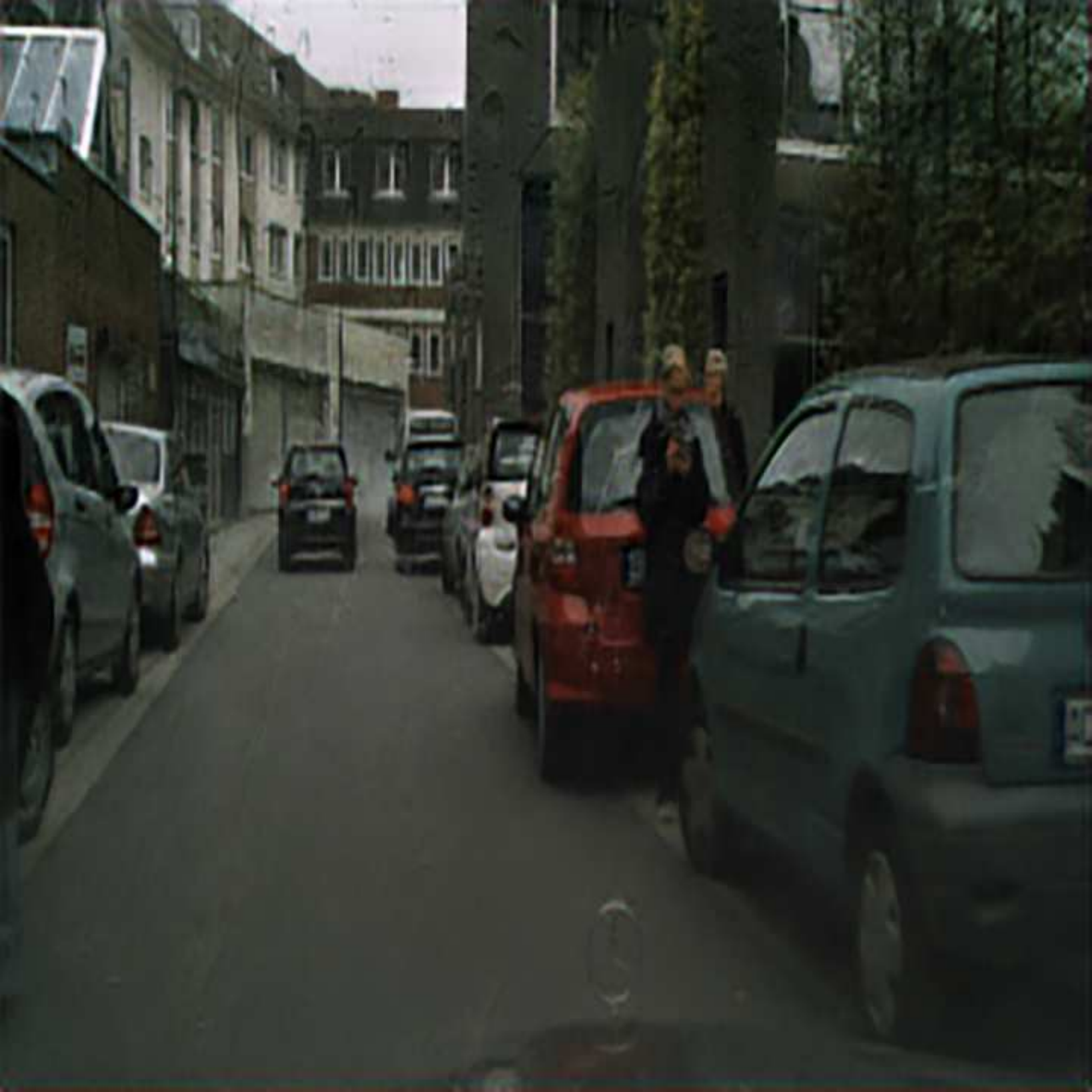}\vspace{2pt}
\includegraphics[width=1\linewidth]{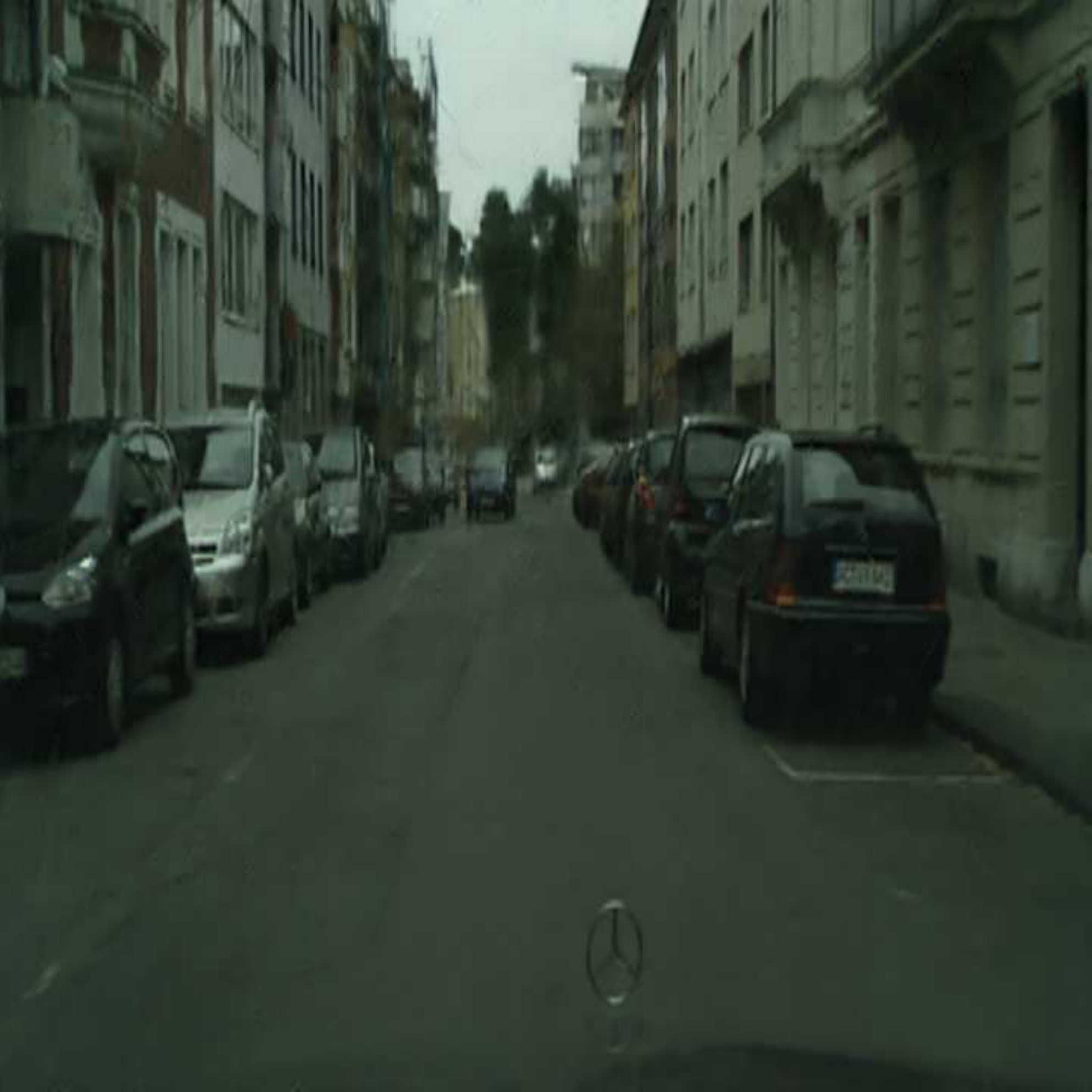}\vspace{2pt}
\includegraphics[width=1\linewidth]{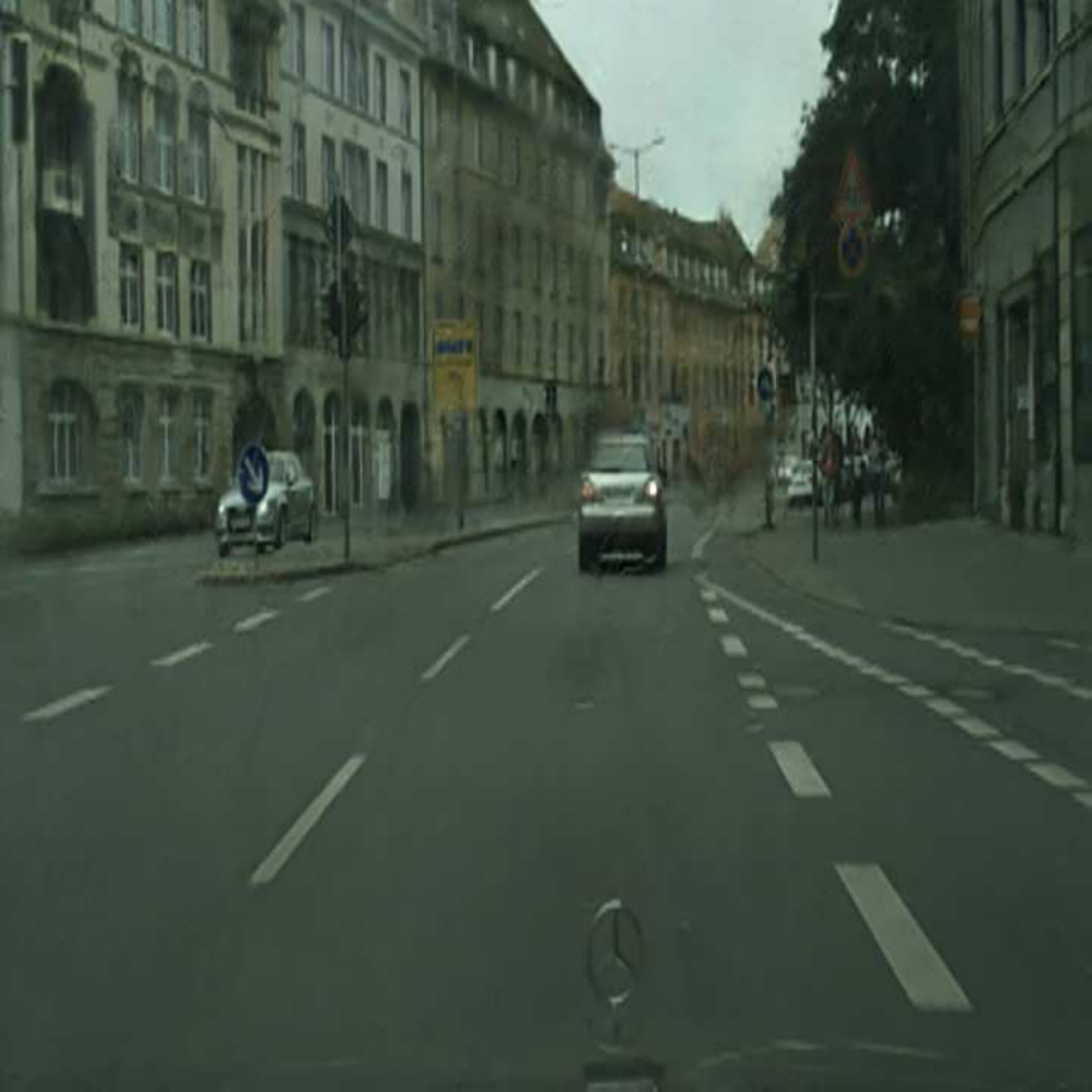}\vspace{2pt}
\end{minipage}}
\caption{Qualitative evaluation of the combination of different methods.} \label{fg:state}
\label{fig:f2}
\end{figure*}

\begin{figure*}
\centering
\subfigure[Input image]{
\begin{minipage}[b]{0.11\linewidth}
\includegraphics[width=1\linewidth]{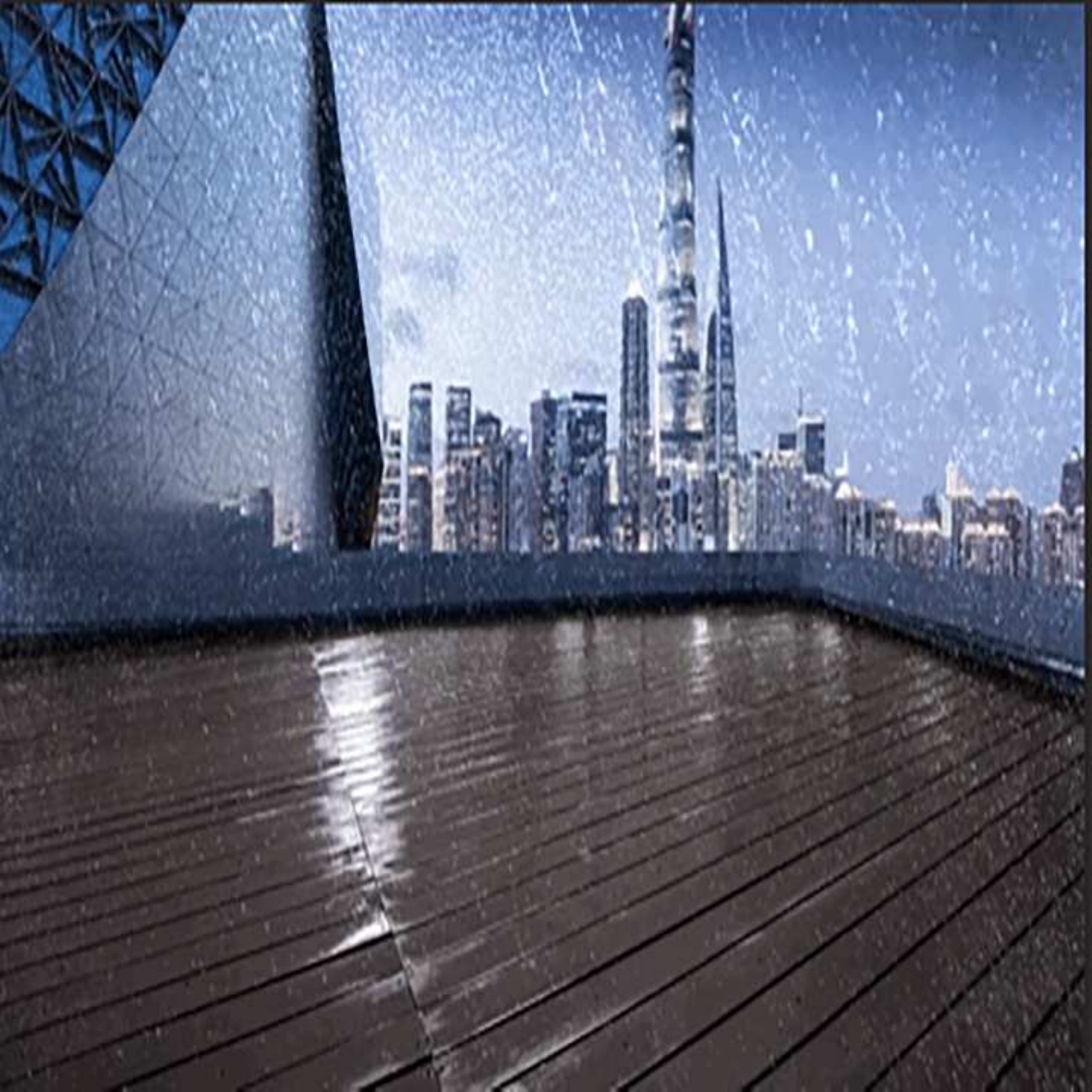}\vspace{2pt}
\includegraphics[width=1\linewidth]{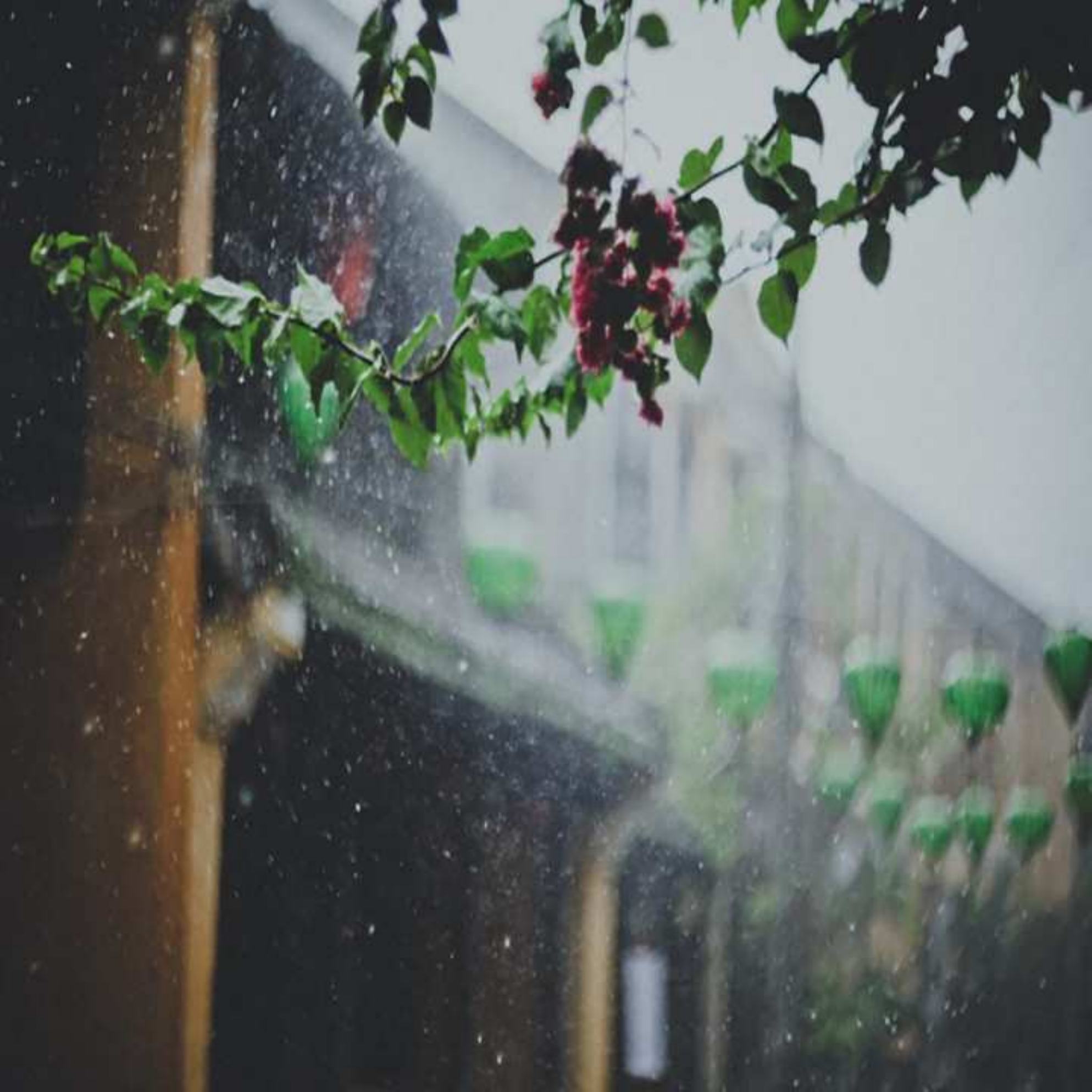}\vspace{2pt}
\includegraphics[width=1\linewidth]{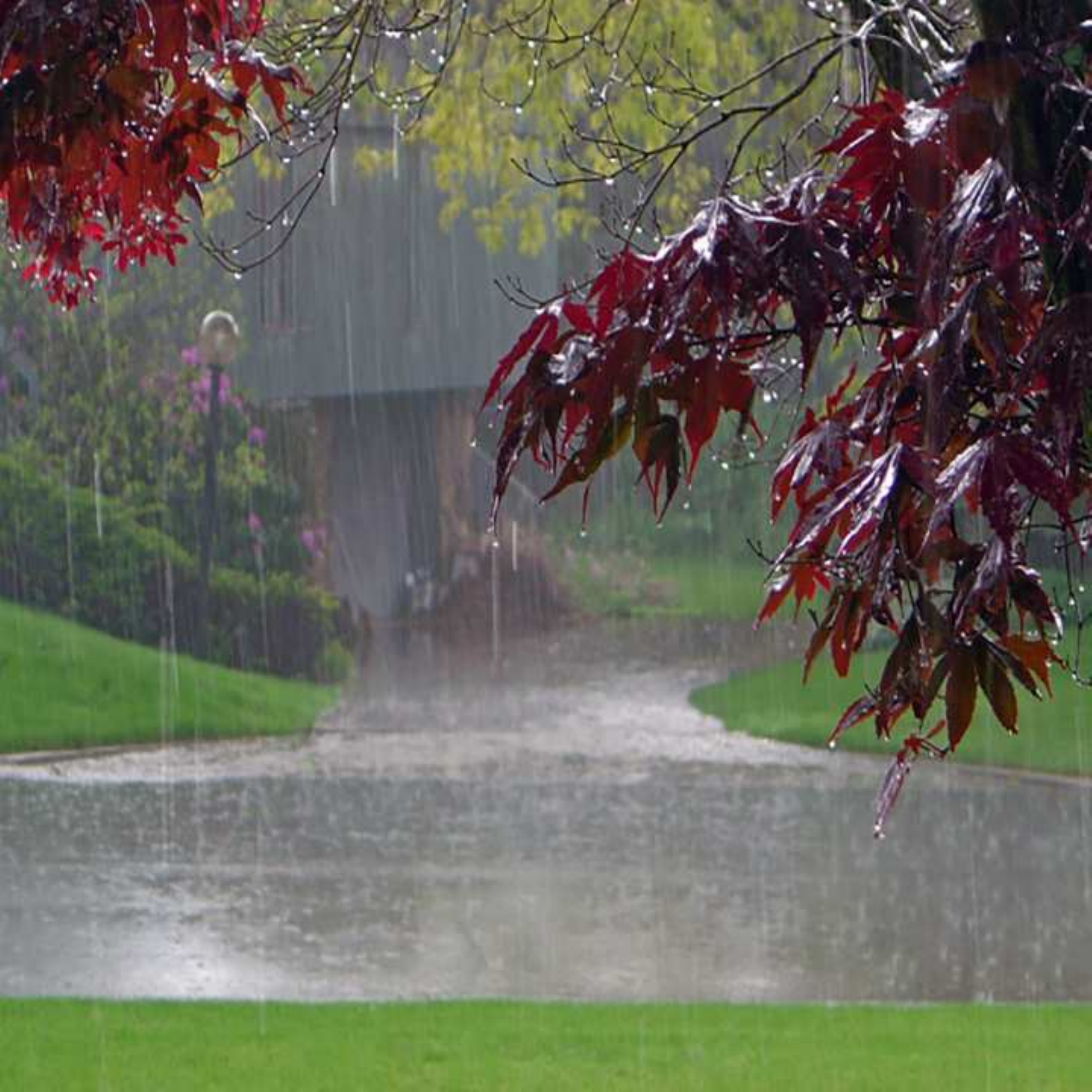}\vspace{2pt}

\end{minipage}}
\subfigure[Li et al.]{
\begin{minipage}[b]{0.11\linewidth}
\includegraphics[width=1\linewidth]{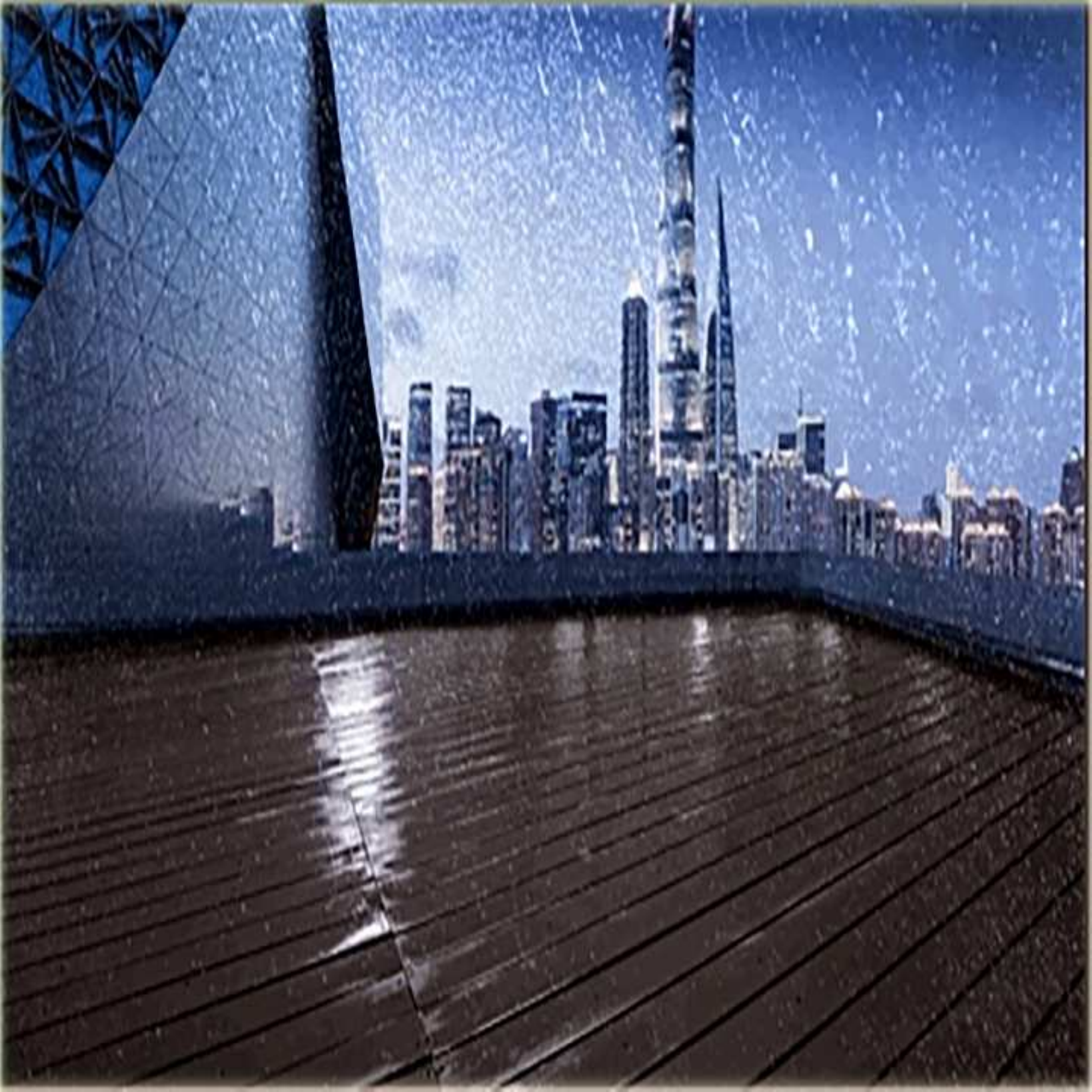}\vspace{2pt}
\includegraphics[width=1\linewidth]{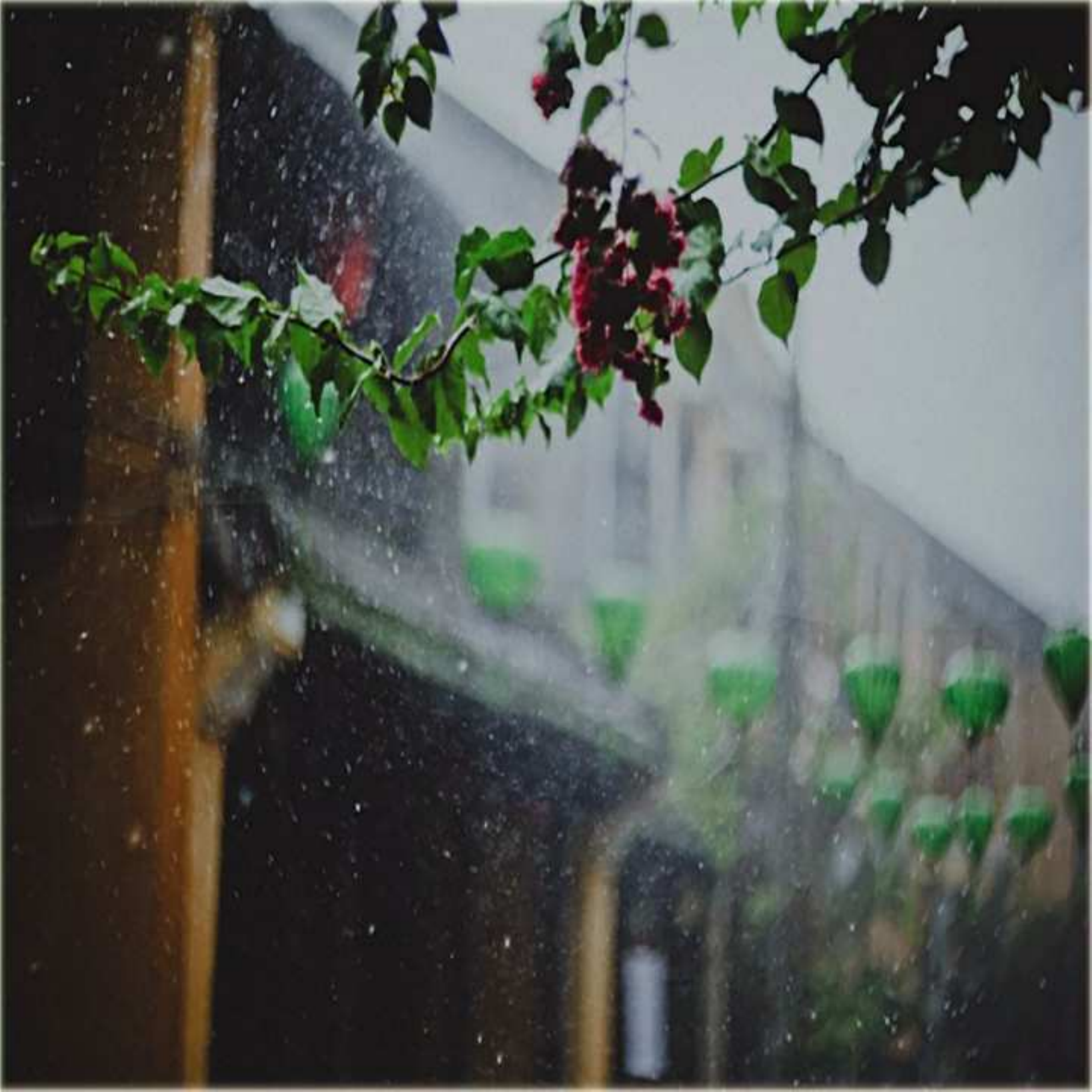}\vspace{2pt}
\includegraphics[width=1\linewidth]{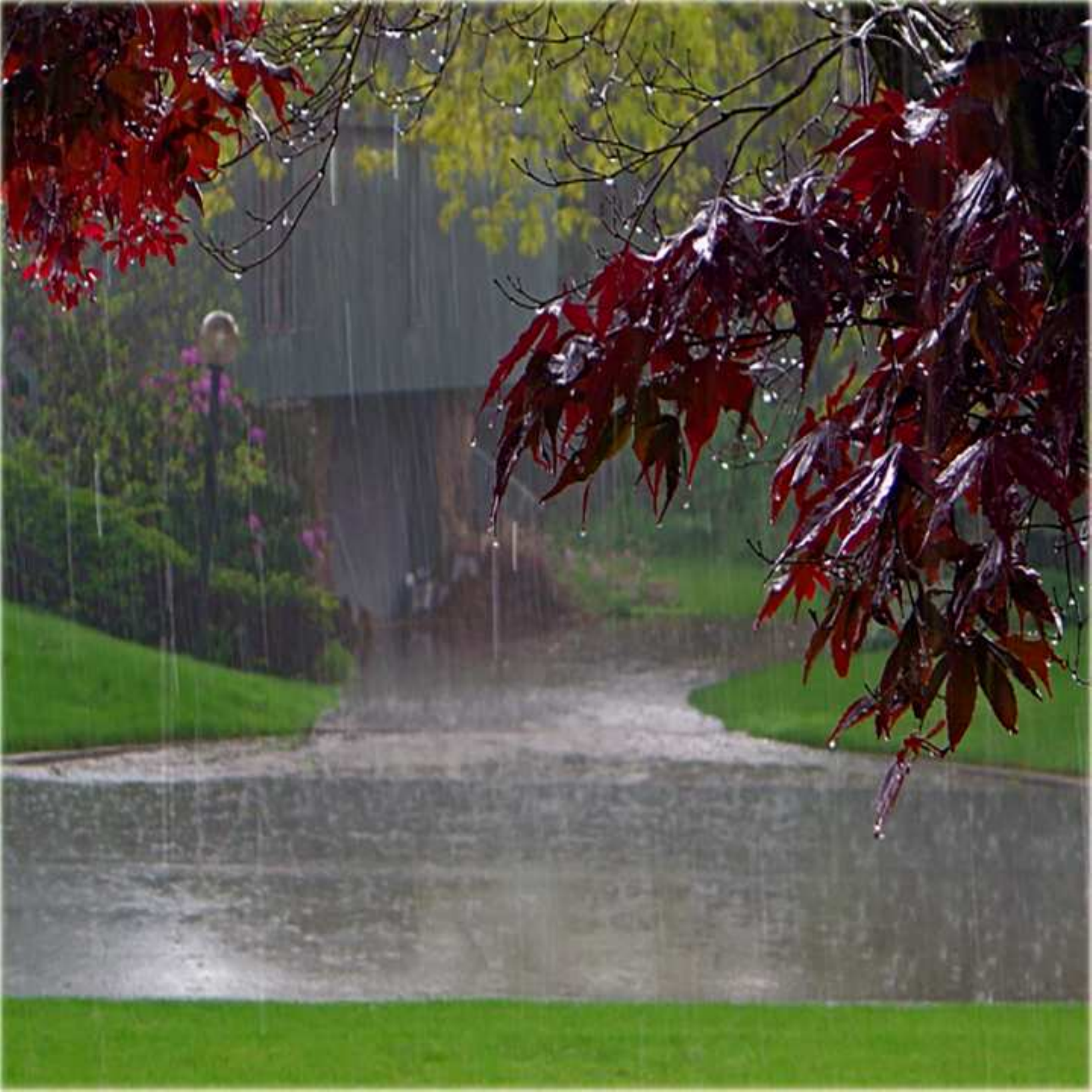}\vspace{2pt}
\end{minipage}}
\subfigure[Ren et al.]{
\begin{minipage}[b]{0.11\linewidth}
\includegraphics[width=1\linewidth]{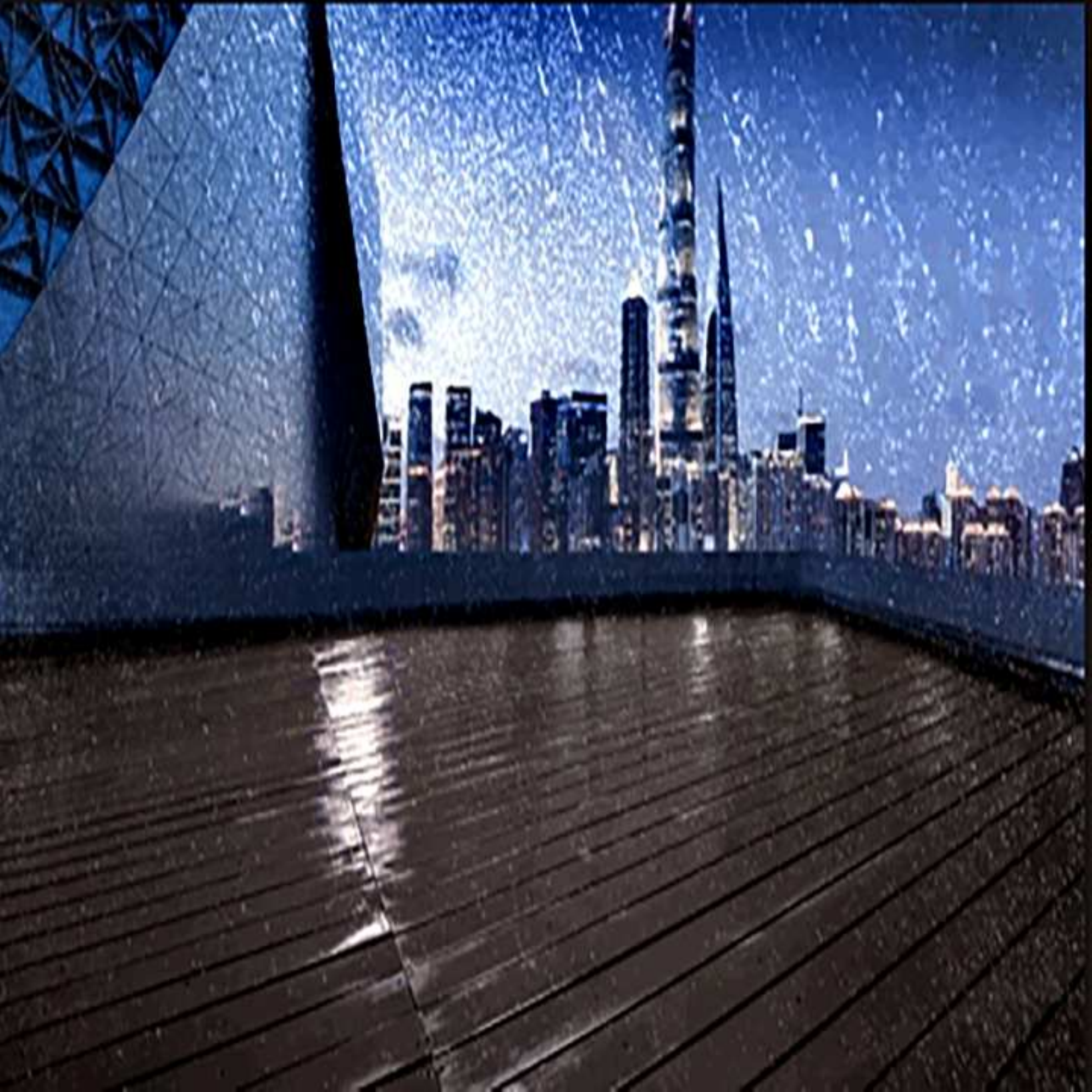}\vspace{2pt}
\includegraphics[width=1\linewidth]{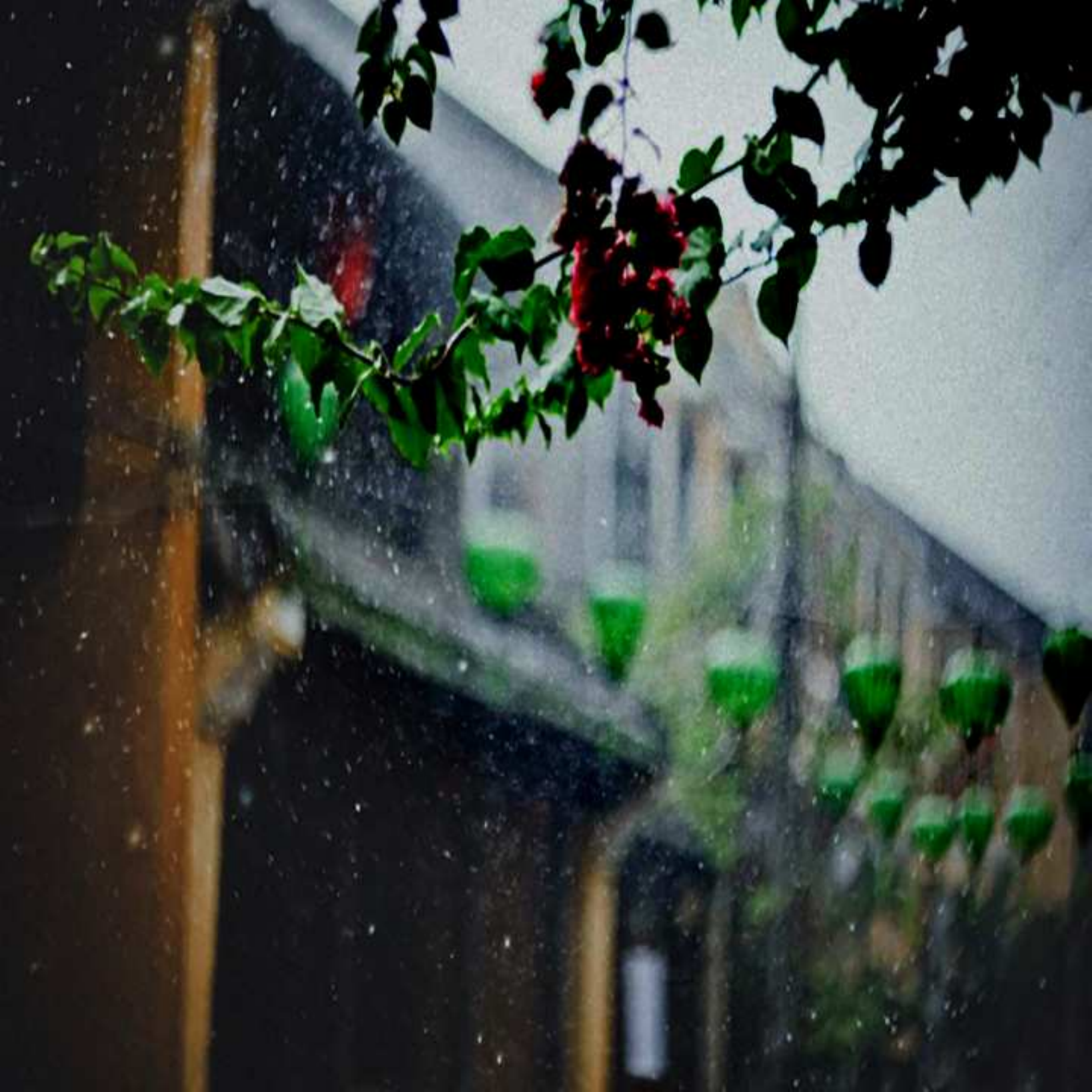}\vspace{2pt}
\includegraphics[width=1\linewidth]{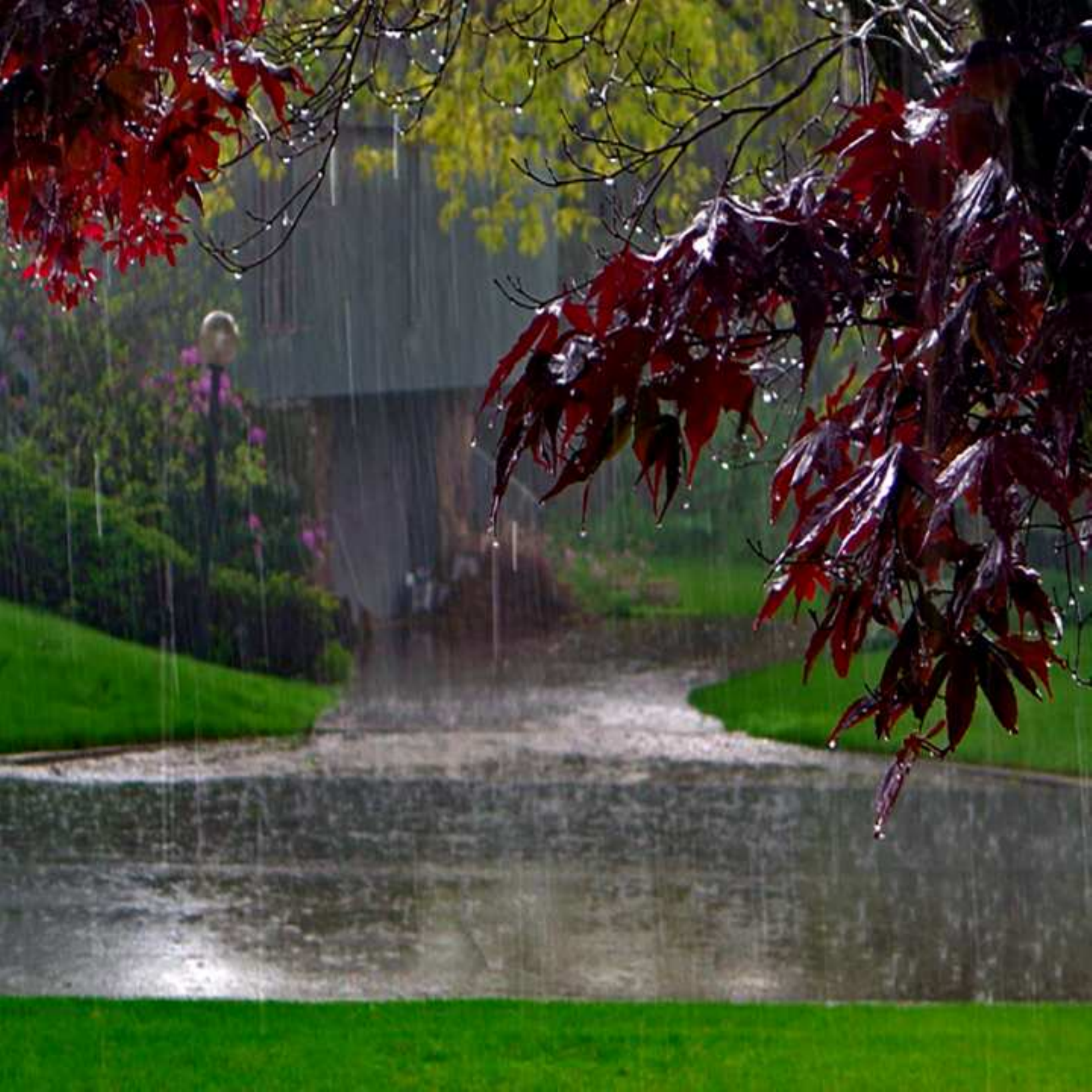}\vspace{2pt}
\end{minipage}}
\subfigure[Eigen et al.]{
\begin{minipage}[b]{0.11\linewidth}
\includegraphics[width=1\linewidth]{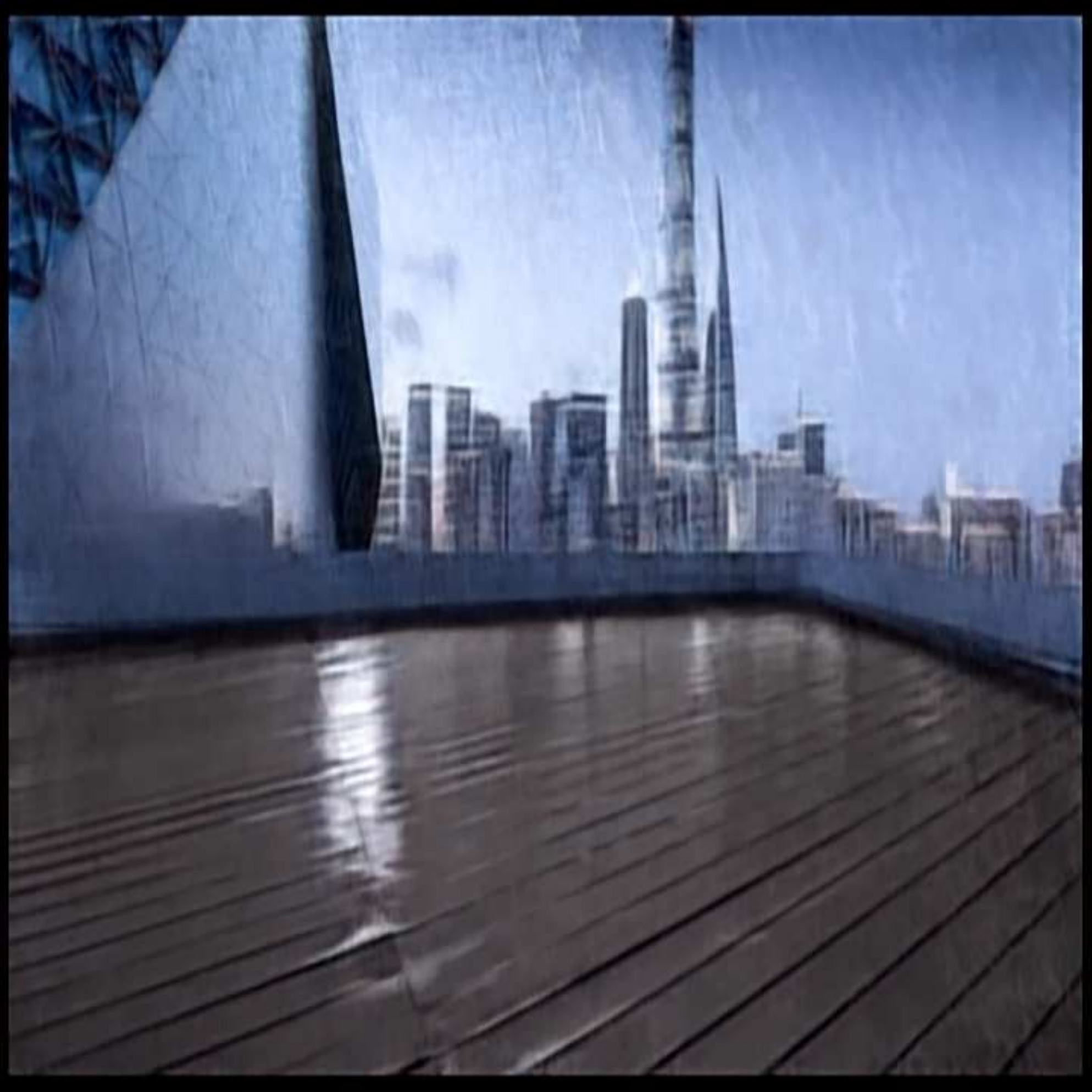}\vspace{2pt}
\includegraphics[width=1\linewidth]{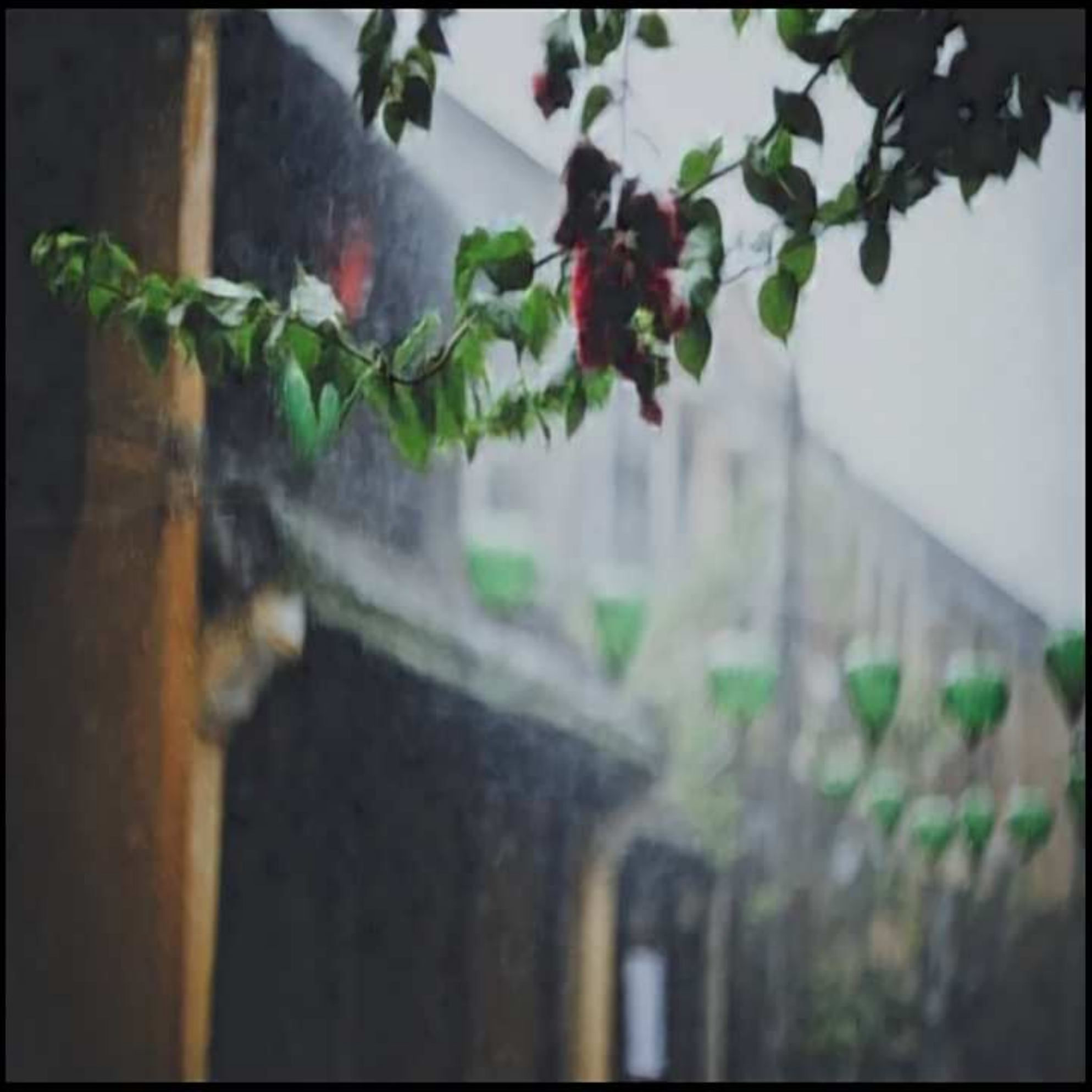}\vspace{2pt}
\includegraphics[width=1\linewidth]{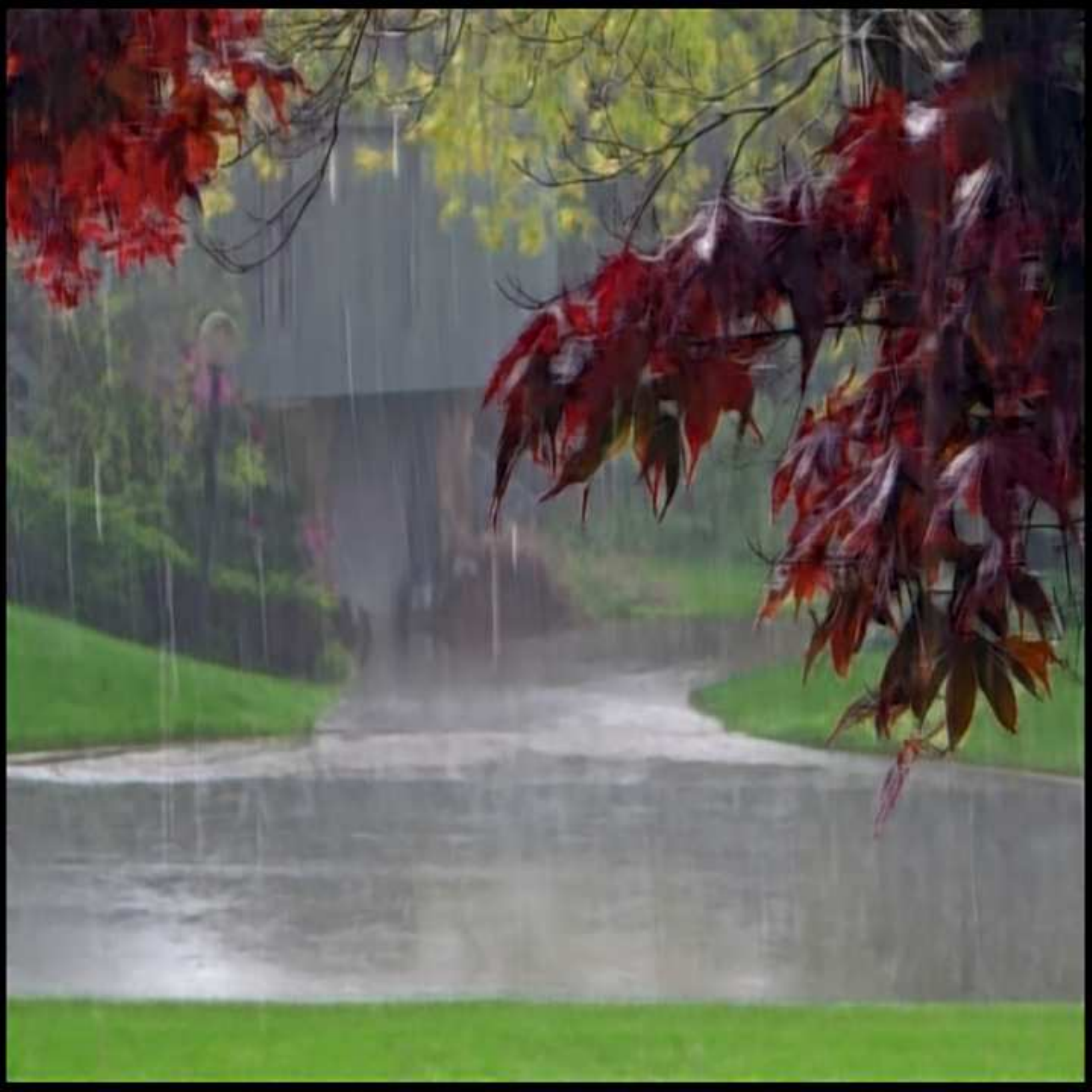}\vspace{2pt}
\end{minipage}}
\subfigure[Wang et al.]{
\begin{minipage}[b]{0.11\linewidth}
\includegraphics[width=1\linewidth]{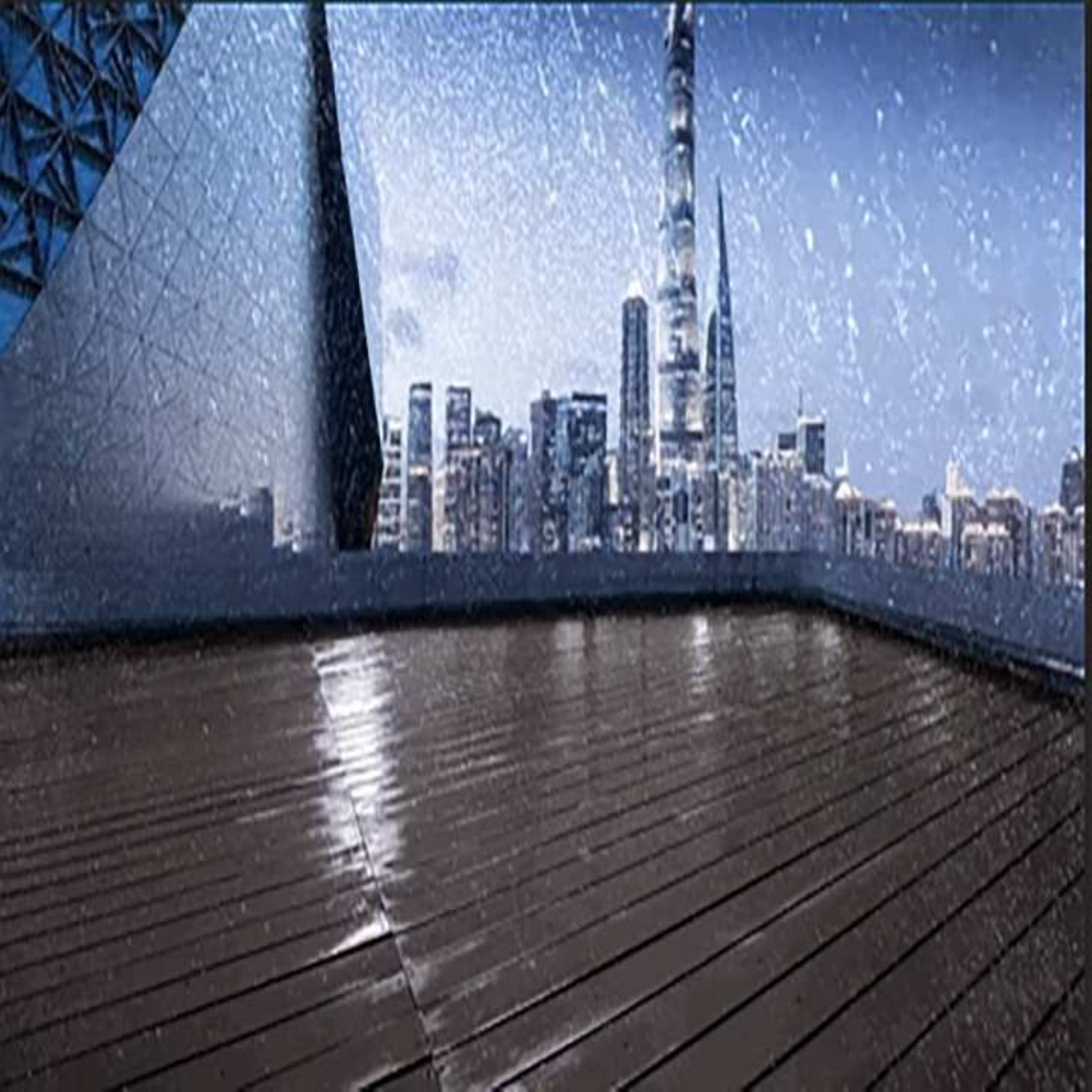}\vspace{2pt}
\includegraphics[width=1\linewidth]{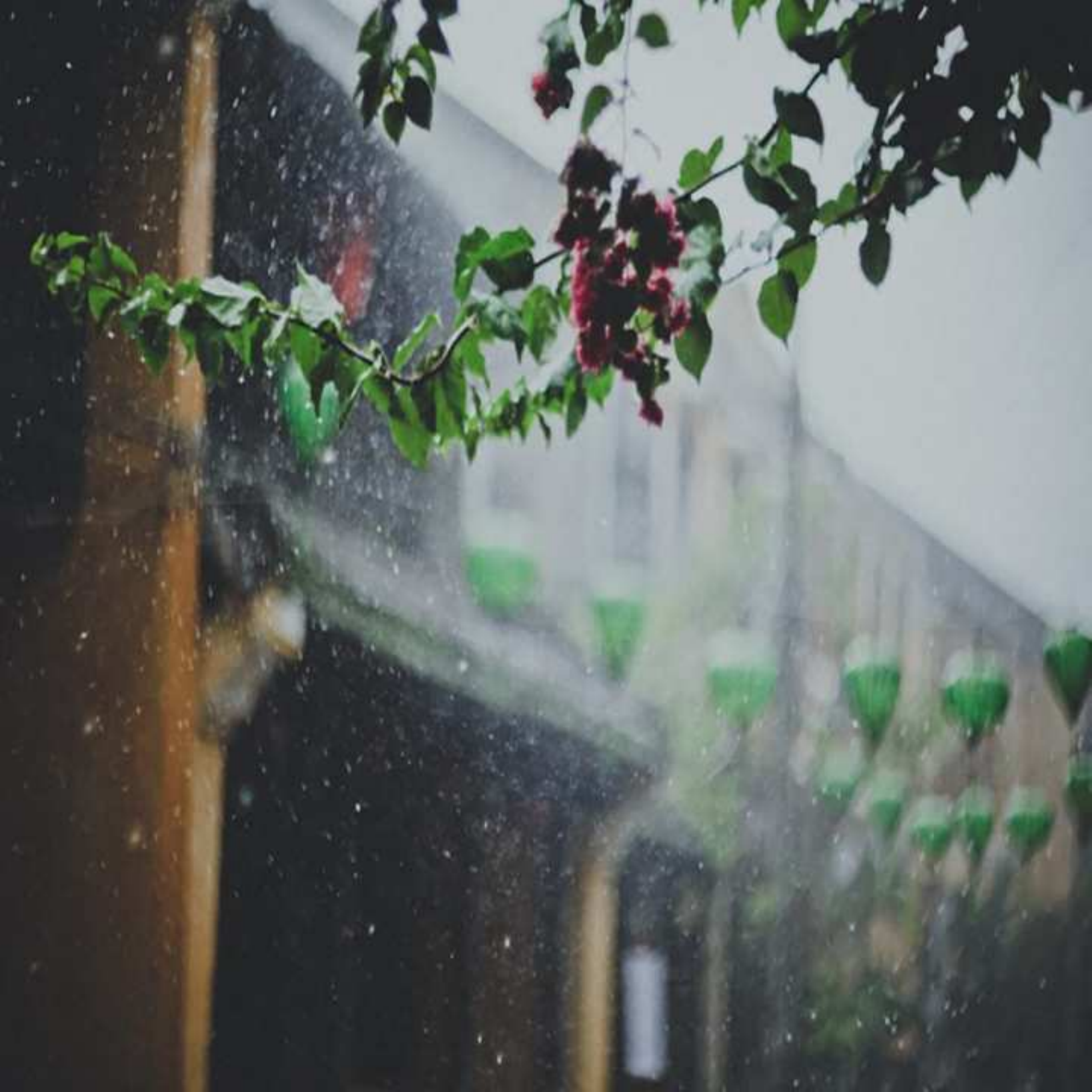}\vspace{2pt}
\includegraphics[width=1\linewidth]{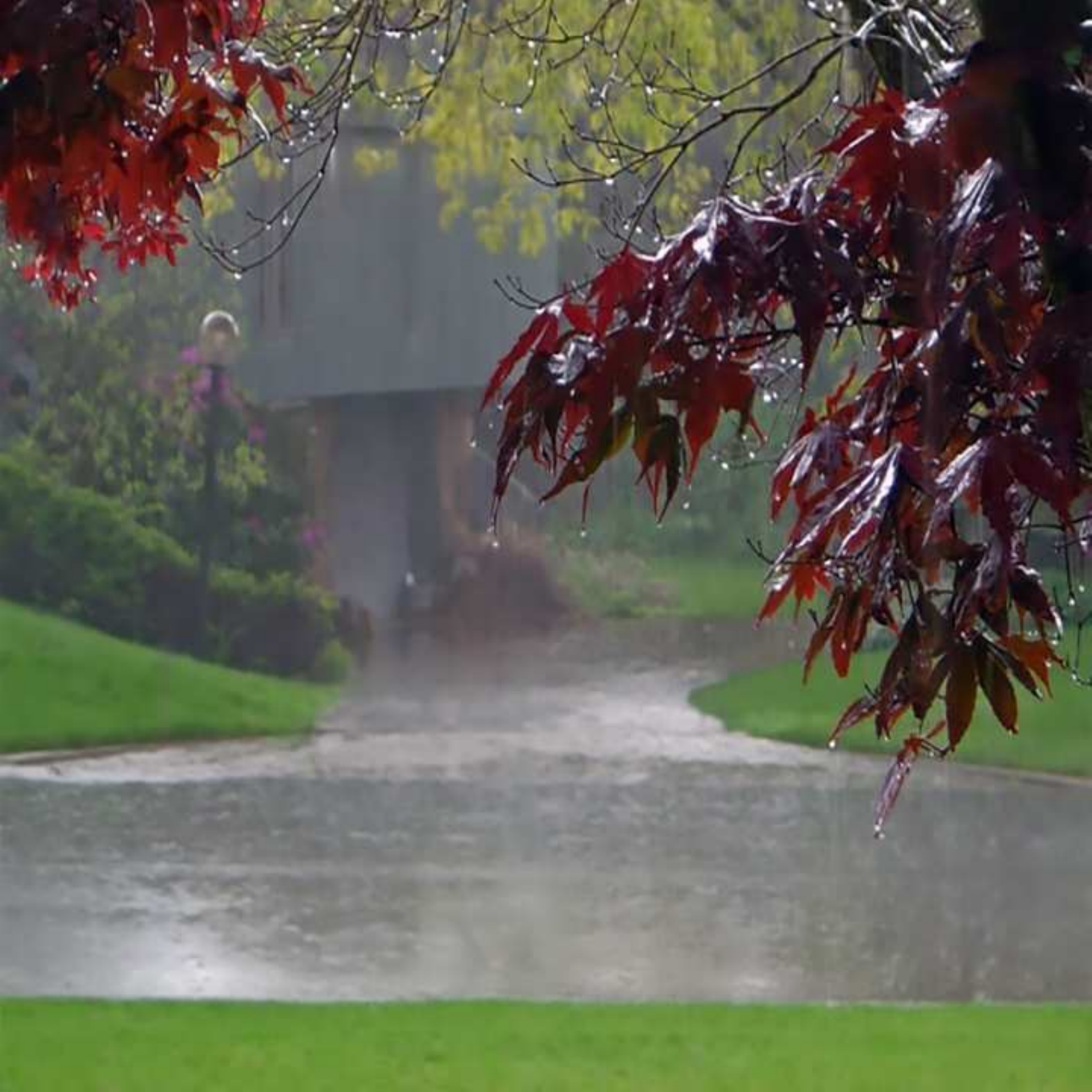}\vspace{2pt}
\end{minipage}}
\subfigure[Qian et al.]{
\begin{minipage}[b]{0.11\linewidth}
\includegraphics[width=1\linewidth]{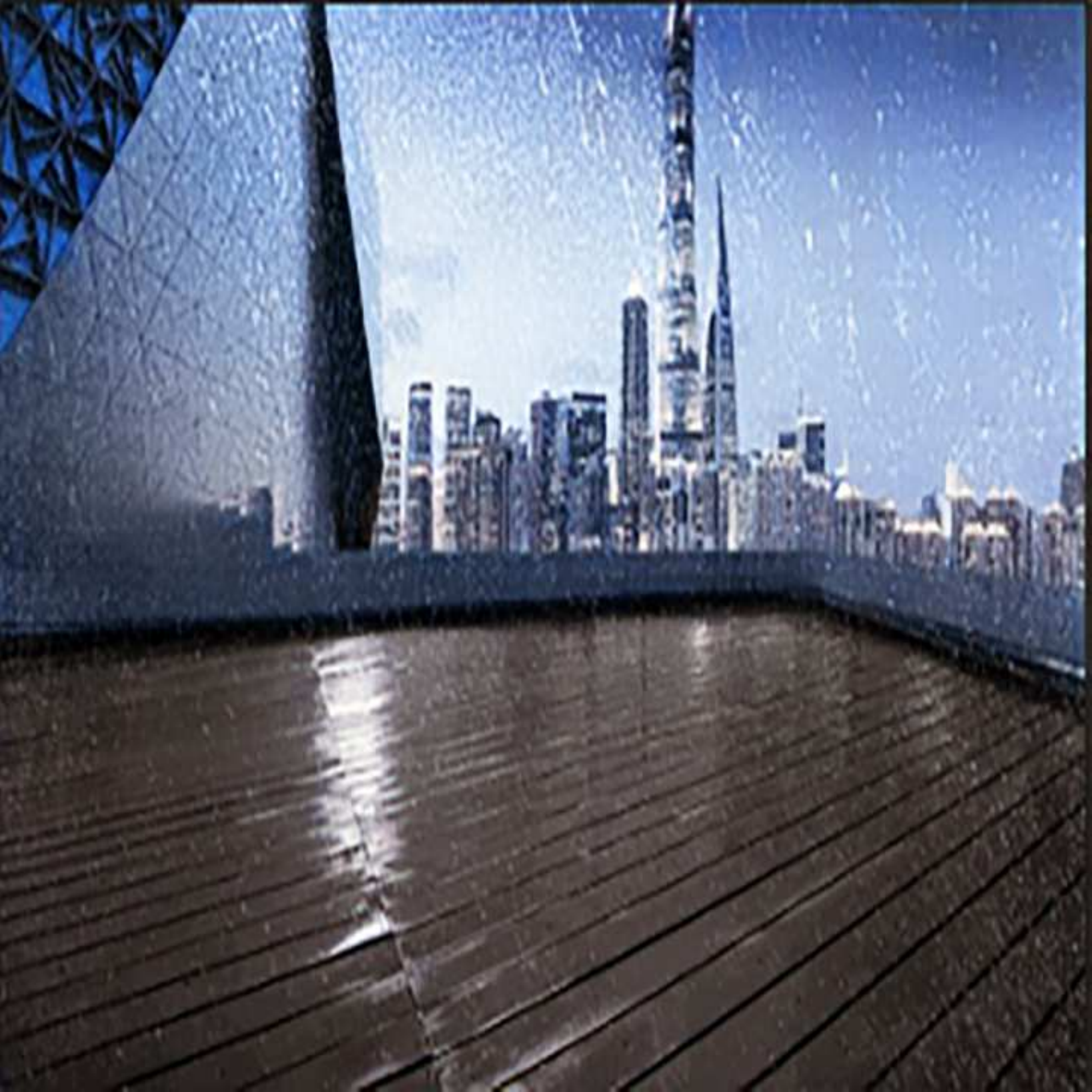}\vspace{2pt}
\includegraphics[width=1\linewidth]{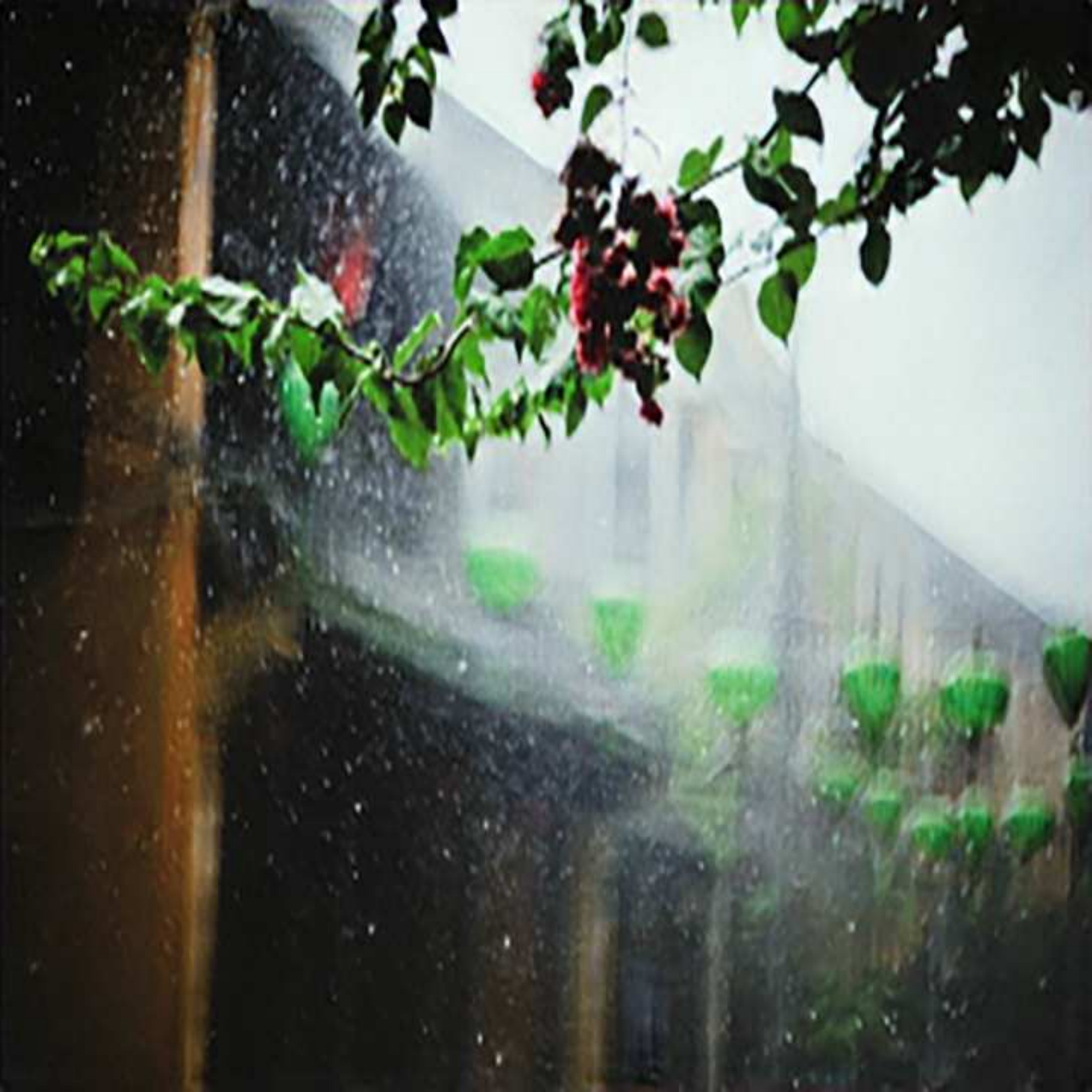}\vspace{2pt}
\includegraphics[width=1\linewidth]{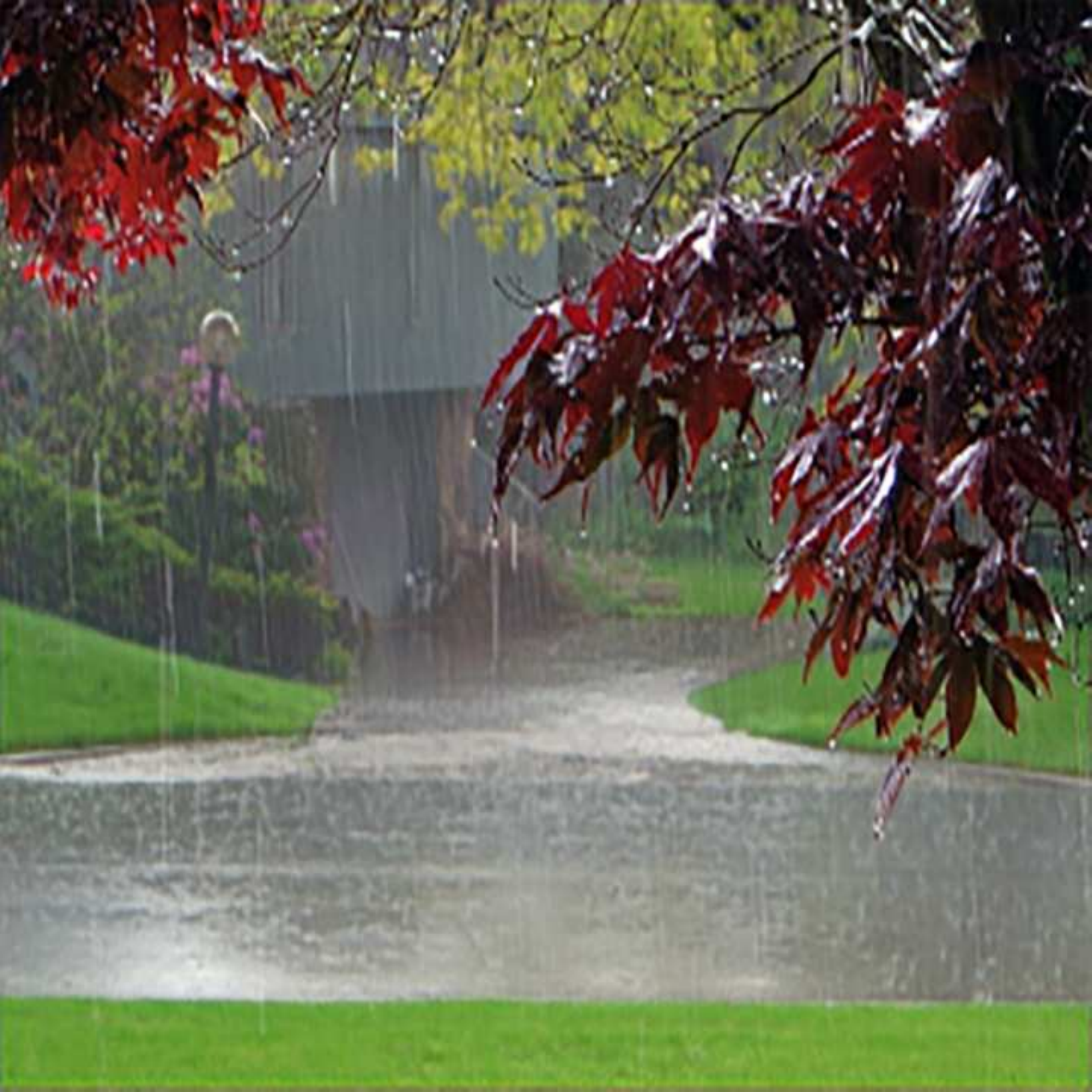}\vspace{2pt}
\end{minipage}}
\subfigure[Hu et al.]{
\begin{minipage}[b]{0.11\linewidth}
\includegraphics[width=1\linewidth]{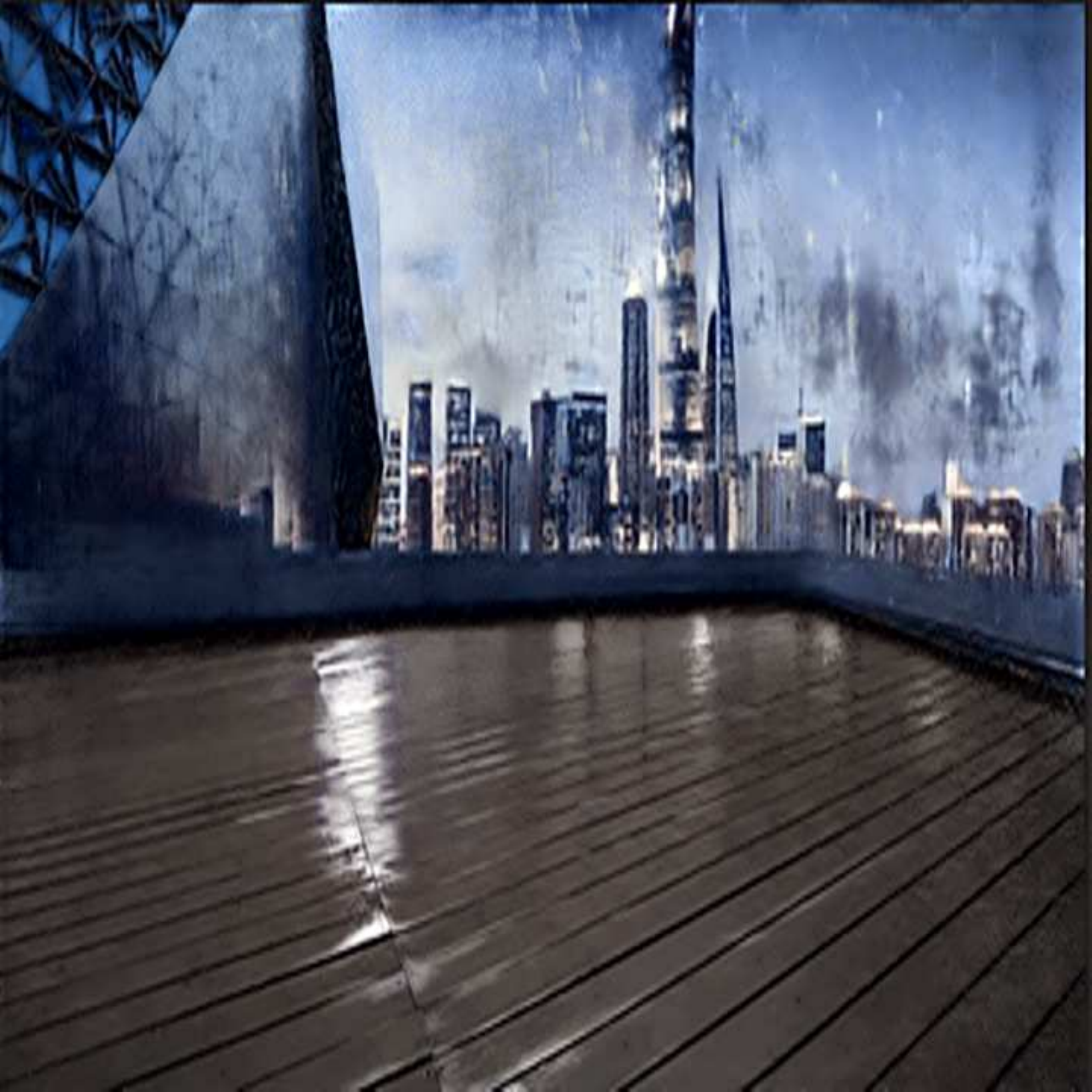}\vspace{2pt}
\includegraphics[width=1\linewidth]{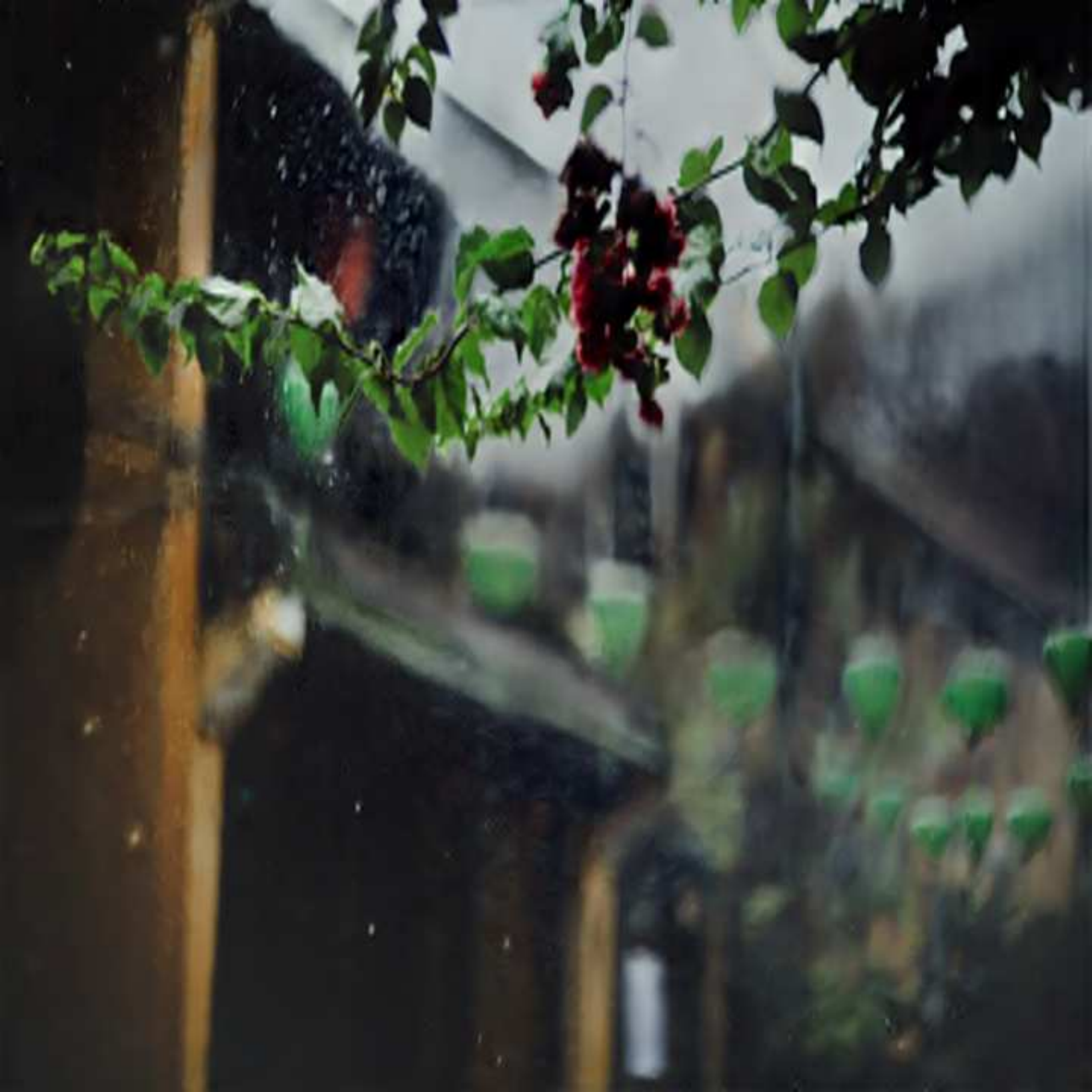}\vspace{2pt}
\includegraphics[width=1\linewidth]{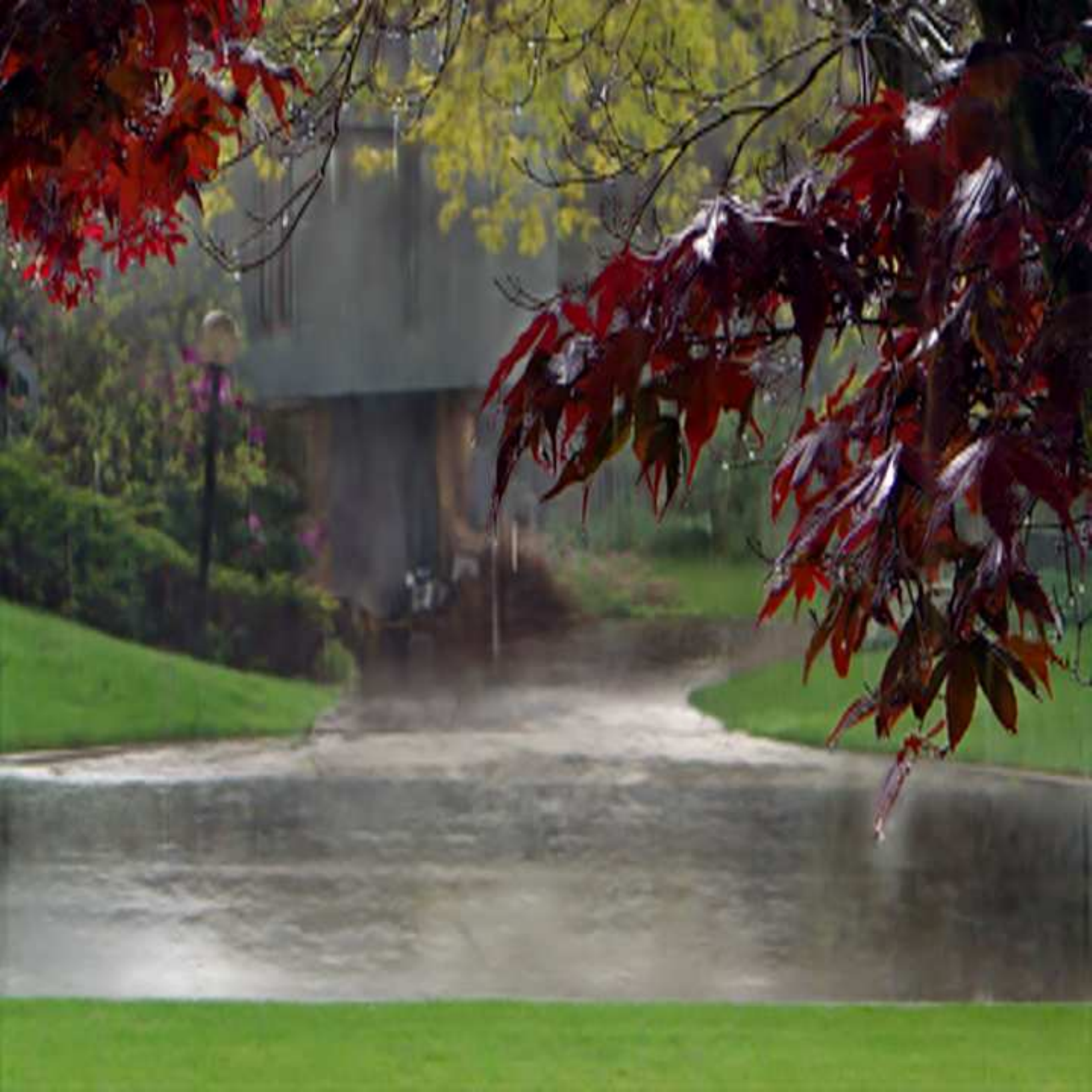}\vspace{2pt}
\end{minipage}}
\subfigure[Ours]{
\begin{minipage}[b]{0.11\linewidth}
\includegraphics[width=1\linewidth]{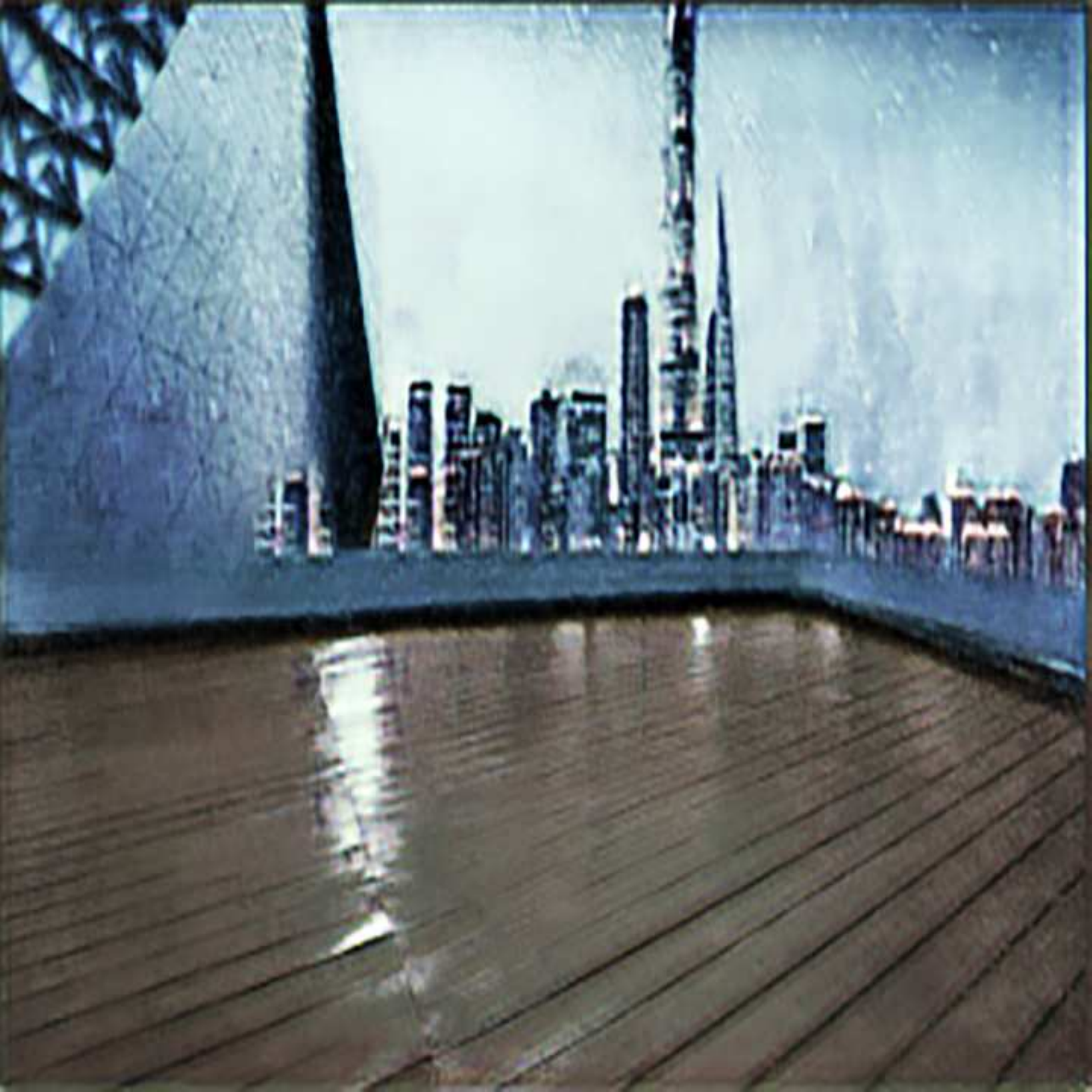}\vspace{2pt}
\includegraphics[width=1\linewidth]{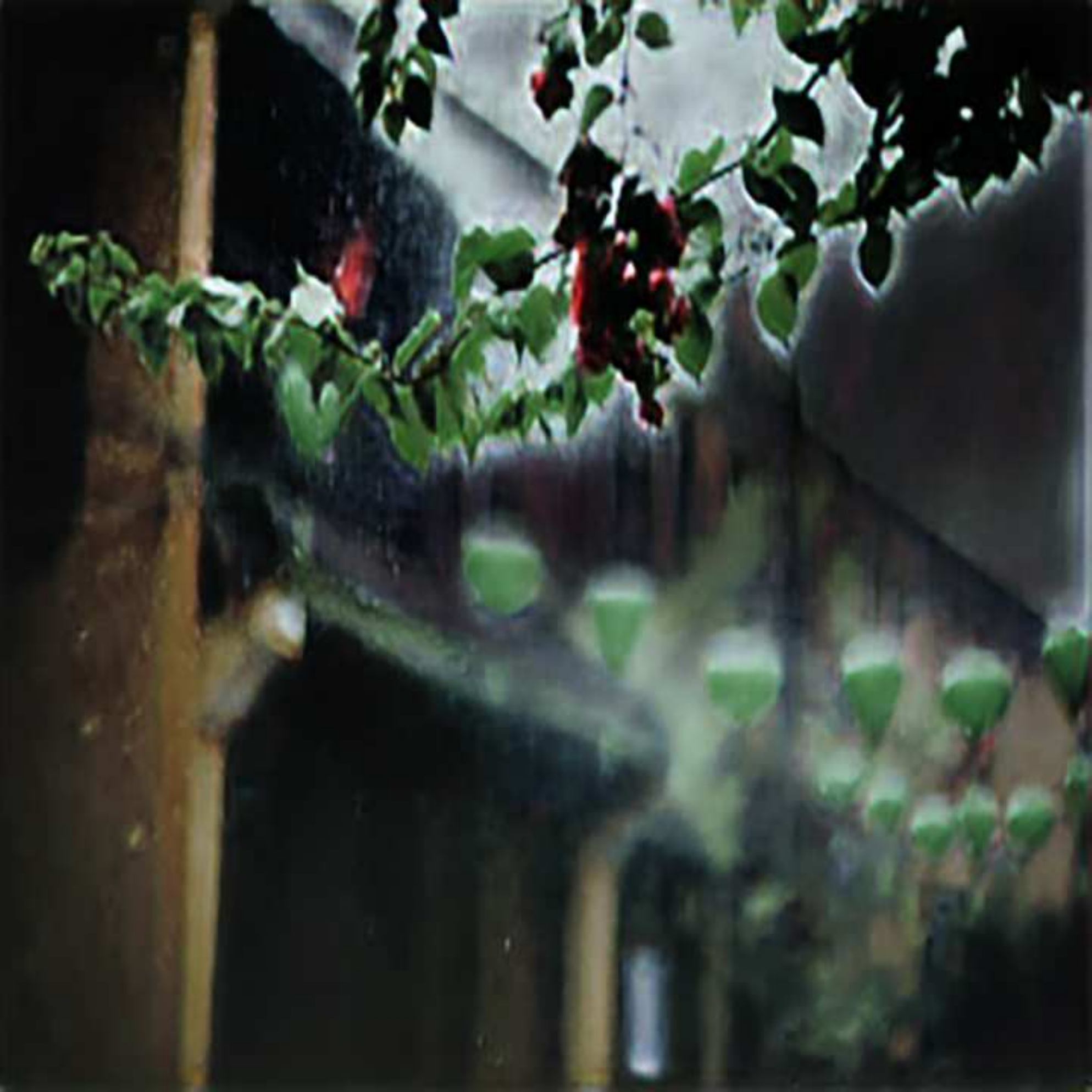}\vspace{2pt}
\includegraphics[width=1\linewidth]{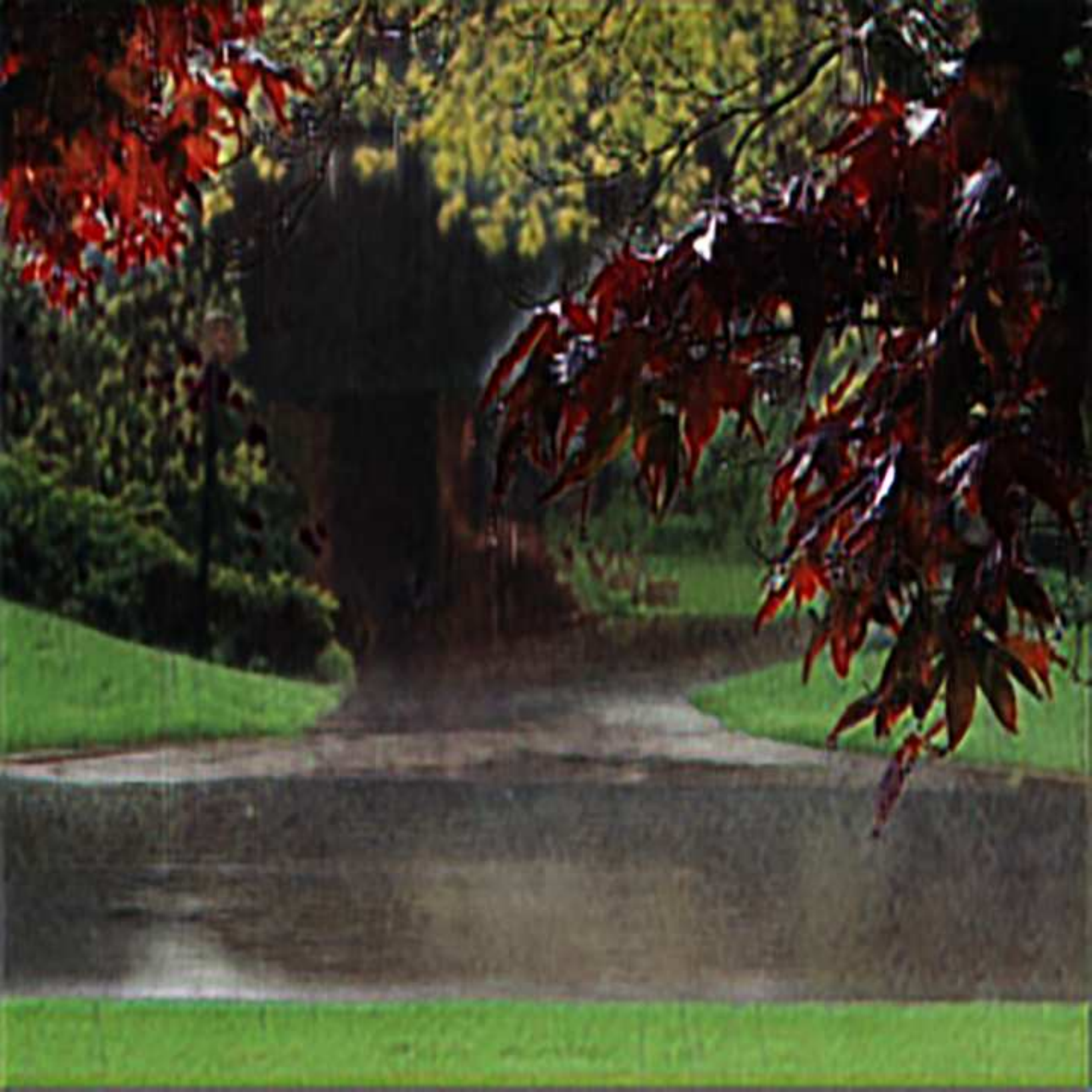}\vspace{2pt}
\end{minipage}}
\caption{Qualitative evaluation on real-world MOR images.}
\label{fig:f3}
\end{figure*}

\begin{table*}[htbp]
 \centering
 \small
\begin{tabular}{cccccccc}
\toprule
Method& Eigen et al. & Qian et al. & Wang et al. & Hu et al. & Li et al. &Ren et al. & Ours\\
\midrule
PSNR& 17.18 & 24.03 & 23.12 & 25.21 & 19.22 & 20.70 & \textbf{28.16}\\
SSIM& 0.71 & 0.84 & 0.82 & 0.85 & 0.76 & 0.77 & \textbf{0.89}\\
\bottomrule
\end{tabular}
\caption{Averaged PNSR and SSIM on RainCityscapes++ of different methods for removing MOR.}
\label{tables1}
\end{table*}

\begin{table}
 \centering
 \small
\begin{tabular}{ccc}
\toprule
Scheme& PSNR& SSIM\\
\midrule
A& 17.79& 0.72\\
H+A& 22.56& 0.79\\
L+A& 24.73& 0.83\\
H+L+A& 26.16 & 0.86\\
H+L+R+A& 26.97 & 0.87\\
H+L+R+A+D& \textbf{28.16}& \textbf{0.89}\\
\bottomrule
\end{tabular}
\caption{The decomposition for ablation study.}
\label{tables3}
\end{table}

\begin{figure*}
\centering
\subfigure[Input image]{
\begin{minipage}[b]{0.12\linewidth}
\includegraphics[width=1\linewidth]{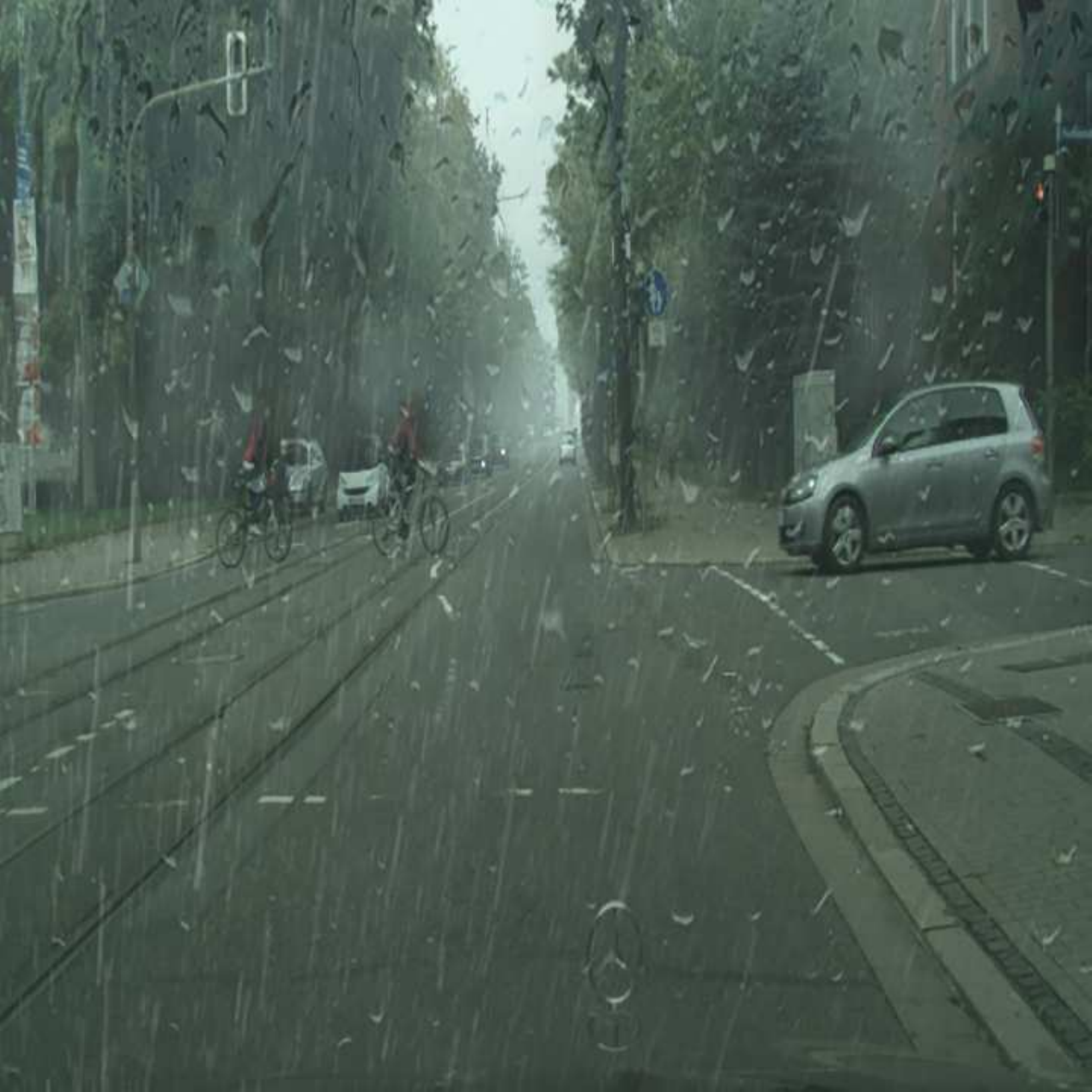}\vspace{2pt}
\includegraphics[width=1\linewidth]{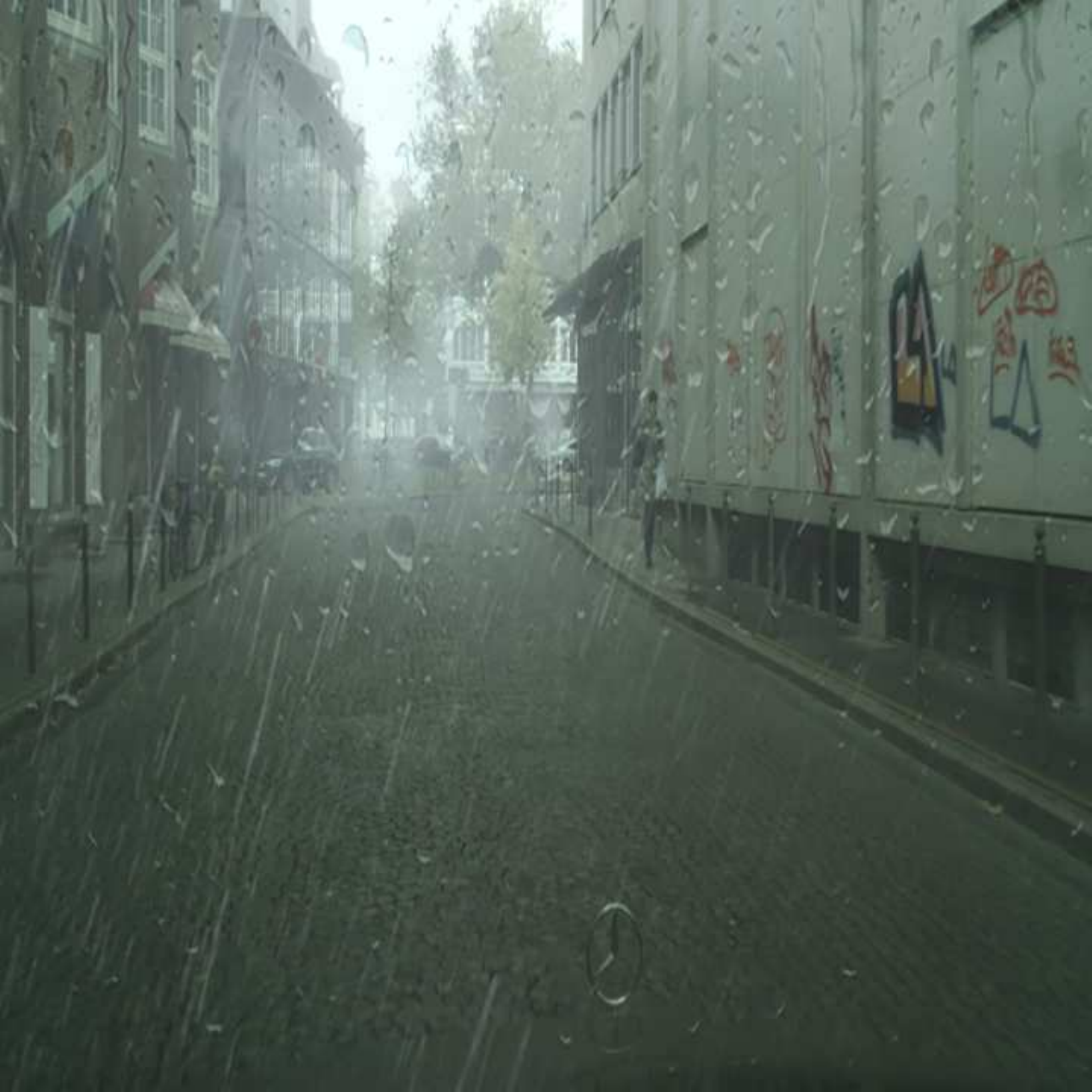}\vspace{2pt}
\includegraphics[width=1\linewidth]{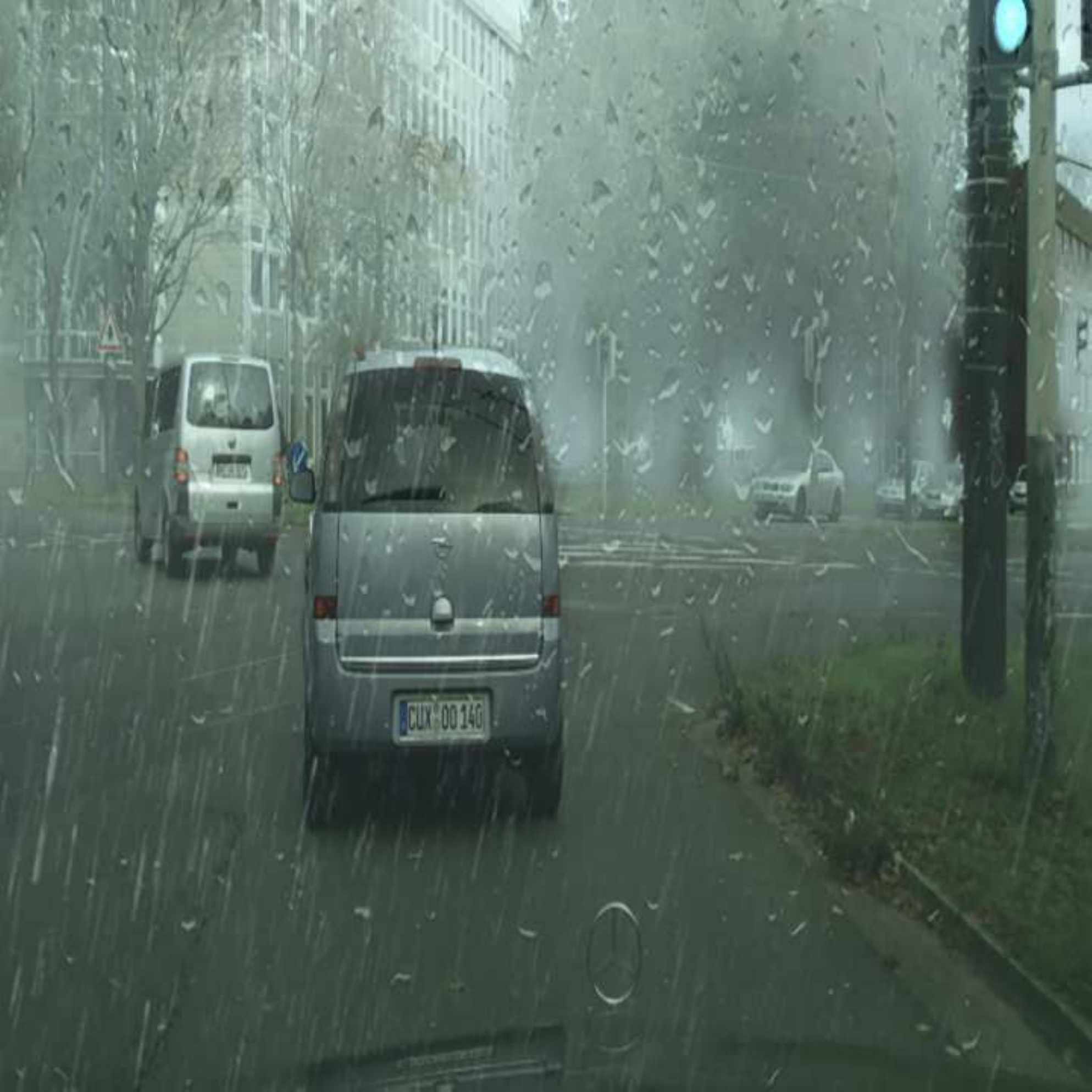}\vspace{2pt}
\end{minipage}}
\subfigure[A]{
\begin{minipage}[b]{0.12\linewidth}
\includegraphics[width=1\linewidth]{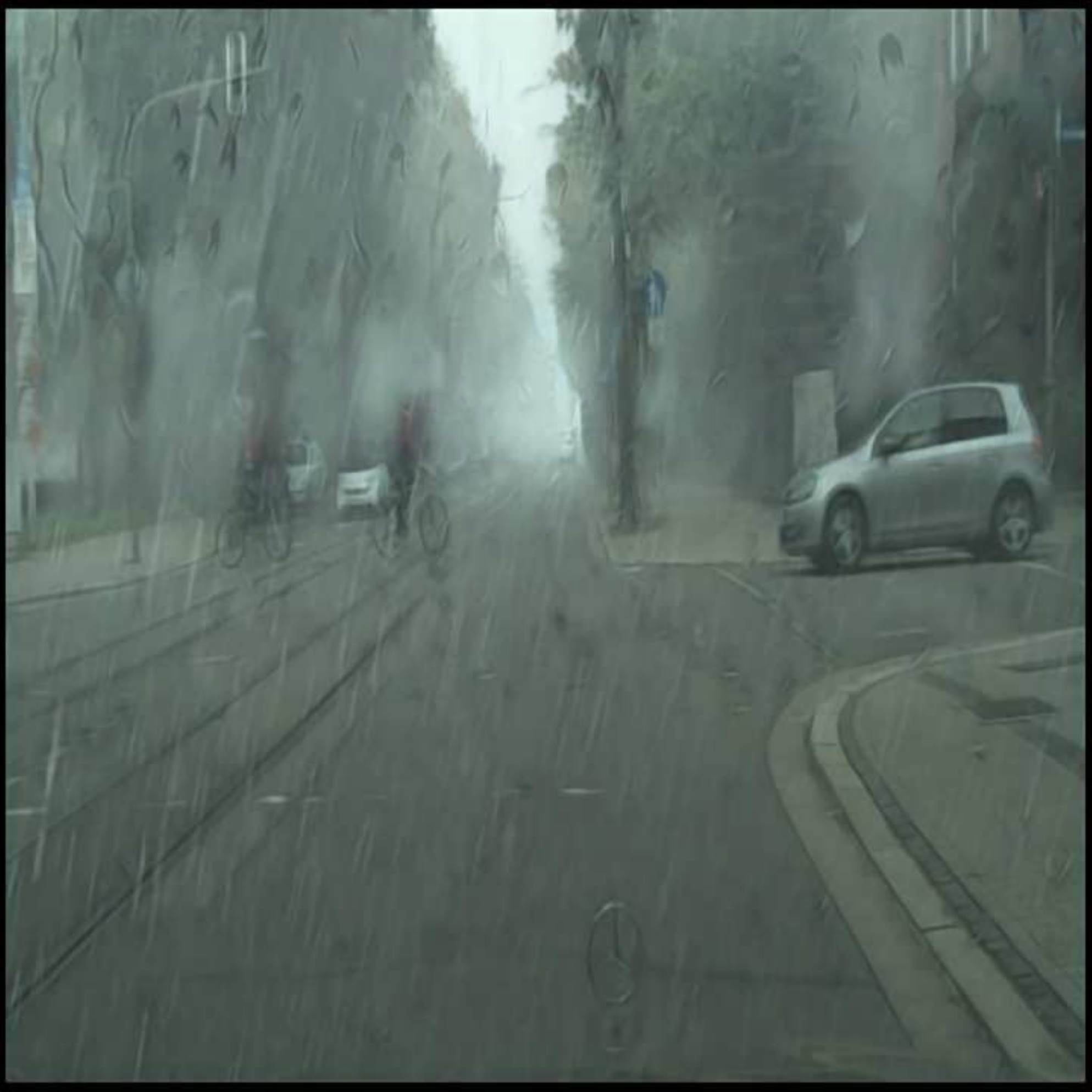}\vspace{2pt}
\includegraphics[width=1\linewidth]{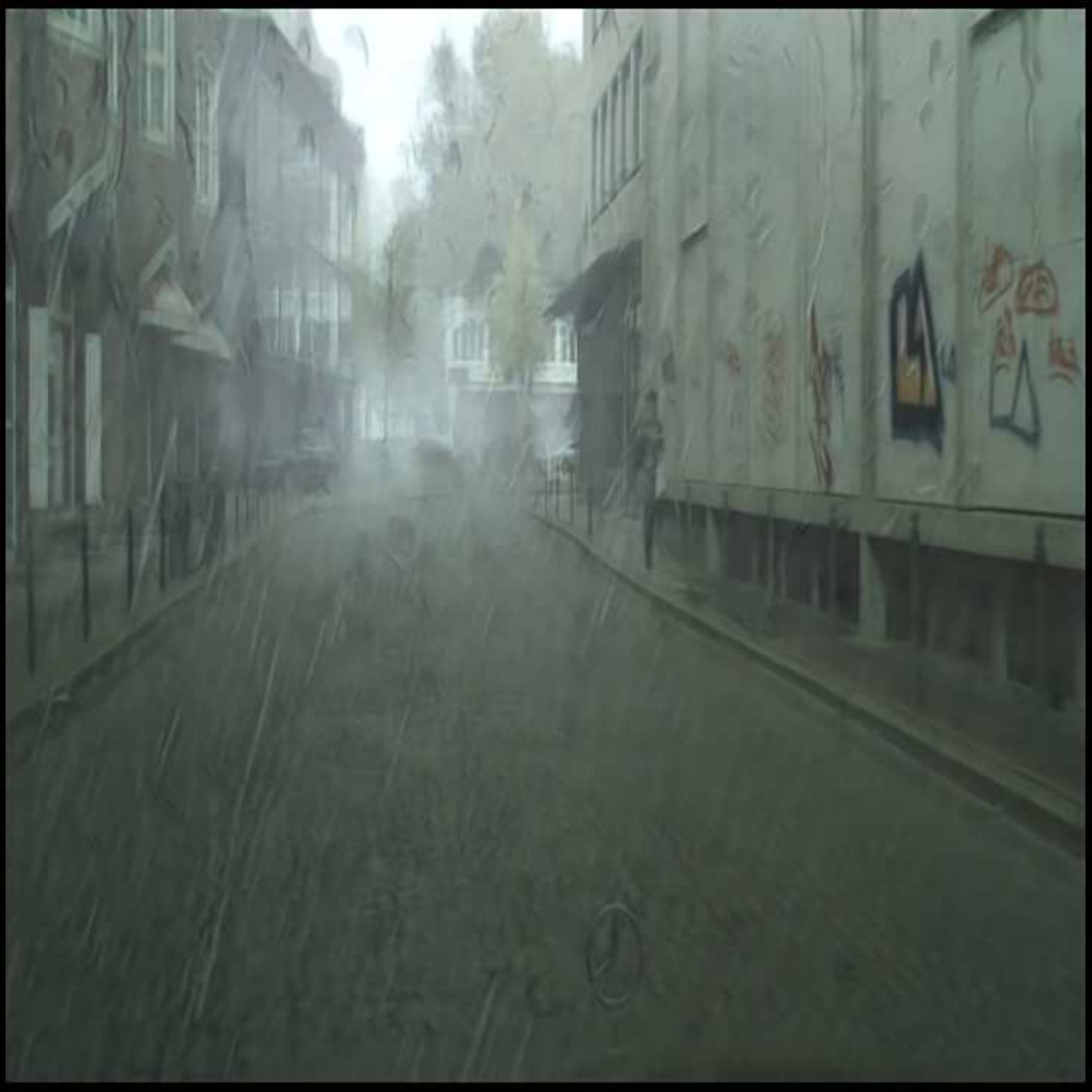}\vspace{2pt}
\includegraphics[width=1\linewidth]{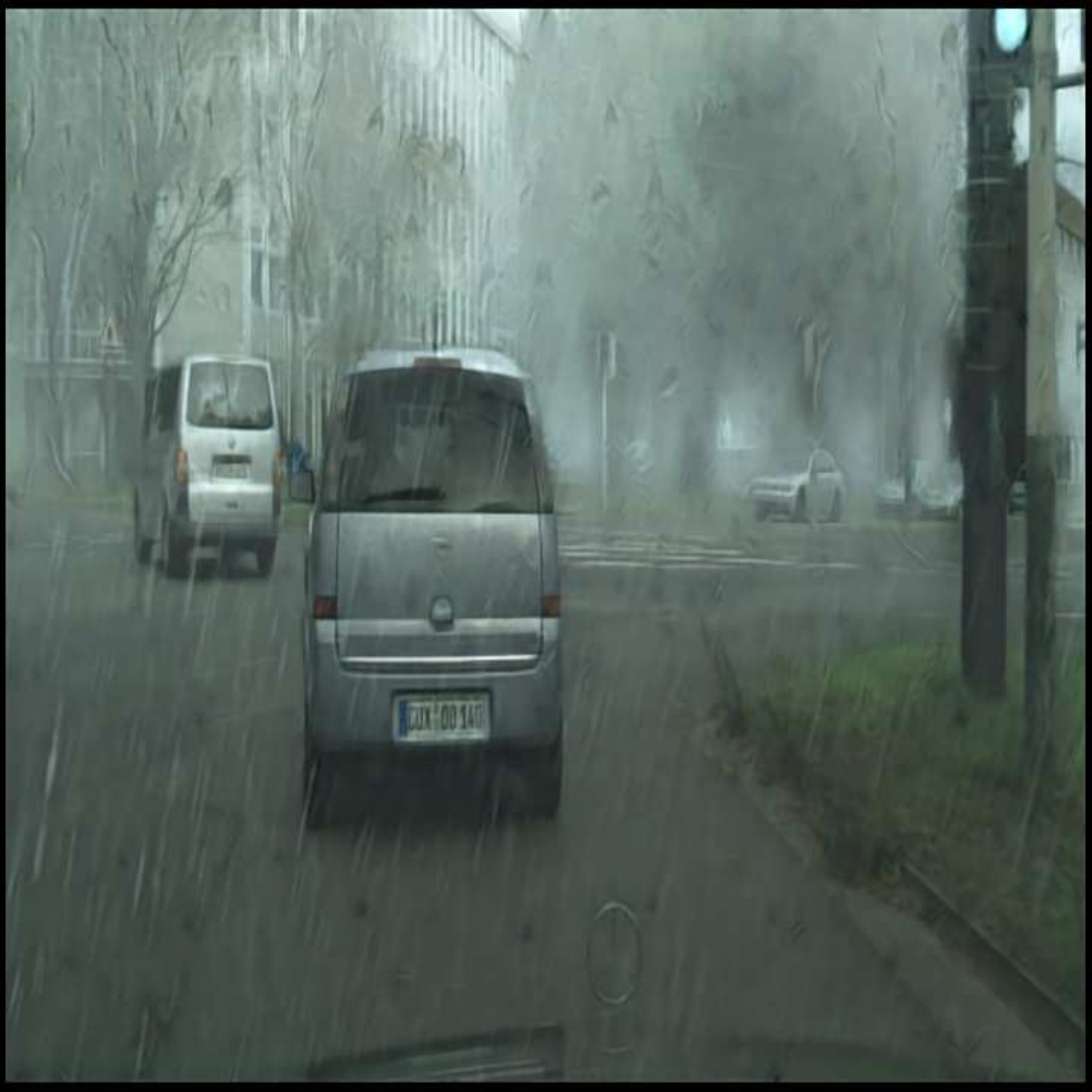}\vspace{2pt}
\end{minipage}}
\subfigure[H+A]{
\begin{minipage}[b]{0.12\linewidth}
\includegraphics[width=1\linewidth]{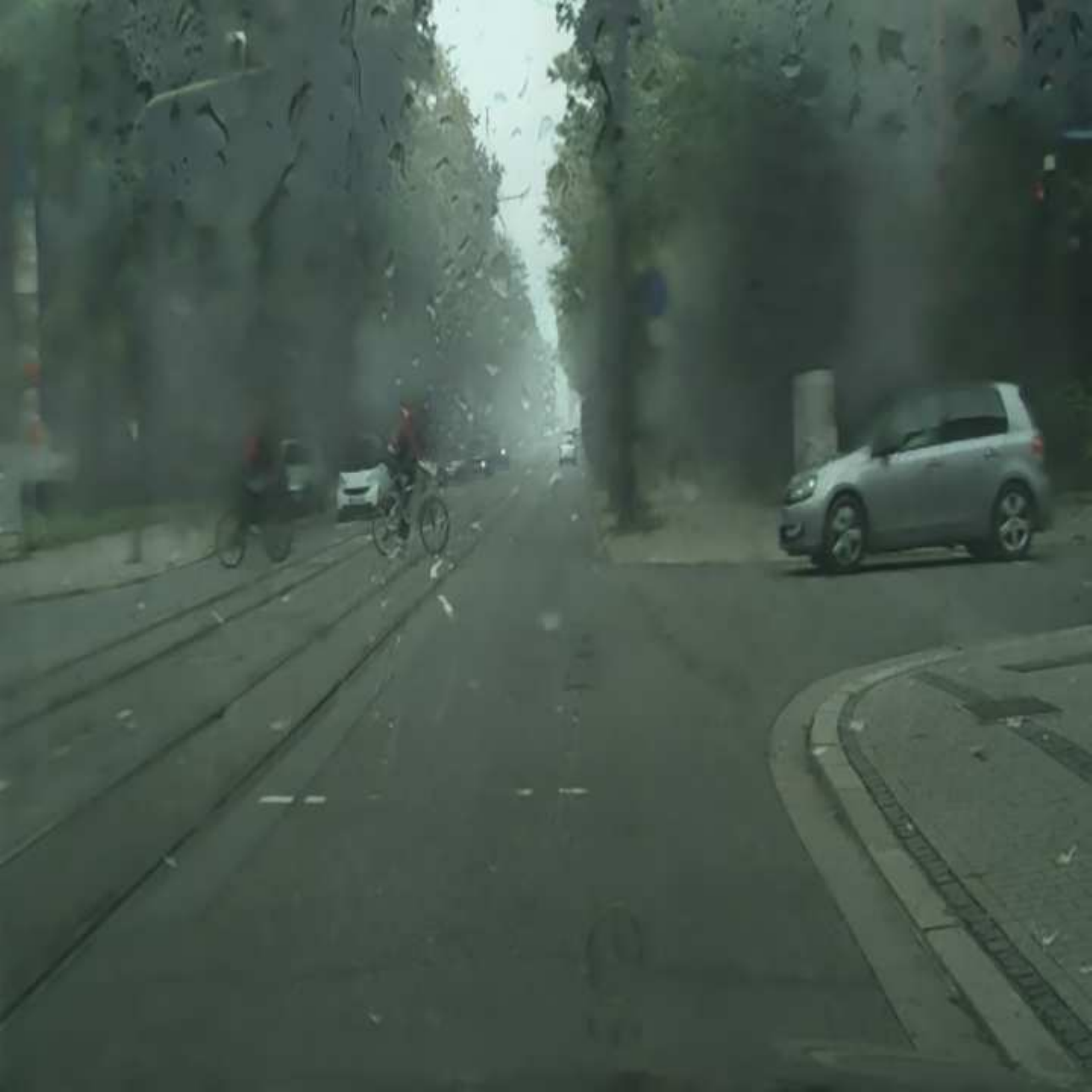}\vspace{2pt}
\includegraphics[width=1\linewidth]{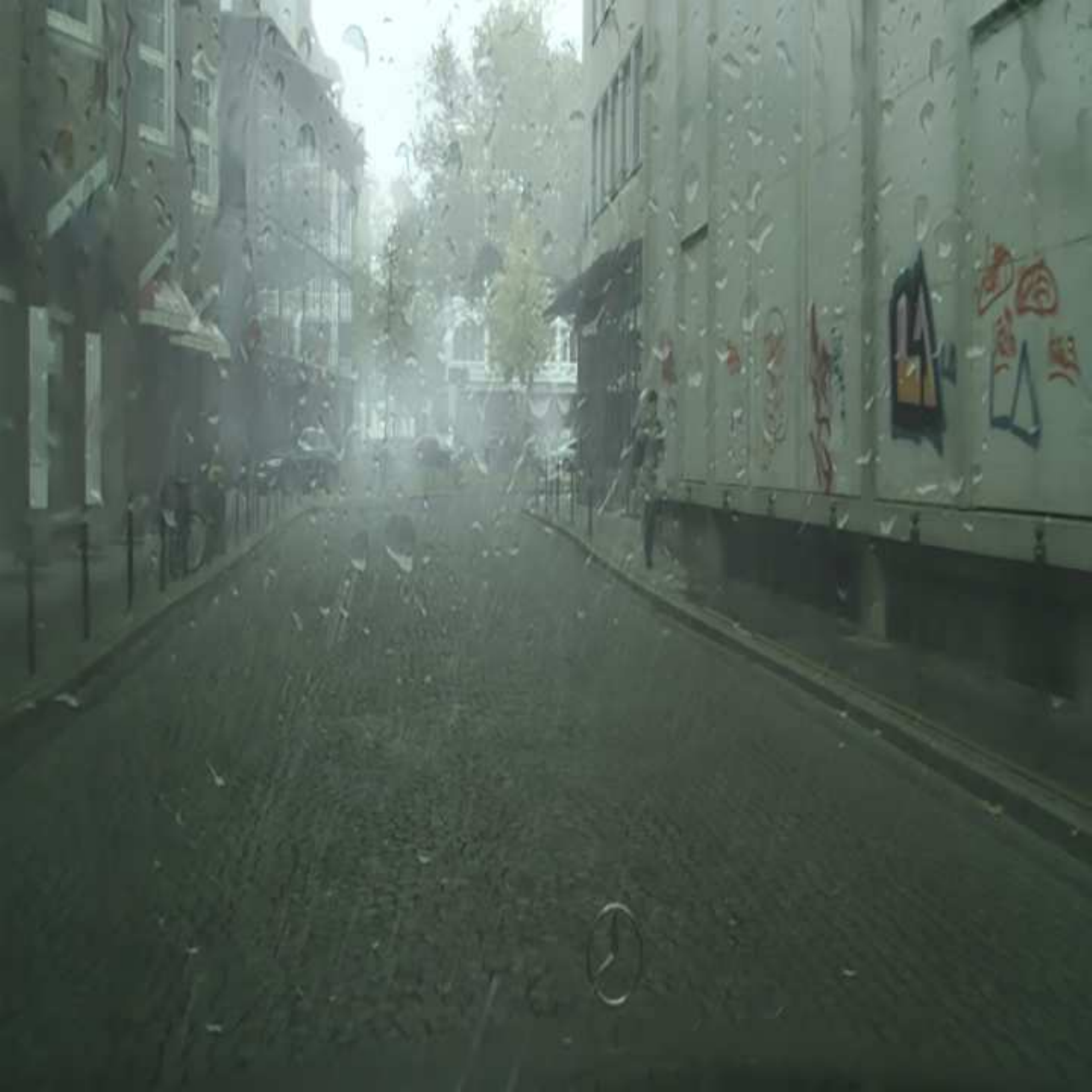}\vspace{2pt}
\includegraphics[width=1\linewidth]{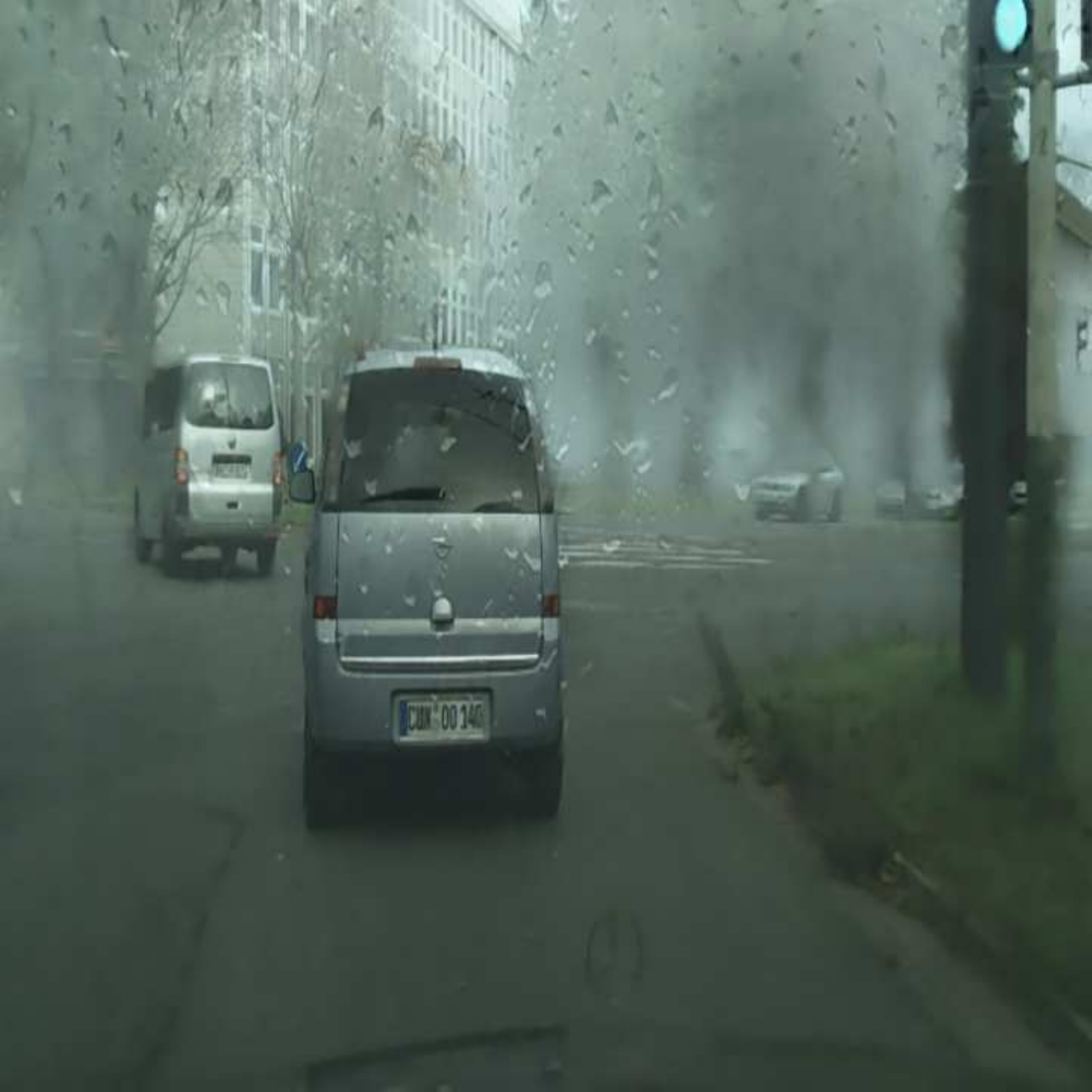}\vspace{2pt}
\end{minipage}}
\subfigure[L+A]{
\begin{minipage}[b]{0.12\linewidth}
\includegraphics[width=1\linewidth]{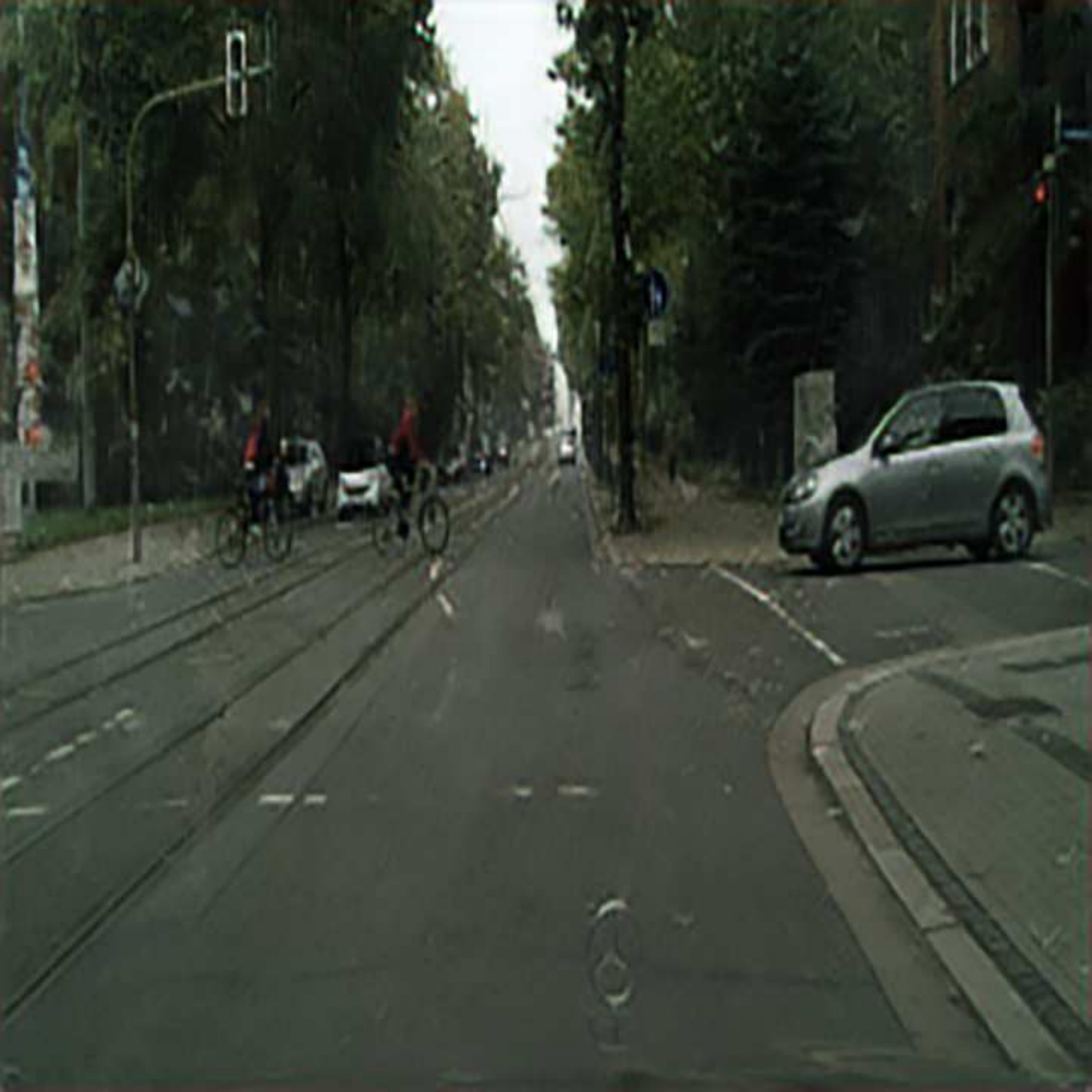}\vspace{2pt}
\includegraphics[width=1\linewidth]{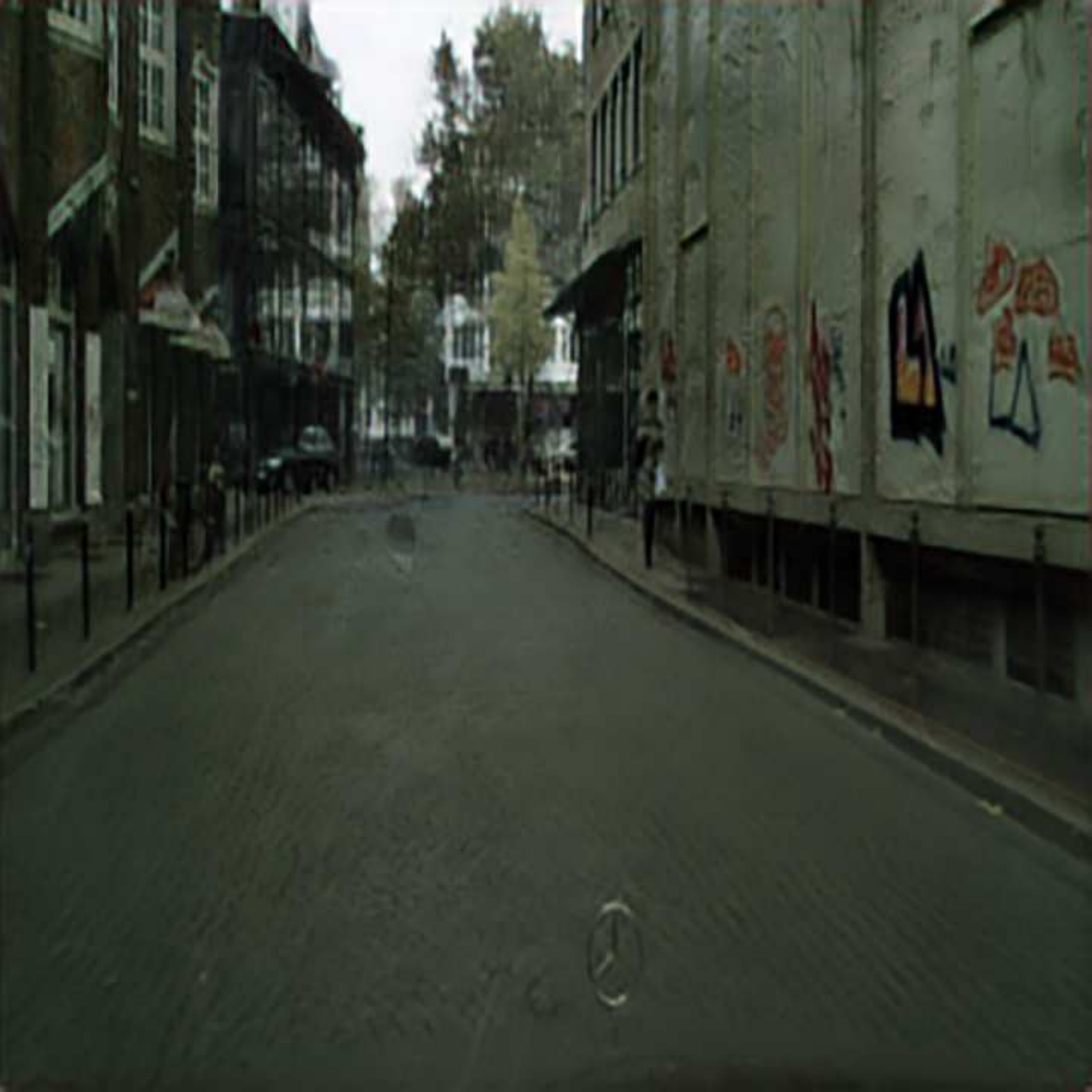}\vspace{2pt}
\includegraphics[width=1\linewidth]{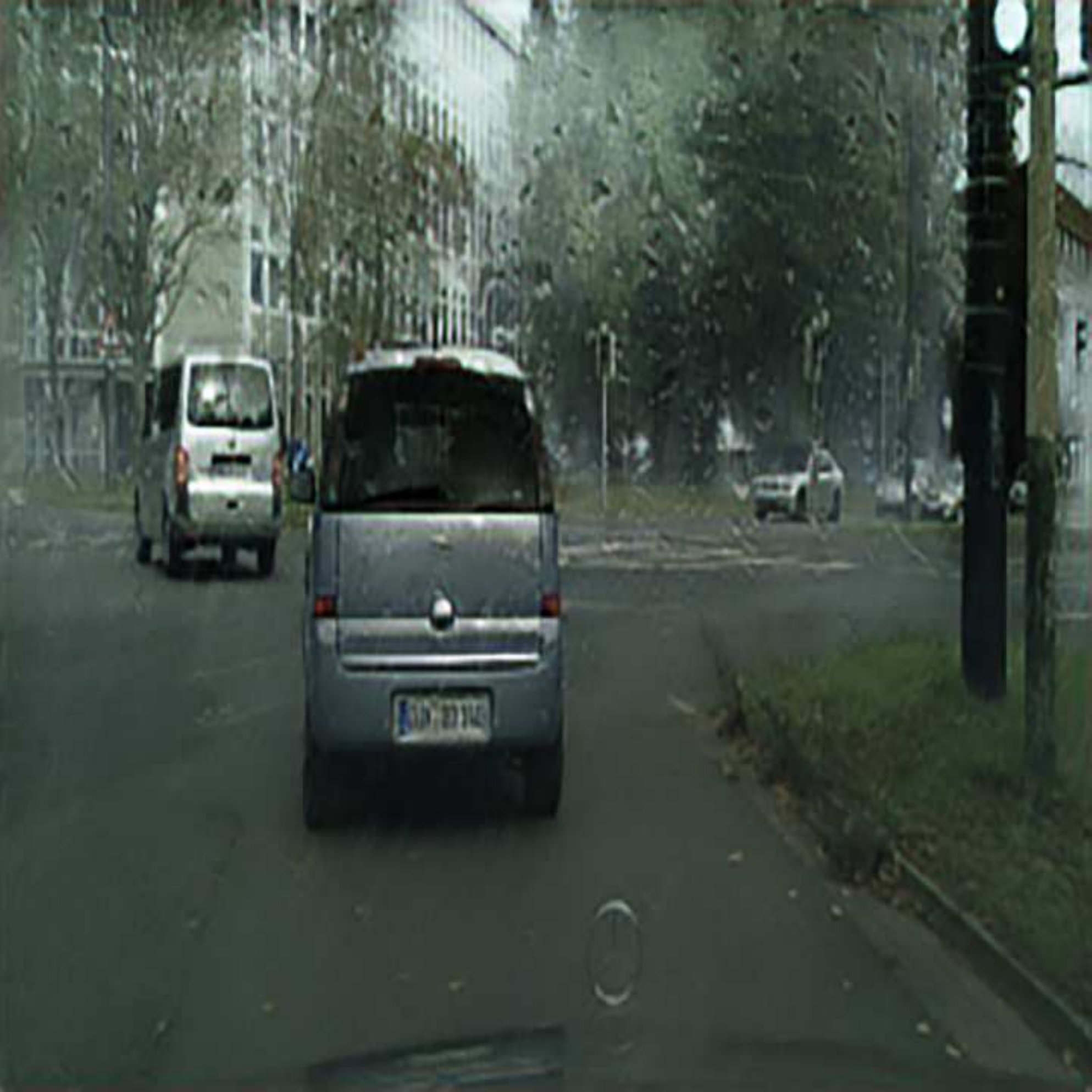}\vspace{2pt}
\end{minipage}}
\subfigure[H+L+A]{
\begin{minipage}[b]{0.12\linewidth}
\includegraphics[width=1\linewidth]{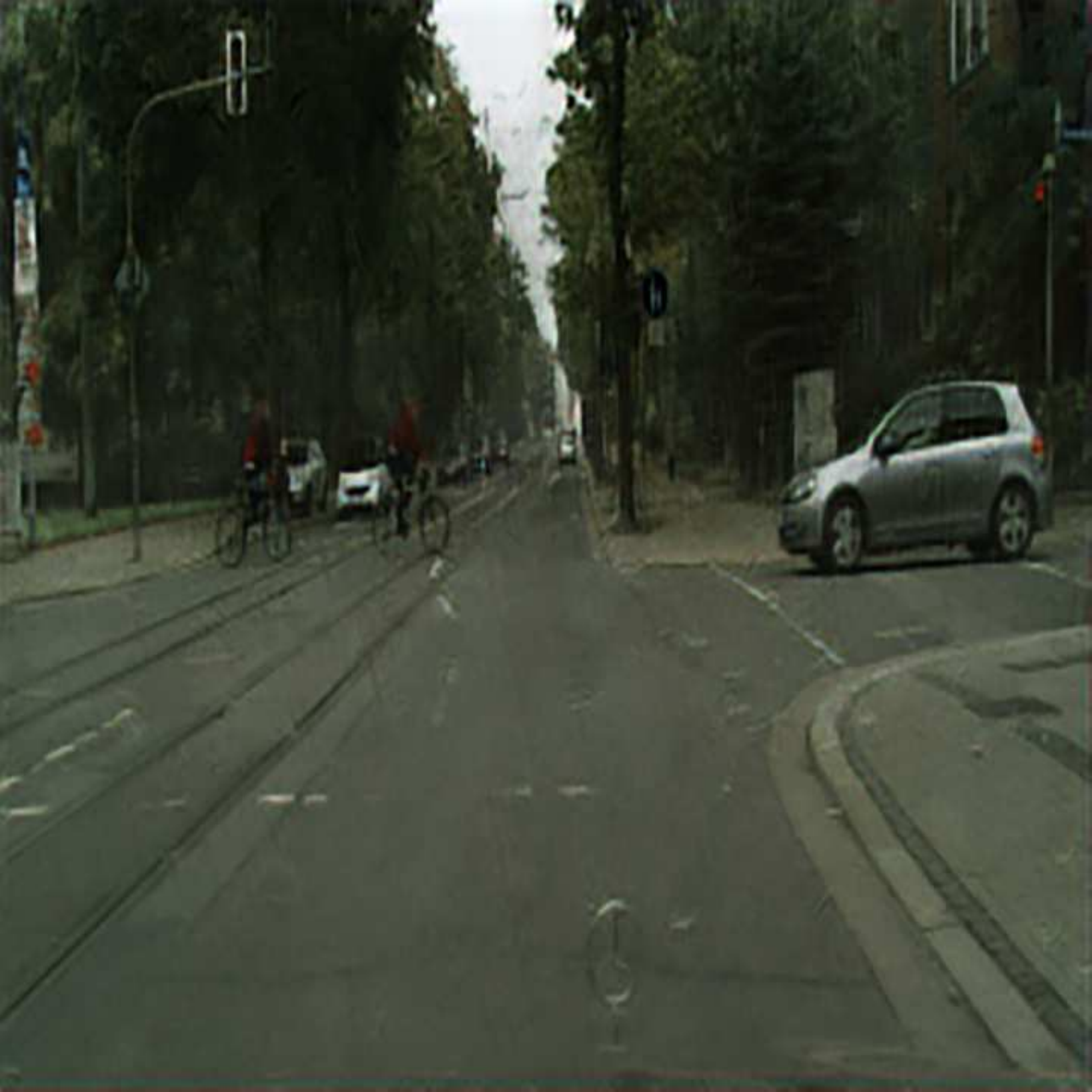}\vspace{2pt}
\includegraphics[width=1\linewidth]{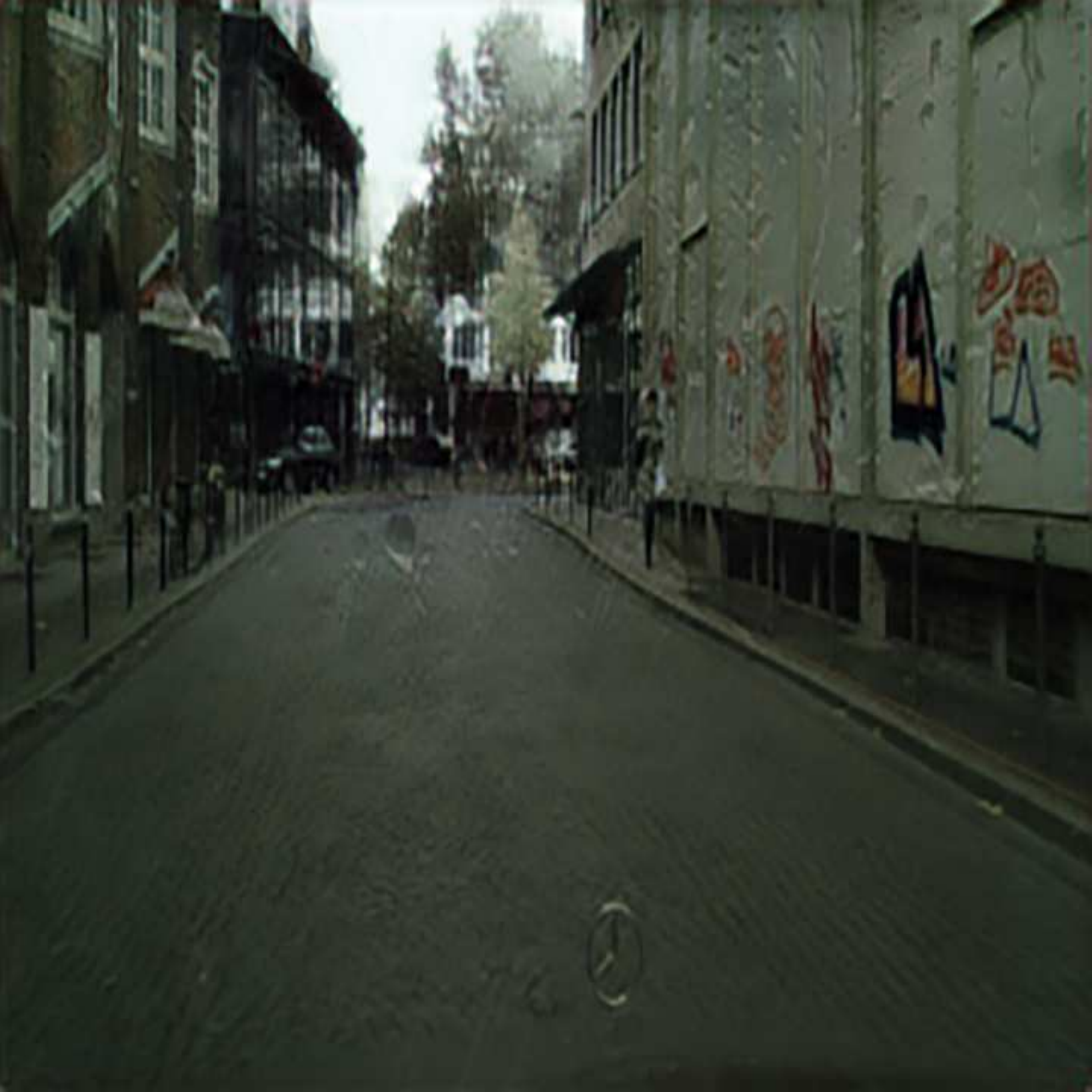}\vspace{2pt}
\includegraphics[width=1\linewidth]{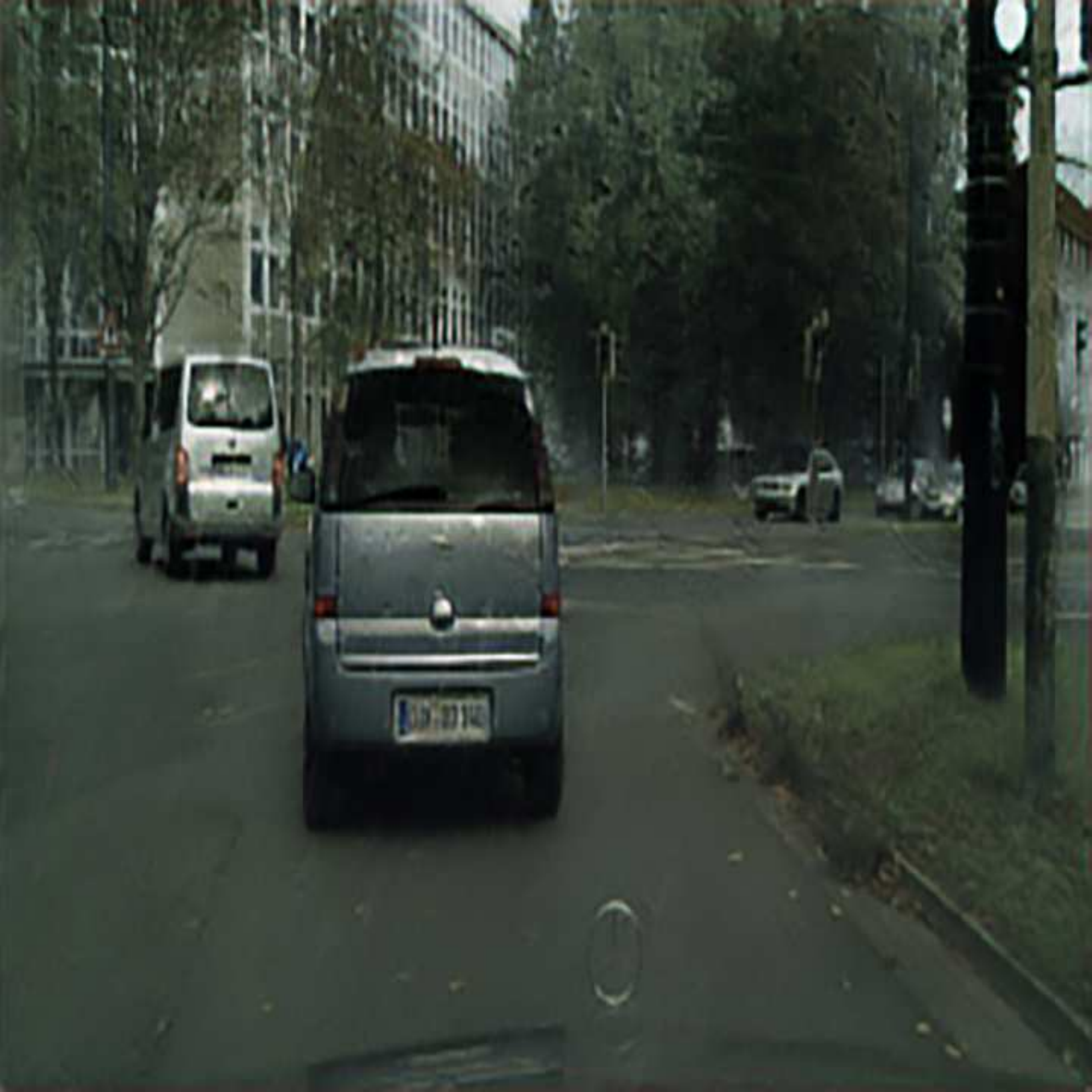}\vspace{2pt}
\end{minipage}}
\subfigure[H+L+R+A]{
\begin{minipage}[b]{0.12\linewidth}
\includegraphics[width=1\linewidth]{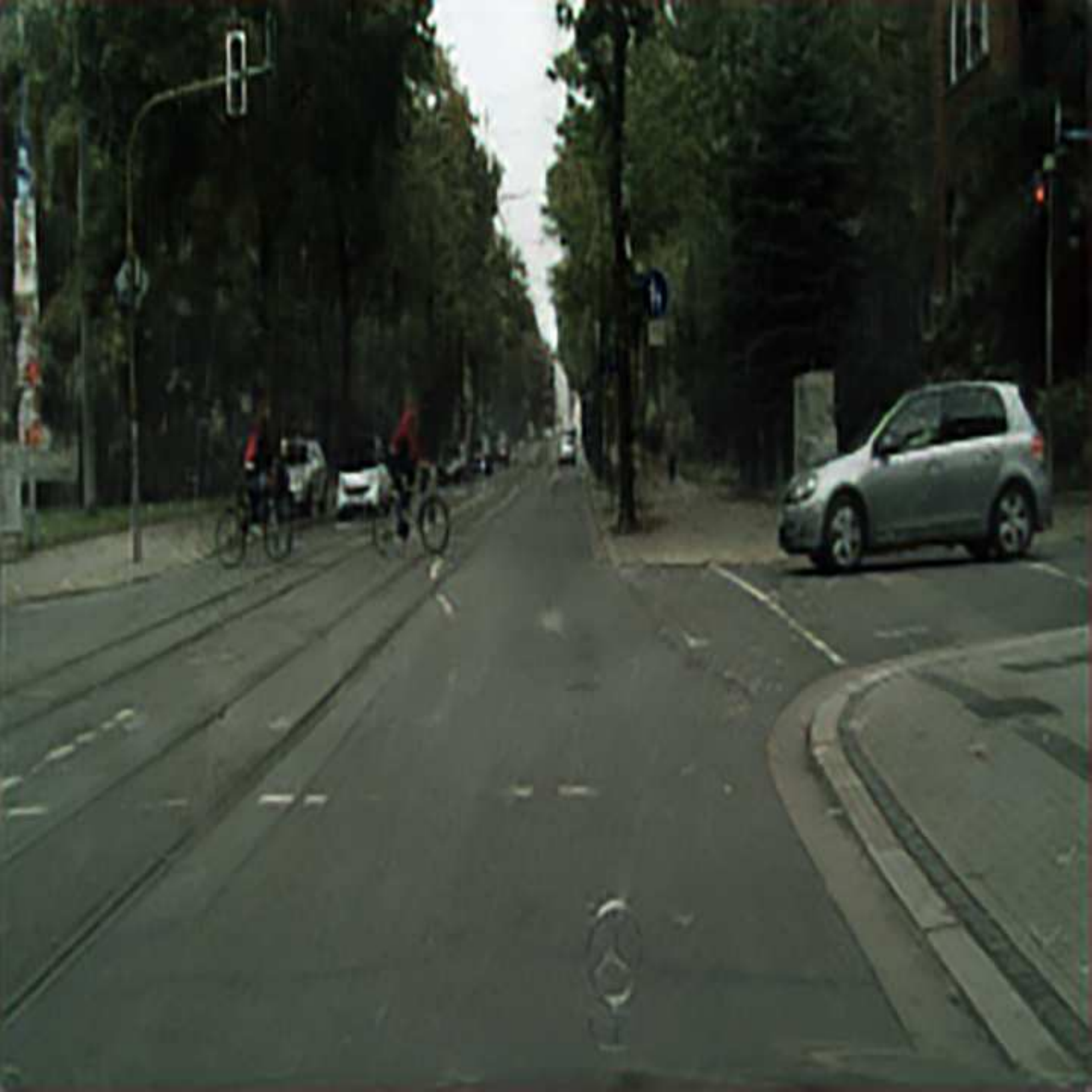}\vspace{2pt}
\includegraphics[width=1\linewidth]{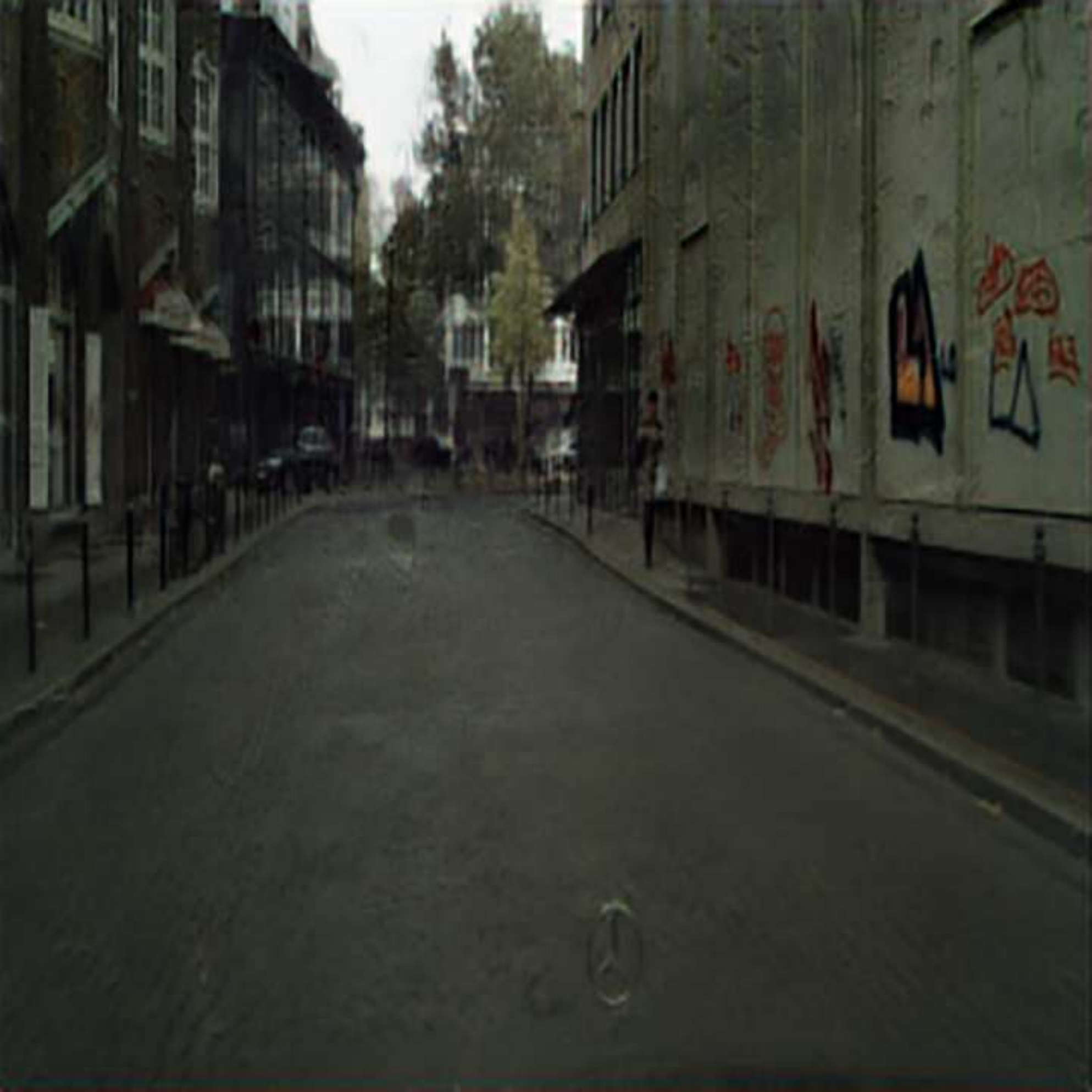}\vspace{2pt}
\includegraphics[width=1\linewidth]{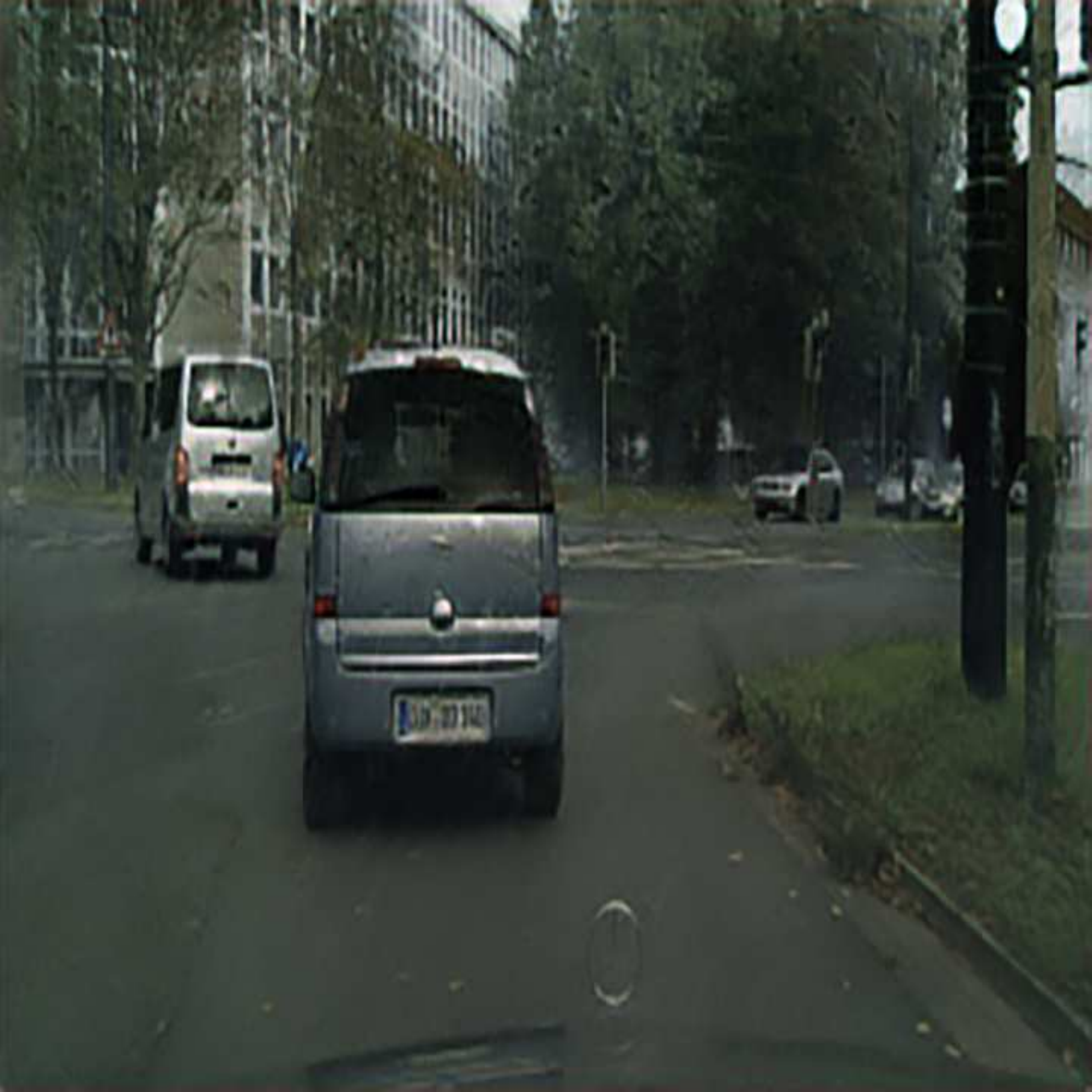}\vspace{2pt}
\end{minipage}}
\subfigure[Full]{
\begin{minipage}[b]{0.12\linewidth}
\includegraphics[width=1\linewidth]{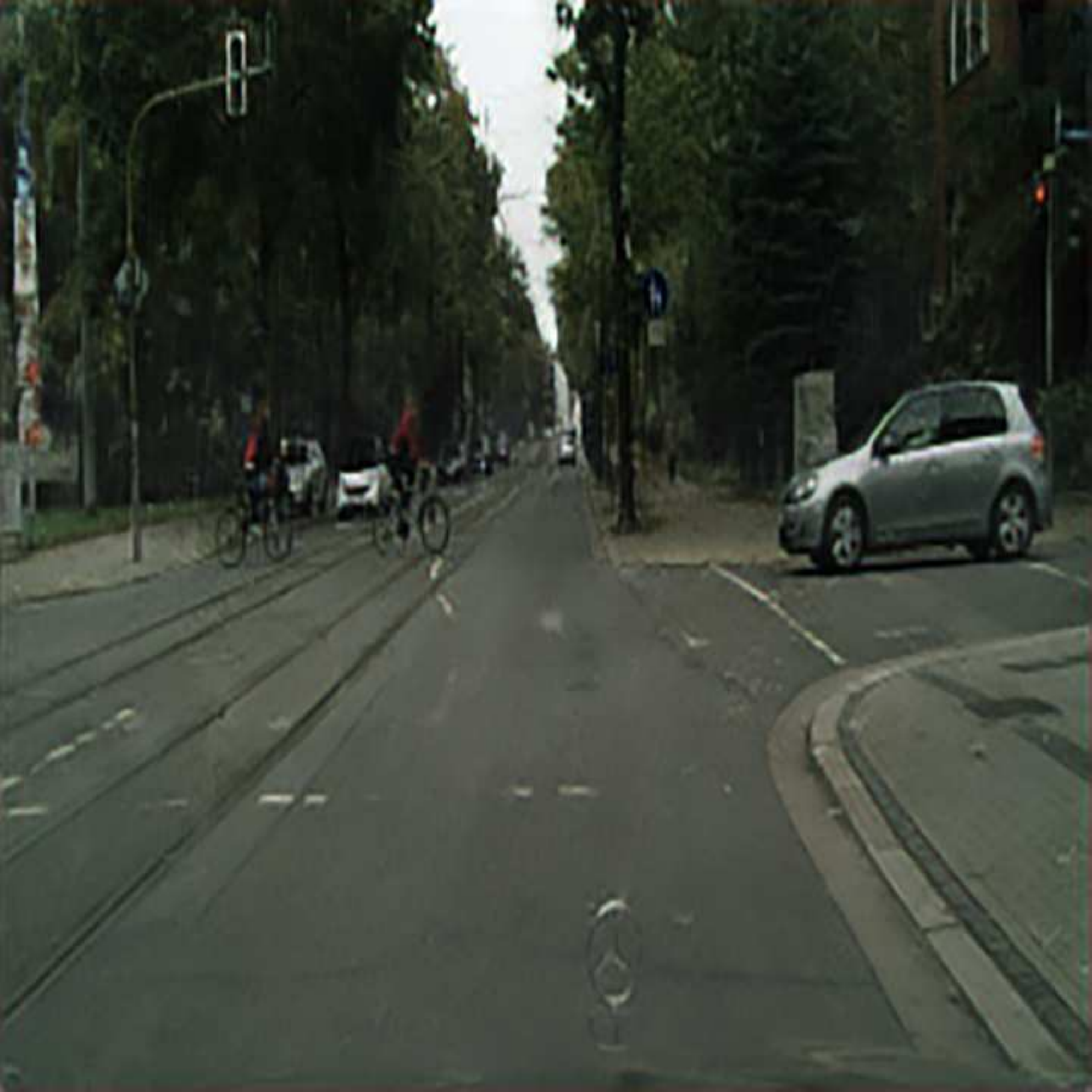}\vspace{2pt}
\includegraphics[width=1\linewidth]{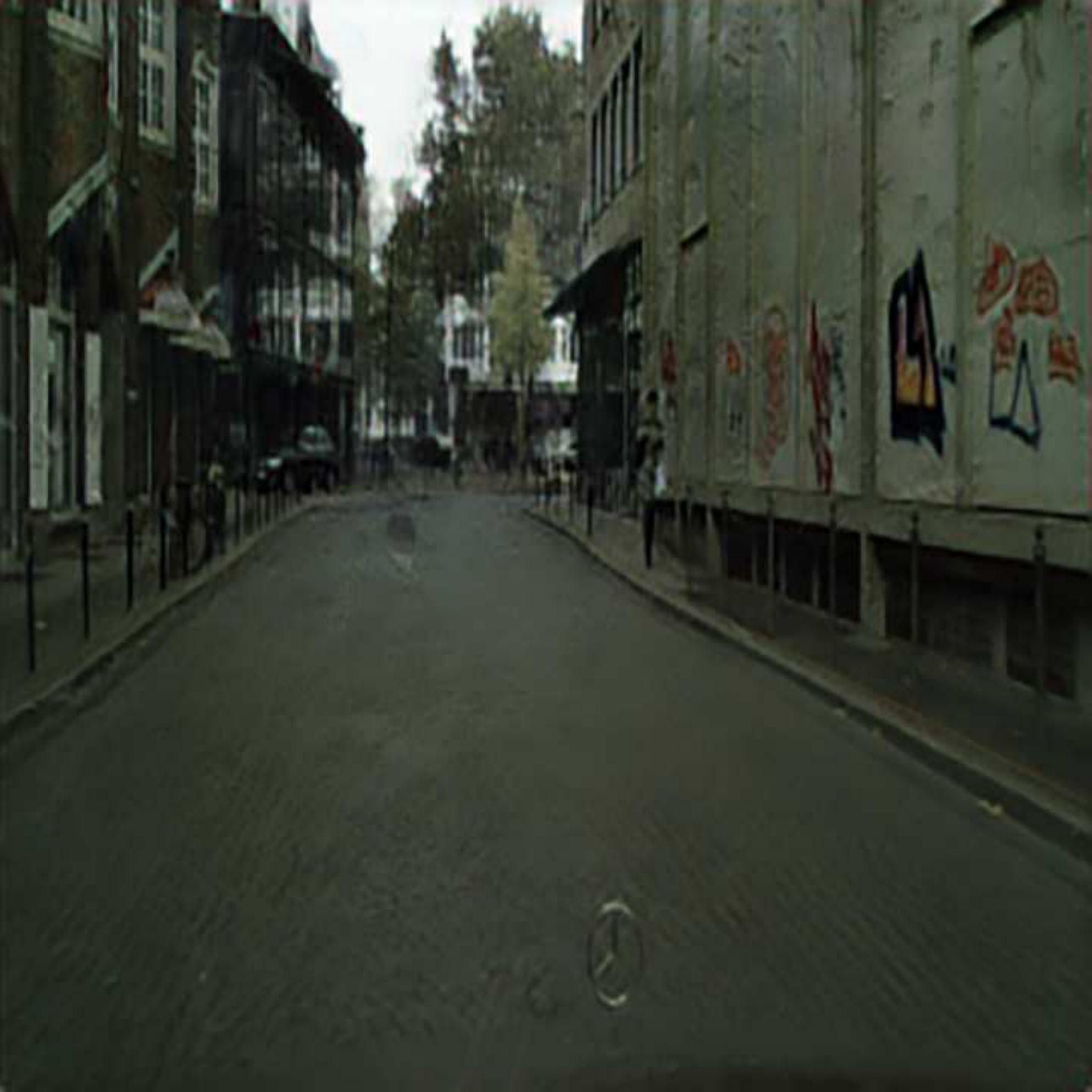}\vspace{2pt}
\includegraphics[width=1\linewidth]{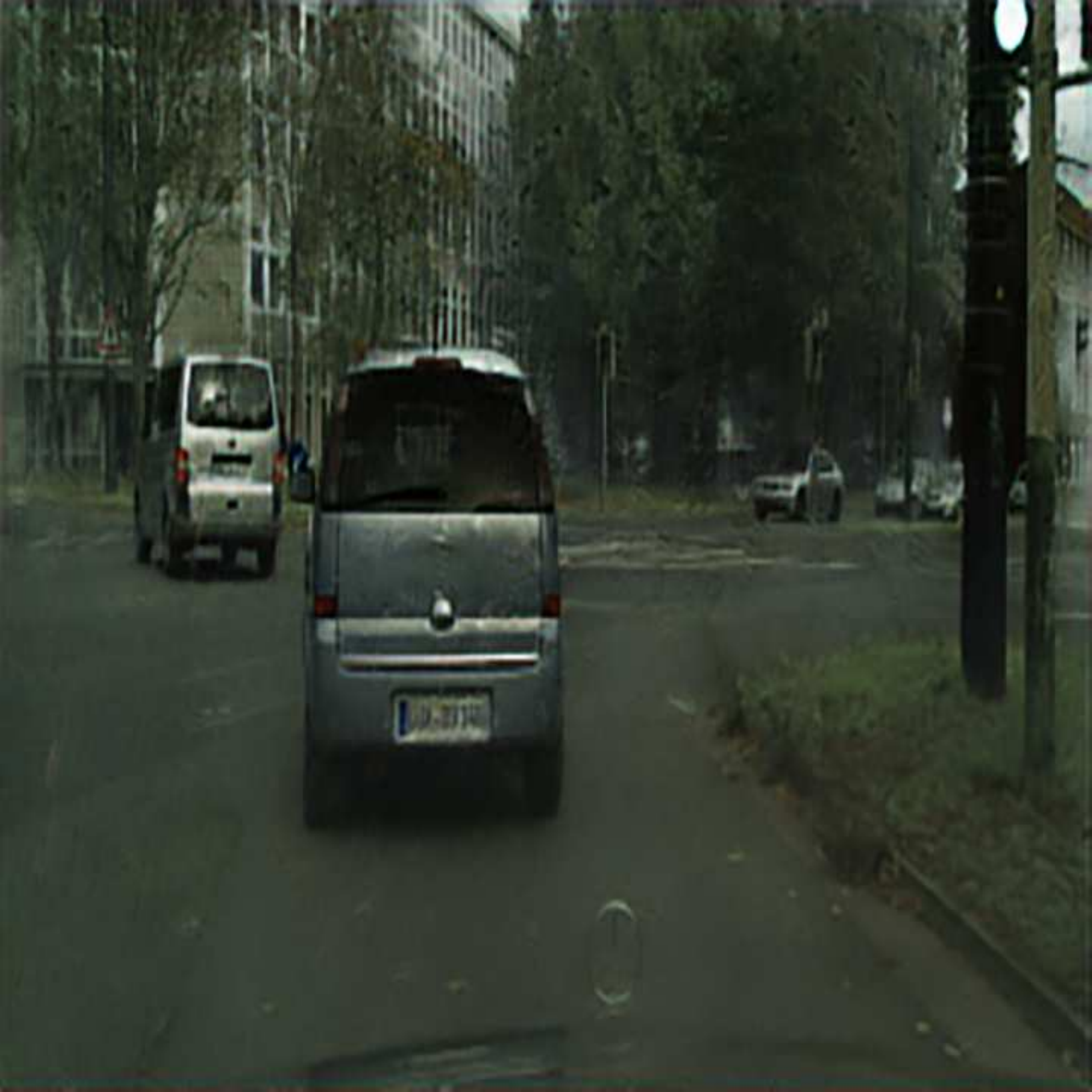}\vspace{2pt}
\end{minipage}}
\caption{Qualitative evaluation of different parts of our network architecture.} \label{fg:state}
\label{fig:f4}
\end{figure*}

\section{Experiments}
We evaluate the performance of our MBA-RainGAN and SOTA deraining network on both the proposed dataset of RainCityscapes++ and real rain images.
In detail, MBA-RainGAN is compared with the raindrop removal methods including Eigen et al. \cite{eigen2013restoring} and Qian et al. \cite{qian2018attentive}, the rain streak removal method, i.e., Wang et al. \cite{wang2019spatial}, and the rainy haze removal methods including Li et al. \cite{li2017aod} and Ren et al. \cite{ren2016single}. Also, we compare with Hu et al. \cite{hu2019depth} which removes both rain streaks and rainy haze. To make a fair comparison, all the methods are trained on the established RainCitysnapes++ dataset. In addition, we conduct an ablation study to validate the effectiveness of the proposed multi-branch attention scheme.

It is worth mentioning that, since there are many similar images in the original RainCityscape dataset, we decide to divide the total 8580 images in RainCityscapes++ into 429 groups, where each group contains 20 similar images. Then, we randomly select 5 different images from each group to construct the training set (contains 2145 images) and the testing set (contains 1030 images). And we also download 65 MOR photos from the Internet by using keyword search with "rain and fog photo" (from which we randomly pick three photos.).

\subsection{Comparison with the State-of-the-arts}
\textbf{Qualitative Evaluation} For fair comparison, firstly, we test the state-of-the-arts individual degradation removal algorithms. As shown in Fig. \ref{fig:f1}, most competitors suffer from distortions in MOR affected areas. Eigen et al. \cite{eigen2013restoring}, Qian et al. \cite{qian2018attentive} and Wang et al. \cite{wang2019spatial} fail to removes the raindrops entangled with dense rainy haze. In contrast, our method achieves the best performance in recovering the clean images. Hu et al. \cite{hu2019depth} removes rain streaks and rainy haze according to scene depth, but it cannot remove the raindrops randomly distributed on the glass. In conclusion, our method performs best in raindrops removal, rain streak removal and rainy haze removal. Then, observing that existing rain removal methods tend to ignore the rainy haze, we are motivated to apply a state-of-the-art haze removal method. See Fig. \ref{fig:f2}, the rain streaks and raindrops (which were not removed) would become more obvious after removing the rainy haze. Therefore, after dehazing, the methods of Qian et al. \cite{qian2018attentive}, Eigen et al. \cite{eigen2013restoring} and Wang et al. \cite{wang2019spatial} will yield better results. In addition, we also combine the methods from Hu et al. \cite{hu2019depth} and Qian et al. \cite{qian2018attentive}, which should be capable in raindrop removal, rain streak removal and rainy haze removal for MOR images. As can be seen, our method is considerably more effective in removing raindrops, rain streaks and rainy haze. Futhermore, we exhibit visual comparisons on real MOR images in Fig. \ref{fig:f3}, which demonstrates the superiority of our method in tackling various degradations in rainy scenes over the state-of-the-arts.

\textbf{Quantitative Evaluation}  As shown in Tab. \ref{tables1}, the average PSNR and SSIM are adopted to numerically assess the results. We can see that, both the PSNR and SSIM values of our MBA-RainGAN are highest among all the methods, which are consistent to the visual results in Fig. \ref{fig:f1}. That means our method outperforms SOTA networks by a large margin in removing MOR artifacts.

\subsection{Ablation Study} \label{ablation}
To validate the effectiveness of the designed network architecture, we conduct the ablation study on key components of MBA-RainGAN: A (autoencoder alone), H+A (autoencoder with HFAM), L+A (autoencoder with LFAM), H+L+A (autoencoder with HFAM and LFAM), H+L+R+A (autoencoder with HFAM, LFAM and RSAM) and H+L+R+A+D (the full version of MBA-RainGAN). As can be observed from Tab. \ref{tables3} and Fig. \ref{fig:f4}, the autoencoder without the attention scheme makes little contribution in removing the entangled MOR effect. When integrating the attention maps, the complete multi-branch attention scheme achieves the best results both quantitatively and qualitatively. Furthermore, the corresponding attentive discriminator enhances the image details and renders our result more realistic for human perception. Compared to Hu et al.'s depth-guided attention mechanism \cite{hu2019depth} (see Tabs. \ref{tables3} and \ref{tables1}), our H+L+A obtains higher SSIM and PSNR, it means that our attention mechanism can obtain better results in removing rain and rainy haze.
\subsection{Limitations} \label{ablation}
Our approach has at least three limitations. 1) We only consider the randomness of raindrop distribution. Hence, the constructed raindrop attention maps are related to the location of raindrop distribution. Maybe a deep consideration between the raindrop and the background will be more desirable. 2) As known, image smoothing is an ill-posed problem. We select the fast RGF to decompose the input image into high- and low-frequency components. Although RGF shows the good decomposition, we believe there are still many other filters available. 3) We use the multi-branch attention network to deal with the problem of removing rain streaks, raindrops and rain haze from single images. Although the proposed MBA-RainGAN can simultaneously derain the image and preserve the details, it requires more parameters and takes a little more time to train the network.


\section{Conclusion and Future Work}
In this work, we explore the visual effects of MOR and formulate a rain imaging model with rain streaks, rainy haze and raindrops. To cope with the MOR problem, we propose a new dataset by adding real raindrops to RainCityscape. Furthermore, we design a multi-branch attention generative adversarial network (termed an MBA-GAB), which adopts a three stage decomposition strategy to disentangle the MOR effects, i.e., the streak-aware decomposition with RGF, attention-level decomposition by multi-branch attentive network and the final image decomposition by the autoencoder, and our final result is refined by an attentive discriminator. Comprehensive experimental evaluations show that our method outperforms SOTA deraining methods in complex rainy scenes both quantitatively and qualitatively. In the future work, the potential of the proposed multi-branch attention scheme can be further explored for mixture degradation problems caused by photographing in various scenarios under bad weathers. \\


\bibliographystyle{ACM-Reference-Format}
\bibliography{refer,dehaze,dl_streak,prior}

\section*{Appendix}
\begin{appendix}
This appendix consists of:
\begin{enumerate}
  \item Formulation of image rain streak \& rainy haze
  \item Training details
  \item Running time
  \item Rolling guidance filter
  \item PSNR \& SSIM of single images
\end{enumerate}
\section{Formulation of image rain streak \& rainy haze}
According to Garg and Nayar \cite{garg2006photorealistic}, the visual intensity of the rain streak and the rainy haze layers depends on the scene depth from the camera to the underlying scene objects behind the rain.
Therefore, we formulate the rain streak layer $S(x)$  as:
\begin{equation}\label{eq:depth}
S(x)=S(x)_\textbf{pattern} \ast t_r(x)
\end{equation}
where $S(x)_\textbf{pattern}\in[0,1]$ is an intensity image of uniformly-distributed rain streaks in the image space, $\ast$ represents a pixel-wise multiplication, and $t_r(x)$ represents the rain streak intensity map defined as:
\begin{equation}\label{eq:trdepth}
t_r(x)=\textbf{e}^{-\alpha\textbf{max}(d_1,d(x))}
\end{equation}
where $\alpha$ is an attenuation coefficient that controls the rain streak intensity, $d(x)$ means the scene depth in the rain model \cite{garg2006photorealistic}. When $t_r(x)$ approaches to the maximum rain streak intensity, it will be equal to to $e^{d1}$, and $t_r(x)$ starts with $t_{r0}$ and gradually drops to zero after $d(x)$ goes beyond $d_1$.

Meanwhile, according to the standard optical model \cite{koschmieder1924theorie} that simulates the image degradation process, the visual intensity of rainy haze increases exponentially with the scene depth. Hence, we model the rainy haze layer $A(x)$ as:
\begin{equation}\label{eq:fogdepth}
A(x)=1 - \textbf{e}^{-\beta d(x)}
\end{equation}
where $\beta$ is an attenuation coefficient that controls the thickness of rainy haze, and a larger $\beta$ means a thicker rainy haze, and vice versa.
\begin{table*}[htbp]
 \centering
 \small
\begin{tabular}{cccccccc}
\toprule
Method& Eigen et al. & Qian et al. & Wang et al. & Hu et al. & Li et al. &Ren et al. & Ours\\
\midrule
Avg time& 1.43s & 1.62s & 1.53s & 1.36s & 1.21s &1.02s & \textbf{1.45s}\\
\bottomrule
\end{tabular}
\caption{Averaged time on RainCityscapes++ of different methods for removing MOR.}
\label{tables1}
\end{table*}
\section{Training details}
During the procedure of our training, a $376\times256$ patch is randomly cropped from each rainy image. For optimizing our network, we use the Adam  with a min-batch size of 10 to train the network. The total number of epochs is set to be 50, and each epoch includes 1000 iterations. We initialize the learning rate as 0.02. All the experiments are running with an Nvidia 2080Ti GPU.

\section{Running time}
We give the averaged statistical time of all methods on the  dataset of RainCityscapes++ in Table \ref{tables1}. It is easily observed that, our method is neither the fastest one nor the slowest one, we think its performance is acceptable.
\begin{figure*}[htbp]
	\centering
	\subfigure[Input image]{
		\begin{minipage}[b]{0.3\linewidth}
			\includegraphics[width=1\linewidth]{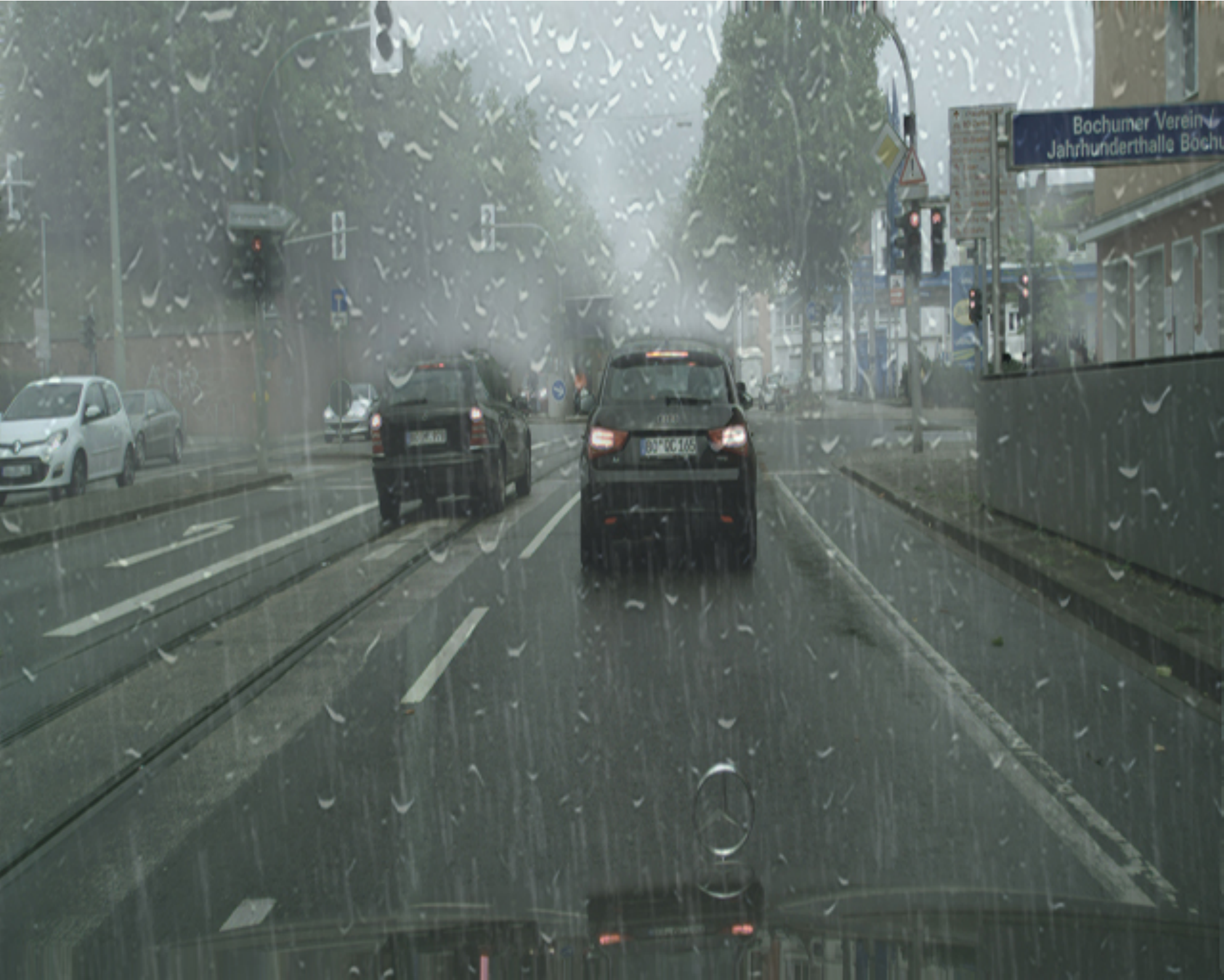}\vspace{4pt}
			\includegraphics[width=1\linewidth]{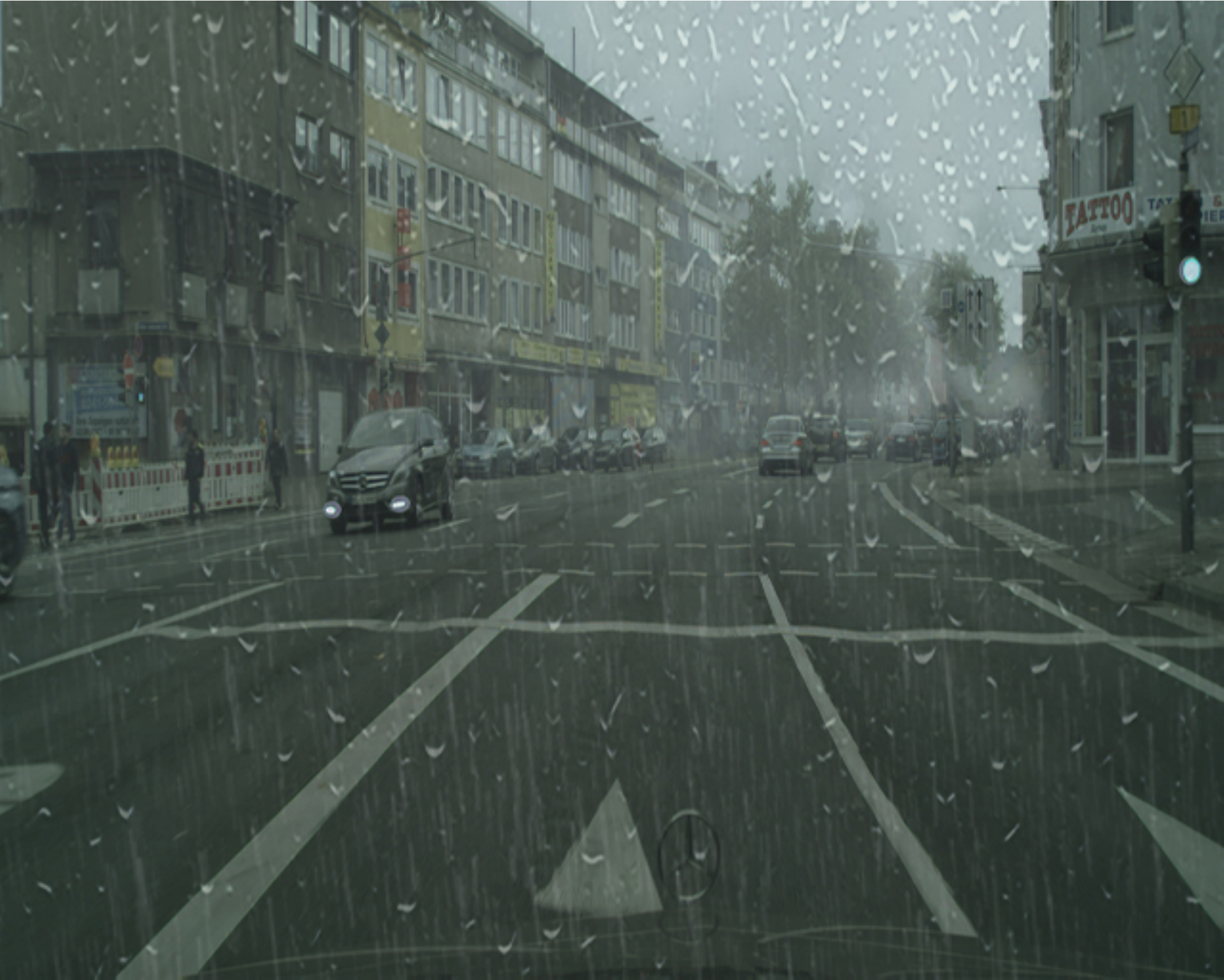}\vspace{4pt}
	\end{minipage}}
	\subfigure[High pass]{
		\begin{minipage}[b]{0.3\linewidth}
			\includegraphics[width=1\linewidth]{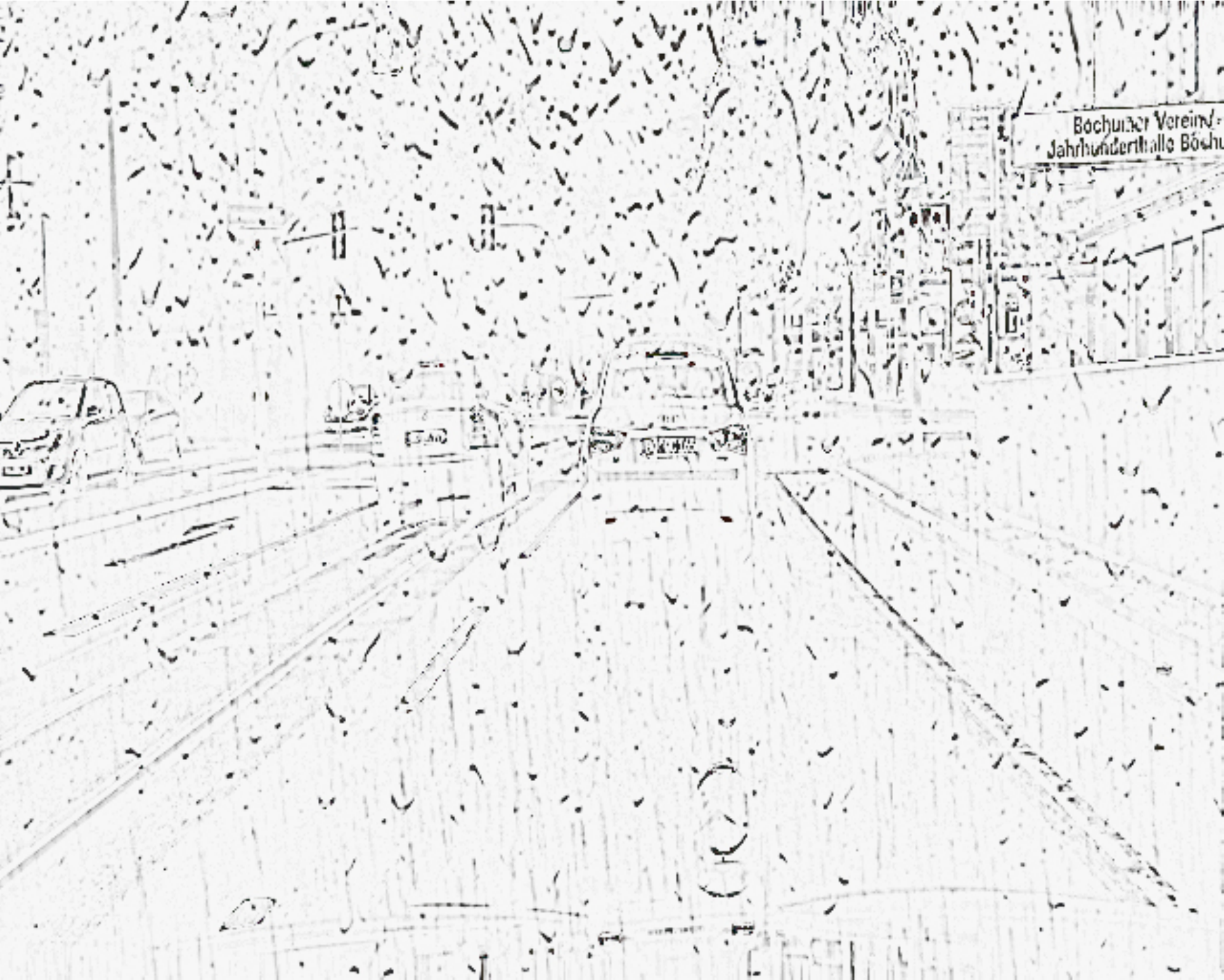}\vspace{4pt}
			\includegraphics[width=1\linewidth]{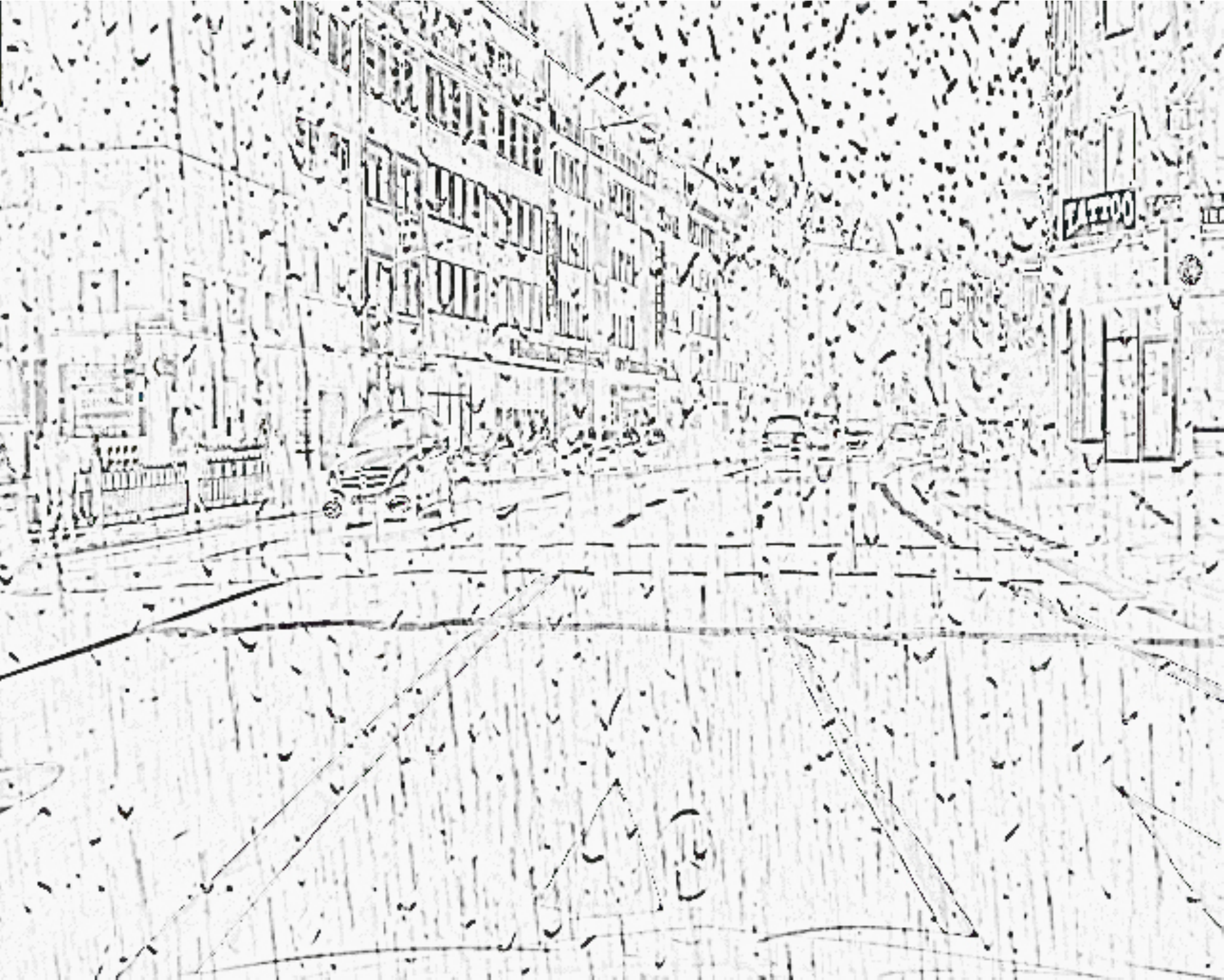}\vspace{4pt}
	\end{minipage}}
	\subfigure[Low pass]{
		\begin{minipage}[b]{0.3\linewidth}
		    \includegraphics[width=1\linewidth]{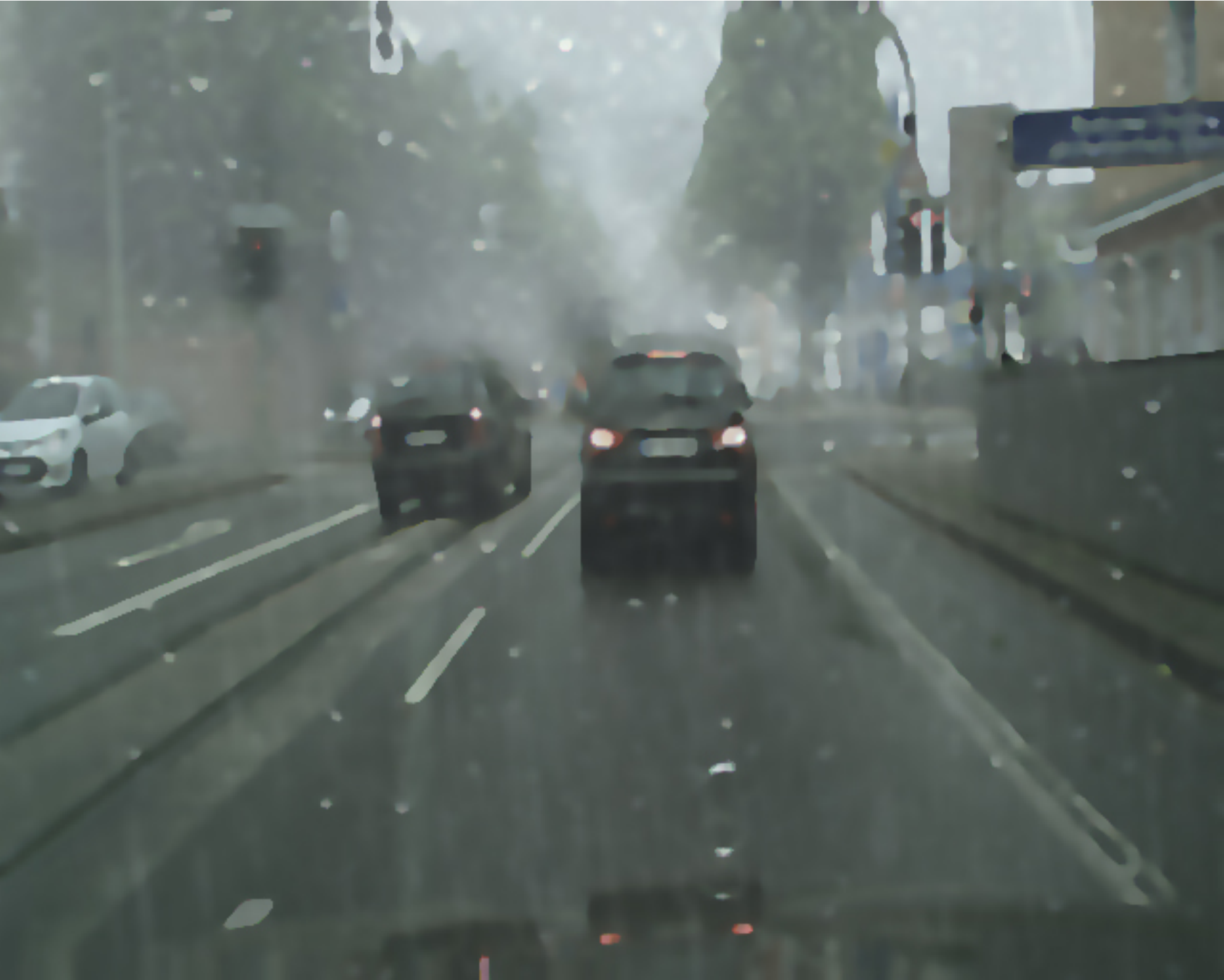}\vspace{4pt}
			\includegraphics[width=1\linewidth]{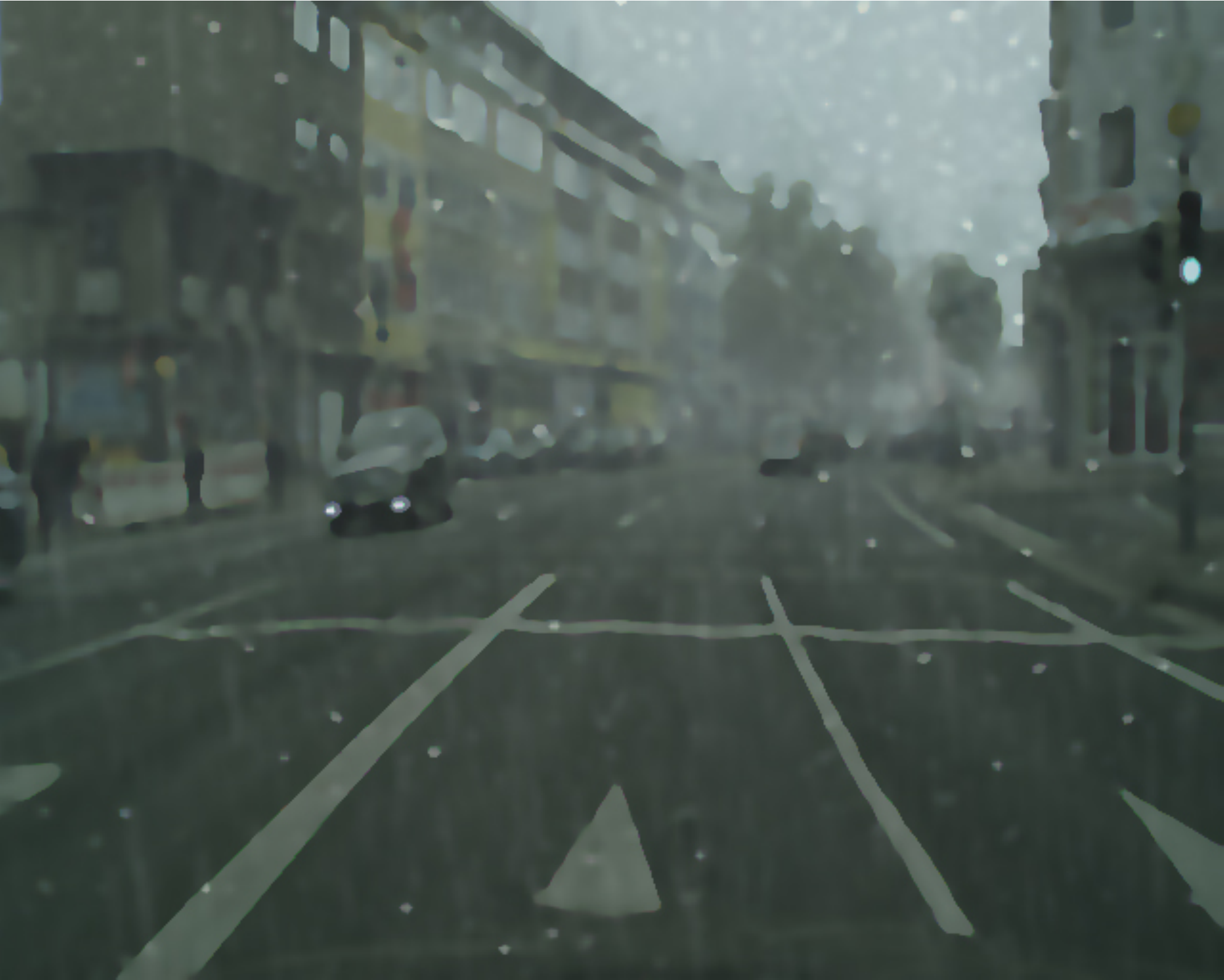}\vspace{4pt}
	\end{minipage}}
	\caption{ Image decomposition on two images by RGF. Structures and details are well decomposed for the later use.}
	\label{fig:label1}
\end{figure*}

\begin{table*}[htbp]
 \centering
 \small
\begin{tabular}{ccccccccc}
\toprule
Method& Li et al. & Ren et al. & Eigen et al. & Wang et al. & Qian et al. & Hu et al. & Ours & Input \\
\midrule
PSNR& 11.32 & 13.96 & 10.10 & 9.87 & 21.37 & 23.00 & \textbf{25.05} & 9.58\\
SSIM& 0.58 & 0.63 & 0.57 & 0.57 & 0.76 & 0.79 & \textbf{0.82} & 0.56\\
\bottomrule
\midrule
PSNR& 20.58 & 22.02 & 20.62 & 21.31 & 23.51 & 26.65 & \textbf{31.45}& 19.8\\
SSIM& 0.75 & 0.74 & 0.76 & 0.77 & 0.80 & 0.86 & \textbf{0.91}& 0.75\\
\bottomrule
\midrule
PSNR& 20.71 & 20.41 & 21.08 & 21.10 & 23.53 & 22.71 & \textbf{32.90}& 20.37\\
SSIM& 0.76 & 0.71 & 0.78 & 0.77 & 0.76 & 0.80 & \textbf{0.91}& 0.7\\
\bottomrule
\midrule
PSNR& 19.29 & 20.86 & 18.2 & 18.17 & 25.22 & 25.60 & \textbf{31.12}& 17.3\\
SSIM& 0.73 & 0.71 & 0.73 & 0.74 & 0.83 & 0.84 & \textbf{0.91}& 0.72\\
\bottomrule
\midrule
PSNR& 18.93 & 21.19 & 18.70 & 18.14 & 24.39 & 25.62 & \textbf{27.85}& 17.0\\
SSIM& 0.75 & 0.76 & 0.73 & 0.75 & 0.83 & 0.85 & \textbf{0.89}& 0.70\\
\bottomrule
\end{tabular}
\caption{PNSR and SSIM of single images used in the Figure 9 of the manuscript.}
\label{tables2}
\end{table*}
\begin{table*}[htbp]
 \centering
 \small
\begin{tabular}{cccccccc}
\toprule
Method& Li+Qian & Ren+Qian & Eigen+Ren & Wang+Li & Hu+Qian & Ours & Input\\
\midrule
PSNR& 19.31 & 18.75 & 21.85 & 20.07 & 24.85& \textbf{25.24}& 18.33\\
SSIM& 0.81 & 0.80 & 0.82 & 0.80 & 0.83 & \textbf{0.84}& 0.79\\
\bottomrule
\midrule
PSNR& 20.34 & 20.17 & 20.41 & 18.45 & 27.42& \textbf{33.18}& 16.22\\
SSIM& 0.78 & 0.78 & 0.80 & 0.76 & 0.87 & \textbf{0.92}& 0.74\\
\bottomrule
\midrule
PSNR& 19.08 & 18.66 & 19.33 & 16.82 & 26.70& \textbf{28.23}& 14.54\\
SSIM& 0.76 & 0.75 & 0.77 & 0.73 & 0.86 & \textbf{0.89}& 0.73\\
\bottomrule
\end{tabular}
\caption{PNSR and SSIM of single images used in the Figure 10 of the manuscript.}
\label{tables3}
\end{table*}
\section{Rolling guidance filter}
We employ the fast and effective rolling guidance filter (RGF) \cite{zhang2014rolling} to decompose MOR images. RGF is essentially an iterative joint bilateral filter, whose guidance image is updated iteratively as
\begin{equation}\label{eq:roll}
J^{k+1}(p)=\frac{1}{K_{p}} \sum_{q\in N(p)} W_{s}(\rVert p-q \rVert) W_{r}(\rVert J^{k}(p)-J^{k}(q) \rVert)I(q)
\end{equation}
where $k$ denotes the iteration number, $J^{k+1}(p)$ denotes the pixel intensity value of the $(k+1)$-th iteration, $N(p)$ denotes the neighboring pixel set of pixel $p$, $W_{s}$ (measuring spatial similarity) and $W_{r}$ (measuring intensity similarity) are two Gaussian functions with the standard deviations $\sigma_s$ and $\sigma_r$, respectively. $J^{0}$ is usually set be \textit{zero}.

The principle of RGF can be understood as follows. In the first iteration, the filter is actually a Gaussian filter since $J^{0}$ = 0. Thus, given some proper values of $\sigma_s$ and $\sigma_r$, image features whose scales are smaller than $\sigma_s$ can be smoothed empirically. Meanwhile, image features whose scales are larger than $\sigma_s$ are also blurred  to some degrees, i.e., edges.
In the later iterations, $J^{t}$ is no longer zero. However, for the small-scale features (details) that have been smoothed out in the first iteration, their guidance pixels are almost equal to each other. Thus, the weight term $W_{r}$ approximately equals to $1$, and the filter is still a Gaussian filter, resulting in small-scale features still being removed.
For large-scale features (structures), they are recovered gradually since the joint bilateral filter sharpens them.
We set $\sigma_s$, $\sigma_r$, $N_{iter}$ (the total iteration number) to be $3.0$, $0.1$, $6$, which behave well in our experiment. As shown in Figure \ref{fig:label1}, we decompose the input image into high- and low-frequency components in advance that is beneficial to the task of image MOR removal.

\section{PSNR \& SSIM of single images}
In the manuscript, we have illustrated the averaged PSNR and SSIM on the established RainCitysnapes++. In this section, more quantitative results on single images are provided.
The quantitative results in Table \ref{tables2} correspond to the visual results (of different methods) in the Figure 9 of the manuscript, and the quantitative results in Table \ref{tables3} correspond to the visual results (of the combination of different methods) in the Figure 10 of the manuscript.
\end{appendix}


\end{document}